\documentclass[nohyperref]{article}

\usepackage{math_commands}
\usepackage{microtype}
\usepackage{graphicx}
\usepackage{subfigure}
\usepackage{stfloats}
\usepackage{booktabs} 
\usepackage{enumitem}
\usepackage{enumerate}
\usepackage{hyperref}

\usepackage[accepted]{icml2022}

\usepackage{amsmath}
\usepackage{amssymb}
\usepackage{mathtools}
\usepackage{amsthm}

\usepackage[capitalize,noabbrev]{cleveref}

\theoremstyle{plain}

\theoremstyle{definition}

\theoremstyle{remark}

\icmltitlerunning{Sparse Double Descent}

\begin{document}

\twocolumn[
\icmltitle{Sparse Double Descent: Where Network Pruning Aggravates Overfitting}



\icmlsetsymbol{equal}{*}

\begin{icmlauthorlist}
\icmlauthor{Zheng He}{buaa}
\icmlauthor{Zeke Xie}{utokyo,riken}
\icmlauthor{Quanzhi Zhu}{buaa}
\icmlauthor{Zengchang Qin}{buaa}
\end{icmlauthorlist}

\icmlaffiliation{buaa}{Intelligent Computing and Machine Learning Lab, School of ASEE, Beihang University, Beijing, China}
\icmlaffiliation{utokyo}{The University of Tokyo}
\icmlaffiliation{riken}{RIKEN Center for AIP}

\icmlcorrespondingauthor{Zheng He}{zhenghe@buaa.edu.cn}
\icmlcorrespondingauthor{Zengchang Qin}{zcqin@buaa.edu.cn}
\icmlkeywords{Deep Learning, Network Pruning, Overfitting, Double Descent}

\vskip 0.3in
]



\printAffiliationsAndNotice{}  

\begin{abstract}
People usually believe that network pruning not only reduces the computational cost of deep networks, but also prevents overfitting by decreasing model capacity. However, our work surprisingly discovers that network pruning sometimes even aggravates overfitting. We report an unexpected \textit{sparse double descent} phenomenon that, as we increase model sparsity via network pruning, test performance first gets worse (due to overfitting), then gets better (due to relieved overfitting), and gets worse at last (due to forgetting useful information). While recent studies focused on the \textit{deep double descent} with respect to model overparameterization, they failed to recognize that sparsity may also cause double descent. In this paper, we have three main contributions. First, we report the novel sparse double descent phenomenon through extensive experiments. Second, for this phenomenon, we propose a novel learning distance interpretation that the curve of $\ell_{2}$ learning distance of sparse models (from initialized parameters to final parameters) may correlate with the sparse double descent curve well and reflect generalization better than minima flatness. Third, in the context of sparse double descent, a winning ticket in the lottery ticket hypothesis surprisingly may not always win.

\end{abstract}

\section{Introduction}
\label{sec:intro}

Deep neural networks (DNNs) have achieved great empirical success in recent years, which are usually trained with much more model parameters than training examples \citep{lecun2015deep}.  
The extreme overparameterization gives DNNs excellent approximation as well as a prohibitively large model capacity \citep{cybenko1989approximation, funahashi1989approximate, hornik1989multilayer, hornik1993some}. Overparameterized DNNs can even easily memorize entire random-labeled dataset \citep{zhang2017rethink_generalization}, which suggests that DNNs are kind of ``good at'' overfitting. 

However, in practice, DNNs do not learn via pure memorization and often achieve higher generalization performance on many tasks than smaller models \citep{szegedy2015going, neyshabur2015search, arpit2017closer}. Recent studies even reported an interesting deep double descent phenomenon \citep{Belkin2019reconciling, nakkiran2020deepdoubledescent, loog2020brief, d2020double, yang2020rethinking} that, as the model capacity increases, test performance first gets better, then gets worse (due to classical overfitting), and gets better at last (due to relieved overfitting). Deep double descent motivated us to rethink the widely held viewpoint that network pruning reduces model capacity thus mitigates overfitting \citep{lecun1990optimal, hassibi1992second, Molchanov2017variational_dropout, hoefler2021sparsity}. 

As the extreme overparameterization of DNNs may cause deep double descent, may the sparsification of DNNs also cause a double descent phenomenon symmetric to the existing ``deep double descent''? Our answer is affirmative. We call such phenomenon ``sparse double descent''. 

\begin{figure*}[ht]
\center
\includegraphics[width=0.32\linewidth]{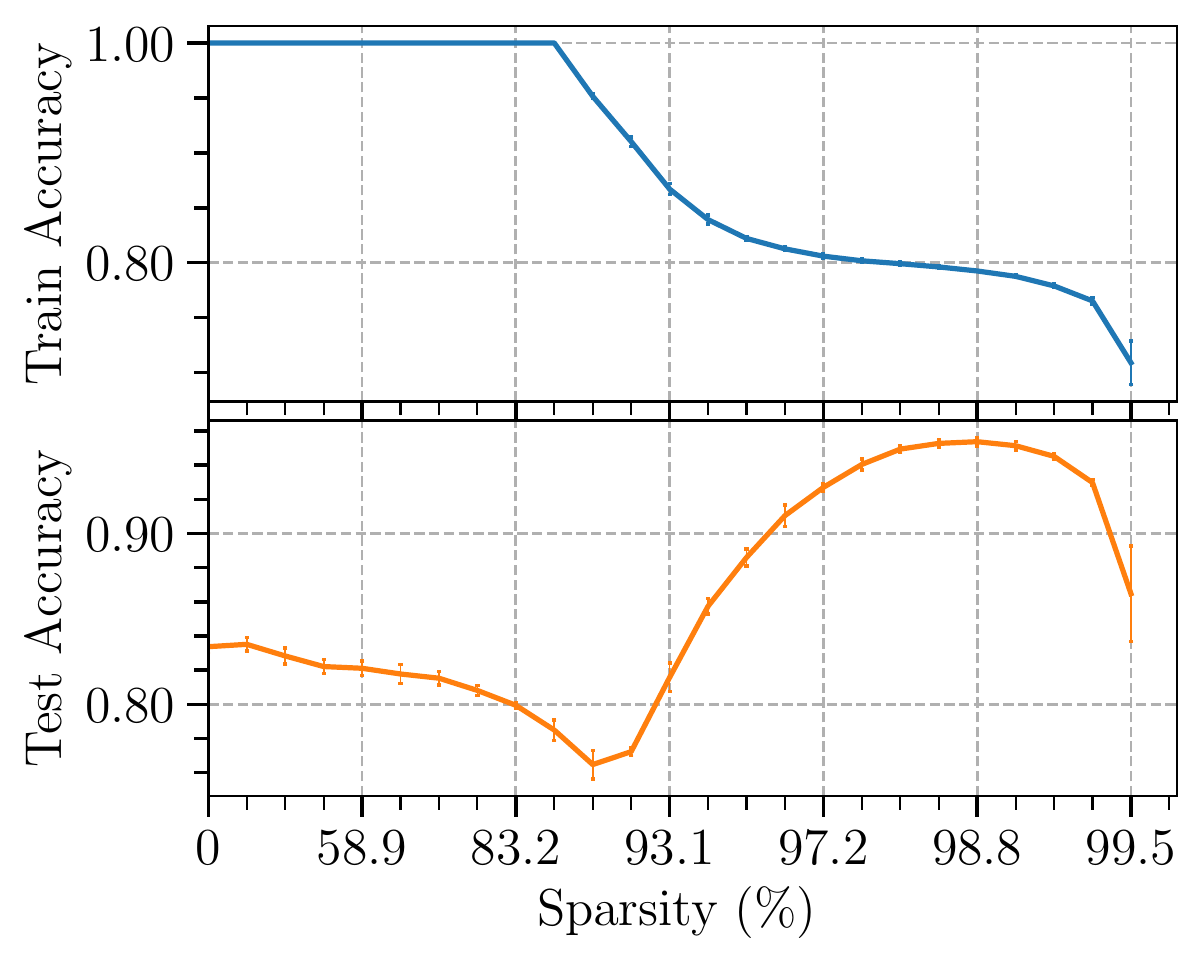}
\includegraphics[width=0.32\linewidth]{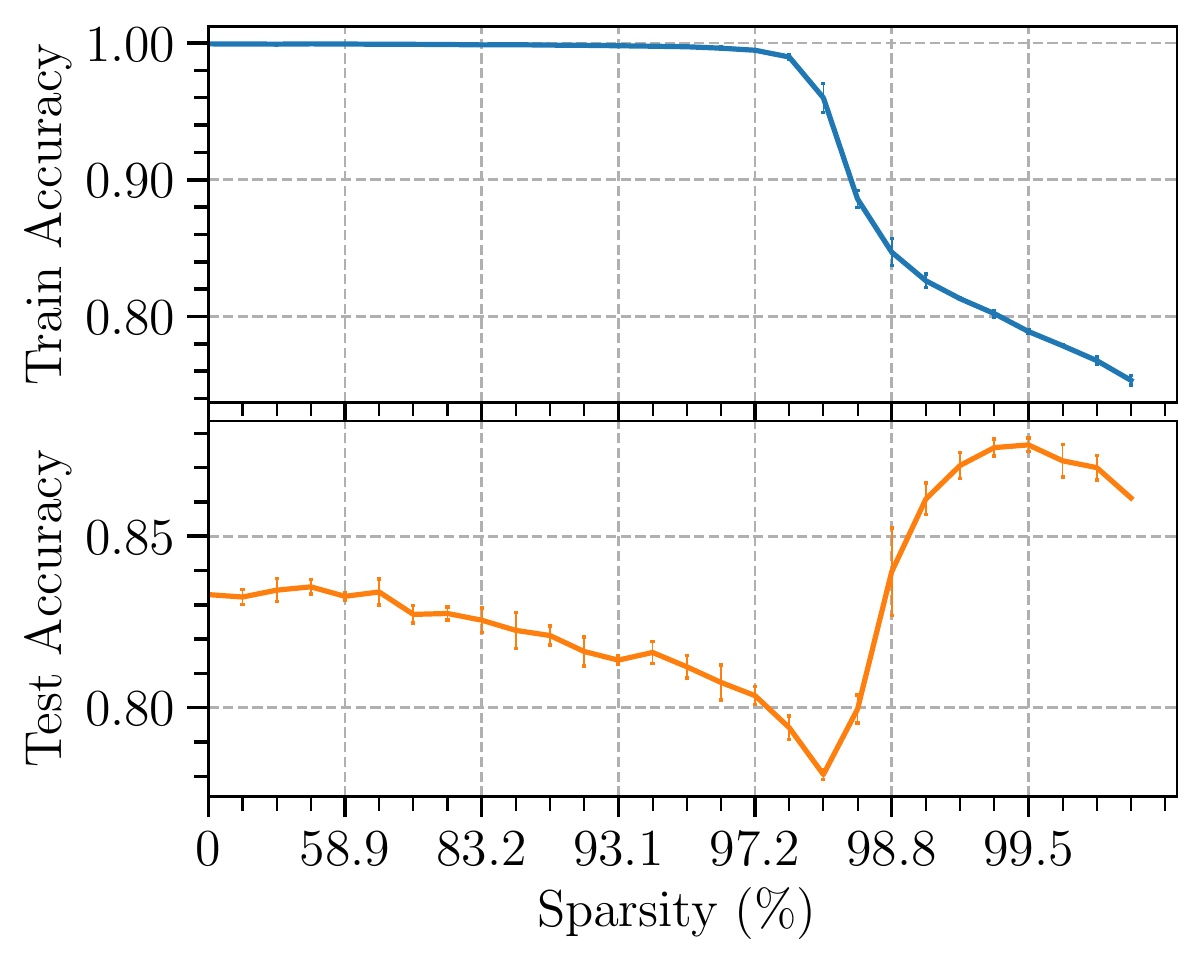}
\includegraphics[width=0.32\linewidth]{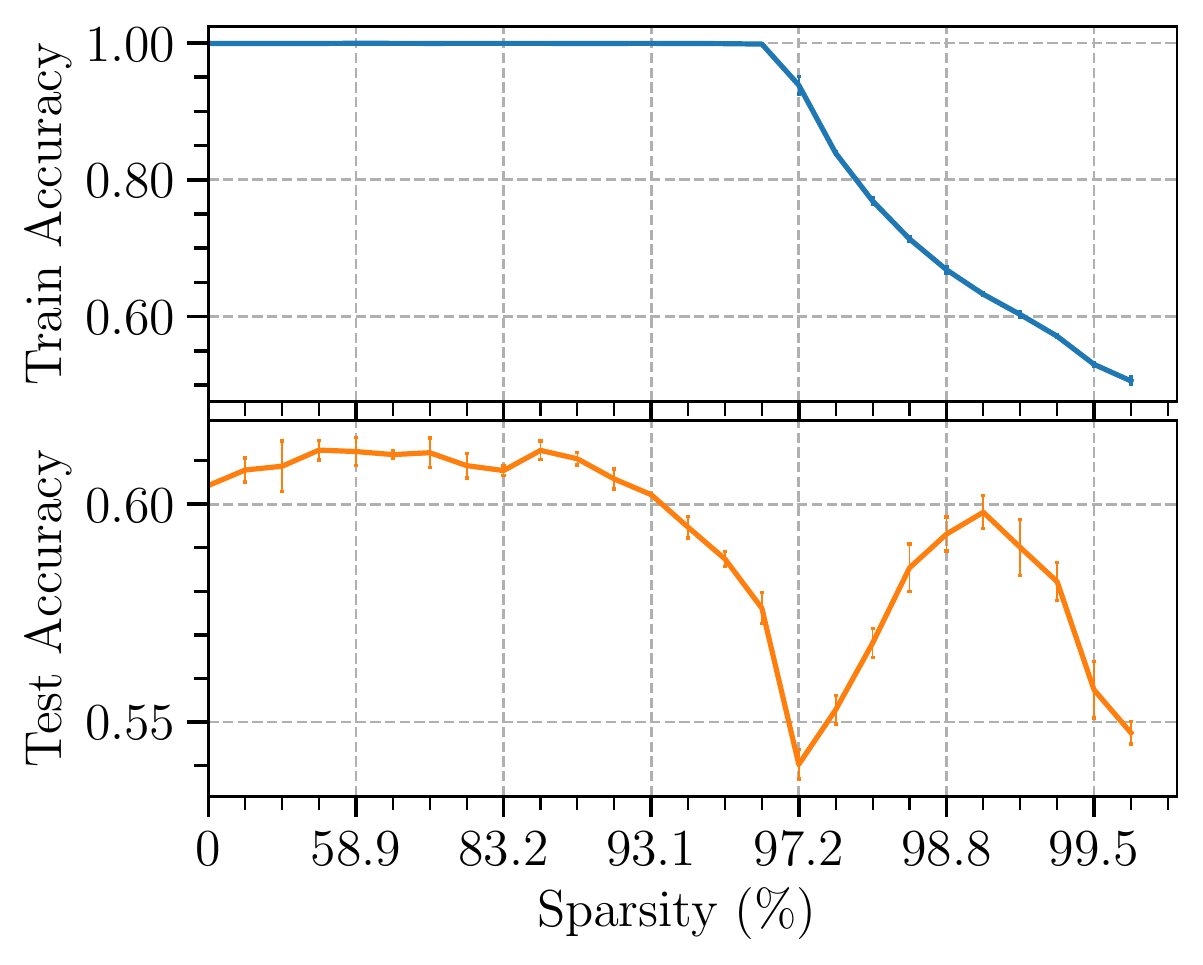}
\caption{Sparse Double Descent of LeNet-300-100 on MNIST and ResNet-18 on CIFAR with $20\%$ symmetric label noise. We evaluate how the train and test performance of networks depend on model sparsity. \textbf{Left}: MNIST. \textbf{Middle}: CIFAR-10. \textbf{Right}: CIFAR-100. }
\label{fig:sparsedd-cifar}
\end{figure*}

Our main contributions are summarized as follows:
\begin{itemize}
\item To the best of our knowledge, our work is the first to report the sparse double descent phenomenon (see Figure \ref{fig:sparsedd-cifar}). More specifically, we demonstrate that high model sparsities may significantly mitigate overfitting, while moderate model sparsities may lead to severer overfitting. And extreme model sparsities ($\to100\%$) tend to lose all learned information.
\item The $\ell_{2}$ learning distance of models (from initialized parameters to final parameters) may be correlated with a double descent curve and reflects generalization better than minima flatness for sparse models.
\item Contrary to the \textit{lottery ticket hypothesis} \citet{frankle2019lottery}, we find the retraining a sparse model from its original initialization may not win at all time. For example, in some cases, a randomly reinitialized pruned model could largely surpass model with the original initialization at certain sparsities. 
\end{itemize}

This paper is organized as follows. In Section \ref{sec:sparsedd}, we demonstrate empirical evidence for sparse double descent through extensive experiments. In Section \ref{sec:learningdistance}, we analyze how learning distance matters to the sparse double descent compared with minima flatness. In Section \ref{sec:discuss}, we include further empirical analysis and discuss the relationship between our results and some related work. In Section \ref{sec:conclusion}, we conclude with our main work.

\section{Sparse Double Descent}
\label{sec:sparsedd}

In this section, we conducted extensive experiments to demonstrate sparse double descent with respect to model sparsity. We also identified four phases of model sparsity in empirical results. In our experiments, we grow model sparsity gradually via network pruning, and use the
train and test accuracy to demonstrate 
how overfitting and generalization depend on model sparsity.

\subsection{Overview}
\label{subsec:overview}

Although network pruning has been widely investigated at the target of storage and computational savings \citep{han2015learning, han2016deep, liu2017slimming, Molchanov2017variational_dropout, li2017filters, louizos2018sparse_l0regularization, frankle2019lottery, liu2019rethinking}, a broad consensus on how pruning will act on generalization has not been achieved yet \citep{hoefler2021sparsity}. 

Prior studies have demonstrated that networks learn simpler patterns first and are less prone to memorize noisy labels with limited capacity \citep{arpit2017closer, li2020gradient}.
Following Occam's razor, network pruning that aims to reduce parameter counts could also be regarded as some kind of regularization on model capacity \citep{lecun1990optimal, hassibi1992second}. 
By restricting a subset of model parameters to a value of zero, pruning imposes sparsity constraints on neural networks and 
penalizes its redundant expressive power. Pruning is thus sometimes considered as a potential regularizer in recent works. \citep{Molchanov2017variational_dropout, ahmad2019how_can_so_dense, xia2021robust, zhang2021subnetwork}

Moreover, there are some other conjectures on how pruning might benefit generalization, e.g., pruning creates sparsified versions of data representation, which introduce noise and encourage flatness into neural networks \citep{han2017DSD, bartoldson2020generalization-stability}. And flatness of minima is usually correlated with good generalization \citep{hochreiter1995simplifying,hochreiter1997flat,hardt2016train,shirish2017large_batch, zhu2019anisotropic_noise,xie2020diffusion,xie2022adaptive}. 
Given the discussions above, it is intuitive to suppose that pruning can enhance model performance and prevent overfitting.

However, our experimental results reveal that moderately sparse networks sometimes generalize even worse than their dense counterpart, meaning sparsity may aggravate overfitting under some circumstances.
In this paper, we particularly evaluate overfitting and generalization in the presence of noisy labels. This setting is common in related work \citep{yang2020rethinking, padhy2021geometry_memorization, xie2021artificial, xie2021positive}, because it is helpful to understanding severe overfitting in real-world problems. As DNNs overfit easily and noisy labels exist pervasively in real-world datasets \citep{shankar2020evaluating, northcutt2021confident, curtis2021pervasive}, learning with noisy labels has been a popular topic recently. Thus, we consider learning with noisy labels as an important problem setting to evaluate overfitting and generalization.

We find that in the presence of label noise, the double descent phenomenon occurs through model sparsification. Namely, when noisy labels are added to the training data, the sparse double descent can be observed robustly across different datasets, architectures, pruning settings and label noise types. 
The existence of sparse double descent raises doubts on conventional wisdom that pruning relieves overfitting, and offers new insights into understanding the relationship between sparsity and generalization.

\subsection{Experimental Setup}
\label{subsec:experi-set-up}

\begin{figure*}[ht]
\center
\includegraphics[width=0.32\linewidth]{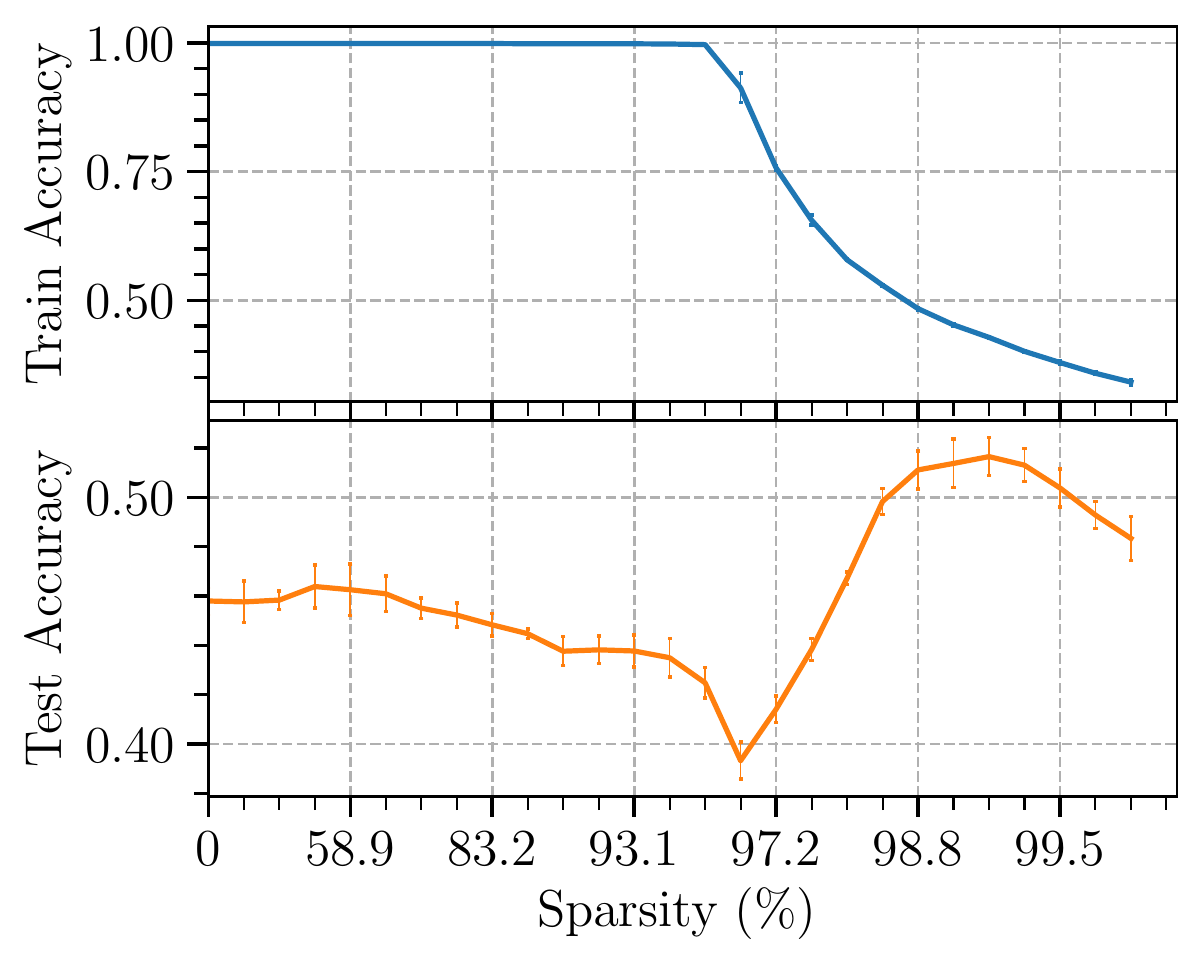}
\includegraphics[width=0.32\linewidth]{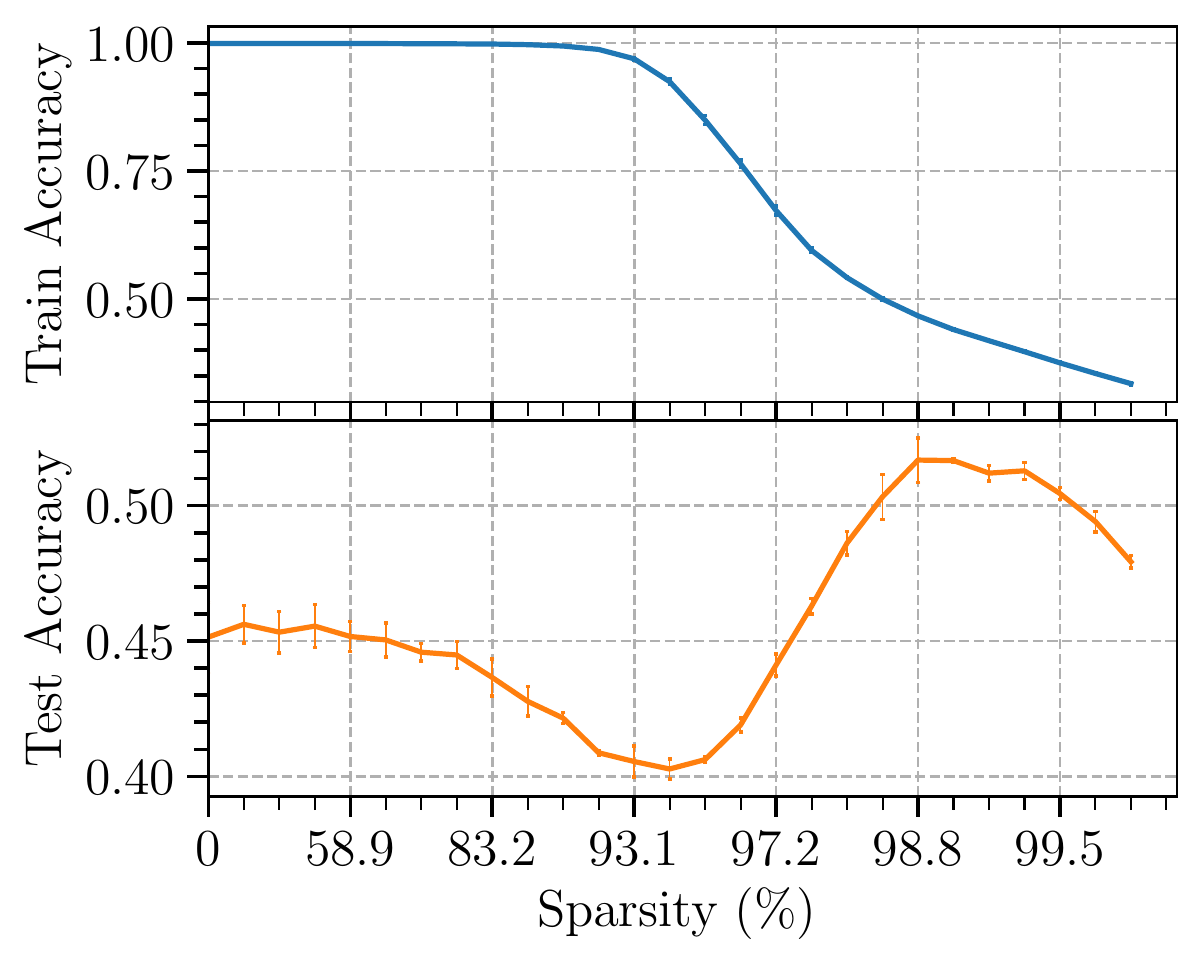}
\includegraphics[width=0.33\linewidth]{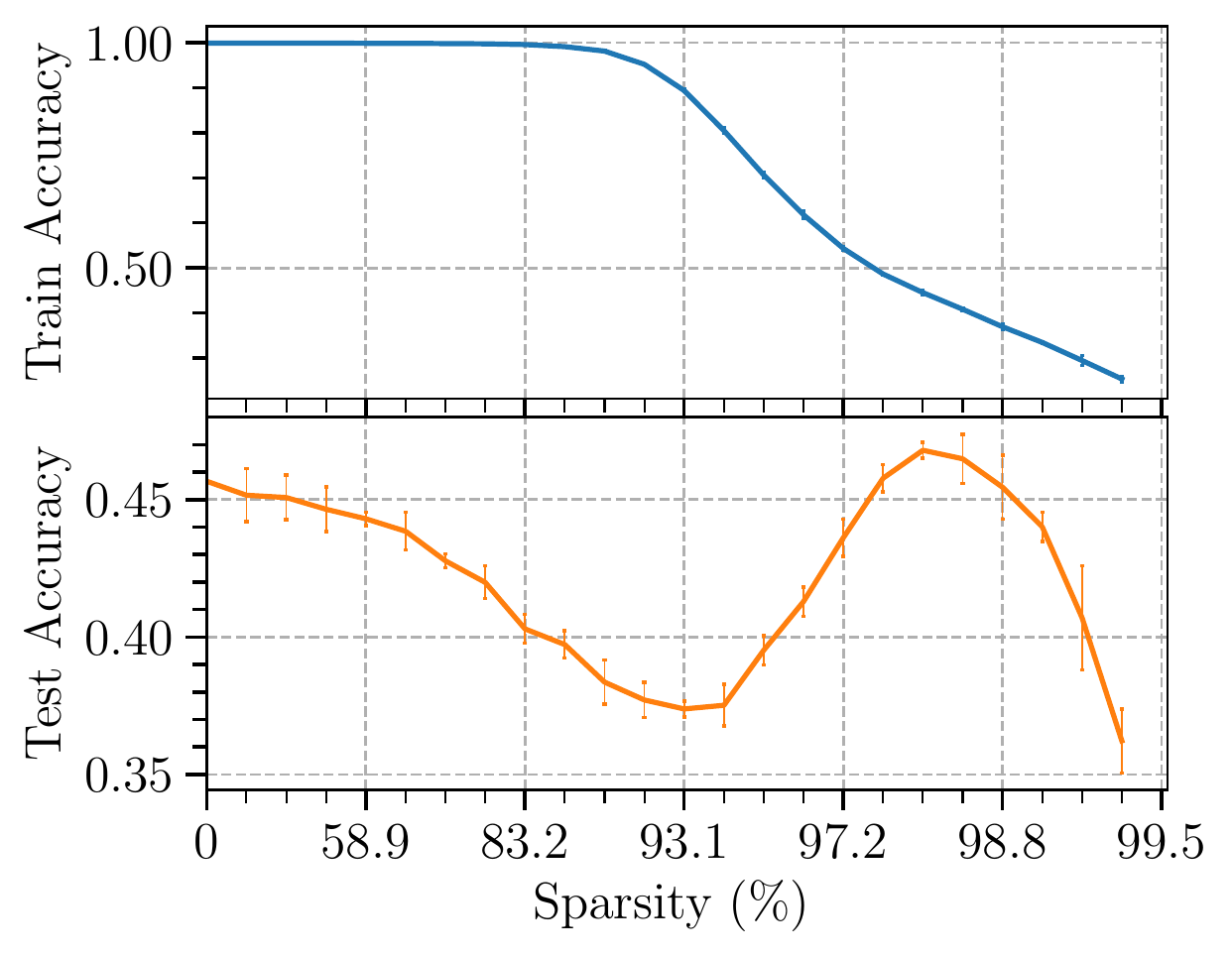}
\caption{Sparse Double Descent of ResNet-18 on CIFAR-100 with $40\%$ symmetric label noise, pruned using different strategies. We plot the train and test accuracy against sparsity. \textbf{Left}: Magnitude-based pruning. \textbf{Middle}: Gradient-based pruning. \textbf{Right}: Random pruning. }
\label{fig:sparsedd-pruning-strategy-cifar100}
\end{figure*}

\begin{figure*}[ht]
\center
\includegraphics[width=0.32\linewidth]{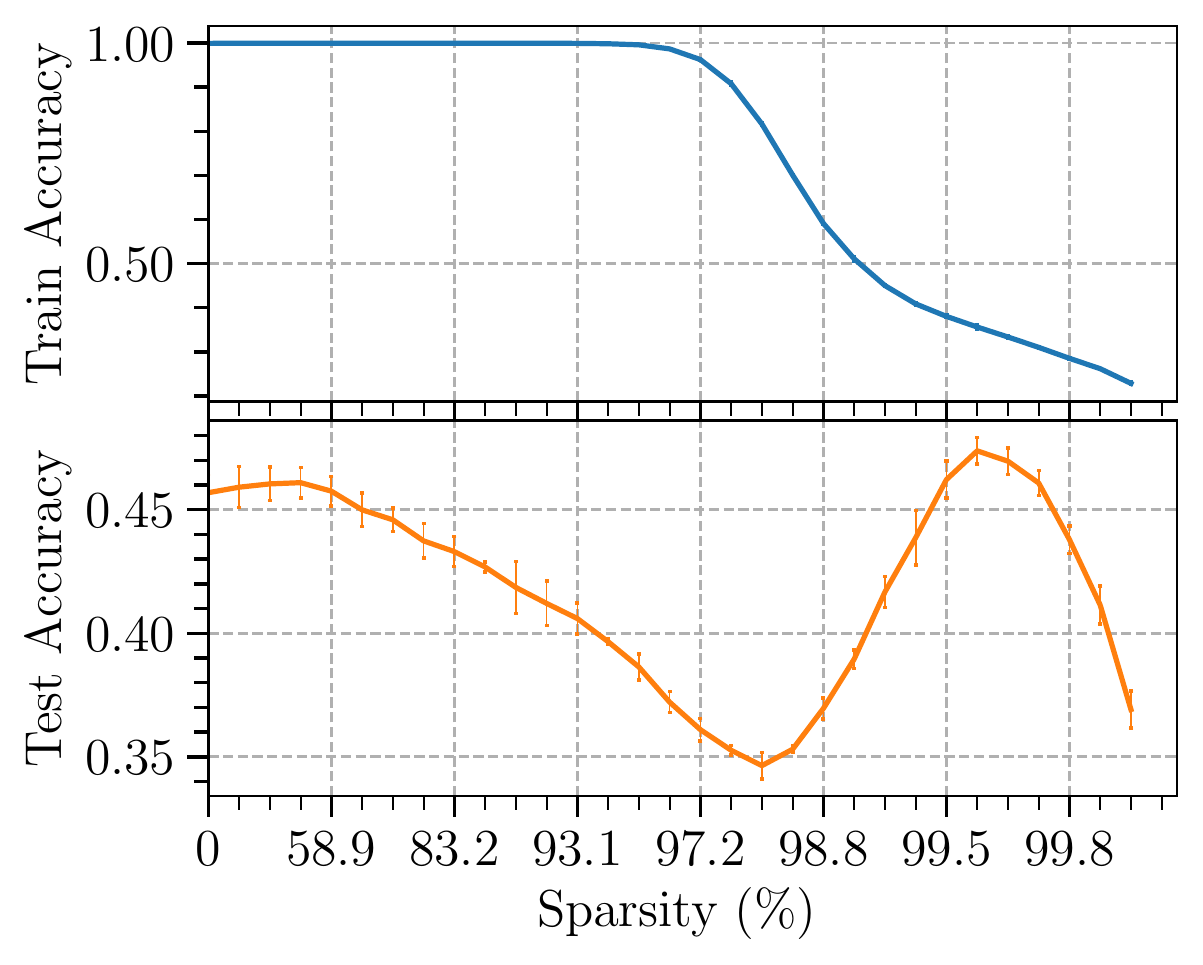}
\includegraphics[width=0.32\linewidth]{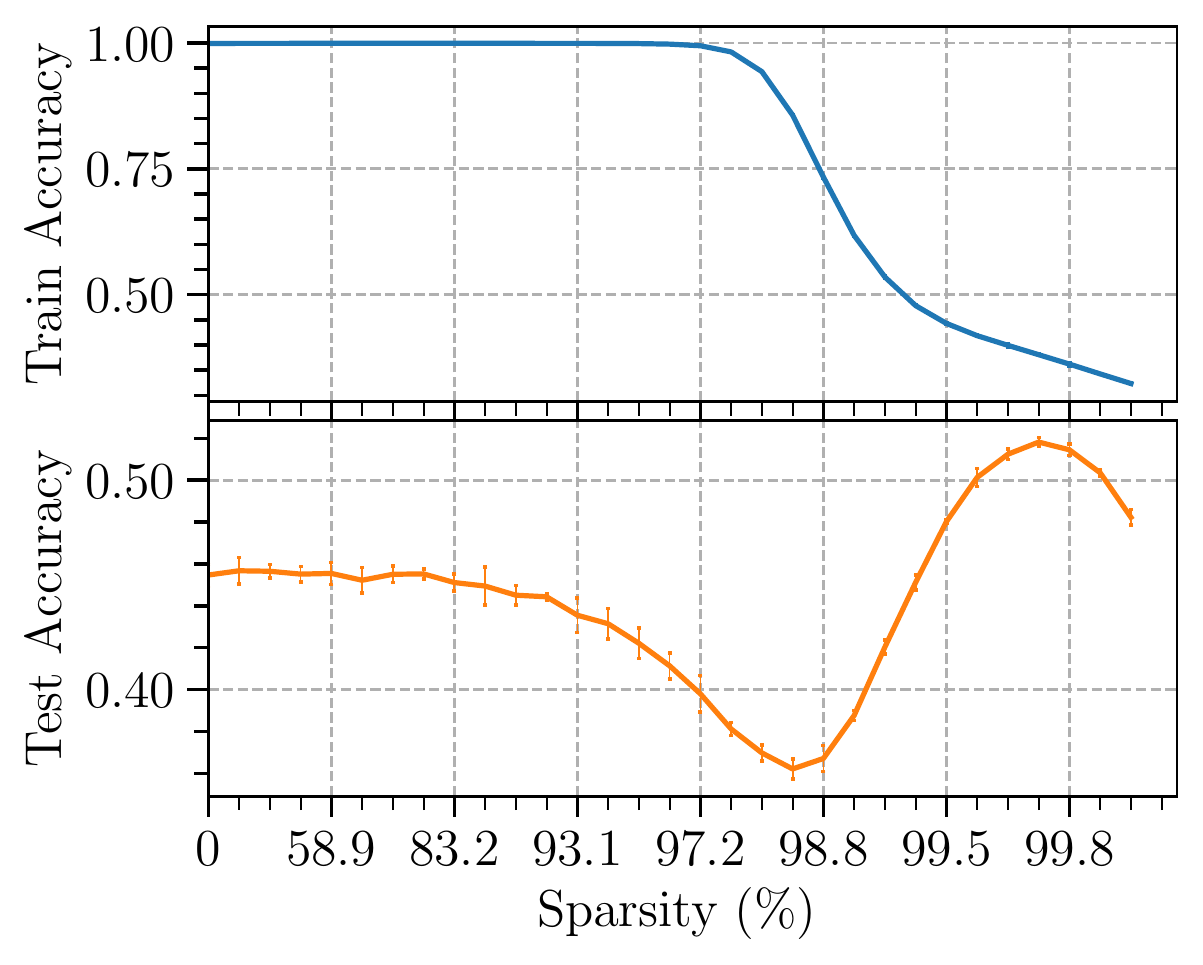}
\includegraphics[width=0.32\linewidth]{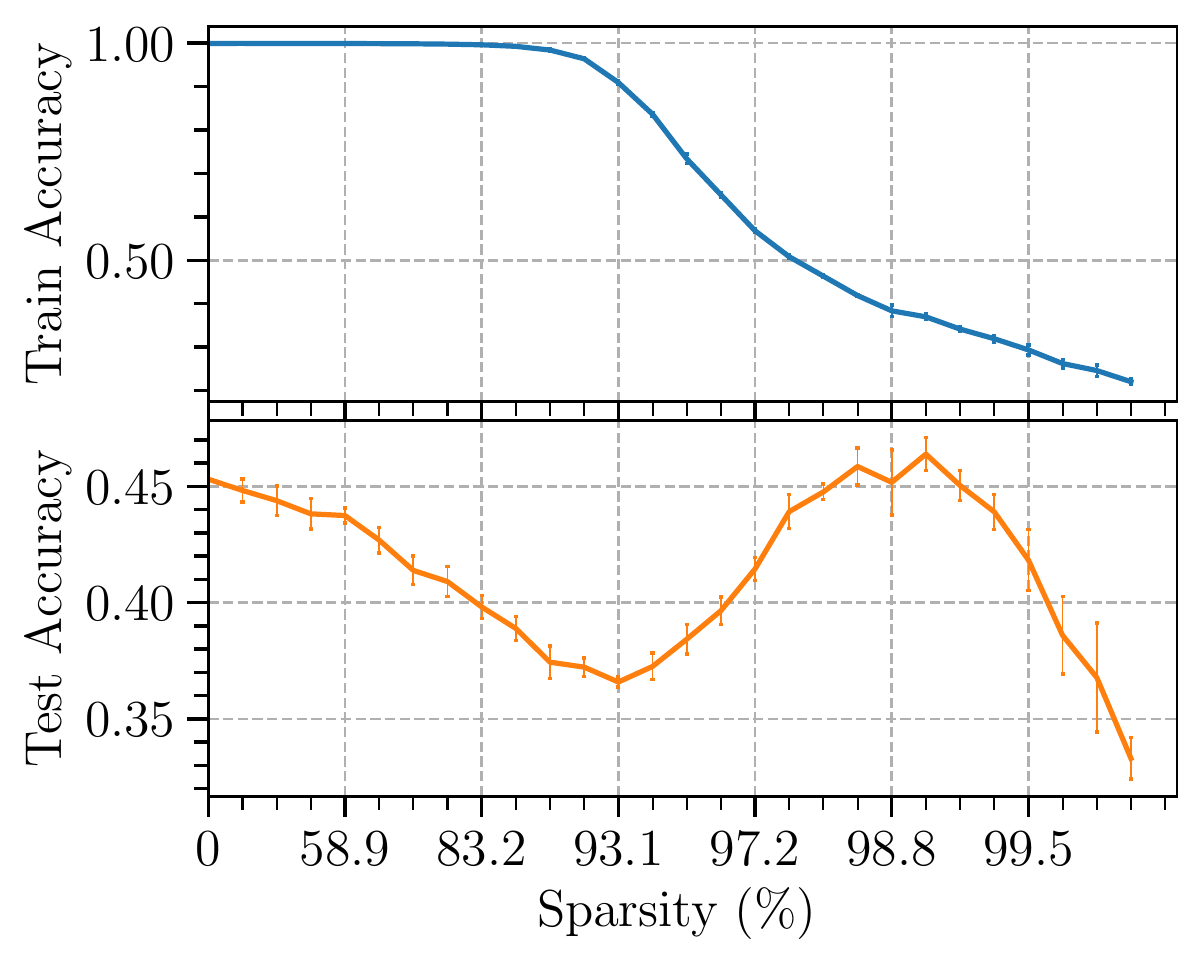}
\caption{Sparse Double Descent of ResNet-18 on CIFAR-100 with $40\%$ symmetric label noise, and different retraining methods.We plot the train and test accuracy against sparsity. \textbf{Left}: Finetuning. \textbf{Middle}: Learning rate rewinding. \textbf{Right}: Scratch retraining. }
\label{fig:sparsedd-retraining-cifar100}
\end{figure*}

We describe the main experimental setup used throughout this paper.
Particularly, we also vary several experimental choices, e.g., models and datasets, pruning strategies, retraining methods and label noise settings, to verify the generalizability of the sparse double descent phenomenon. Preliminaries, experimental details as well as more experimental results are given in the Appendix.

\textbf{Models and datasets.} We train a fully-connected LeNet-300-100 \citep{lecun1998gradient} on MNIST \citep{lecun1998gradient}, a ResNet-18 \citep{he2016deep} on CIFAR-10 or CIFAR-100 \citep{cifar}.
We use the SGD optimizer with momentum 0.9 and adopt commonly used hyperparameters for training and pruning. We repeat experiments five (on MNIST) or three (on other datasets) times with different seeds and plot the mean and standard deviation. We also test a VGG-16 \citep{simonyan2015vgg} on CIFAR datasets (see Figure \ref{fig:sparsedd-cifar10-pruning-strategy-0.2-vgg} and \ref{fig:sparsedd-cifar100-pruning-strategy-0.4-vgg}). In the context of larger model and dataset, we train a ResNet-101 on Tiny ImageNet dataset\footnote{This dataset is from
the Tiny ImageNet Challenge:
\href{https://tiny-imagenet.herokuapp.com/}{https://tiny-imagenet.herokuapp.com/}} (see Figure \ref{fig:sparsedd-tinyimagenet}), which is a reduced version of ImageNet \citep{deng2009imagenet}.

\textbf{Network pruning.} Pruning is an effective technique to enhance the efficiency of deep networks with limited computational budget, by removing dispensable weights, filters or other structures from neural networks \citep{lecun1990optimal, han2015learning, li2017filters, liu2017slimming}. In this work, we do not chase state-of-the-art accuracy nor the computing resource efficiency; thus we simply remove each weights individually (i.e., unstructured pruning), for this method can adjust easily to different tasks and architectures.

Given a dataset $\mathcal{D} = \{(\rvx_i,\rvy_i)\}^n_{i=1}$, 
we define a neural network classifier function as $f(\rvw; \mathcal{D})$, where $\rvw \in \sR^d$ is the set of weights, and $d$ is the total number of weights. As weight pruning removes parameters individually, we introduce a binary masks $\rvm \in \{0,1\}^d$ as auxiliary to represent the remained weights $\rvw \odot \rvm$. The classifier function under sparsity constraints is denoted as $f(\rvw \odot \rvm; \mathcal{D})$, where $\odot$ is the element-wise product. The model sparsity of a pruned network is defined as $1-\sum_{i=1}^d \rvm_i/ d$.

\textbf{Pruning strategies.}
We use three existing pruning heuristics summed up by \citet{blalock2020state}, i.e., magnitude-based pruning, gradient-based pruning and random pruning. Magnitude-based pruning is one of the most commonly used baselines, and has been shown to achieve comparable performance to many complex techniques \citep{han2015learning, han2016deep, gale2019state_sparsity}. Gradient-based pruning preserves training dynamics and provides possibility to prune a network early in training \citep{lee2019snip, lee2020signal}. And random pruning is often regraded as a naive method, setting the performance benchmark that any elaborately designed method should surpass \citep{frankle2021missing_mark}.

We prune weights in a network globally by comparing them across layers with the mentioned heuristics.
For mainline experiments, we utilize the magnitude-based pruning if not otherwise noted. Weights with the lowest absolute magnitudes in a network will be removed after one pruning operation.

\textbf{Retraining.} 
A common approach to recover network performance after pruning is retraining, which means training the pruned networks for some extra epochs.
Along side the sparse structures induced by different pruning strategies, re-training methods also affect network performance by determining which point on the optimization landscape to start training from, i.e., near initialization or close to the final weights; or which learning rate schedule to utilize.
We mainly utilized the technique \textit{lottery ticket rewinding} (LTR) proposed in \textit{lottery ticket hypothesis} \citep{frankle2019lottery} to retrain a network from near initialization.

In addition to LTR, we also consider other existing retraining techniques like finetuning, learning rate rewinding, and scratch retraining. Interestingly, the sparse double descent phenomenon exists consistently despite different retraining settings (see Figure \ref{fig:sparsedd-retraining-cifar100}). We mainly demonstrate the results of LTR unless otherwise specified, and leave more experimental results in the Appendix.
In all experiments, networks are pruned, retrained and pruned iteratively, and 20\% of weights will be removed during each pruning.

\begin{figure*}[t]
\center
\includegraphics[width=0.32\linewidth]{img/cifar10/0.2_acc_magnitude.pdf}
\includegraphics[width=0.32\linewidth]{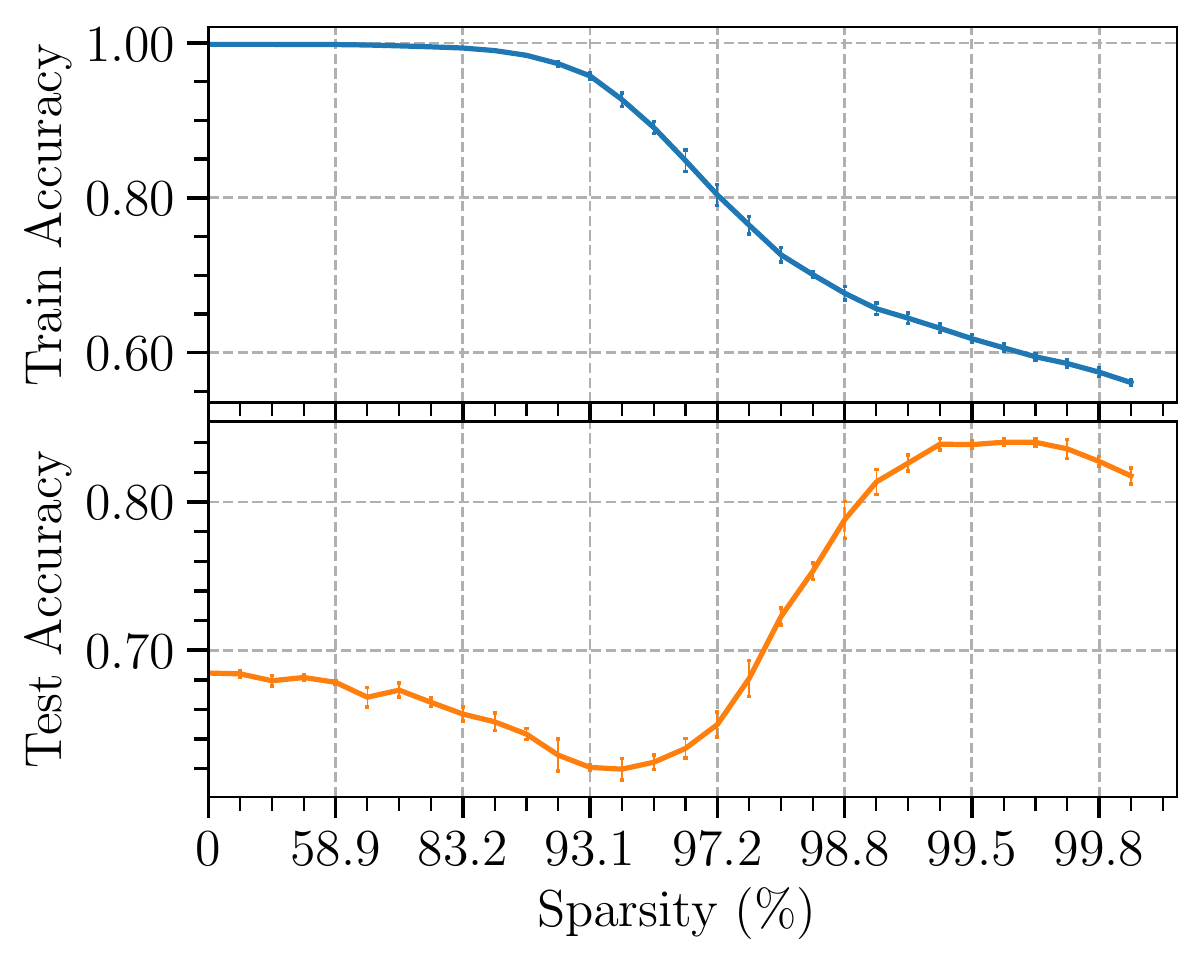}
\includegraphics[width=0.32\linewidth]{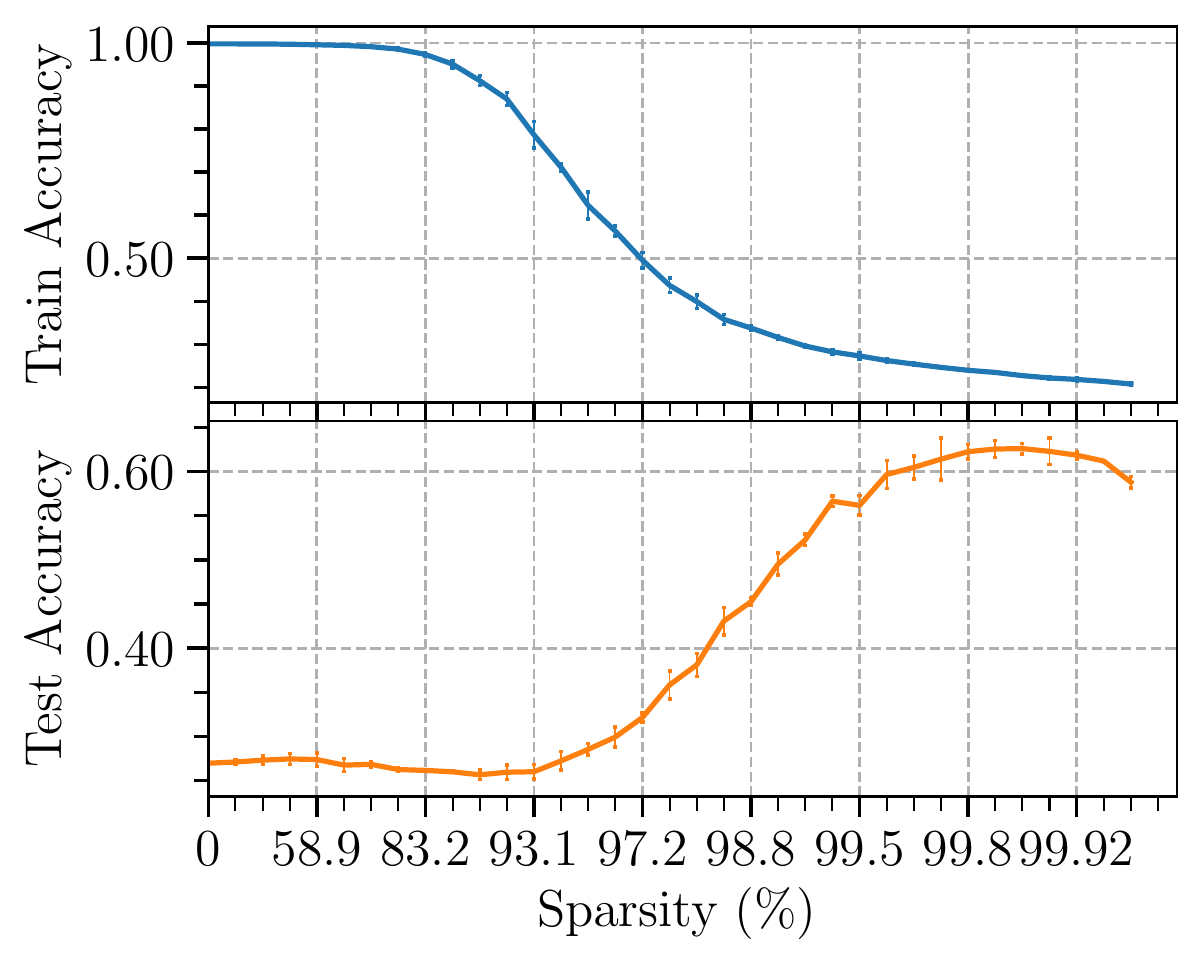}
\caption{Sparse Double Descent of ResNet-18 on CIFAR-10 with various symmetric label noise rate. We evaluate how the train and test performance of networks depend on model sparsity. \textbf{Left}: $\epsilon=20\%$. \textbf{Middle}: $\epsilon=40\%$. \textbf{Right}: $\epsilon=80\%$. }
\label{fig:sparsedd-noise-rate-cifar10}
\end{figure*}

\begin{figure}[t]
\begin{center}
    \includegraphics[width=0.7\linewidth]{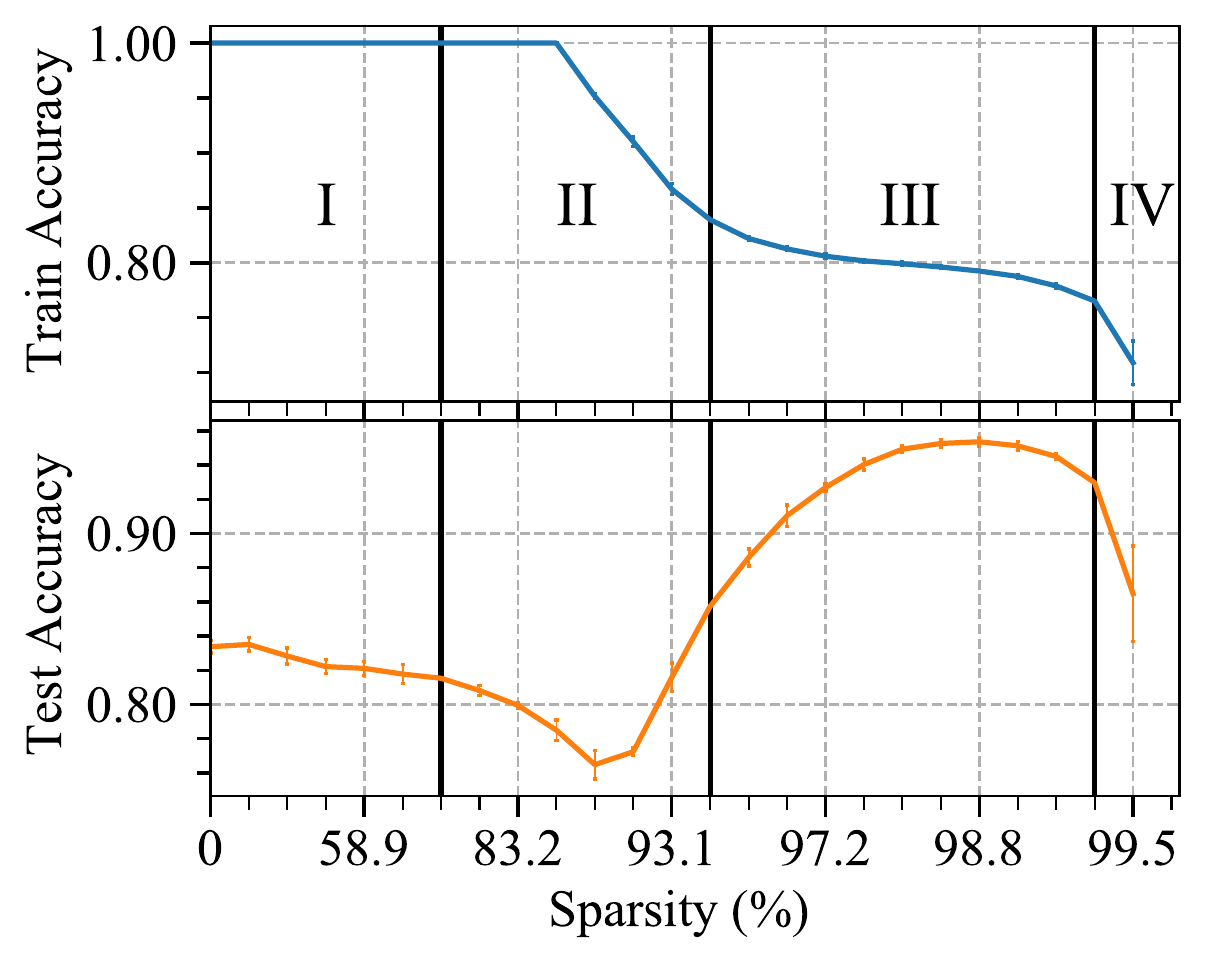}
\end{center}
\caption{Illustration of four phases using the result of LeNet-300-100 on MNIST
with $20\%$ symmetric label noise. I: Light Phase. II: Critical Phase. III: Sweet Phase. IV: Collapsed Phase.
}
\label{fig:4phases}
\end{figure}

\textbf{Label noise settings.} 
To verify the generalizability of our observations, we run experiments on three types of synthetic label noise, i.e., symmetric noise, asymmetric noise and pairflip noise, which are widely used in prior works \citep{ma2018dimensionality, li2020dividemix, xia2021robust}. The noise rate $\epsilon$ is set to 20\%, 40\% and 80\%. Most results and discussions are based on symmetric label noise except where otherwise provided.

\subsection{Effect of Model Sparsity on Overfitting}
\label{subsec:effect-of-sparsity}

\begin{figure*}[htb]
\begin{center}
    \includegraphics[width=1\textwidth]{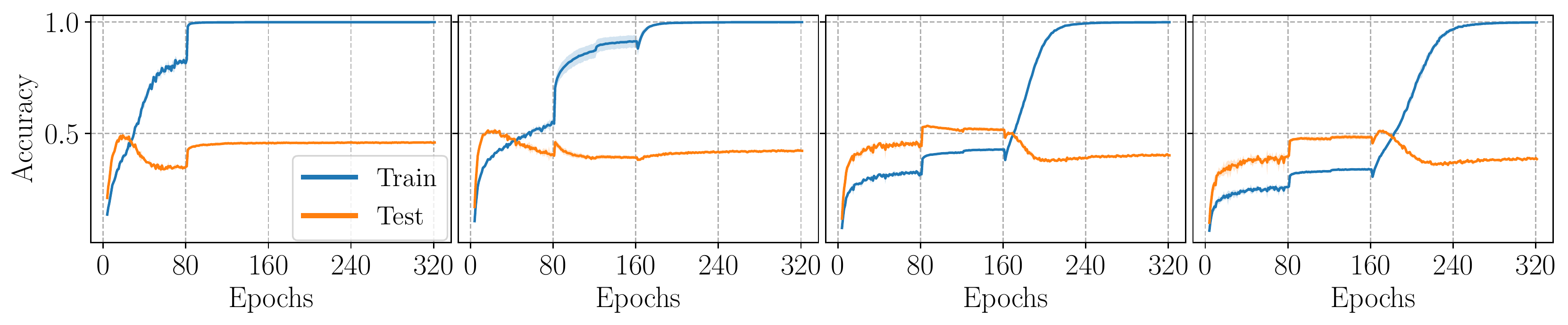}
\end{center}
\caption{Accuracy as a function of epochs during sparse and re-dense training process. Results are from ResNet-18 networks trained on CIFAR-100 with $\epsilon=40\%$. Pruned weights are recovered and trained from zero at epoch 160. The re-dense networks are optimized using learning rate of 0.001 for another 160 epoch. From left to the right, we display results of dense model, and models in Critical Phase, Sweet Phase and Collapsed Phase respectively.}
\label{fig:re-dense-training-cifar100}
\end{figure*}

\textbf{Empirical evidence of sparse double descent.} From Figure \ref{fig:sparsedd-cifar}, we empirically observe
the sparse double descent for both ResNet-18 on CIFAR and LeNet-300-100 on MNIST in the presence of noisy labels. This suggests that sparse double descent may generally exist in various models and datasets, when overfitting is severe.
Figures in section \ref{appendix:sparsedd} summarize the double descent behavior of sparse neural networks across different datasets, architectures and settings of pruning and label noise. 
In most cases, increasing sparsity of networks results in a first decrease then increase, and then decrease again in the test accuracy. Note that this increase is not caused by incomplete training of models, for models across all sparsities are able to converge to steady states (Figure \ref{fig:epoch-wies-train-test-accuracy-mnist}, \ref{fig:epoch-wies-train-test-accuracy-cifar}). Particularly, the right plot in Figure \ref{fig:sparsedd-pruning-strategy-cifar100} shows that even random pruning may cause sparse double descent.

\textbf{Four phases of model sparsity.} Given the empirical observations, we identified four phases of model sparsity (see Figure \ref{fig:4phases}). 
Different from previous work \citep{frankle2020linear, frankle2021missing_mark}, we define these four phases in terms of training accuracy and test accuracy.

First, \textit{Light Phase} indicates low sparsities where the network is so overparameterized that pruned network can still reach similar accuracy to the dense model.
Second, \textit{Critical Phase} means an interval around the \textit{interpolation threshold} \citep{nakkiran2020deepdoubledescent} where training accuracy is going to drop and test accuracy is about to first decrease then increase as sparsity grows. 
Third, \textit{Sweet Phase} are high sparsities where test accuracy can significantly be boosted.
Fourth, \textit{Collapsed Phase} are those beyond, where both training accuracy and test accuracy drops significantly.

Here we illustrate how the sparse double descent is affected by data itself. 
As shown in Figure \ref{fig:sparsedd-noise-rate-cifar10}, increasing the fraction of label noise shifts the Critical Phase towards models with larger capacity, which is to say, lower sparsities. 
On the other hand, in order to combat the side effects brought by the existence of heavier labels noise, more parameters in the network need to be pruned. Thus, Sweet Phase generally moves to higher sparsities as noisy labels increase.

The most interesting discovery is the existence of Critical Phase, where overfitting becomes the most severe. To the best of our knowledge, we are the first to report the novel Critical Phase, where pruning, no matter reasonable heuristic-based pruning or random pruning, may decrease test accuracy while maintaining training accuracy, thus hurting generalization. 

\textbf{Discussion of related work.} The observation that the Critical Phase may exist in random pruning contradicts with the conventional wisdom that random parameter perturbations can relieve overfitting \citep{harutyunyan2020improving,xie2021artificial}. On the other hand, a number of papers on sparsity \citep{Molchanov2017variational_dropout, lee2019snip, hooker2019compressed_forget, goel2021robustness} only revealed that sparsity impairs memorization, but failed to uncover the unexpected loss of generalization performance as well. \citet{liebenwein2021lost} studied the performance of pruned networks under distributional shifts, but did not reveal the nontrivial changing curve of test accuracy against sparsity.

A recent work \citet{chang2021provable_benefits_compression} studied double descent of pruned models with respect to the number of original model parameters. However, \citet{chang2021provable_benefits_compression} still focused on how model performance depends on overparameterization of original models rather than pruned models, while our work focused on the influence of sparsity on overfitting and generalization.

In summary, previous studies did not expect the existence of Critical Phase with respect to model sparsity, and thus, failed to discover sparse double descent.

\begin{figure}[t]
\begin{center}
    \includegraphics[width=0.75\linewidth]{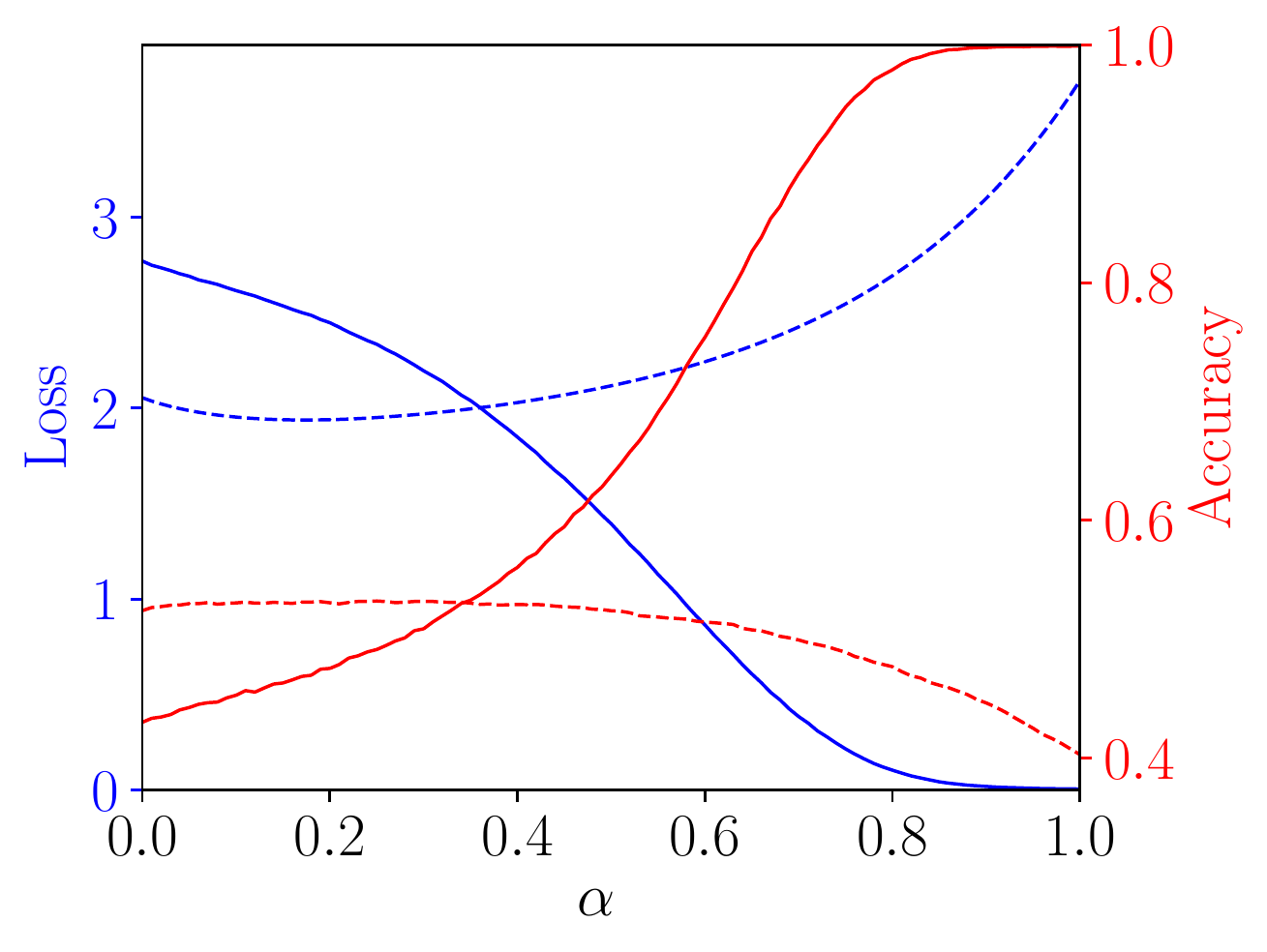}
\end{center}
\vspace{-0.3cm}
\caption{Linear interpolation plots for ResNet-18 on CIFAR-100 
in Sweet Phase 
with $\epsilon=40\%$. Note that $\alpha=0$ corresponds to sparse solutions, while $\alpha=1$ corresponds to the re-dense solutions (see methodology in Appendix \ref{apx:interpolation}). The blue lines are loss curves and the red lines are accuracy curves; solid lines are about training data and dashed lines are about testing data. 
}
\label{fig:linear-interpolation-cifar10}
\end{figure}

\section{Why Sparse Double Descent Occurs}
\label{sec:learningdistance}

In this section, we try to understand and analyze why sparse double descent occurs.
We believe that the mechanism behind sparse double descent can help to understand the conventional deep double descent with respect to model overparameterization from a new perspective.

\subsection{Minima Flatness Cannot Explain Sparse Double Descent} 
Flatness is a widely used measure to capture the generalization behavior of neural networks \citep{jiang2020fantastic}. Flatter minima are usually believed to imply robustness to perturbations in model parameters \citep{xie2021artificial,foret2020sharpness}, low complexity \citep{blier2018description,xie2022power}, and better generalization \citep{shirish2017large_batch, wu2017towards, chaudhari2019entropy}. 
Previous works hypothesized that pruning could encourage the optimizer to move towards flatter minima that benefit generalization \citep{han2017DSD, bartoldson2020generalization-stability}. May such minima flatness hypothesis explain sparse double descent? 

\begin{figure}[t]
\center
\includegraphics[width=0.85\linewidth]{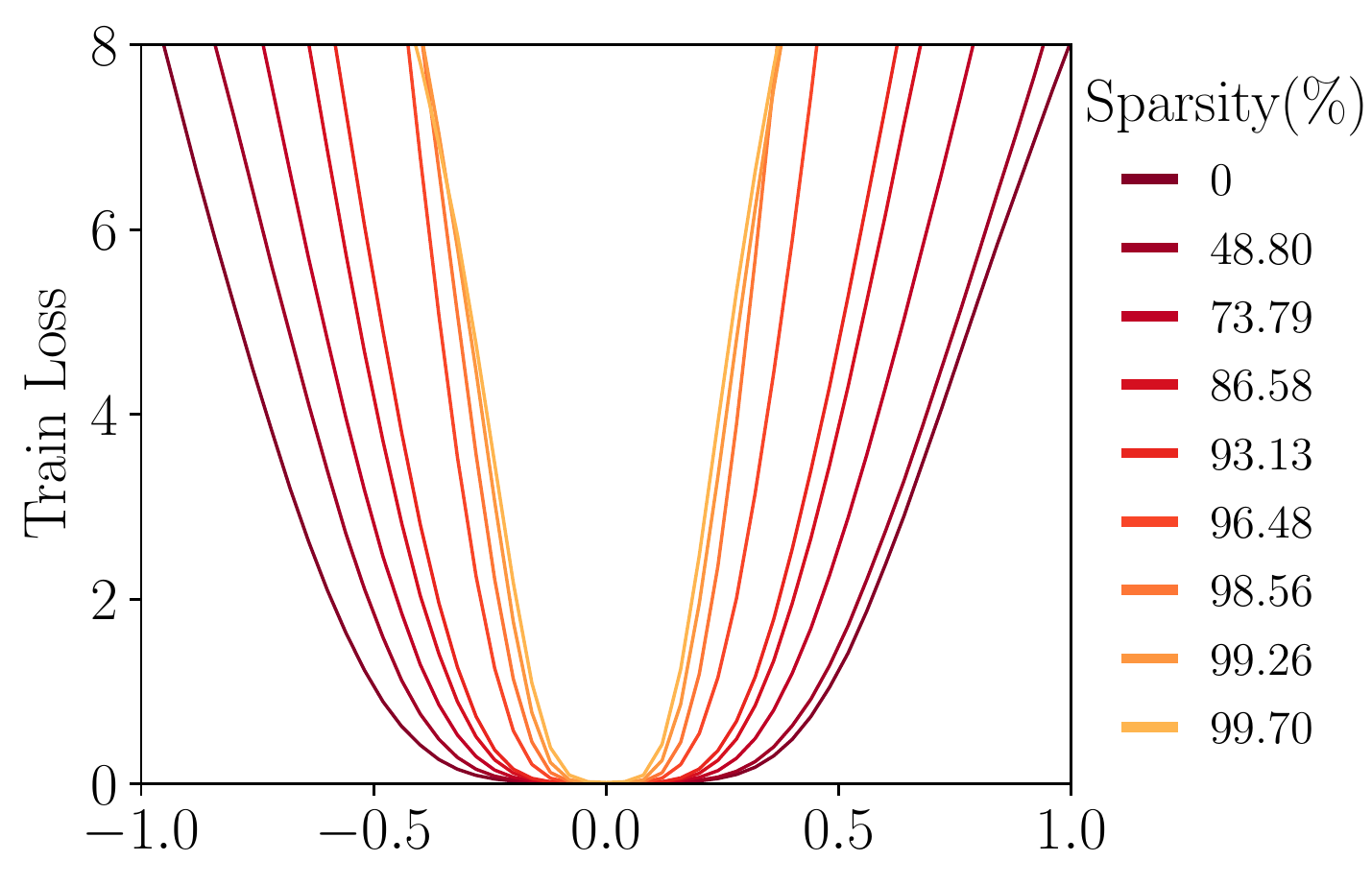}
\caption{The 1-D loss visualization of minima found by re-dense training using filter normalization \citep{li2018landscape}. Result is of ResNet-18 on CIFAR-100 with $\epsilon=40\%$. The sparsity of particular model is measured before the re-dense step. The sparser the original pruned network is, the sharper minima it might converge to after re-dense training. 
}
\label{fig:1d-loss-cifar}
\end{figure}

We note that by removing a quantity of parameters, pruning restricts the movement of optimizer in low-dimension space, and drastically changes the loss landscape.
Thus, measuring the minima flatness of pruned networks and comparing them across different sparsities possibly 
leads to unfair comparison. We are therefore motivated to seek indirect evidence that can estimate flatness with the same dimensions.

We apply the re-dense training approach to connect a sparse network with a dense one, which is introduced in Appendix \ref{apx:re-dense}: after training a sparse network to reach convergence, we recover its pruned weights, and further retrain the whole network for certain epochs. If the optimizer reaches a flat basin of loss landscape during sparse training, we may suspect that a small learning rate in the re-dense training stage will continually attract optimizer around this basin. Then the final re-dense solutions will have comparable generalization performance to the sparse ones.

To demonstrate our points, we will focus on four model sparsities in the following analysis: 1) the zero sparsity (dense model), 2) the sparsity where the last test accuracy degrades the most (in Critical Phase), 3) the sparsity where the test accuracy reaches a peak (in Sweet Phase), and 4) the sparsity where both the train accuracy and test accuracy suffer (in Collapsed Phase).

Figure \ref{fig:re-dense-training-cifar100} surprisingly shows that sparse solutions are not stable in dense subspace. Once the sparsity constraints are removed, the model will escape from pruned solutions and overfit severely. The phenomenon that neural networks escape from highly sparse solutions during re-dense training, even with a small learning rate, raises doubts on the conjecture regarding minima flatness. 

With linear interpolation of loss function, we find a monotonically decreasing path from the high-loss point to low-loss point in Figure \ref{fig:linear-interpolation-cifar10}. The existence of such path also demonstrates that these highly sparse solutions may no longer be minimizers in the loss landscape of dense model, thus allowing for the escape phenomenon during re-dense training process. Given this observation, we conjecture that sparsity may restrict the movement of optimizers, and trap them near initialization, which would be normally skipped when training dense networks.

Furthermore, the final solutions of re-dense training do not generalize well, and even have higher sharpness than the original dense models (see the 1-D visualization of re-dense solutions at various sparsities in Figure \ref{fig:1d-loss-cifar}).
The empirical results refute that highly sparse solutions stick around flat basins of minimizers. 

\begin{figure}[t]
\begin{center}
\includegraphics[width=0.85\linewidth]{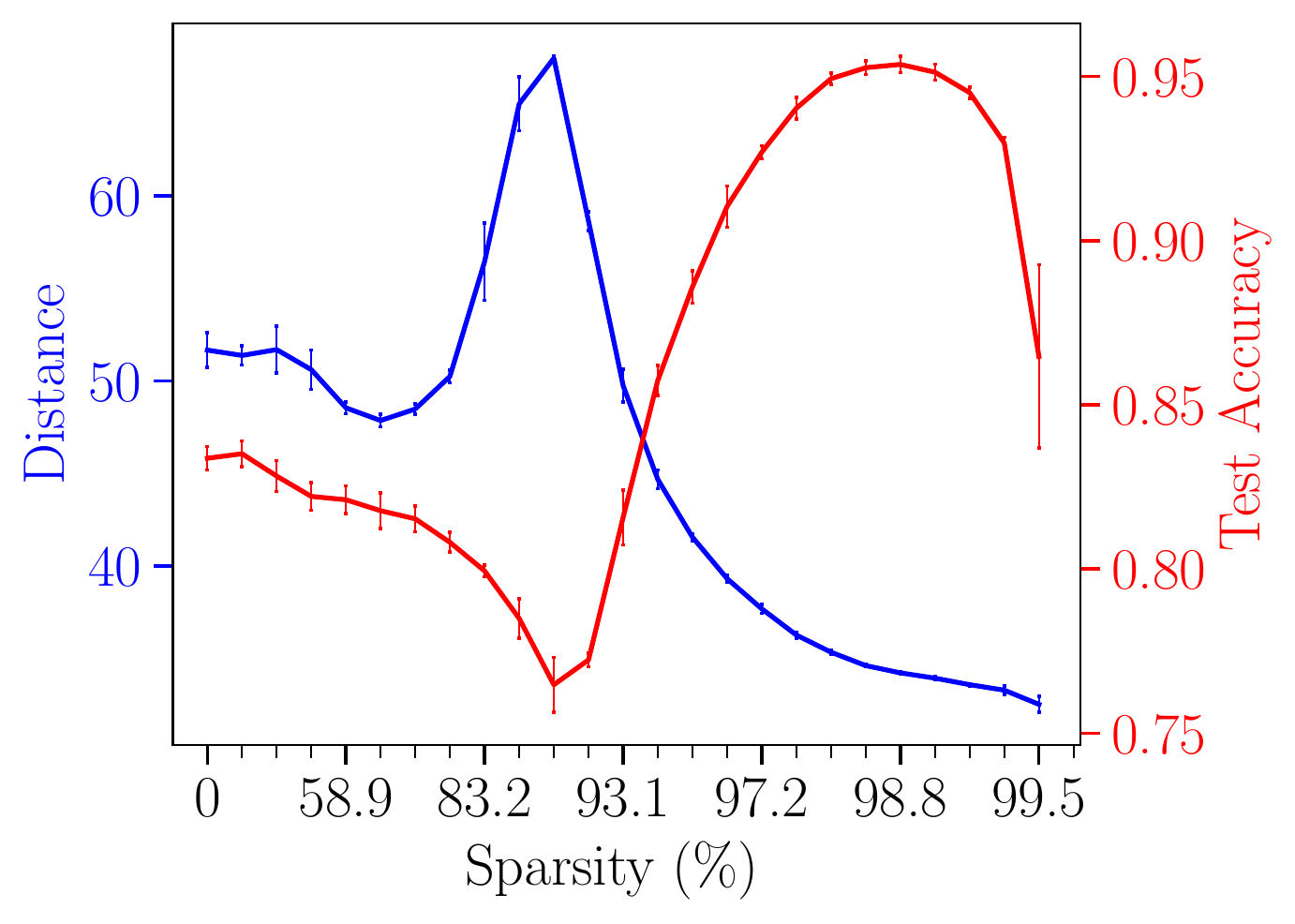}
\caption{The curve of learning distance for LeNet-300-100 on MNIST with $\epsilon=20\%$ may explain the double descent of test accuracy. As model sparsity increases, learning distance coincides the changes of test accuracy. The blue lines refer to $\ell_2$ learning distance and the red lines are test accuracy.
}
\label{fig:distance-mnist}
\end{center}
\end{figure}

\subsection{The Learning Distance Hypothesis for Sparse Double Descent}

Learning distance has been observed to be very related to generalization in deep learning \citep{neyshabur2018role, nagarajan2019generalization}. Several studies suggested that neural networks need to move far from initialization to have large model capacity and overfit noisy labels, while avoid overfitting when staying close \citep{li2020gradient, wei2020simple, padhy2021geometry_memorization}. And prior work \citet{nagarajan2019generalization} provided empirical and theoretical evidences that model capacity could be restricted by the $\ell_{2}$ distance from initialization.

Given the failure of minima flatness hypothesis, we propose a novel learning distance hypothesis for sparse double descent that sparsity affects the learning distance, while learning distance correlates with model capacity and generalization ability.

For illustration, we denote weights at initialization by $\rvw_{init} \in \sR^d$, and weights trained after the $i_{th}$ pruning by $\rvw^i_{learned} \in \sR^d$, where the pruned weights are regarded as zero weights.
We define the learning distance $D$ as the $\ell_{2}$ distance from the original initialization to learned parameters, namely, $D(\rvw_{init}, \rvw^i_{learned})= ||\rvw_{init} - \rvw^i_{learned}||_2$. 
Note that, the weights at initialization refer to the initialized parameters of the original dense model, instead of the sparse initialization of the pruned model.

\begin{figure}[t]
\begin{center}
\includegraphics[width=0.8\linewidth]{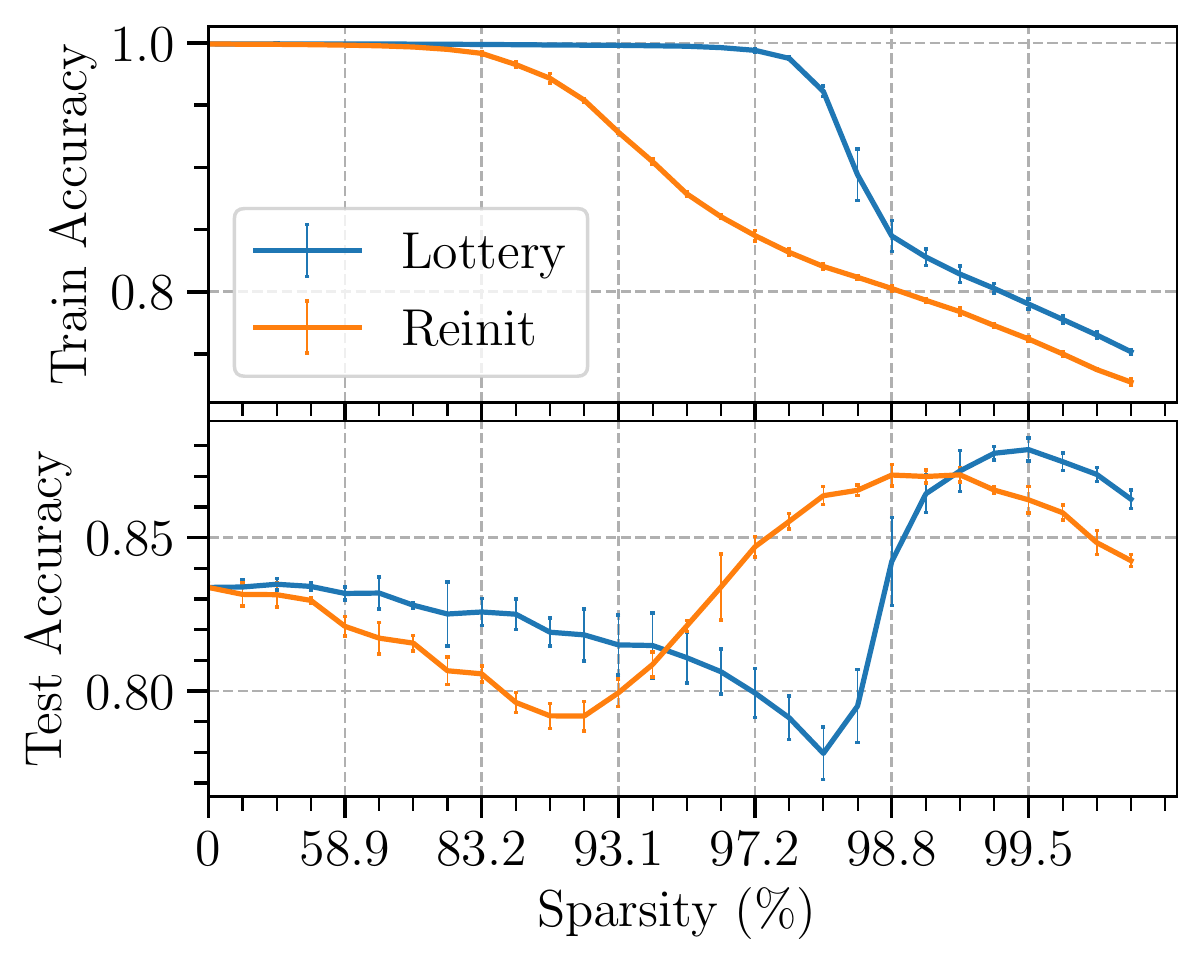} \hspace{1.5cm}
\end{center}
\caption{Performance of ResNet-18 on CIFAR-10 with $\epsilon=20\%$ when retrained from either the original initialization (lottery tickets), or a random reinitialization. Reinitialization results sometimes surpass lottery results. 
}
\label{fig:reinitlization-lottery}
\end{figure}

In order to demonstrate the relationship between the learning distance and model performance, we plot both learning distance and test accuracy against sparsity in Figure \ref{fig:distance-mnist}. We surprisingly find that the changing curve of learning distance correlates with test accuracy.
Our experimental results suggest that staying closer to initialization coincides with better robustness:
in Critical Phase, sparse solutions are located farther from initialization, and present an inferior performance; while at high sparsities in Sweet Phase, sparse minimizers stay closer and manifest robustness (although in Collapsed Phase, too few parameters remain trainable that learning distance becomes less informative). 
Furthermore, the turning point of test accuracy from decreasing to increasing is usually consistent with a peak in learning distance. This suggests a relatively strong correlation between learning distance and sparse double descent.

While a number of generalization measures have been studied in deep learning, previous studies rarely touched how these generalization measures including learning distance may reflect double descent in deep learning. Our finding may shed light on theoretically understanding sparse double descent from the perspective of learning distance.

\section{Empirical Analysis and Discussion}
\label{sec:discuss}

In this section, we present more empirical results about spare double descent. We discuss their meaning and some related work.

\begin{figure}[t]
\center
    \includegraphics[width=0.8\linewidth]{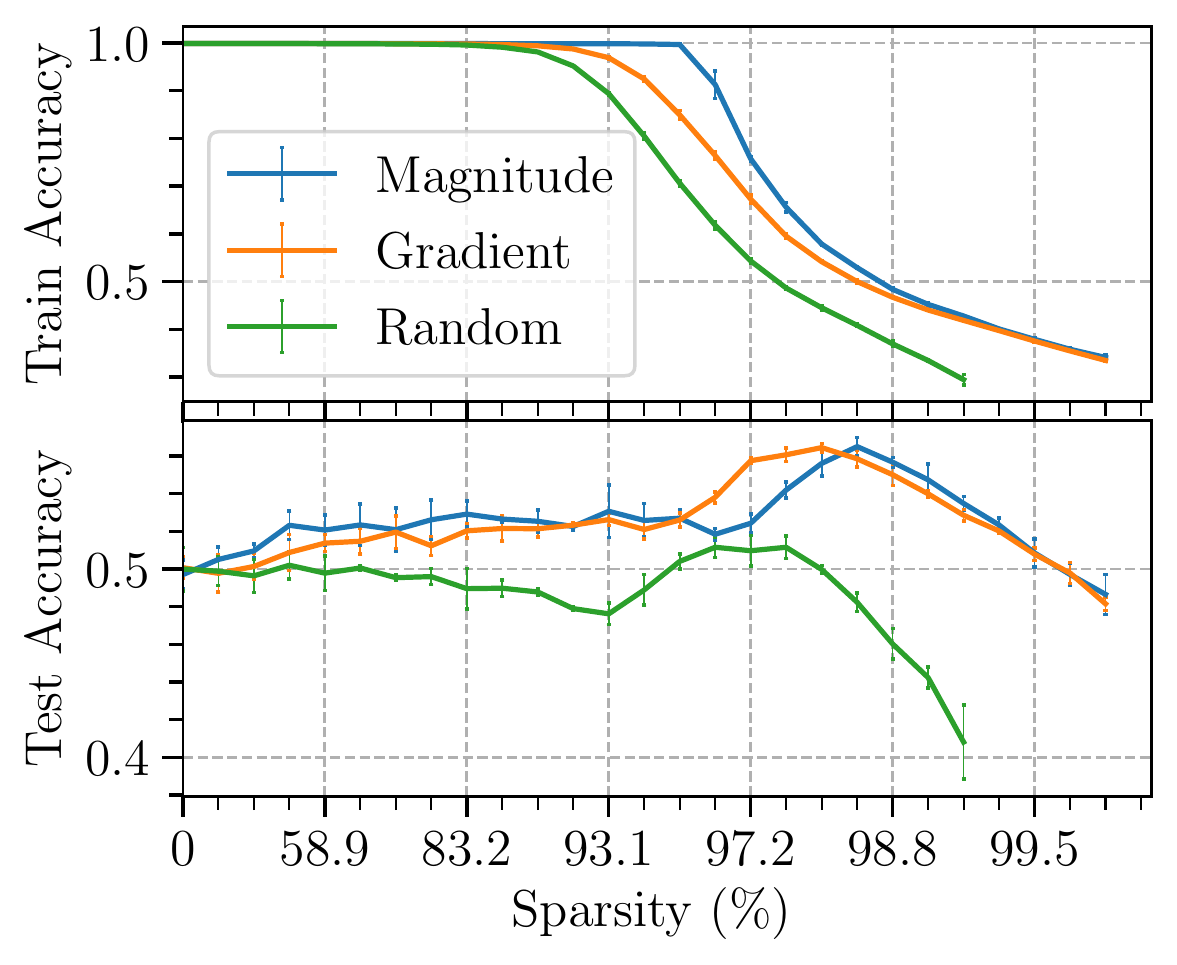}
\caption{The train accuracy and best test accuracy as functions of sparsities. Models are ResNet-18 on CIFAR-100 with $\epsilon=40\%$, pruned with three different pruning strategies.
}
\label{fig:best-test-cifar-0.2}
\end{figure}

\textbf{Lottery tickets may not win at all time.} 
The \textit{lottery ticket hypothesis} \citep{frankle2019lottery} supposes that a large network contains a subnetwork, which could be trained effectively together with the original initialization, while performs far worse if reinitialized.
However, against the well-known lottery ticket hypothesis, we identify that reinitialized sparse models could outperform lottery ticket models in some circumstances (see Figure \ref{fig:reinitlization-lottery}, \ref{fig:sparsedd-cifar10-lottery-reinit} and \ref{fig:sparsedd-cifar100-lottery-reinit}).
For the reinitialized models, the sparse double descent still holds, but with the Critical Phase and Sweet Phase both moved to left. This way, reinitialized models could beat lottery ticket models at the same sparsity but different phases. 
The results show that models with the same sparse structure but different initialization behave distinctively. Our finding suggests that a winning lottery ticket may not always win in the presence of sparse double descent.


\textbf{The best test accuracy benefits from pruning.} Although the test accuracy at the last epoch shows double descent with respect to pruning, we find that the best test accuracy across all epochs in training may not possess such well-defined trend. Instead, the best accuracy might increase in general from Light Phase till early Sweet Phase before finally drops down, 
with only a slight drop in Critical Phase (Figure \ref{fig:best-test-cifar-0.2}). So why the best test accuracy sometimes does not show a double descent trend? As deep double descent actually also depends on the number of training epochs \citep{nakkiran2020deepdoubledescent}, we conjecture that training epochs of DNNs also plays a key role in model capacity in the context of sparse double descent. 

\begin{figure}[t]
\center
    \includegraphics[width=0.8\linewidth]{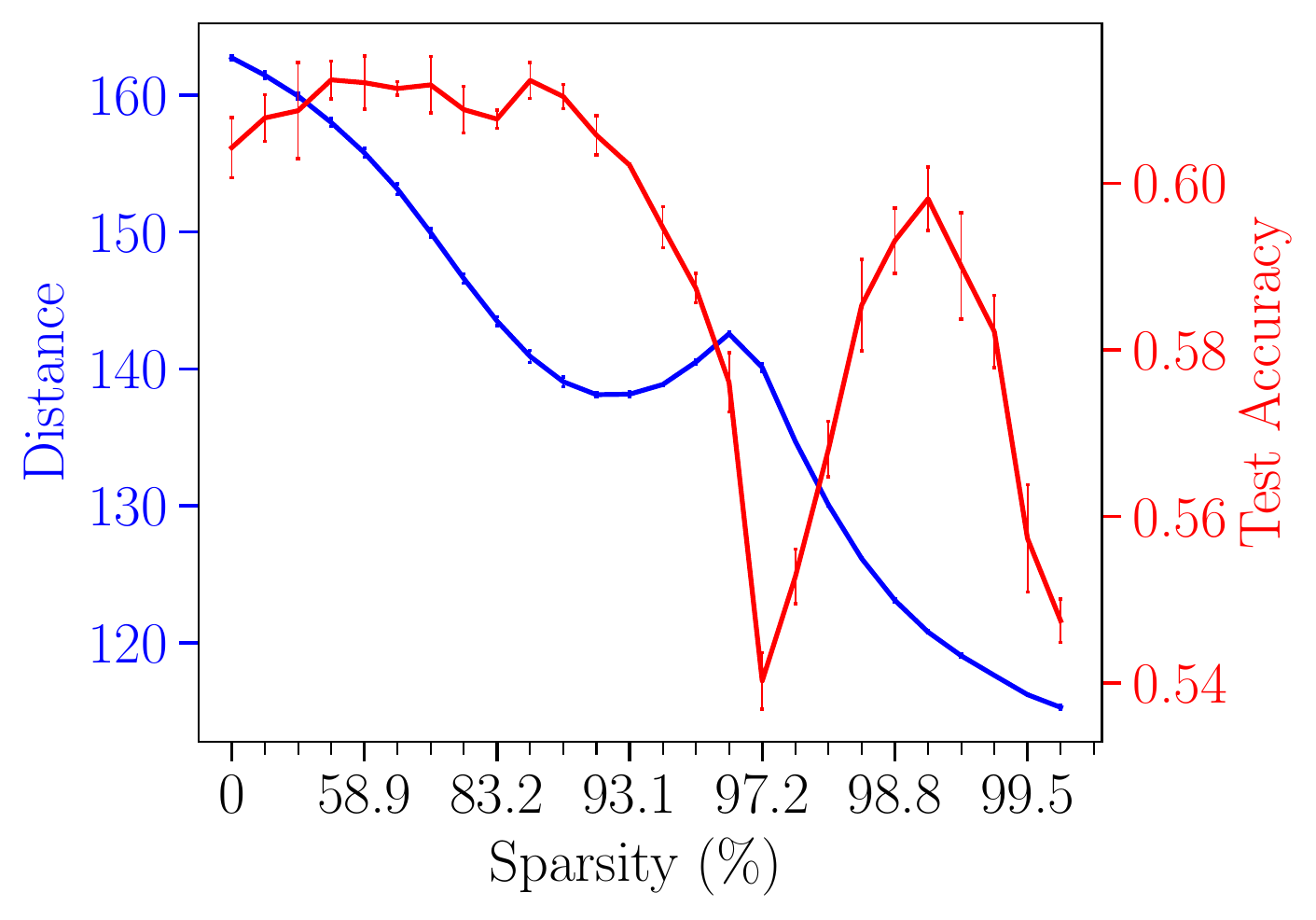}
\caption{The $\ell_2$ learning distance and test accuracy as functions of sparsity for CIFAR-100 with $\epsilon=20\%$. 
The blue lines are $\ell_2$ learning distance curves and the red lines are accuracy curves. }
\label{fig:distance-from-initialization-cifar}
\end{figure}

\textbf{Does learning distance tell the whole story?} Results in Figure \ref{fig:distance-mnist}, \ref{fig:distance-from-initialization-cifar} and Appendix \ref{apx:distance} show that when sparsity is higher than critical (interpolation threshold) sparsity, learning distance of sparse networks declines continuously as pruning, and test accuracy climbs up to a peak before finally goes down. 
This observation supports our proposed hypothesis to some extent, and is consistent with prior theoretical studies that small distance from initialization may relieve overfitting \citep{nagarajan2019generalization, li2020gradient}.
However, in Figure \ref{fig:distance-from-initialization-cifar}, the test accuracy of Critical Phase model is inferior to that of dense model, while the critical learning distance is not higher to the dense one (which differs from the MNIST results). It is natural to wonder why the learning distance curves of CIFAR do not present a peak in Critical Phase as distinct as those of MNIST? 

One possible reason might be the inherent property of rectified neural networks: \textit{scale invariance}. As we utilize ReLU as non-linearities activation, the network remains unchanged if we  multiply the weights in one layer by $t$, and the next layer by $1/t$. In this way, the network’s behavior remains unchanged but its distance from original parameters increases. Furthermore, this invariance is even more prominent when batch normalization is used, where the output of each layer will be re-scaled \citep{dinh2017sharp, li2018landscape}. In this case, the influence made by parameter removal during pruning might be offset by reparametrization effect of batch normalization, as such, changing the behavior of learning distance curve.

We suggest the learning distance as a potentially helpful generalization measure for sparse models, because this hypothesis correlates with the sparse double descent results to a certain extent. Proving it theoretically and rigorously for sparse neural networks remains as an open question. We will leave it for future work.

\textbf{Possible explanations for the imperception of sparse double descent without label noise.}
We'd like to further discuss why the double descent phenomenon is rarely exhibited in existing pruning literature as well as our experiments under zero label noise (see Figure \ref{fig:sparsedd-zero-noise}). Several possible explanations might account for its imperception: 

\begin{itemize}
    \item[(1)] The double descent phenomenon itself is not stable \citep{ba2020Generalization, nakkiran2020deepdoubledescent, yang2020rethinking}. To observe it in modern network architectures, we sometimes need to inject label noise \citep{nakkiran2020deepdoubledescent, yang2020rethinking}. 
    Moreover, even though a ``plateau'' or a small peak in test error around the interpolation point of dense models is reported on the noiseless CIFAR dataset \citep{nakkiran2020deepdoubledescent}, it will still be too hard to distinguish from the irregular fluctuations of test accuracy in pruning cases \citep{frankle2019lottery, liu2019rethinking, renda2020comparing}.

    \item[(2)] The Critical Phase often occur within quite thin sparsity ranges, and the Sweet Phase relies on relatively high sparsities (generally greater than 90\% under label noise). 
    Many studies might not report enough points in the accuracy-sparsity tradeoff curve \citep{Molchanov2017variational_dropout, liu2017slimming, le2021retraning_variants};
    
    \item[(3)] The regularization of pruning may come from other sources except for parameter reduction, like noise injection in model parameters\citep{bartoldson2020generalization-stability}, which might offset the performance loss brought by pruning when training without label noise.
\end{itemize} 
Nevertheless, under label noise settings, we can amplify the impact of reduced capacity on model performance brought about by sparsity, and reconcile the conventional understanding and the modern practice of network pruning.

\textbf{Relation with theory of sparse learning.}
So far, the relationship between sparsity, learning dynamics and generalization remains as open question and has received growing attention from researchers. 
Emerging studies from the perspective of loss landscape provide enlightening insight into 
understanding the behaviors of sparse regimes. \citet{evci2020difficulty} revealed that the existence of bad solutions in \textit{sparse subspace} (namely, the sparsity pattern found by pruning), and illustrate the difficulty of escaping from bad solutions to good ones. And \citet{lin2021sparse_landscape} provide theoretical justification that sparsity can deteriorate the loss landscape by creating spurious local minima or spurious valleys. Our work is motivated by these findings, and what's more, moves a step further by empirically demonstrating that the reshaping effect on loss landscape at high sparsities by network pruning is actually beneficial in the presence of label noise. 

\textbf{Connection to robust learning with label noise.} While our focus has been on the characteristics of sparse neural networks under noisy labels, there are other research hot-spots concerning label-noise learning, e.g., designing state-of-the-art robust training algorithms \citep{han2018coteaching, jiang2018mentornet, li2019learning2learn, li2020dividemix}. Among these methods, we find CDR proposed by \citet{xia2021robust} particularly related regarding the way to hinder memorization. Using a similar criterion to gradient-based pruning, they identify non-critical parameters and penalize them during optimization. By deactivating redundant parameters, memorization of noisy labels is hindered, and test performance before early stopping is enhanced. While our results reveal that, with a large proportion of parameters being removed permanently, performance after early stopping could also be boosted greatly.


\section{Conclusion}
\label{sec:conclusion}

In this paper, we reassess some common beliefs concerning the generalization properties of sparse networks, and illustrate the inapplicability of these viewpoints.
Instead, we report an unexpected sparse double descent phenomenon. And our proposed learning distance hypothesis correlates with previous theoretical studies, and accounts for this phenomenon to a certain extent.
We provide some insight into 
the optimization dynamics and generalization ability of sparse regimes, which we hope will guide progress towards further understanding for the theory of both sparse learning and deep learning.


\bibliography{sparsedd}
\bibliographystyle{icml2022}

\newpage
\appendix
\onecolumn
\section{Methodology}
\label{apx:methodology}

Here, we introduce the methods used in this paper in detail.
\subsection{Pruning Strategies}
\label{apx:pruning-strategy}
We used three network pruning strategies in the paper. Unless particularly specified, results and discussions are based on the magnitude-based pruning.
\begin{itemize}
\setlength{\itemsep}{0pt}
\item \textit{Magnitude-based pruning}: prunes the weights with the lowest absolute magnitudes $\lvert \rvw \rvert$.
\item \textit{Gradient-based pruning}: prunes the weights with the lowest absolute values of magnitude multiplies gradient $\lvert \frac{\partial L}{\partial \rvw} \odot \rvw \rvert$, with $L$ be the loss function evaluated on a random batch of inputs. 
\item \textit{Random pruning}: issues each weight with a random score sampled independently from the uniform distribution $\mathcal{U}(0,1)$, and prunes the weights with the lowest scores.
\end{itemize}

\subsection{Retraining Techniques}
\label{apx:retraining}
The typical retraining based pruning procedure consists of three stages \citep{liu2019rethinking}: 1) train a large, dense neural network to completion, 2) prune structures of the trained network according to certain heuristic, 3) retrain the network for $T$ epochs to mitigate accuracy loss. In this paper, we run experiments using the following four retraining methods to verify the robustness of sparse double descent, and implementation details like rewinding iteration and learning rate of finetuning are listed in \ref{table:training-details}:
\begin{itemize}
\setlength{\itemsep}{0pt}
    \item Lottery ticket rewinding: rewind the parameters and the learning rate of pruned network to the iteration $t$ at an early training stage, and subsequently retrain the models from there \citep{frankle2019lottery} for $T-t$ epochs.
    \item Finetuning: retrains the pruned model from the final values of parameters, for $T$ epochs \citep{han2015learning}. Usually, finetuning will use a fixed small learning rate, i.e., the last learning rate of the original training schedule \citep{li2017filters, liu2019rethinking}.
    \item Learning rate rewinding: retrains the pruned model for $T-t$ epochs from the final parameters, but reuse the learning rate schedule from the iteration $t$ at the early training phase \citep{renda2020comparing}. Learning rate rewinding is kind of a hybrid between lottery ticket rewinding and finetuning.
    \item Scratch retraining: rewind the learning rate schedule of pruned model to the beginning, but use a random reinitialization of parameters, and retrain the model for $T$ epochs. This method is often used as baseline to make comparison with lottery ticket rewinding method. In our paper, during the pruning and scratch training iteration, we use a different initialization each time when we retrain.
\end{itemize}

\subsection{Label Noise Types}
\label{apx:label-noise}
Here, we introduce the different types of synthetic label noise used in our paper. Following \citet{xia2021robust}, we manually generate symmetric, asymmetric and pairflip label noise, and the details are listed below. Most results and discussions are based on symmetric label noise unless otherwise specified. 
\begin{itemize}
\setlength{\itemsep}{0pt}
    \item Symmetric noise: generated by randomly permuted the labels for a fraction $\epsilon$ of the training data.
    \item Asymmetric noise: generated by flipping $\epsilon$ of labels within a set of similar classes. In our work, for MNIST, flip labels $2 \to 7, 3 \to 8, 5 \leftrightarrow 6$. For CIFAR-10,  BIRD $\to$ AIRPLANE, DEER $\to$ HORSE, TRUCK $\to$ AUTOMOBILE, CAT $\leftrightarrow$ DOG. And for CIFAR-100, all 100 classes will be divided into 20 super-classes, and then labels in every class will be flipped to the next class with the same super-class.
    \item Pairflip noise: generated by flipping labels in each class to its adjacent class for $\epsilon$ of training data.
\end{itemize}

\subsection{Re-dense Training}
\label{apx:re-dense}
Pruning induces sparsity constraints into the objective function optimization problem, which move the optimization to a lower-dimension space.
To empirically investigate the impact of sparsity constraints, we present studies which allow pruned weights to return to the model, which is motivated by DSD training \citep{han2017DSD} and noted as re-dense training in this paper.

The re-dense training follows the training process of a pruned network after $t$ epochs. During re-dense training, we recover pruned weights in the network, initialized them to zero, and retrain them together with the unpruned weights at the last epoch, for another $t$ epochs with a fixed learning rate. We set the learning rate in re-dense training equal to the last learning rate of sparse training. Other learning hyperparameters (batch size, momentum, weight decay, etc.) are kept the same as original training process.

\subsection{1-D Linear Interpolation.} 
\label{apx:interpolation}
Visualizing the loss landscape can provide an empirical characterization of the geometry of neural network minimizers (e.g., their sharpness/flatness, or the structures of surrounding parameter space). We present linear interpolation plots of the training loss function along a line segment $\rvw$ between sparse solutions $\rvw_s$ and re-dense solutions $\rvw_r$, using the strategy proposed by \citet{goodfellow2015qualitatively}. We define $\rvw(\alpha) = (1-\alpha) \rvw_s + \alpha \rvw_r$ for $\alpha \in [0,1]$ with increment of 0.01. And we compute the loss and accuracy of model with parameters $\rvw(\alpha)$ respectively. If there exists a monotonically decreasing loss objective from sparse solutions to re-dense solutions, we may conjecture that sparsity obstructs the optimization process with less trainable parameters.

\subsection{1-D Loss Visualization}
\label{apx:visualization}
We also plot 1-D loss function over a center minimizer using filter-wise normalized directions, to visualize the loss curvature and make comparisons between different minimizers. More explanation about this technique can be found in \citet{li2018landscape}.

\section{Experimental details}
\label{experimental-details}

Here, we present the implementation details used in this paper.

We adopt standard implementations of LeNet-300-100 from OpenLTH\footnote{\href{https://github.com/facebookresearch/open\_lth}{https://github.com/facebookresearch/open\_lth}}. LeNet-300-100 is a fully-connected network with 300 units in the first layer and 100 units in the second hidden layer, and ReLU activations.

For ResNet-18 network, we utilize a modified version of PyTorch model. To adapt ResNet-18 for CIFAR-10 and CIFAR-100, the first convolutional layer is equipped with filter of size $3\times 3$ and the max-pooling layer that follows has been eliminated. CIFAR-10 and CIFAR-100 are augmented with per-channel normalization, randomly horizontal flipping, and randomly shifting by up to four pixels in any direction. And for VGG-16 network, we follow the settings from OpenLTH. 

The ResNet-101 network for Tiny ImageNet is modified in the same way as the ResNet-18 network. Training instances in Tiny ImageNet are augmented with per-channel normalization, randomly cropping and resizing to a size of $32 \times 32$, and randomly horizontal flipping.

In pruning experiments, for LeNet-300-100, we consider all weights from linear layers except for the last layer as prunable parameters; for ResNets, all weights from convolutional and linear layers are set as prunable; while for VGG-16, we prune convolutional weights. We do not prune biases nor the batch normalization parameters. For convolutional and linear layers, the weights are initialized with Kaiming normal strategy and biases are initialized to be zero. Note that, all these settings are adopted from prior work \citet{frankle2019lottery} and never particularly tuned for the sake of fairness.

We run all the MNIST and CIFAR experiments on single GPU and Tiny ImageNet experiments on four GPUs with CUDA 10.1. The training hyperparameters used in our experiments is given as follows. Our code is available in \href{https://github.com/hezheug/sparse-double-descent}{https://github.com/hezheug/sparse-double-descent}.

\begin{table}[ht]
\renewcommand\arraystretch{1.2}
\scriptsize
\label{table:training-details}
\begin{center}
\setlength{\tabcolsep}{1mm}{
\begin{tabular}{ccccccccccccc}
\hline
\multicolumn{1}{c}{Network}  &\multicolumn{1}{c}{Dataset} &\multicolumn{1}{c}{Epochs} &\multicolumn{1}{c}{Batch} &\multicolumn{1}{c}{Opt.} &\multicolumn{1}{c}{Mom.} &\multicolumn{1}{c}{LR} &\multicolumn{1}{c}{LR Drop} &\multicolumn{1}{c}{Drop Factor}  &\multicolumn{1}{c}{Weight Decay} &\multicolumn{1}{c}{Rewind Iter} 
&\multicolumn{1}{c}{LR(finetune)}
 &\multicolumn{1}{c}{LR(re-dense)}
\\ \hline
LeNet-300-100 & MNIST & 200 & 128 & SGD & --- & 0.1 & ---  & --- & --- & 0 & 0.1 & 0.1 \\
ResNet-18 & CIFAR-10 & 160 & 128 & SGD & 0.9 & 0.1 & 80, 120 & 0.1 & 1e-4 & 1000 & 0.001 & 0.001\\
ResNet-18 & CIFAR-100 & 160 & 128 & SGD & 0.9 & 0.1 & 80, 120 & 0.1  & 1e-4 & 1000 & 0.001 & 0.001  \\
VGG-16 & CIFAR-10 & 160 & 128 & SGD & 0.9 & 0.1 & 80, 120 & 0.1 & 1e-4 & 2000 & 0.001 & 0.001 \\
VGG-16 & CIFAR-100 & 160 & 128 & SGD & 0.9 & 0.1 & 80, 120 & 0.1  & 1e-4 & 2000 & 0.001 & 0.001 \\
ResNet-101 & Tiny ImageNet & 200 & 512 & SGD & 0.9 & 0.2 & 100, 150 & 0.1  & 1e-4 & 1000 & --- & ---  \\
\hline
\end{tabular}}
\end{center}
\end{table}


\section{Additional experiment results and discussion}
\label{sec:supresults}
Here, we will present additional results that are not included in the main body for page limit.

\subsection{Sparse double descent phenomenon}
\label{appendix:sparsedd}

\begin{figure}[H]
\setlength{\abovecaptionskip}{0pt} 
\setlength{\belowcaptionskip}{0pt} 
\begin{center}
\includegraphics[width=0.3\textwidth]{img/mnist/0.2_acc_magnitude.pdf}
\includegraphics[width=0.3\textwidth]{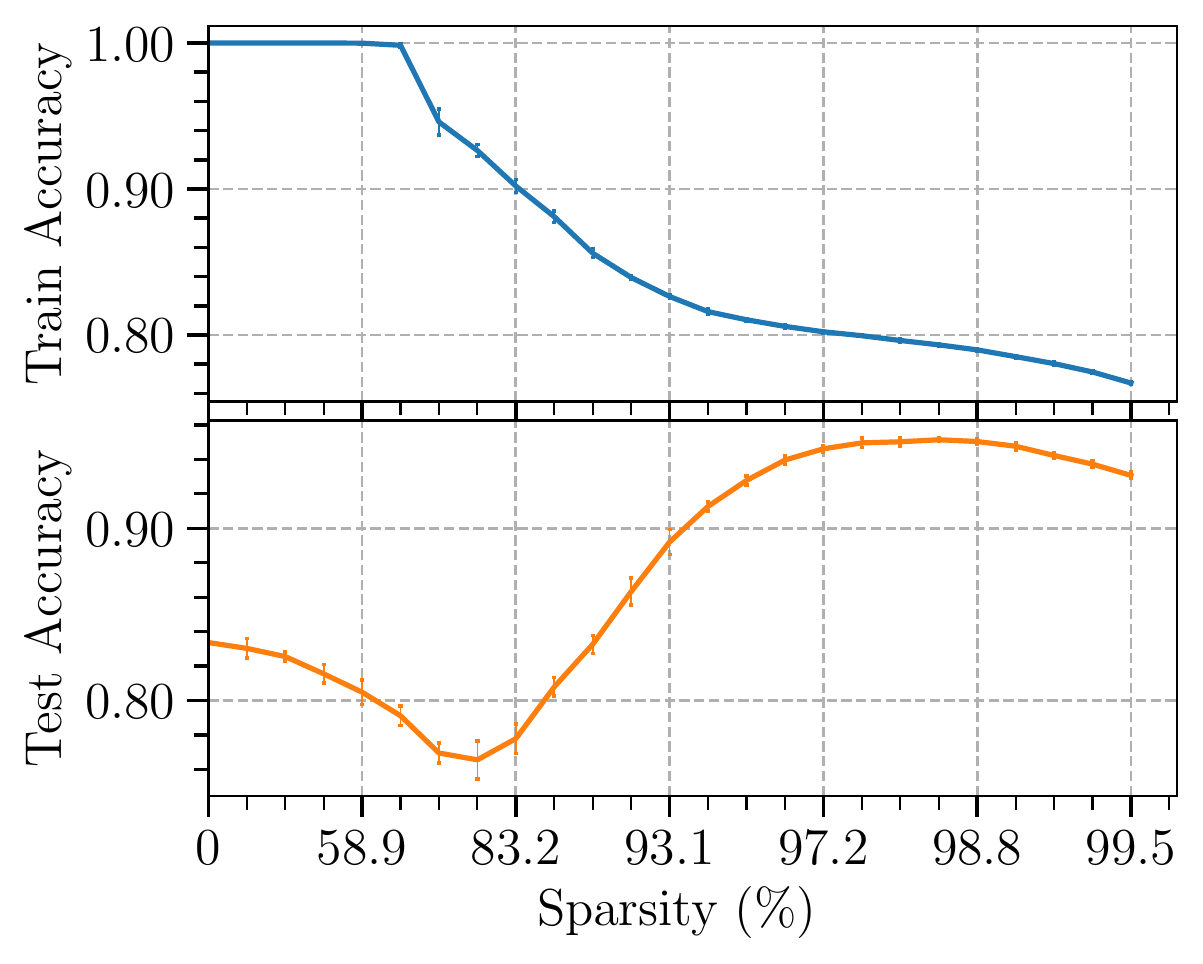}
\includegraphics[width=0.31\textwidth]{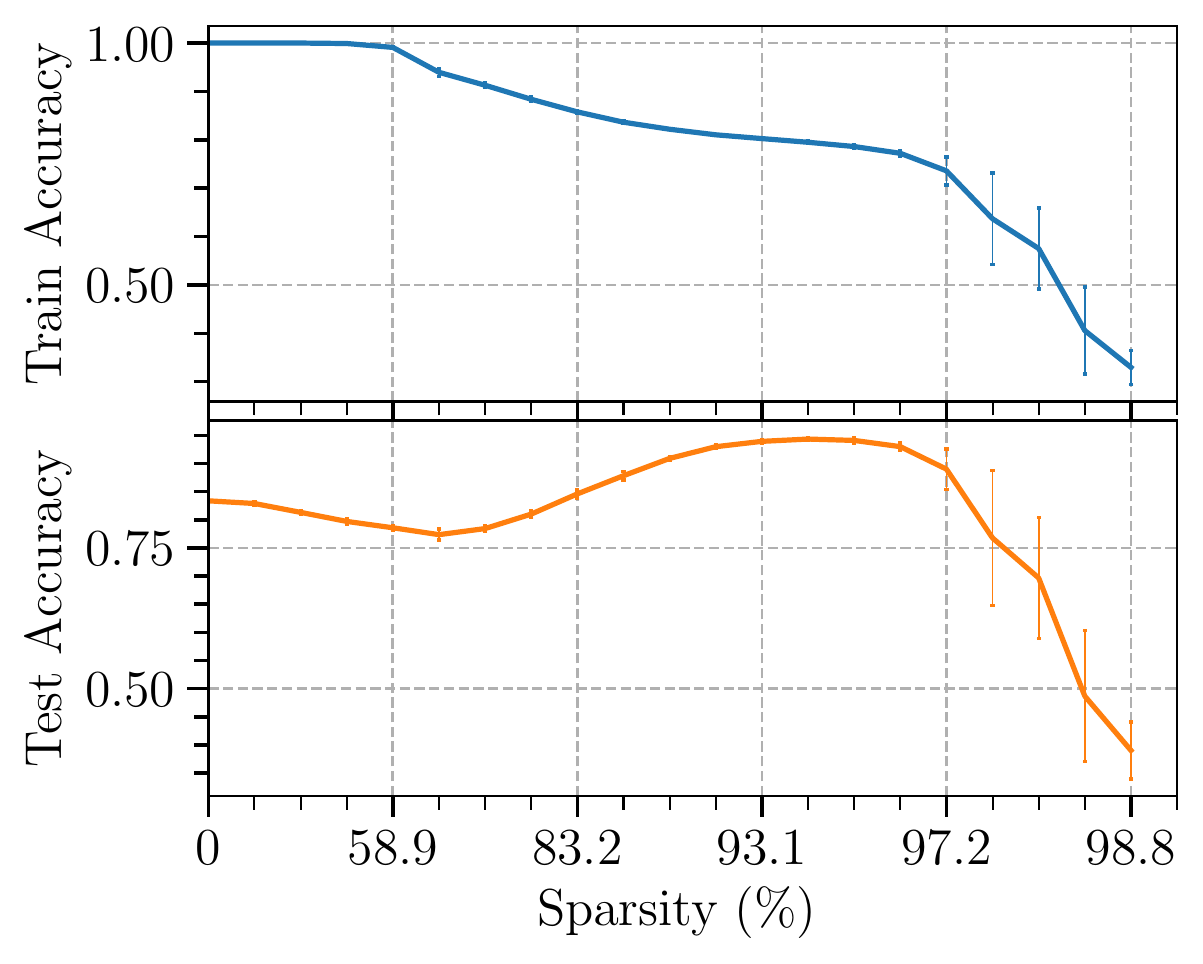}
\end{center}
\caption{Sparse Double descend phenomenon with different pruning strategies for LeNet-300-100 on MNIST with symmetric label noise, $\epsilon=20\%$. \textbf{Left}: Magnitude-based pruning. \textbf{Middle}: Gradient-based pruning. \textbf{Right}: Random pruning.}
\label{fig:sparsedd-mnist-pruning-strategy-0.2}
\end{figure}

\begin{figure}[H]
\vspace{-0.3cm}
\setlength{\abovecaptionskip}{0pt} 
\setlength{\belowcaptionskip}{0pt} 
\begin{center}
\includegraphics[width=0.3\textwidth]{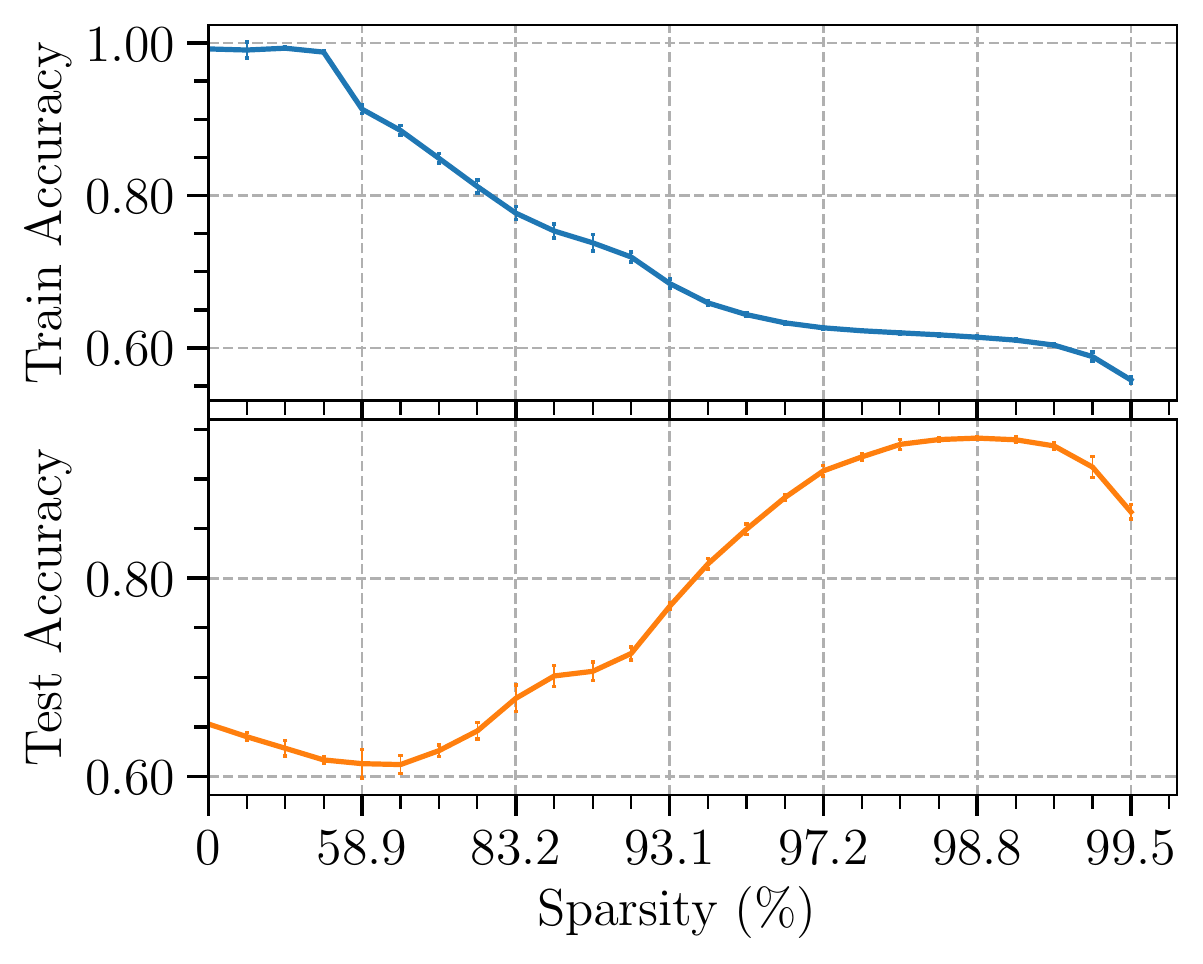}
\includegraphics[width=0.3\textwidth]{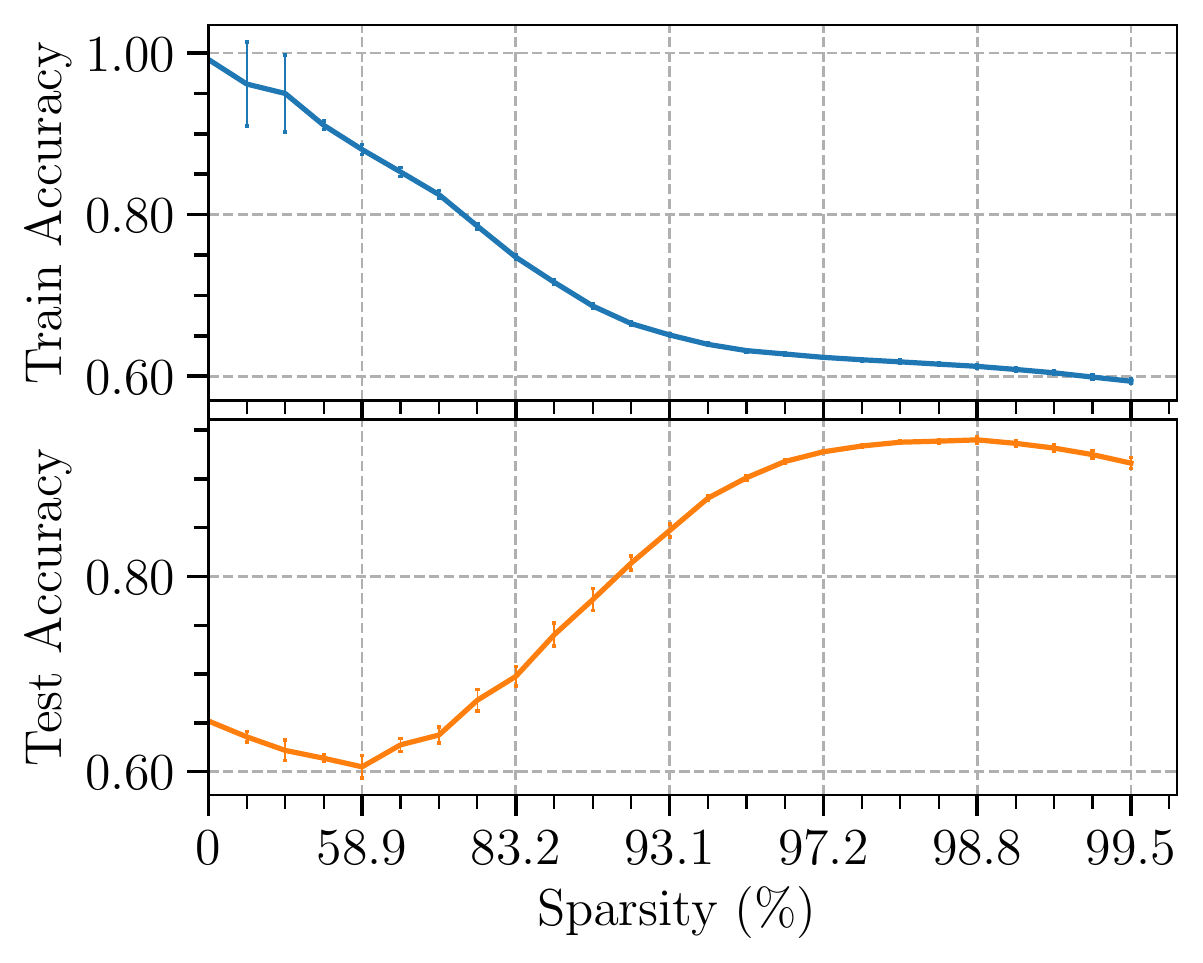}
\includegraphics[width=0.31\textwidth]{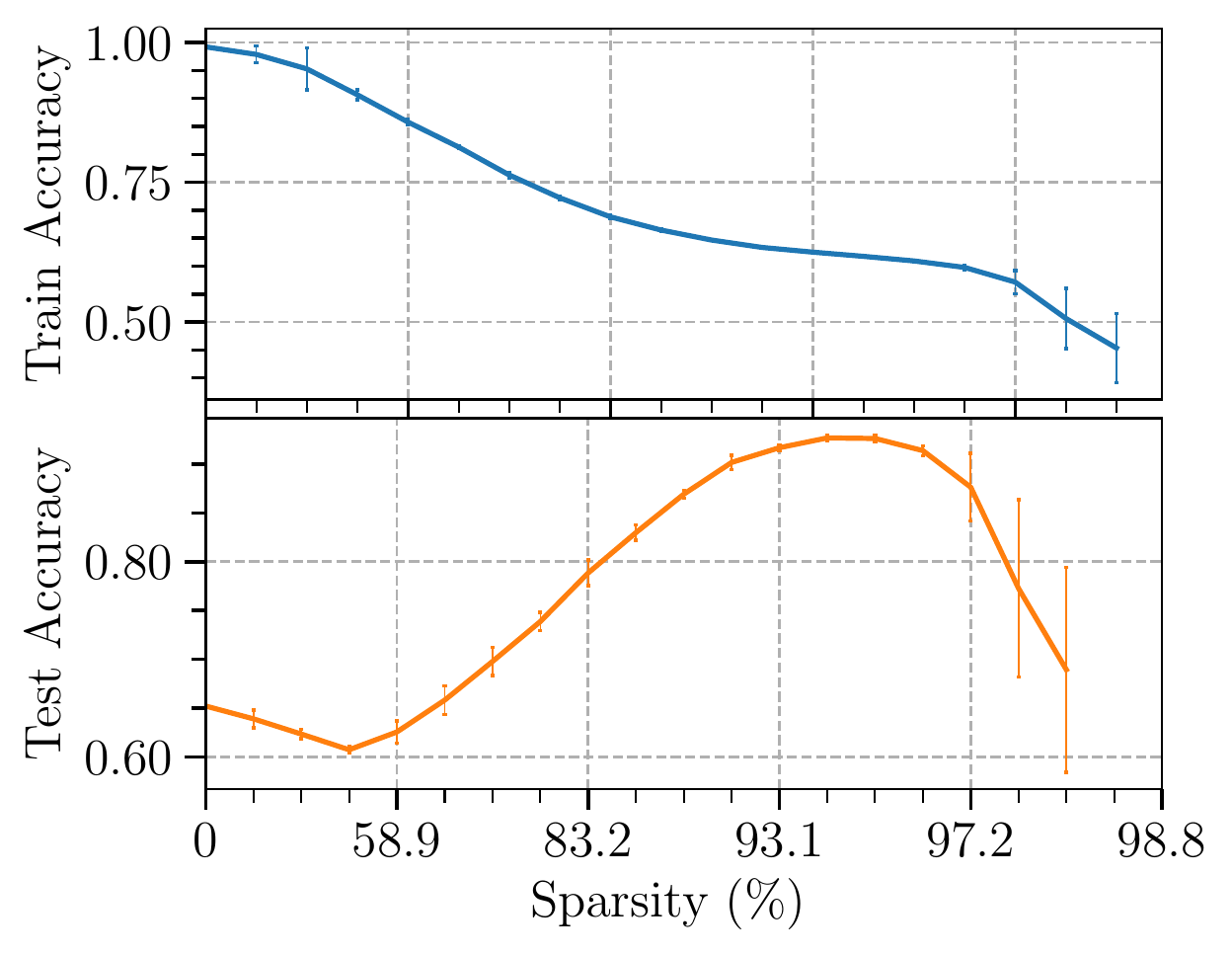}
\end{center}
\caption{Sparse Double descend phenomenon with different pruning strategies for LeNet-300-100 on MNIST with symmetric label noise, $\epsilon=40\%$. \textbf{Left}: Magnitude-based pruning. \textbf{Middle}: Gradient-based pruning. \textbf{Right}: Random pruning.}
\label{fig:sparsedd-mnist-pruning-strategy-0.4}
\end{figure}

\begin{figure}[H]
\vspace{-0.3cm}
\setlength{\abovecaptionskip}{0pt} 
\setlength{\belowcaptionskip}{0pt} 
\begin{center}
\includegraphics[width=0.3\textwidth]{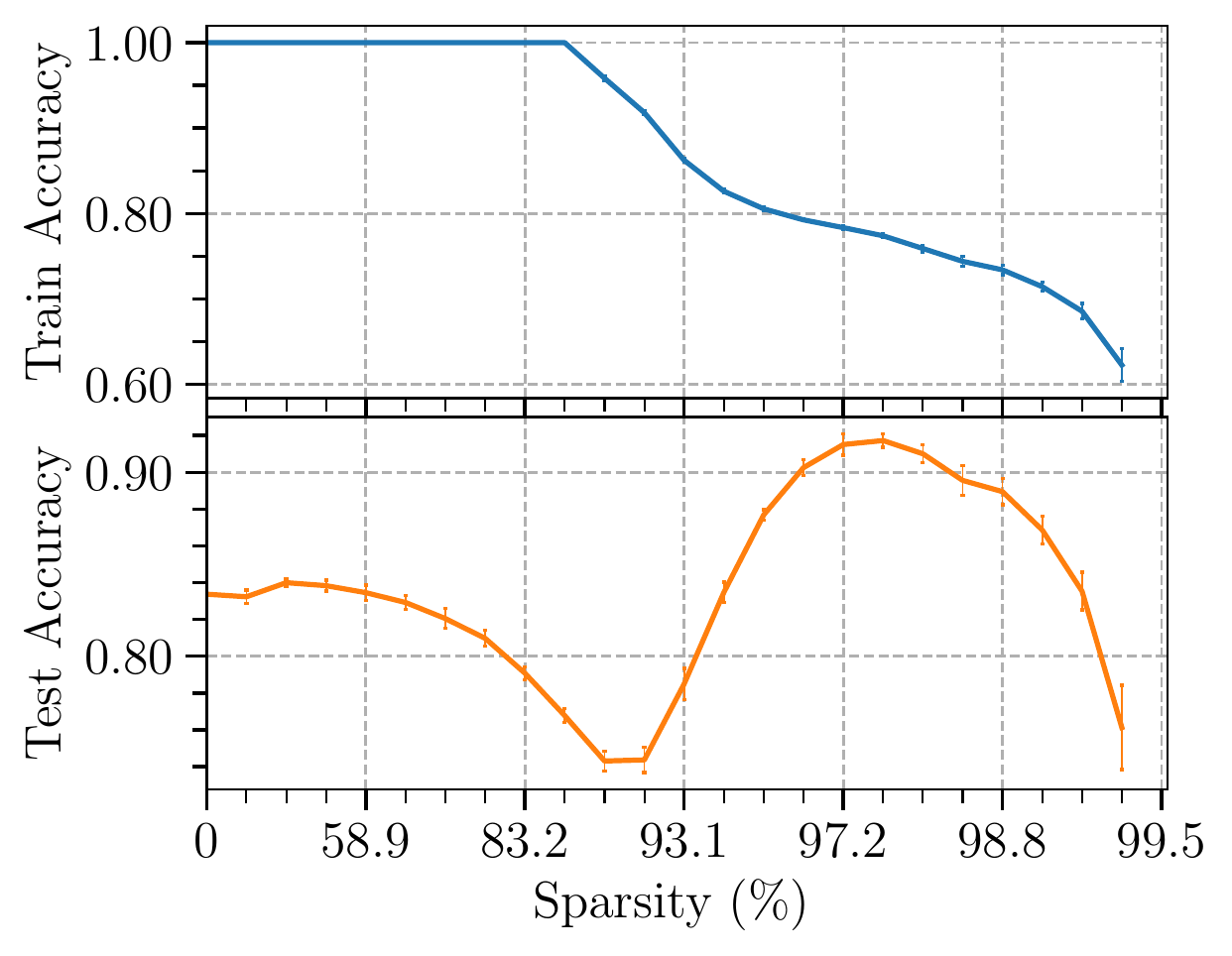}
\includegraphics[width=0.3\textwidth]{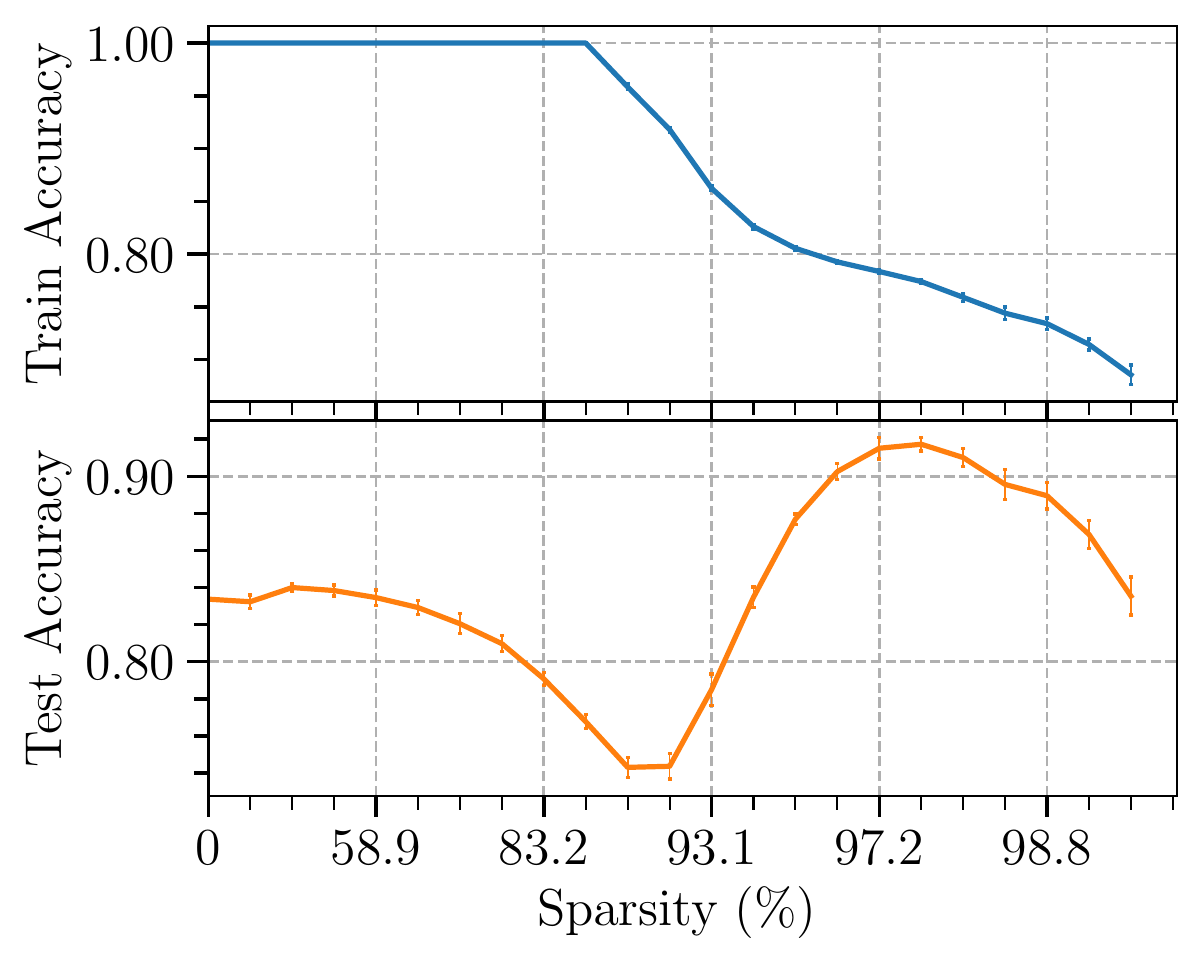}
\includegraphics[width=0.31\textwidth]{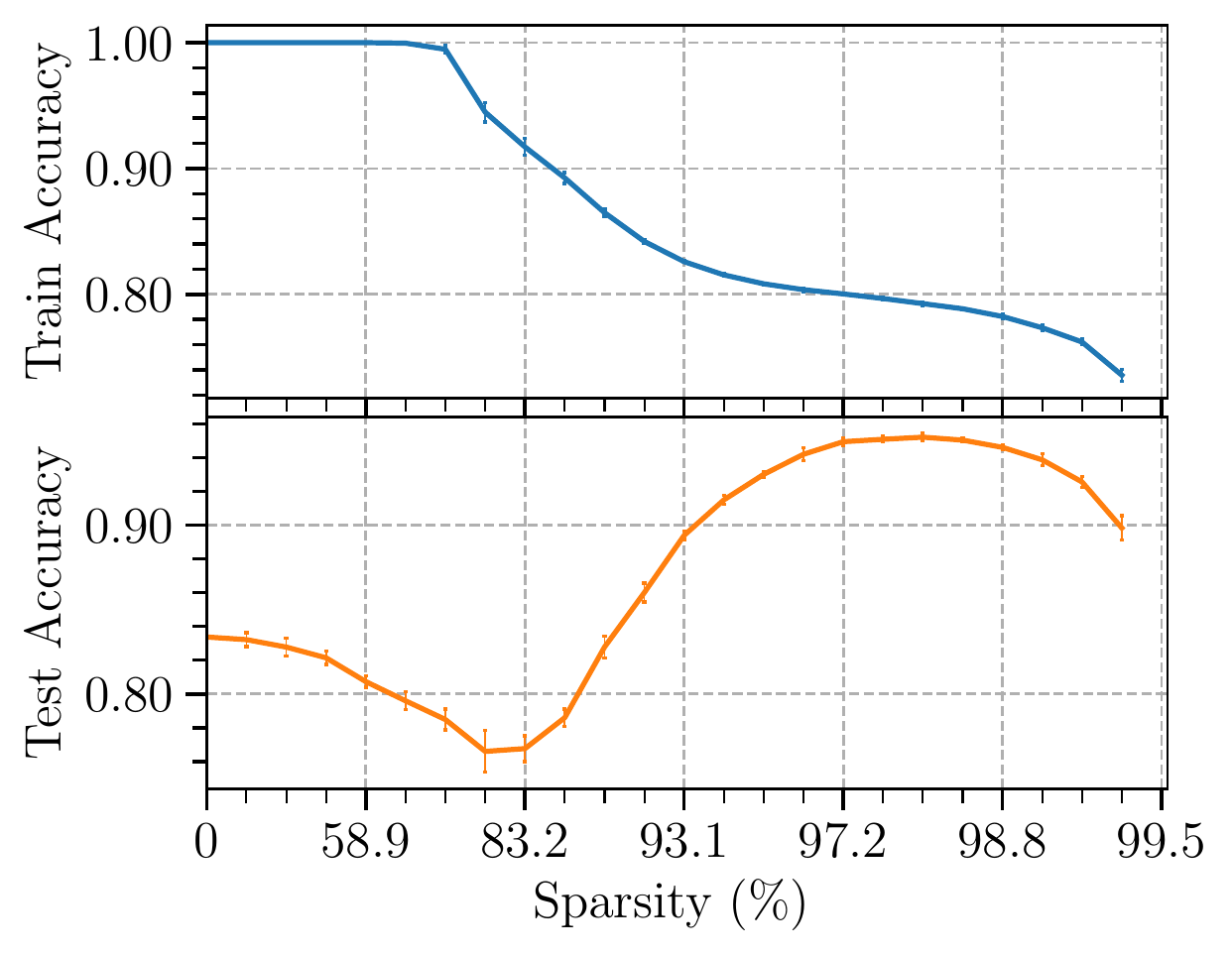}
\end{center}
\caption{Sparse Double descend phenomenon with different retraining methods for LeNet-300-100 on MNIST with symmetric label noise, $\epsilon=20\%$. \textbf{Left}: Finetuning. \textbf{Middle}: Learning Rate Rewinding. \textbf{Right}: Scratch retraining.}
\label{fig:sparsedd-mnist-retrain-0.2}
\end{figure}

\begin{figure}[H]
\vspace{-0.3cm}
\setlength{\abovecaptionskip}{0pt} 
\setlength{\belowcaptionskip}{0pt} 
\begin{center}
    \includegraphics[width=0.3\textwidth]{img/cifar10/0.2_acc_magnitude.pdf}
    \includegraphics[width=0.3\textwidth]{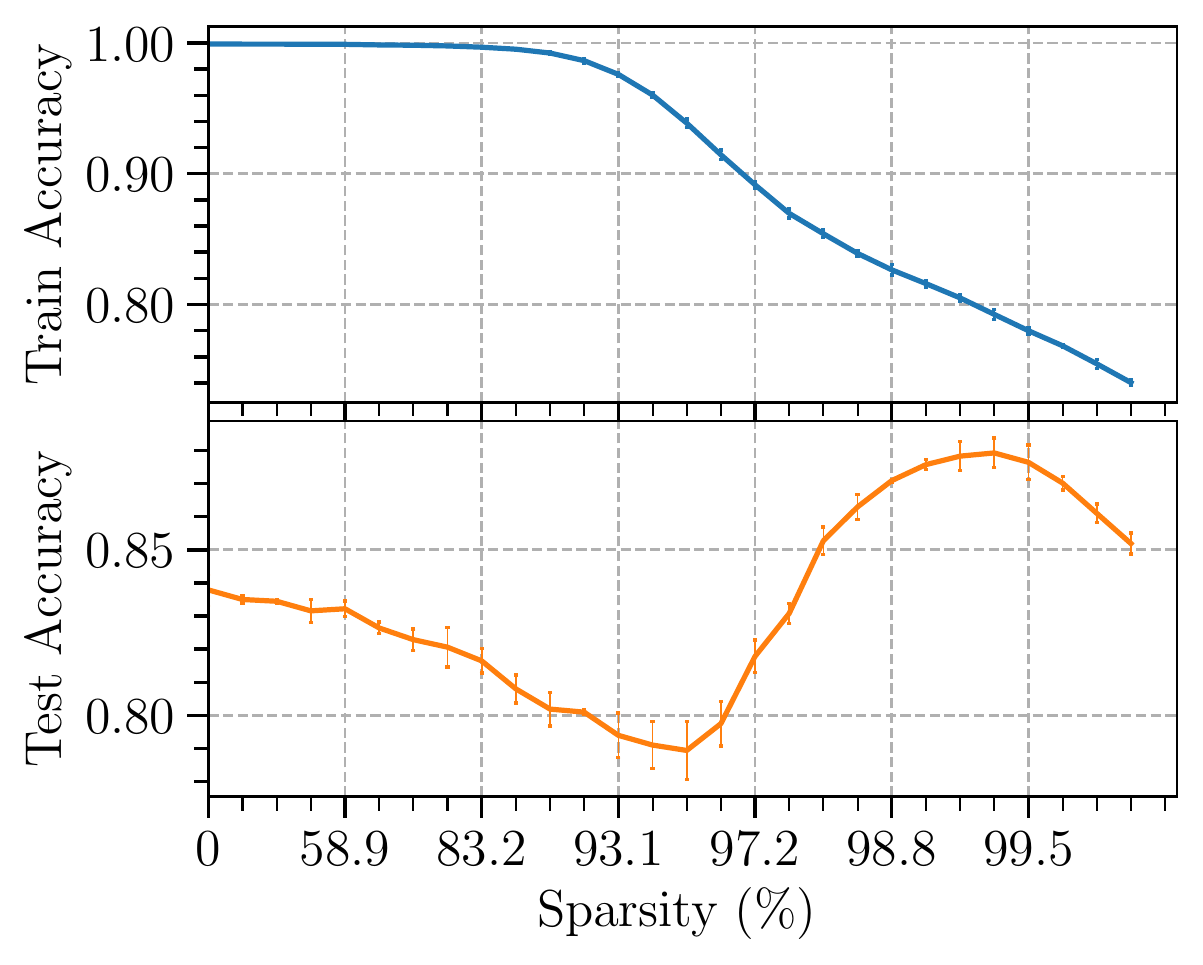}
    \includegraphics[width=0.3\textwidth]{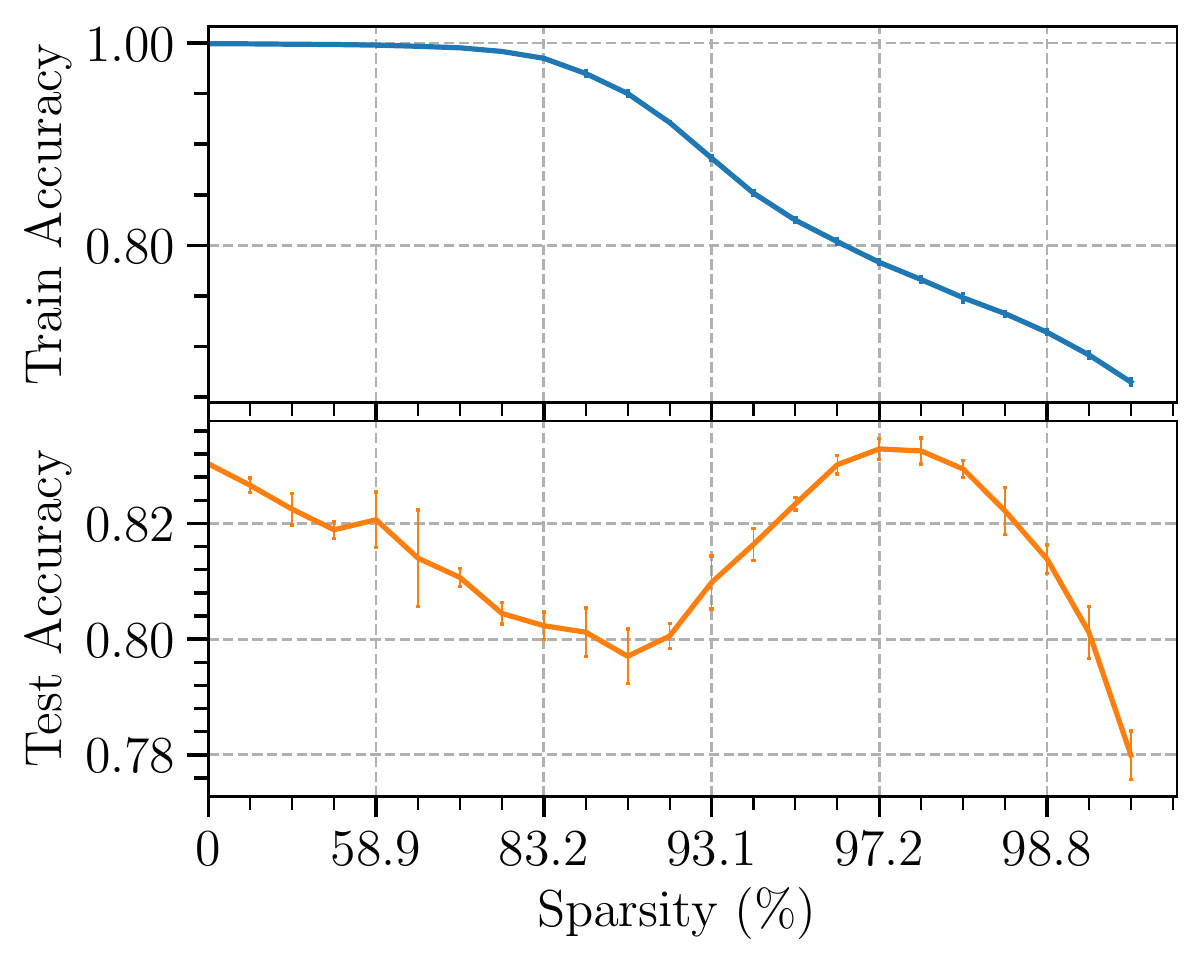}
\end{center}
\vspace{-0.3cm}
\caption{Sparse Double descend phenomenon with different pruning strategies for ResNet-18 on CIFAR-10  with symmetric label noise, $\epsilon=20\%$. \textbf{Left}: Magnitude-based pruning. \textbf{Middle}: Gradient-based pruning. \textbf{Right}: Random pruning.}
\label{fig:sparsedd-cifar10-pruning-strategy-0.2}
\end{figure}

\begin{figure}[H]
\vspace{-0.3cm}
\setlength{\abovecaptionskip}{0pt} 
\setlength{\belowcaptionskip}{0pt} 
\begin{center}
    \includegraphics[width=0.3\textwidth]{img/cifar10/0.4_acc_magnitude.pdf}
    \includegraphics[width=0.3\textwidth]{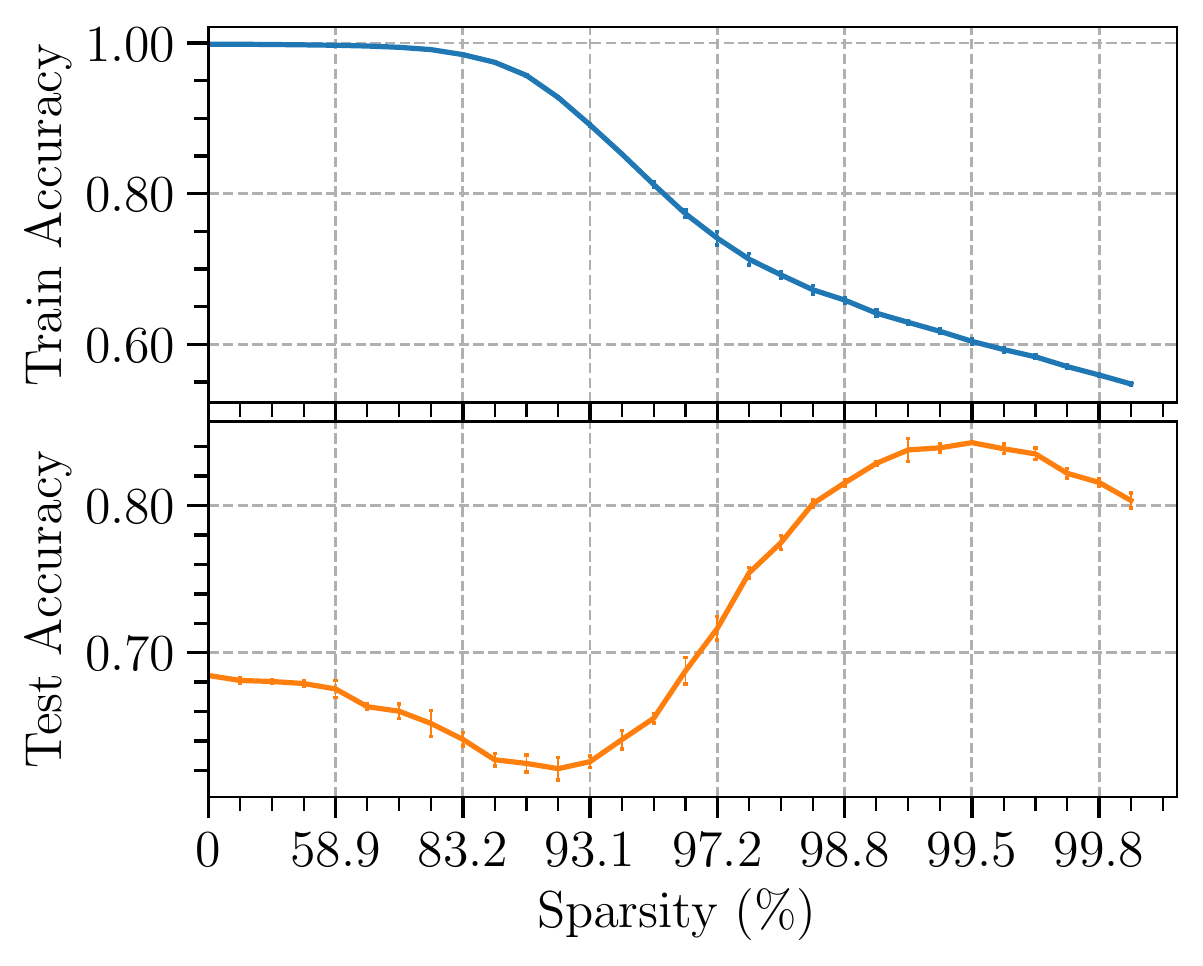}
    \includegraphics[width=0.3\textwidth]{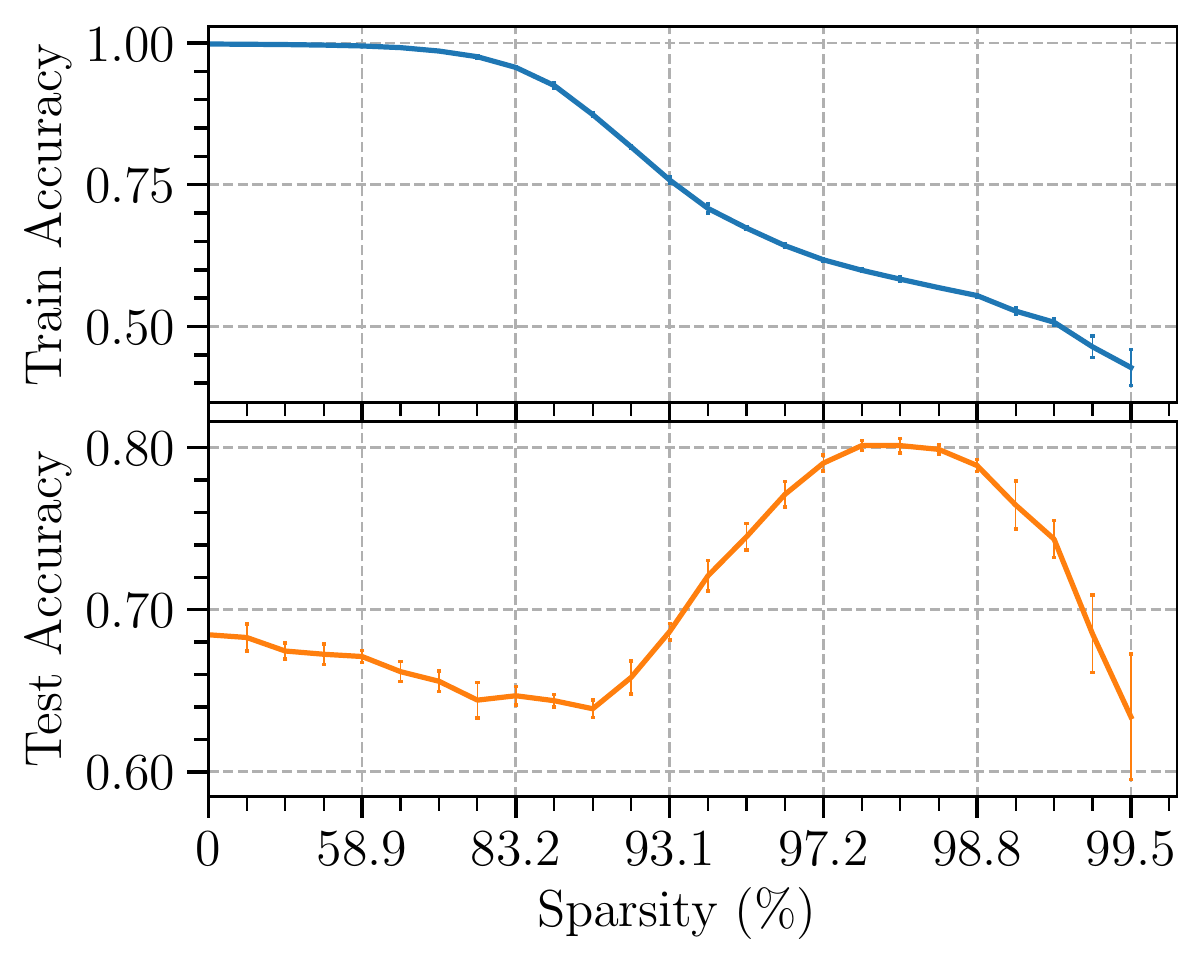}
\end{center}
\vspace{-0.3cm}
\caption{Sparse Double descend phenomenon with different pruning strategies for ResNet-18 on CIFAR-10  with symmetric label noise, $\epsilon=40\%$. \textbf{Left}: Magnitude-based pruning. \textbf{Middle}: Gradient-based pruning. \textbf{Right}: Random pruning.}
\label{fig:sparsedd-cifar10-pruning-strategy-0.4}
\end{figure}

\begin{figure}[H]
\vspace{-0.3cm}
\setlength{\abovecaptionskip}{0pt} 
\setlength{\belowcaptionskip}{0pt} 
\begin{center}
    \includegraphics[width=0.3\textwidth]{img/cifar10/0.8_acc_magnitude.pdf}
    \includegraphics[width=0.3\textwidth]{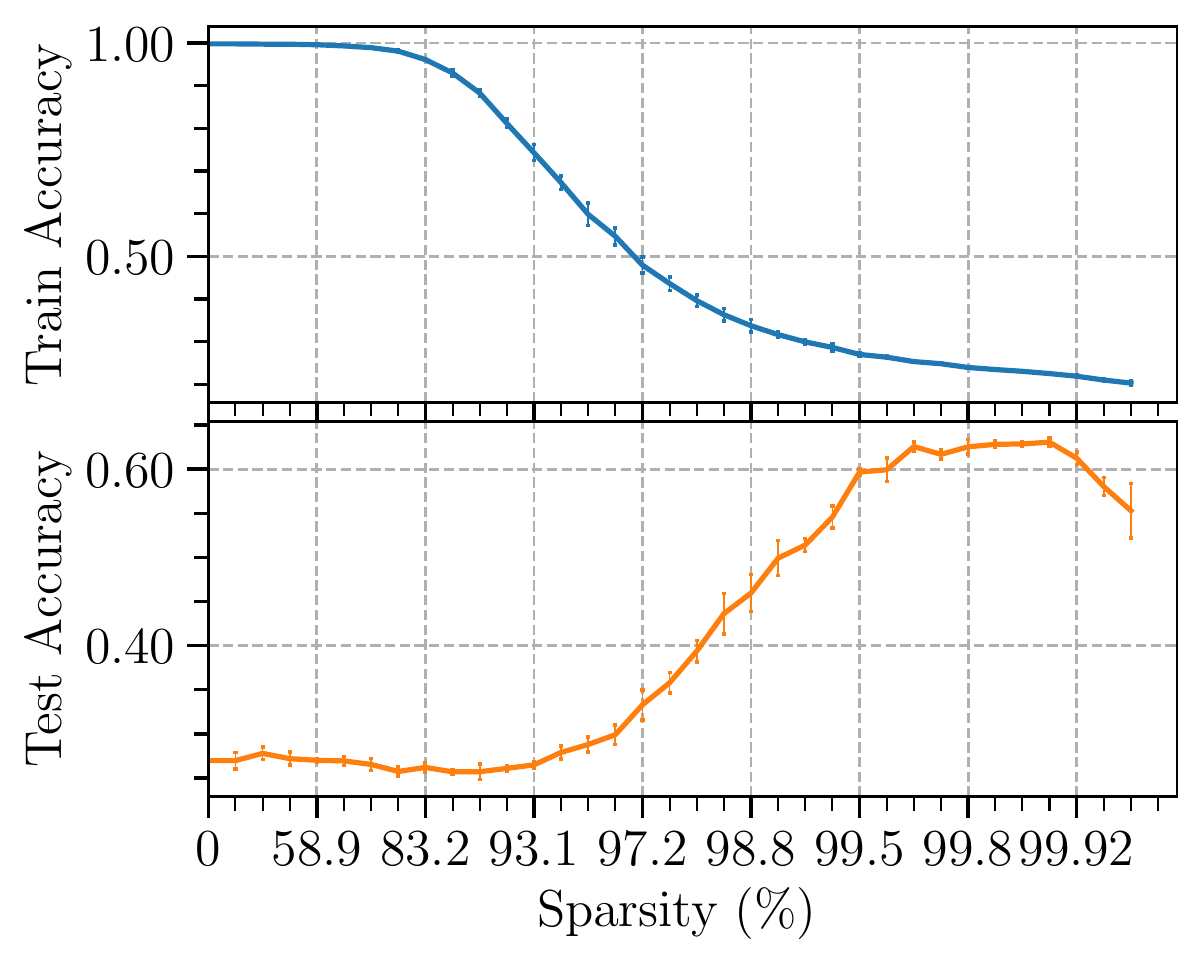}
    \includegraphics[width=0.3\textwidth]{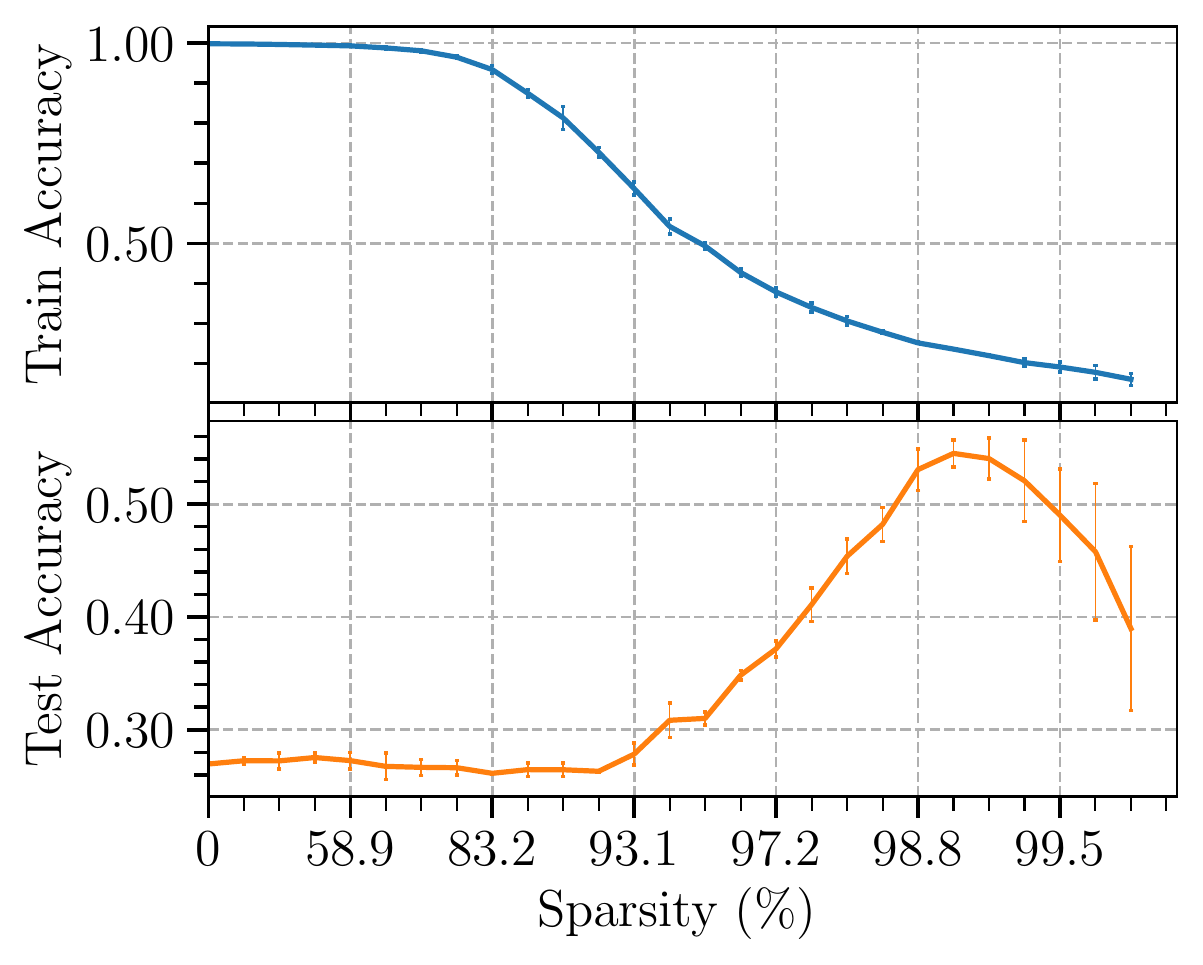}
\end{center}
\vspace{-0.3cm}
\caption{Sparse Double descend phenomenon with different pruning strategies for ResNet-18 on CIFAR-10  with symmetric label noise, $\epsilon=80\%$. \textbf{Left}: Magnitude-based pruning. \textbf{Middle}: Gradient-based pruning. \textbf{Right}: Random pruning.}
\label{fig:sparsedd-cifar10-pruning-strategy-0.8}
\end{figure}

\begin{figure}[H]
\vspace{-0.3cm}
\setlength{\abovecaptionskip}{0pt} 
\setlength{\belowcaptionskip}{0pt} 
\begin{center}
\includegraphics[width=0.3\textwidth]{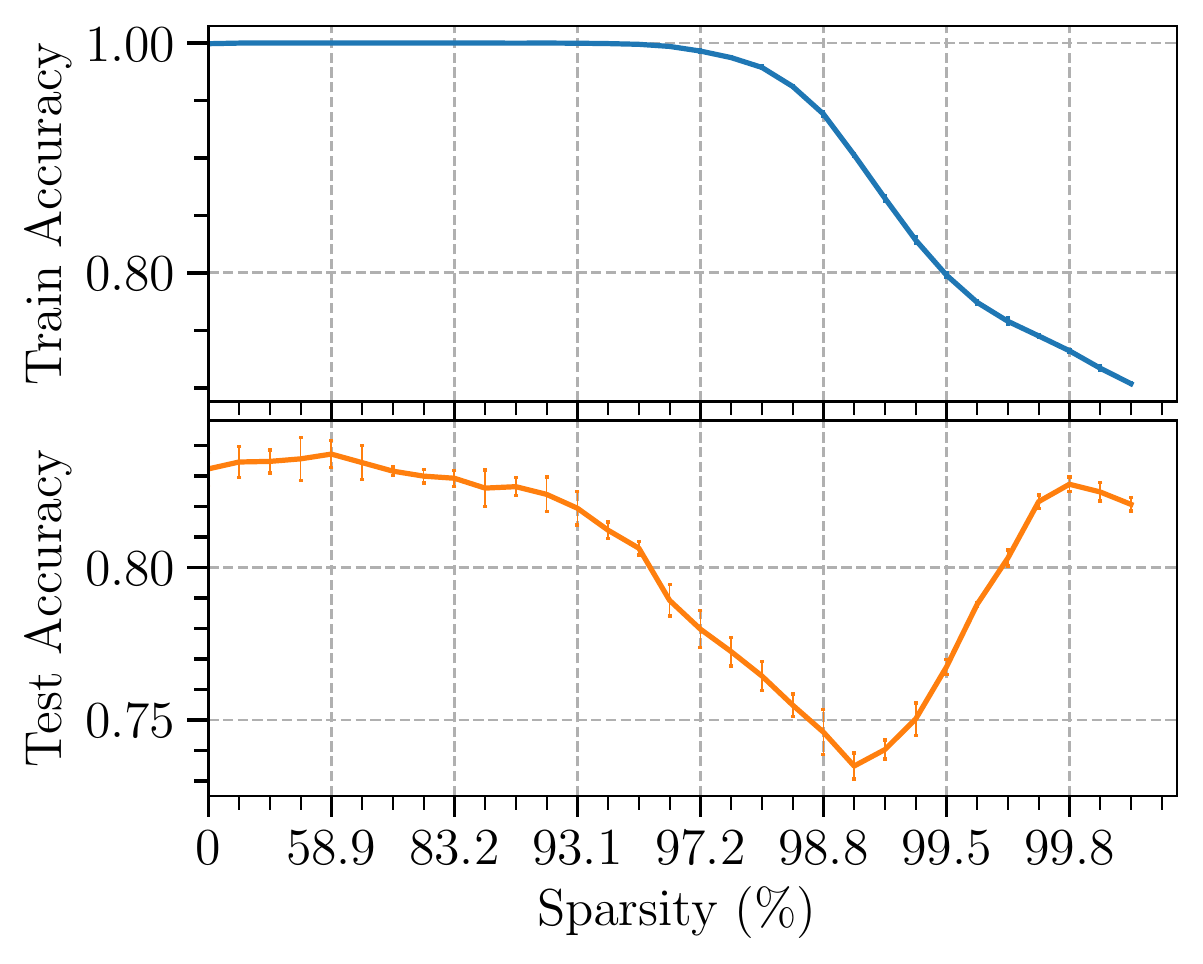}
\includegraphics[width=0.3\textwidth]{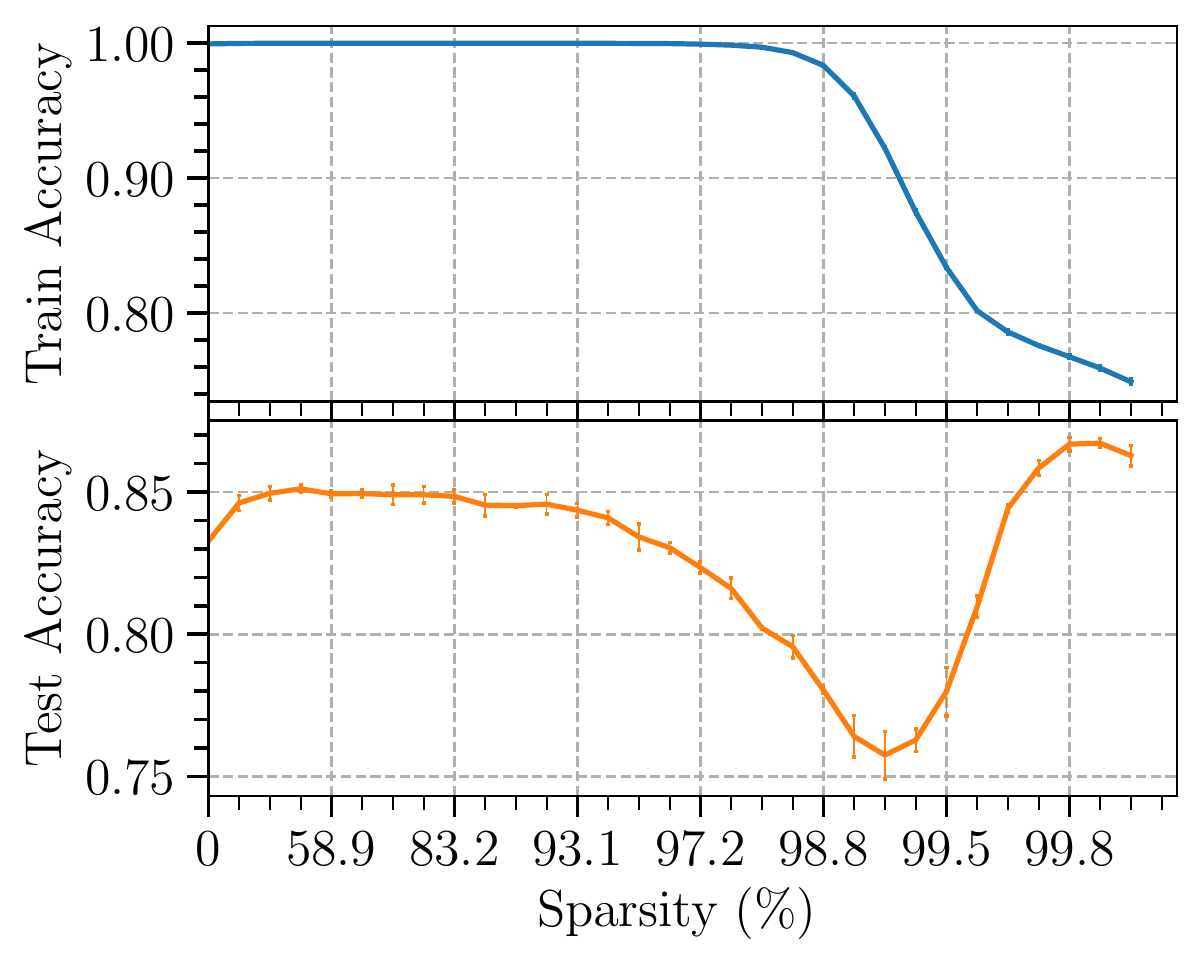}
\includegraphics[width=0.31\textwidth]{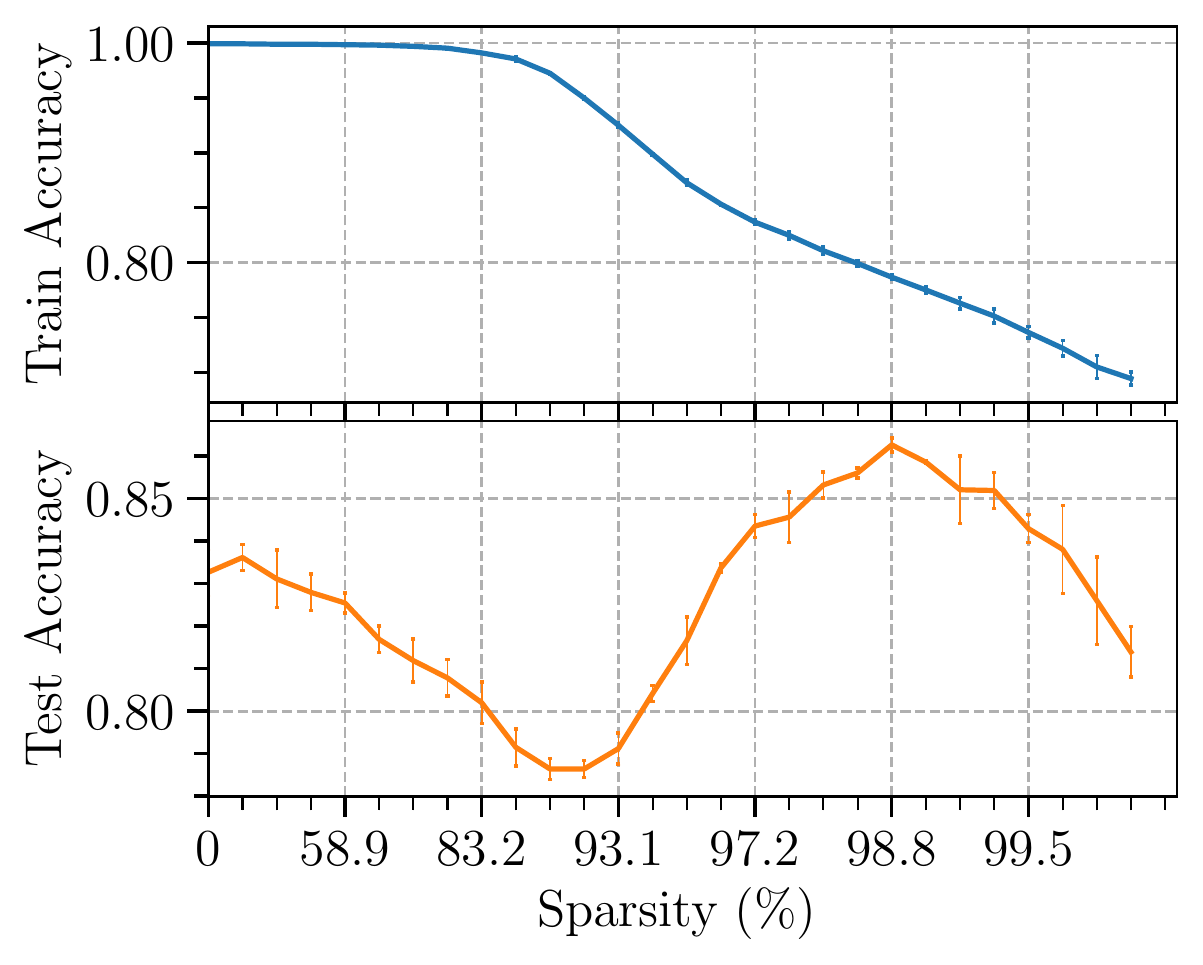}
\end{center}
\vspace{-0.3cm}
\caption{Sparse Double descend phenomenon with different retraining methods for ResNet-18 on CIFFAR-10 with symmetric label noise, $\epsilon=20\%$. \textbf{Left}: Finetuning. \textbf{Middle}: Learning Rate Rewinding. \textbf{Right}: Scratch retraining.}
\label{fig:sparsedd-cifar10-retrain-0.2}
\end{figure}

\begin{figure}[H]
\vspace{-0.3cm}
\setlength{\abovecaptionskip}{0pt} 
\setlength{\belowcaptionskip}{0pt} 
\begin{center}
    \includegraphics[width=0.3\textwidth]{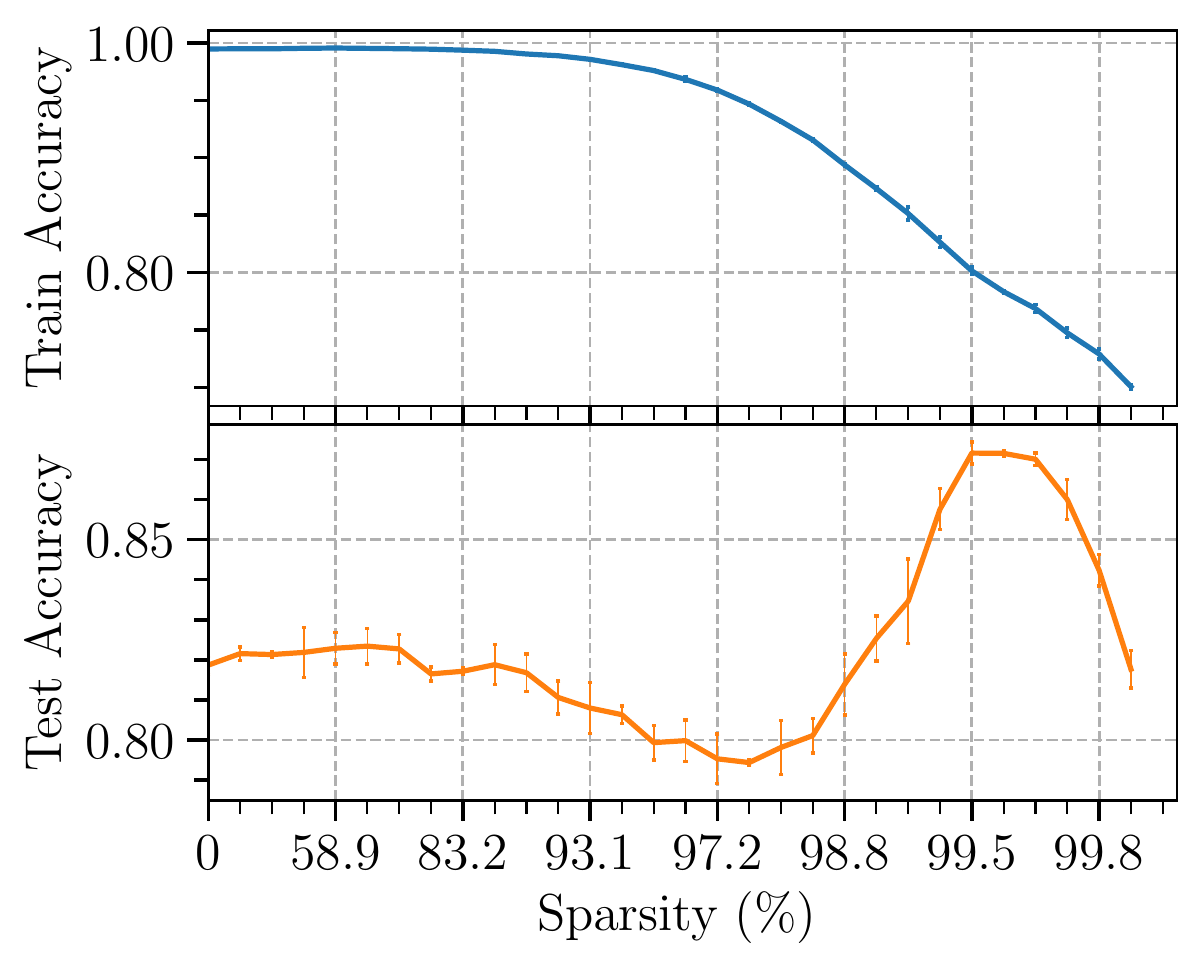}
    \includegraphics[width=0.3\textwidth]{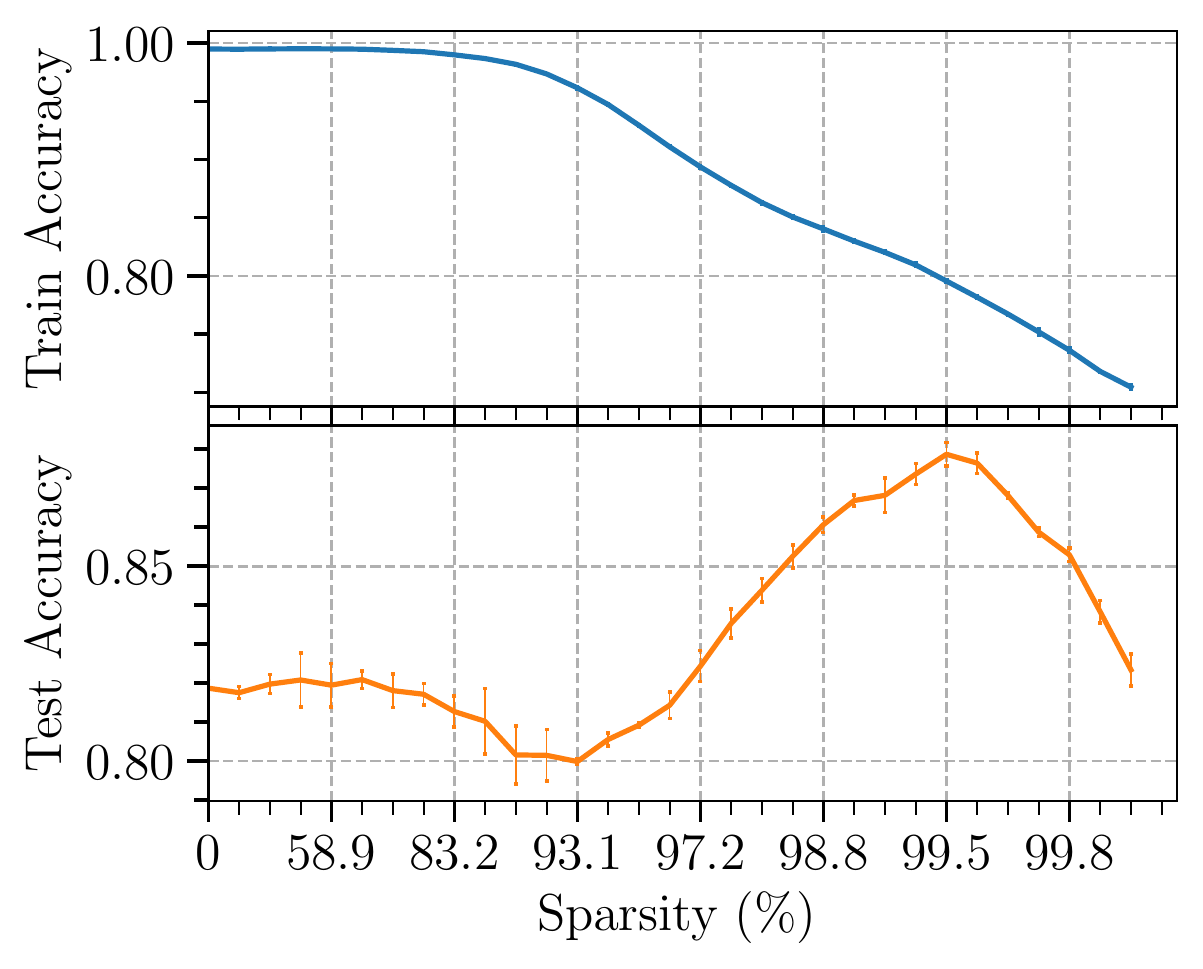}
    \includegraphics[width=0.3\textwidth]{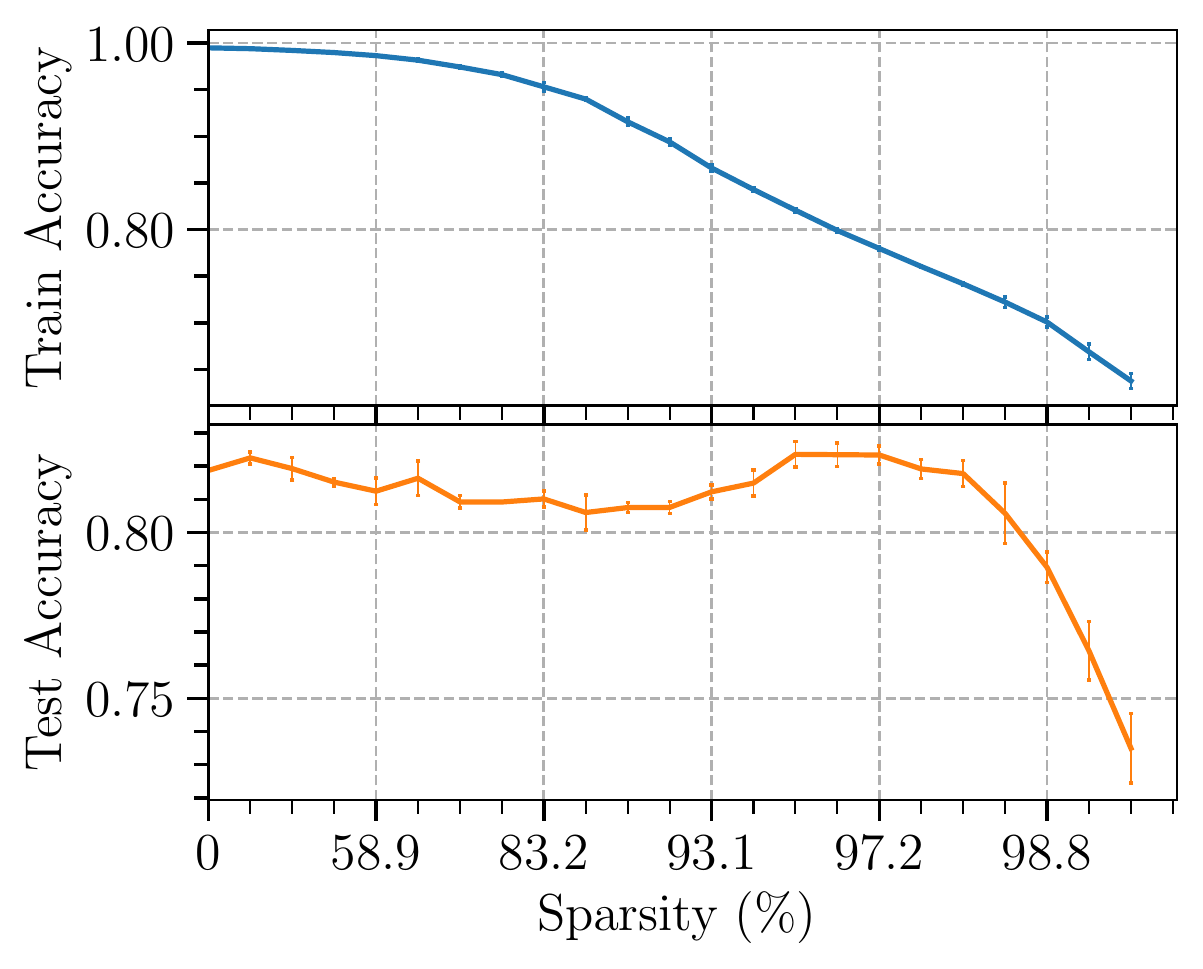}
\end{center}
\vspace{-0.3cm}
\caption{Sparse Double descend phenomenon with different pruning strategies for VGG-16 on CIFAR-10  with symmetric label noise, $\epsilon=20\%$. \textbf{Left}: Magnitude-based pruning. \textbf{Middle}: Gradient-based pruning. \textbf{Right}: Random pruning.}
\label{fig:sparsedd-cifar10-pruning-strategy-0.2-vgg}
\end{figure}

\begin{figure}[H]
\setlength{\abovecaptionskip}{0pt} 
\setlength{\belowcaptionskip}{0pt} 
\begin{center}
    \includegraphics[width=0.3\textwidth]{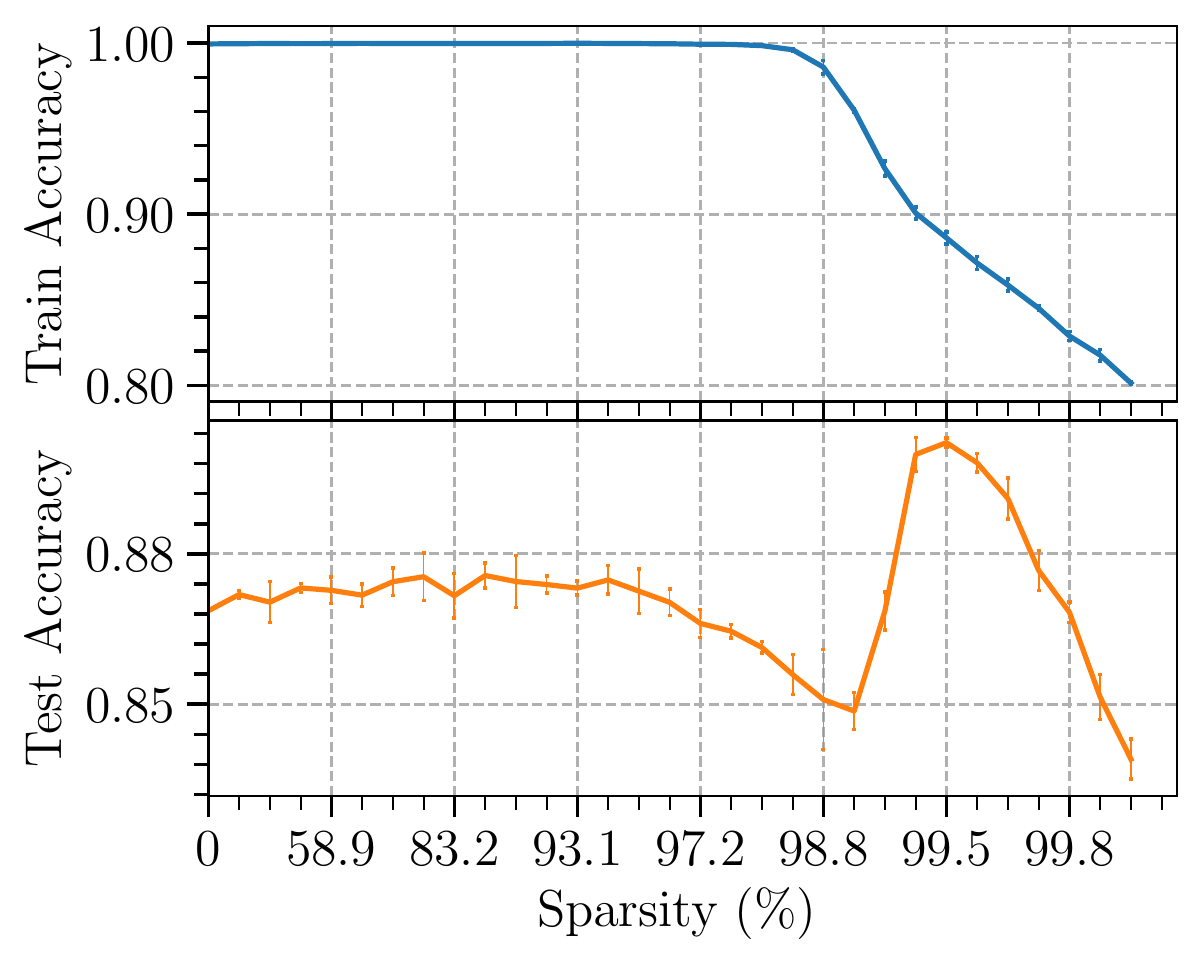}
    \includegraphics[width=0.3\textwidth]{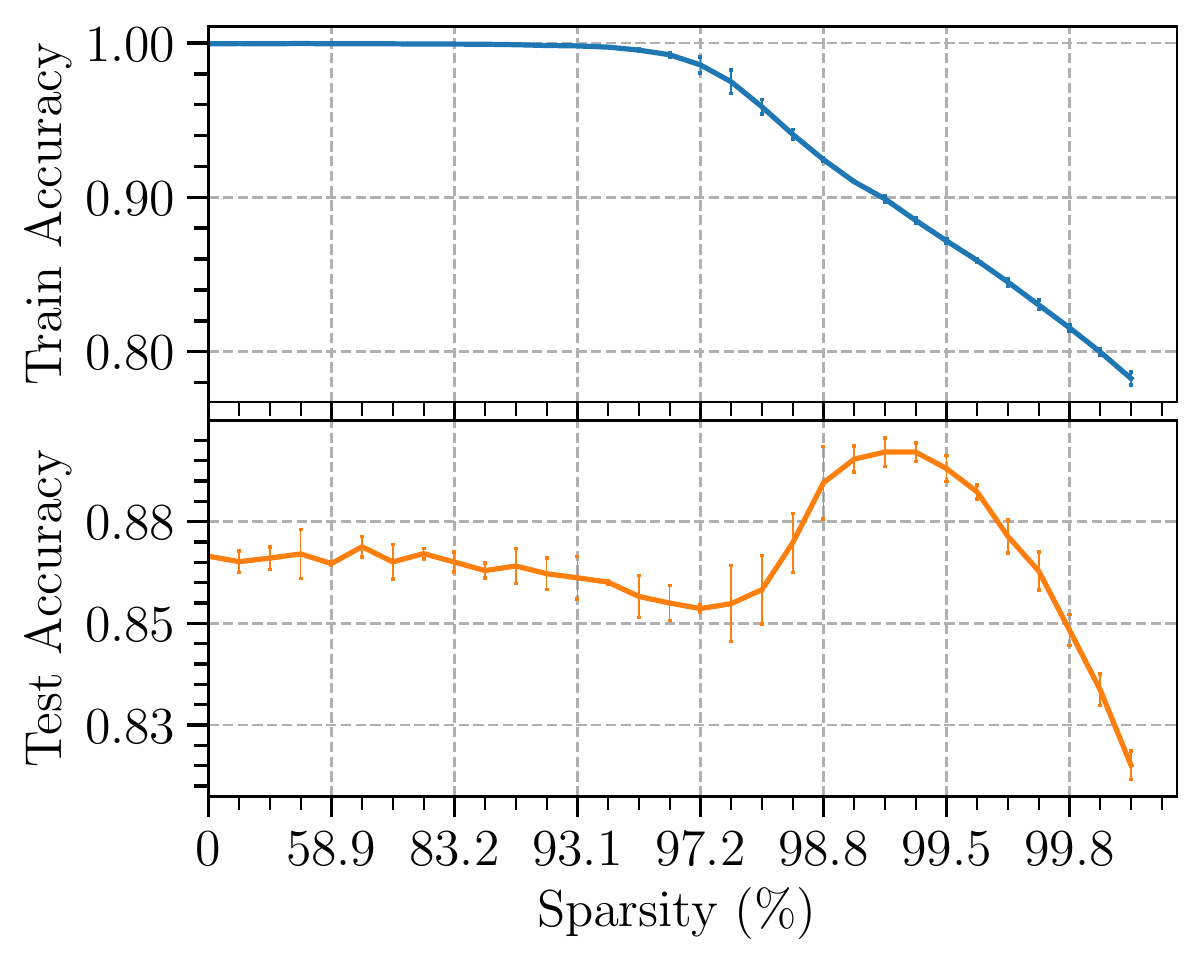}
    \includegraphics[width=0.3\textwidth]{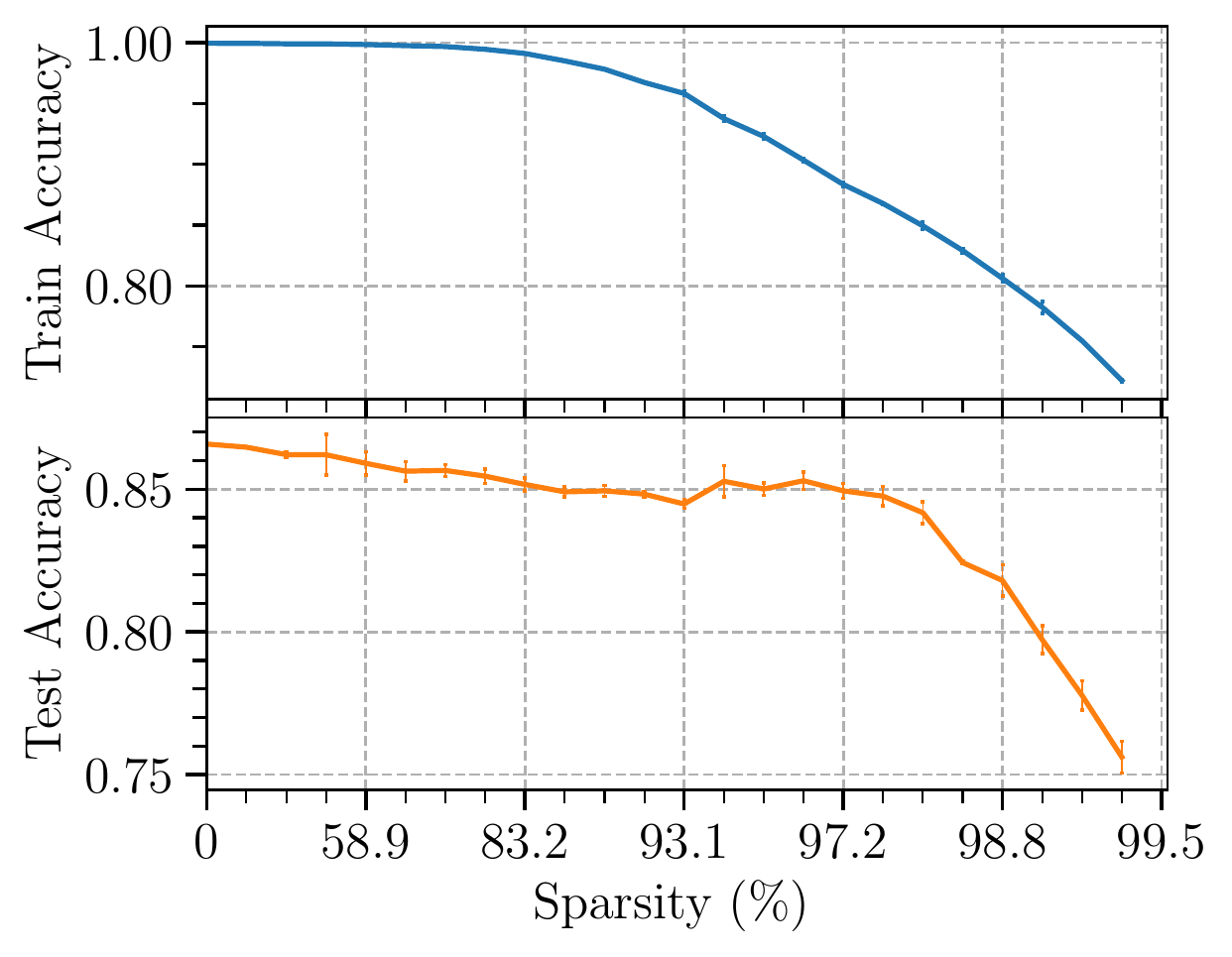}
\end{center}
\vspace{-0.3cm}
\caption{Sparse Double descend phenomenon with different pruning strategies for ResNet-18 on CIFAR-10  with asymmetric label noise, $\epsilon=20\%$. \textbf{Left}: Magnitude-based pruning. \textbf{Middle}: Gradient-based pruning. \textbf{Right}: Random pruning.}
\label{fig:sparsedd-cifar10-asy-0.2}
\end{figure}

\begin{figure}[H]
\vspace{-0.3cm}
\setlength{\abovecaptionskip}{0pt} 
\setlength{\belowcaptionskip}{0pt} 
\begin{center}
    \includegraphics[width=0.3\textwidth]{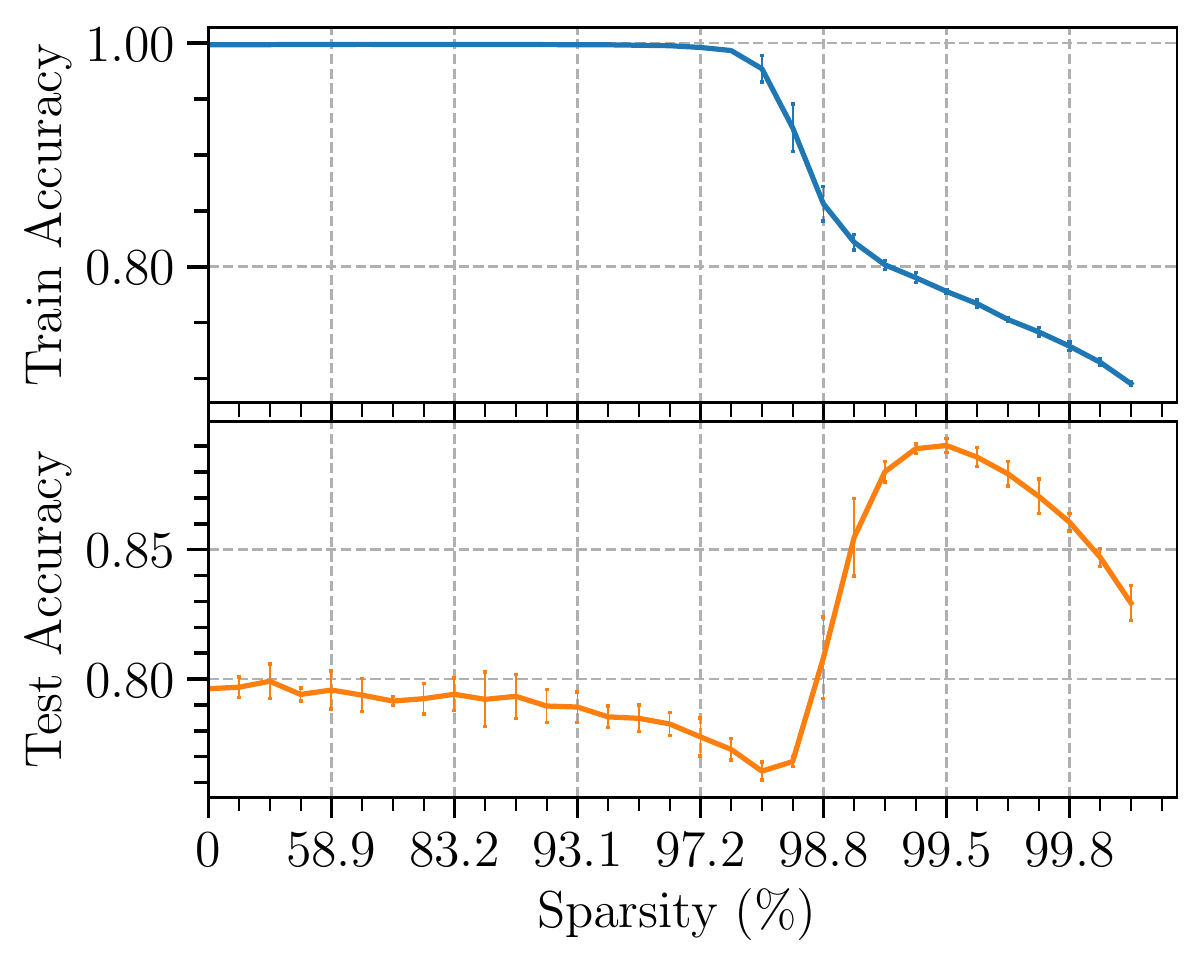}
    \includegraphics[width=0.3\textwidth]{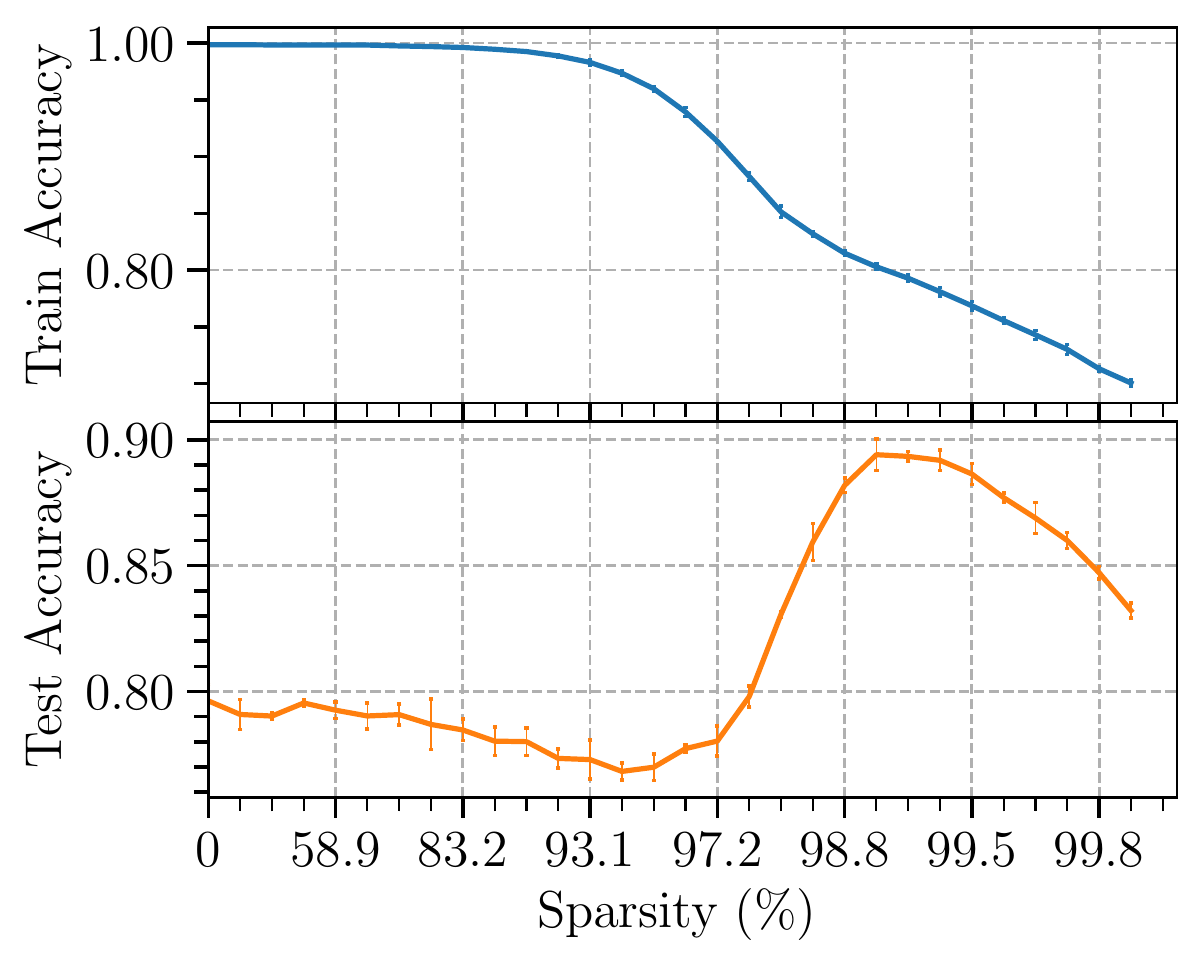}
    \includegraphics[width=0.3\textwidth]{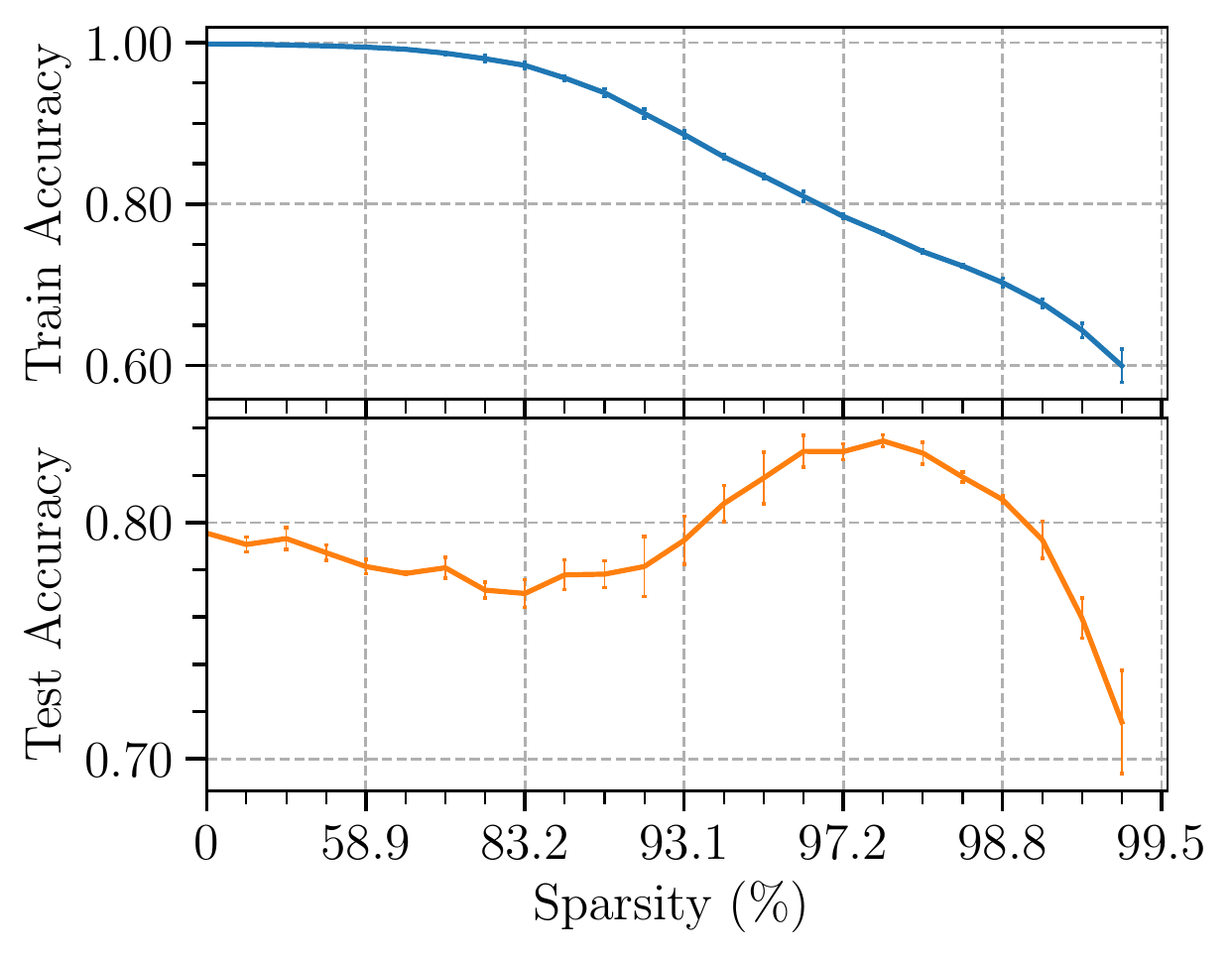}
\end{center}
\vspace{-0.3cm}
\caption{Sparse Double descend phenomenon with different pruning strategies for ResNet-18 on CIFAR-10  with pairflip label noise, $\epsilon=20\%$. \textbf{Left}: Magnitude-based pruning. \textbf{Middle}: Gradient-based pruning. \textbf{Right}: Random pruning.}
\label{fig:sparsedd-cifar10-pairflip-0.2}
\end{figure}

\begin{figure}[H]
\vspace{-0.3cm}
\setlength{\abovecaptionskip}{0pt} 
\setlength{\belowcaptionskip}{0pt} 
\begin{center}
    \includegraphics[width=0.3\textwidth]{img/cifar100/0.2_acc_magnitude.pdf}
    \includegraphics[width=0.3\textwidth]{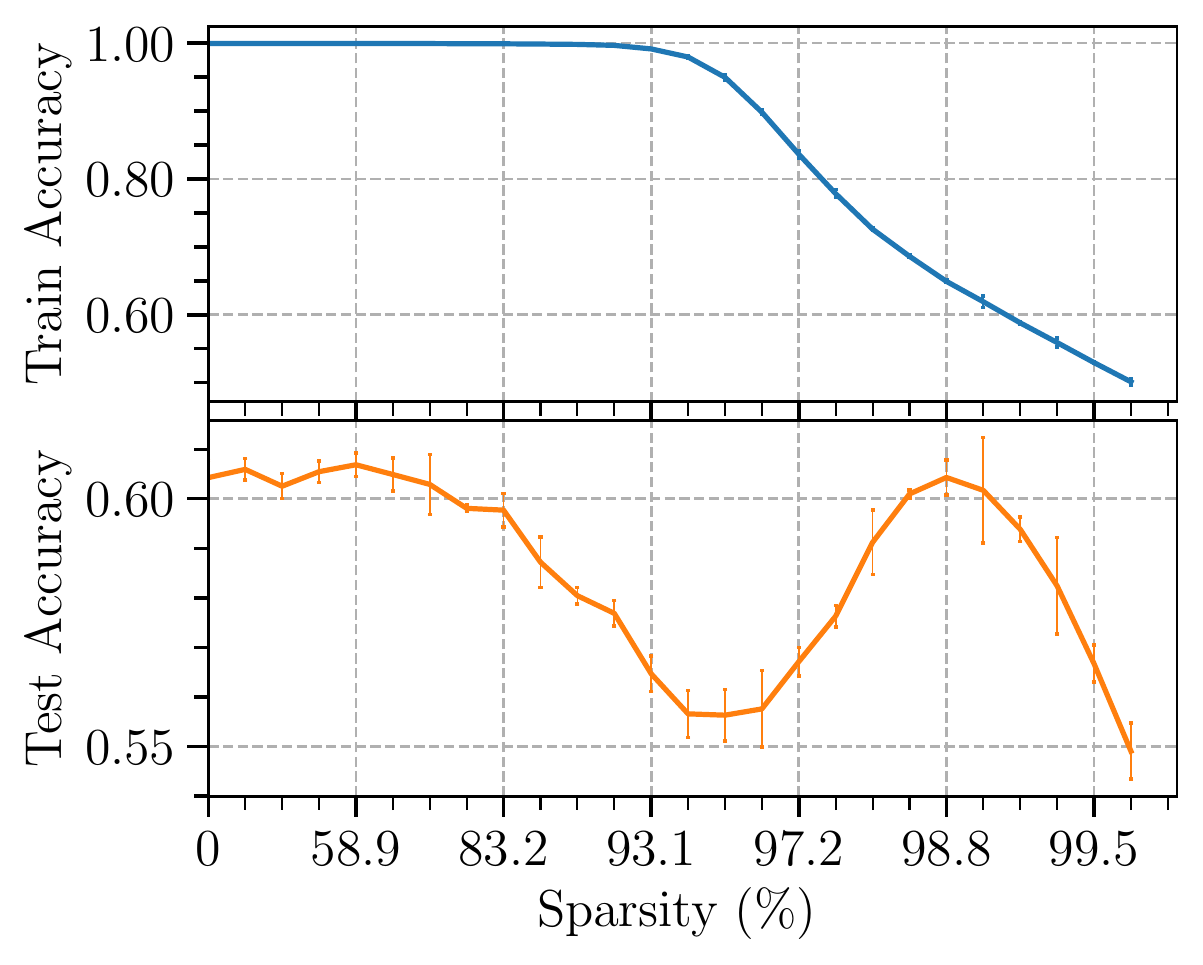} \includegraphics[width=0.3\textwidth]{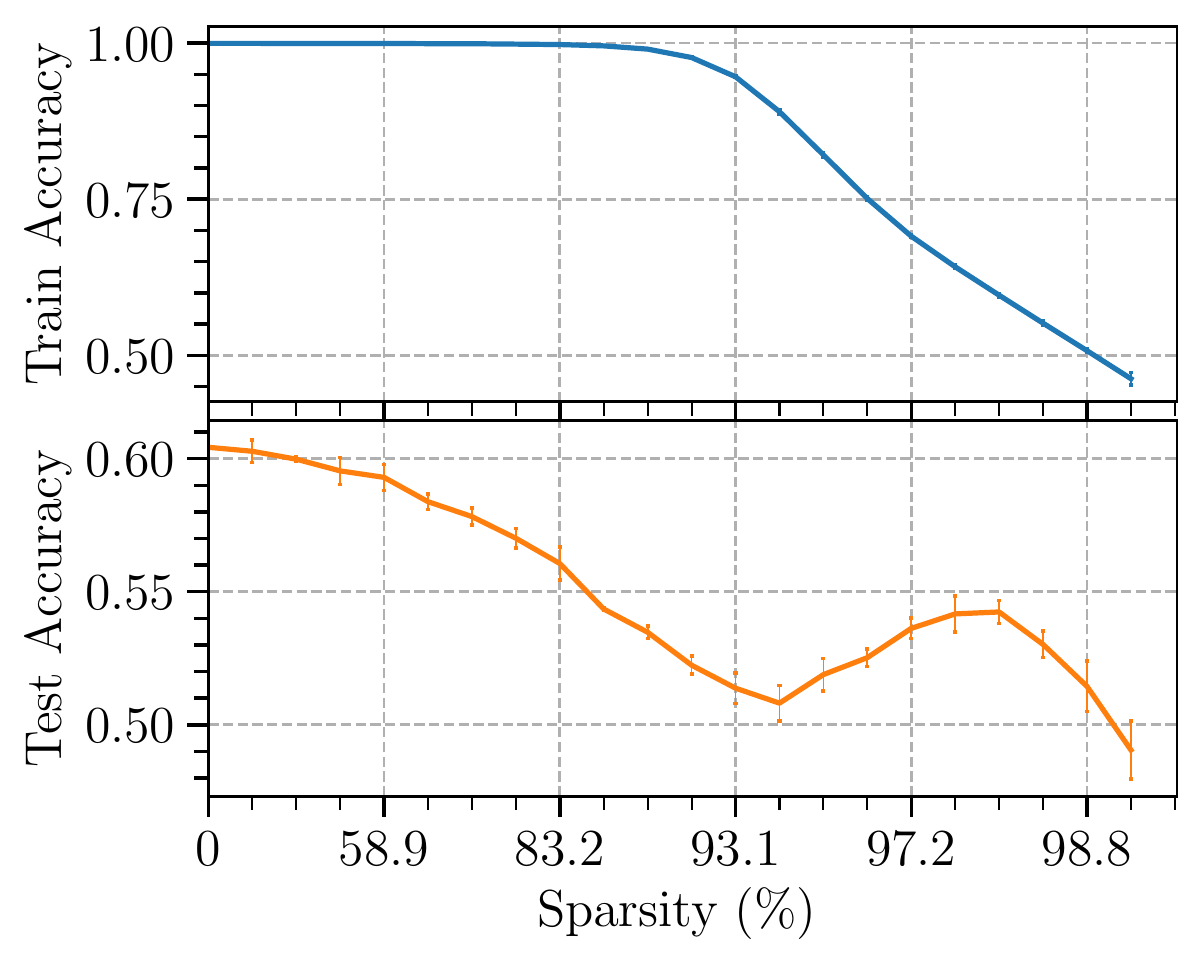}
\end{center}
\vspace{-0.3cm}
\caption{Sparse Double descend phenomenon with different pruning strategies for ResNet-18 on CIFAR-100  with symmetric label noise, $\epsilon=20\%$. \textbf{Left}: Magnitude-based pruning. \textbf{Middle}: Gradient-based pruning. \textbf{Right}: Random pruning.}
\label{fig:sparsedd-cifar100-pruning-strategy-0.2}
\end{figure}

\begin{figure}[H]
\vspace{-0.3cm}
\setlength{\abovecaptionskip}{0pt} 
\setlength{\belowcaptionskip}{0pt} 
\begin{center}
    \includegraphics[width=0.3\textwidth]{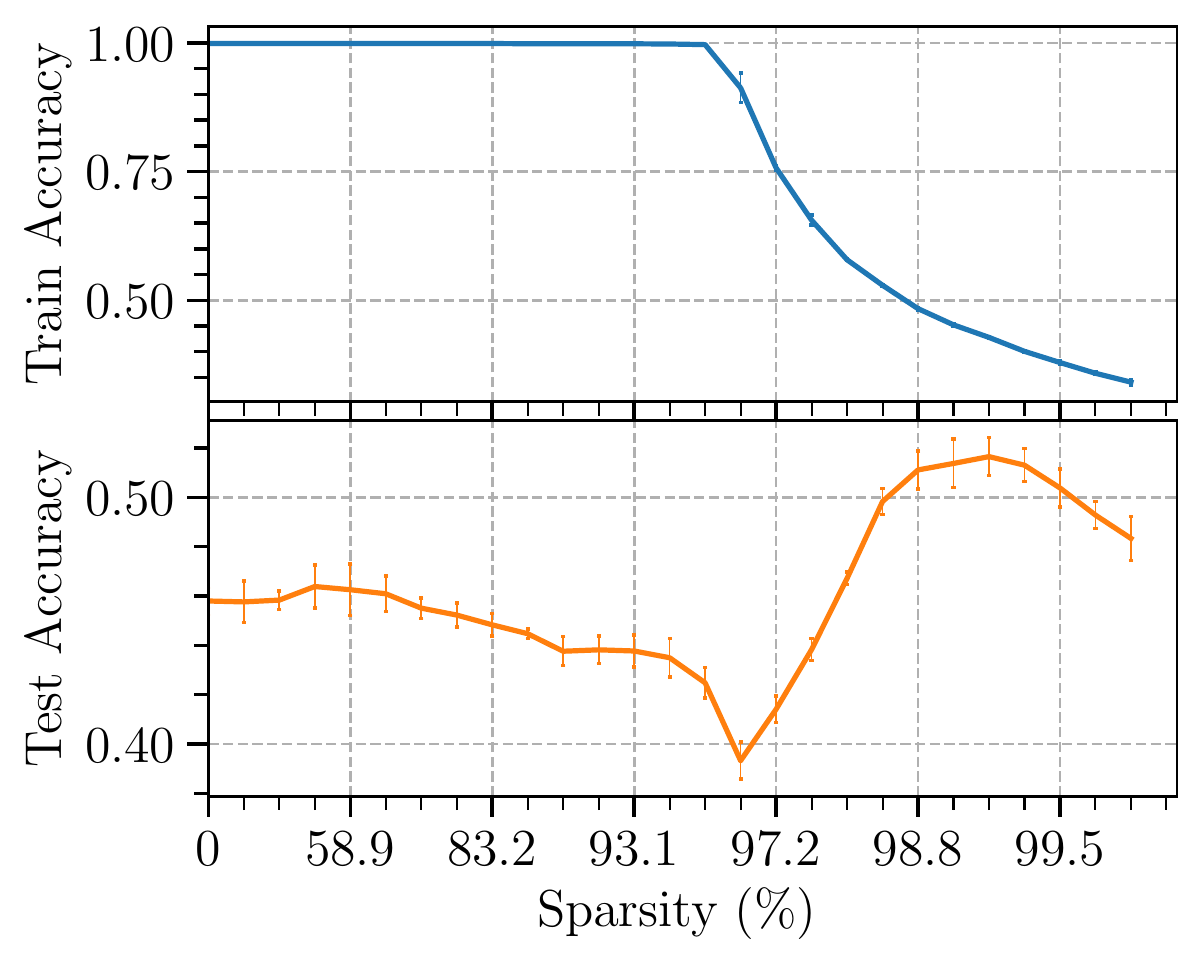}
    \includegraphics[width=0.3\textwidth]{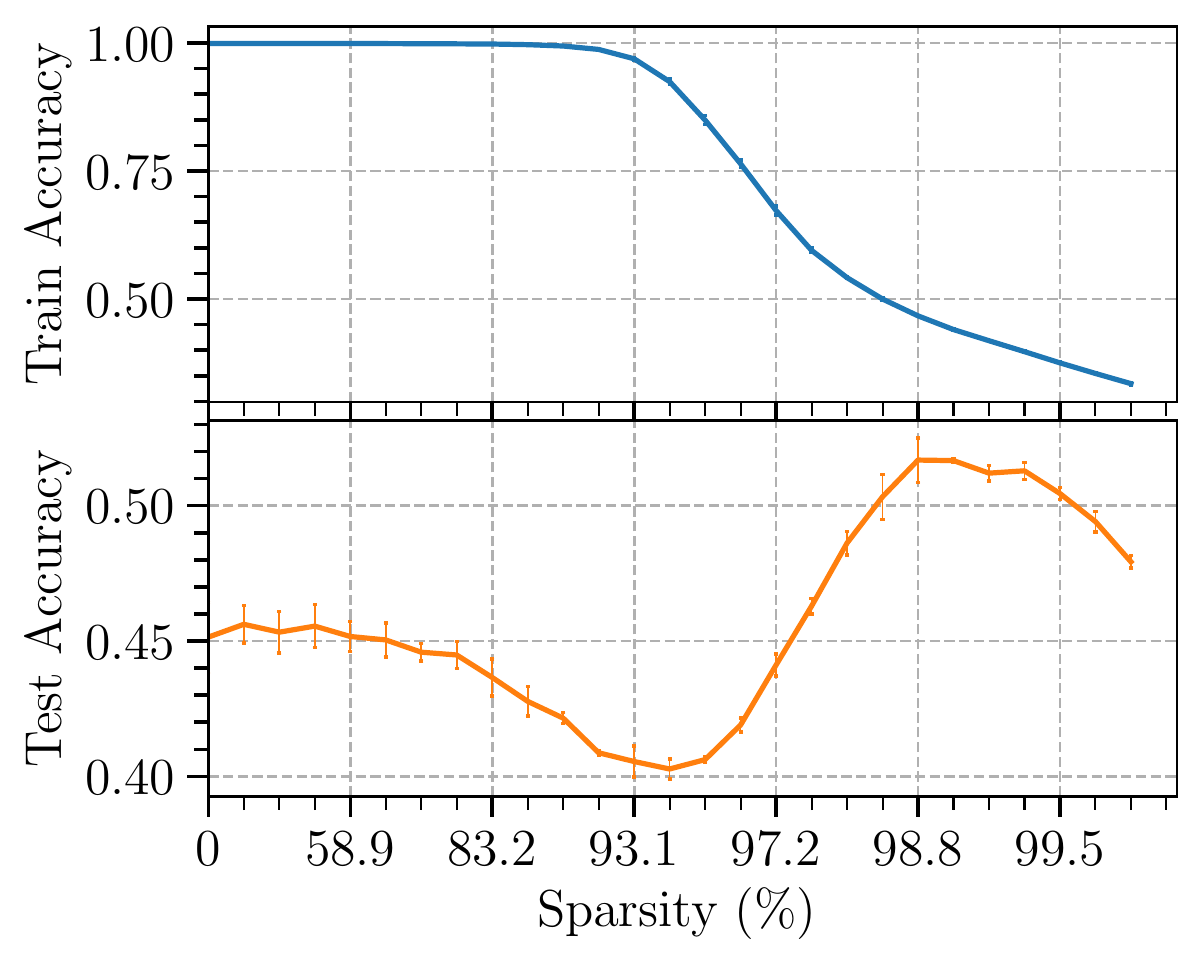} \includegraphics[width=0.3\textwidth]{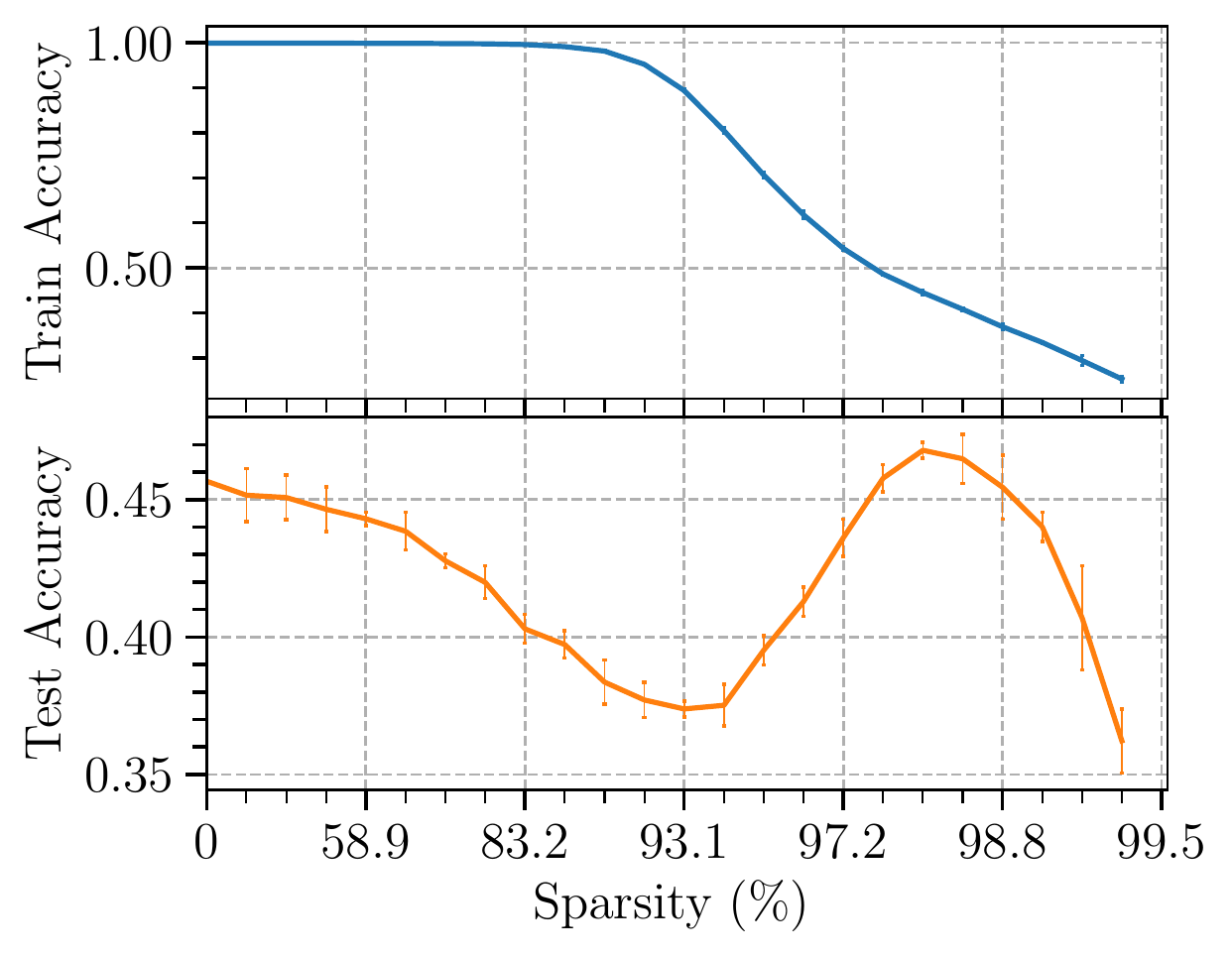}
\end{center}
\vspace{-0.3cm}
\caption{Sparse Double descend phenomenon with different pruning strategies for ResNet-18 on CIFAR-100  with symmetric label noise, $\epsilon=40\%$. \textbf{Left}: Magnitude-based pruning. \textbf{Middle}: Gradient-based pruning. \textbf{Right}: Random pruning.}
\label{fig:sparsedd-cifar100-pruning-strategy-0.4}
\end{figure}

\begin{figure}[H]
\vspace{-0.3cm}
\setlength{\abovecaptionskip}{0pt} 
\setlength{\belowcaptionskip}{0pt} 
\begin{center}
    \includegraphics[width=0.3\textwidth]{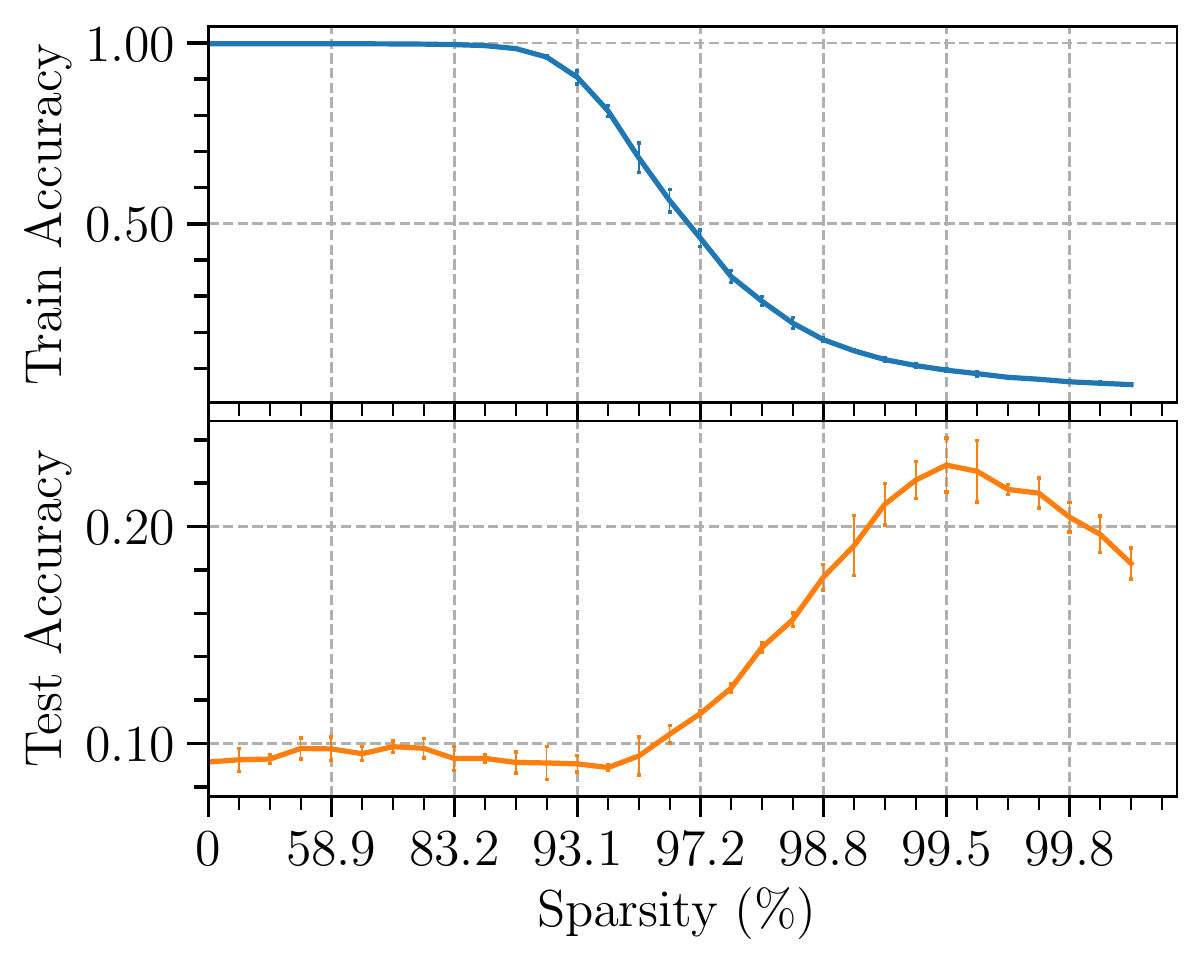}
    \includegraphics[width=0.3\textwidth]{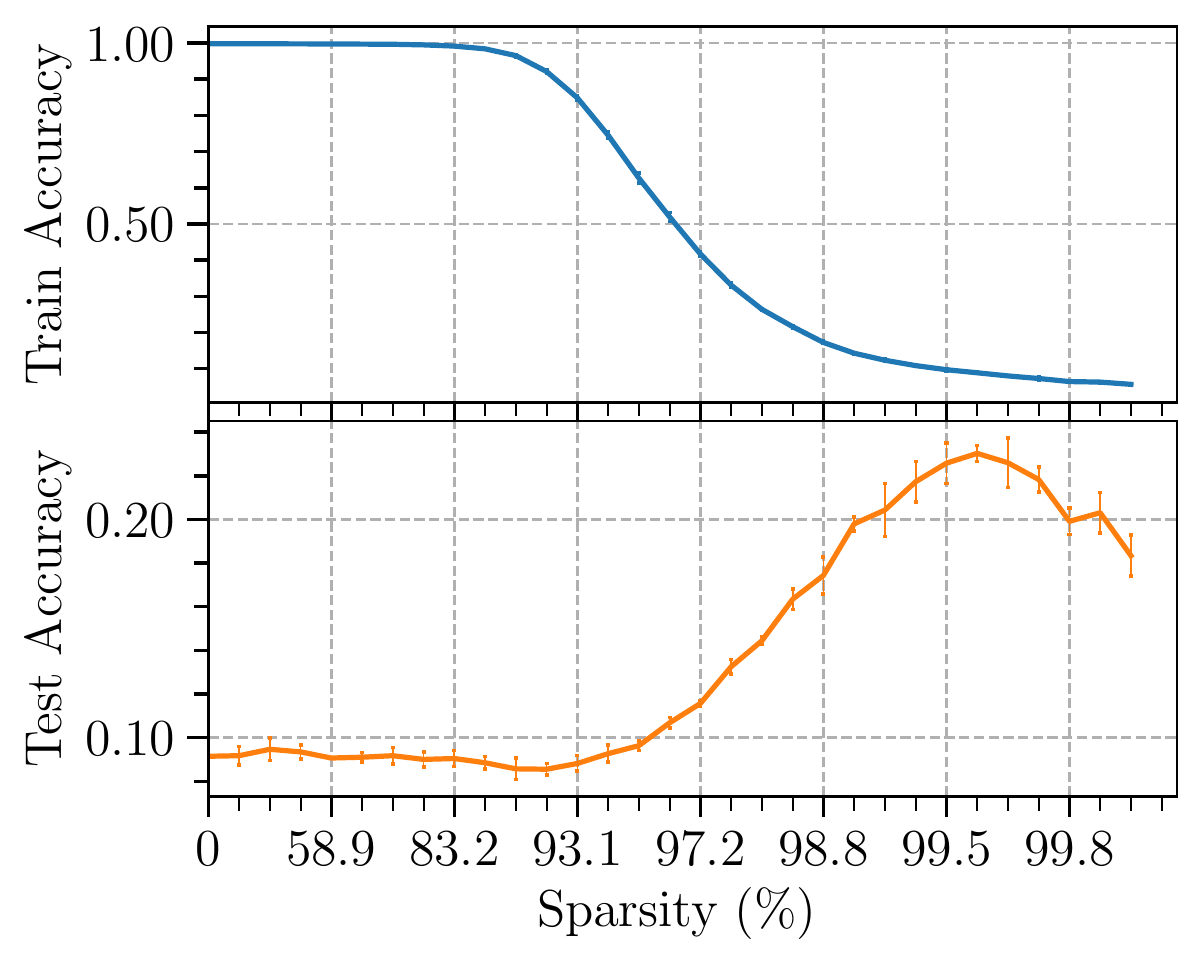} \includegraphics[width=0.3\textwidth]{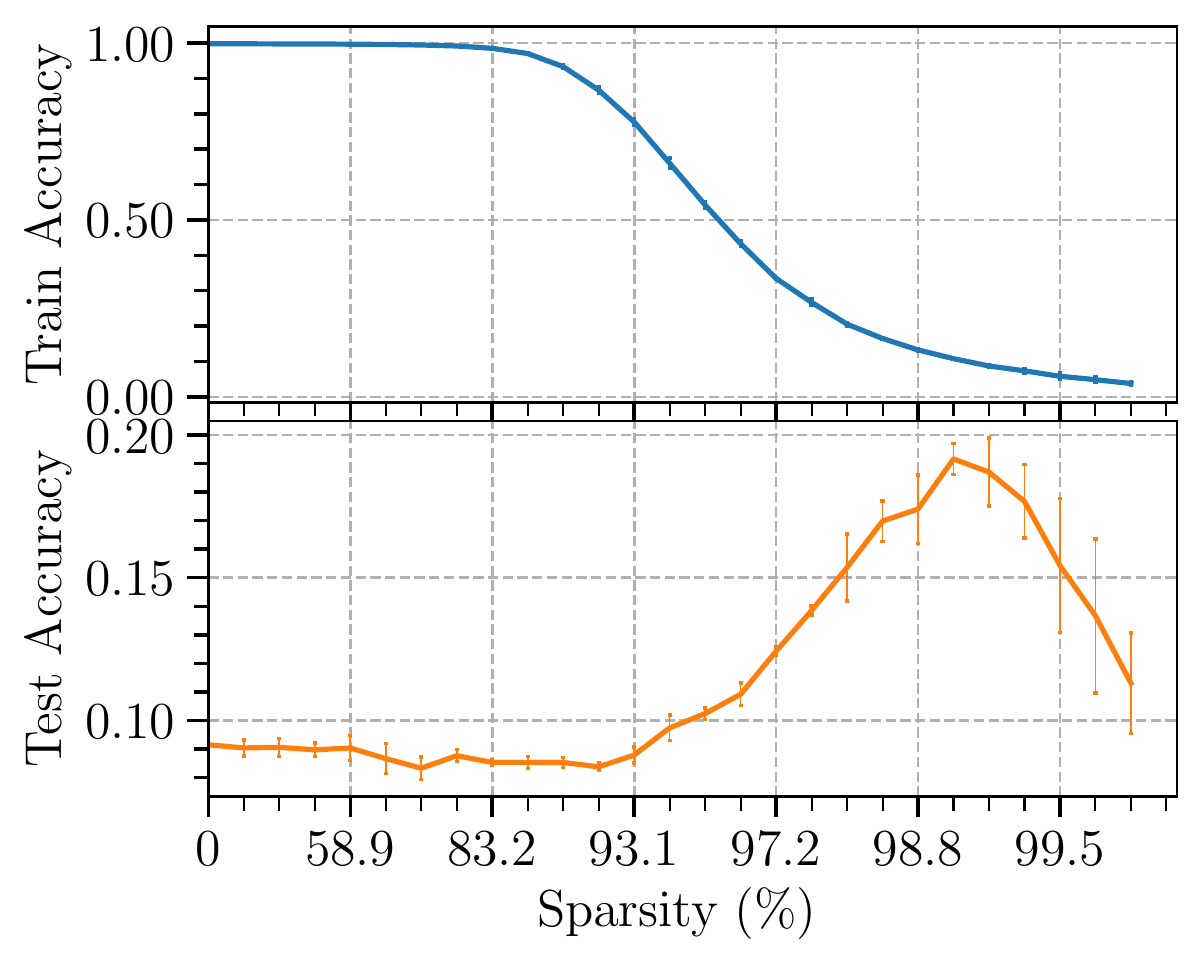}
\end{center}
\vspace{-0.3cm}
\caption{Sparse Double descend phenomenon with different pruning strategies for ResNet-18 on CIFAR-100  with symmetric label noise, $\epsilon=80\%$. \textbf{Left}: Magnitude-based pruning. \textbf{Middle}: Gradient-based pruning. \textbf{Right}: Random pruning.}
\label{fig:sparsedd-cifar100-pruning-strategy-0.8}
\end{figure}

\begin{figure}[H]
\vspace{-0.3cm}
\setlength{\abovecaptionskip}{0pt} 
\setlength{\belowcaptionskip}{0pt} 
\begin{center}
    \includegraphics[width=0.3\textwidth]{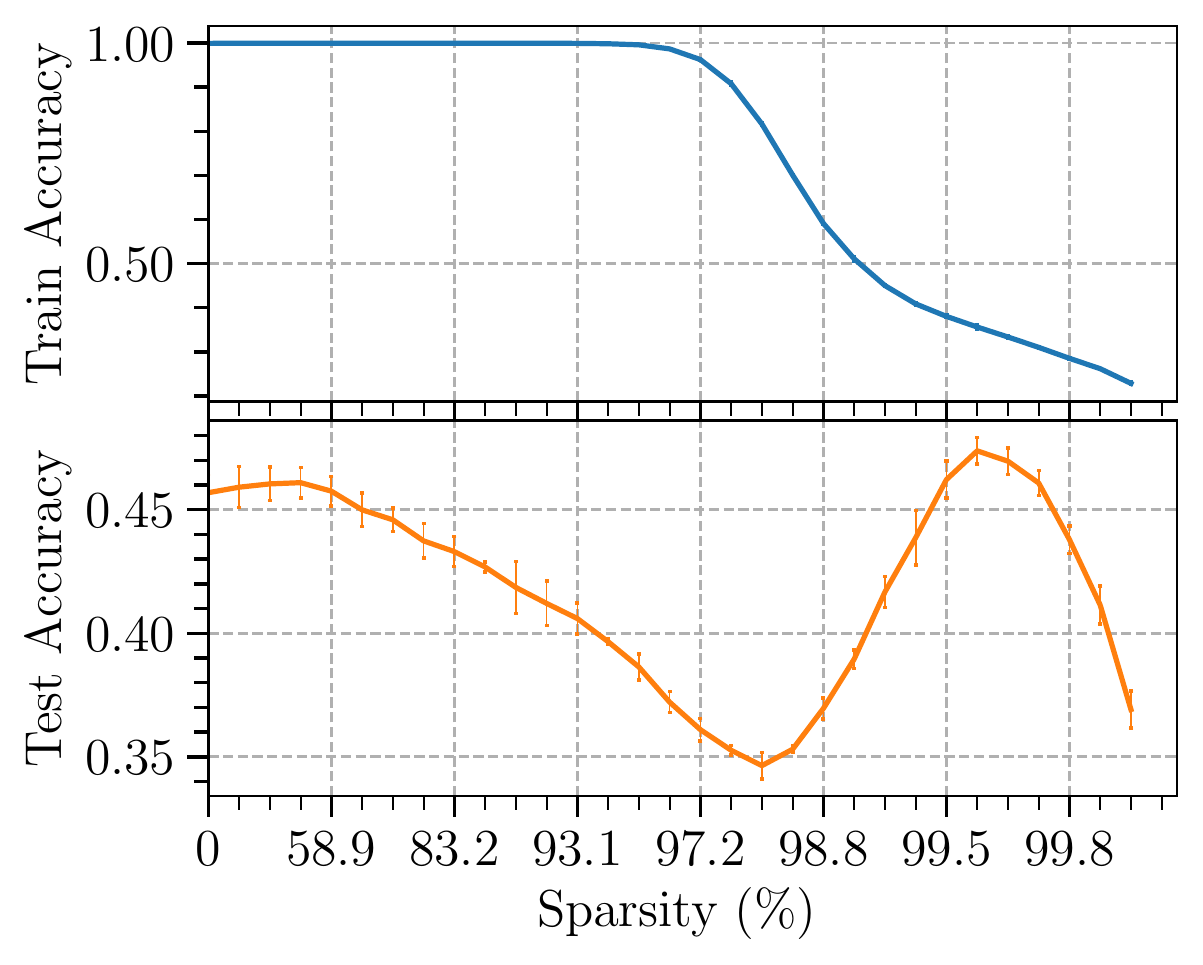}
    \includegraphics[width=0.3\textwidth]{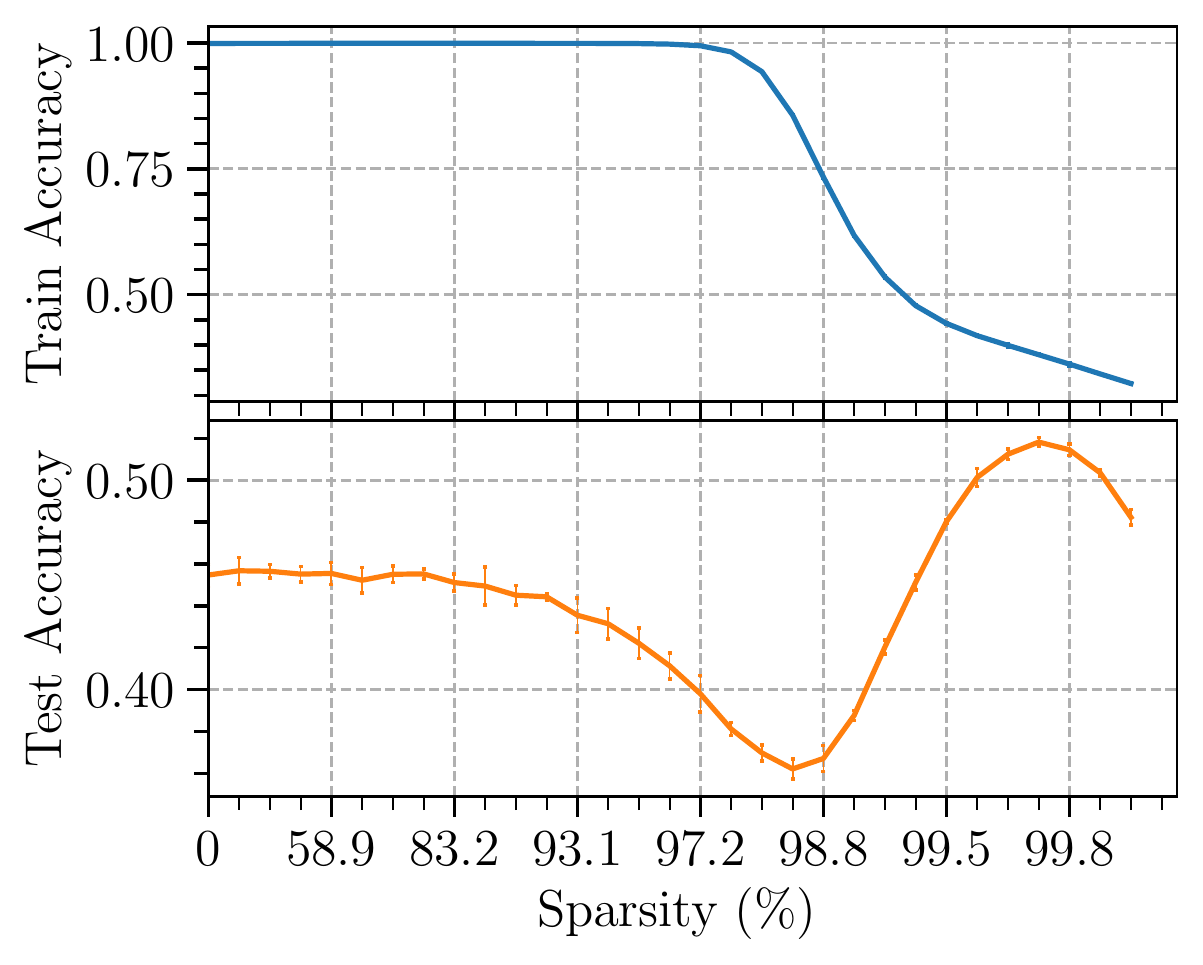} \includegraphics[width=0.3\textwidth]{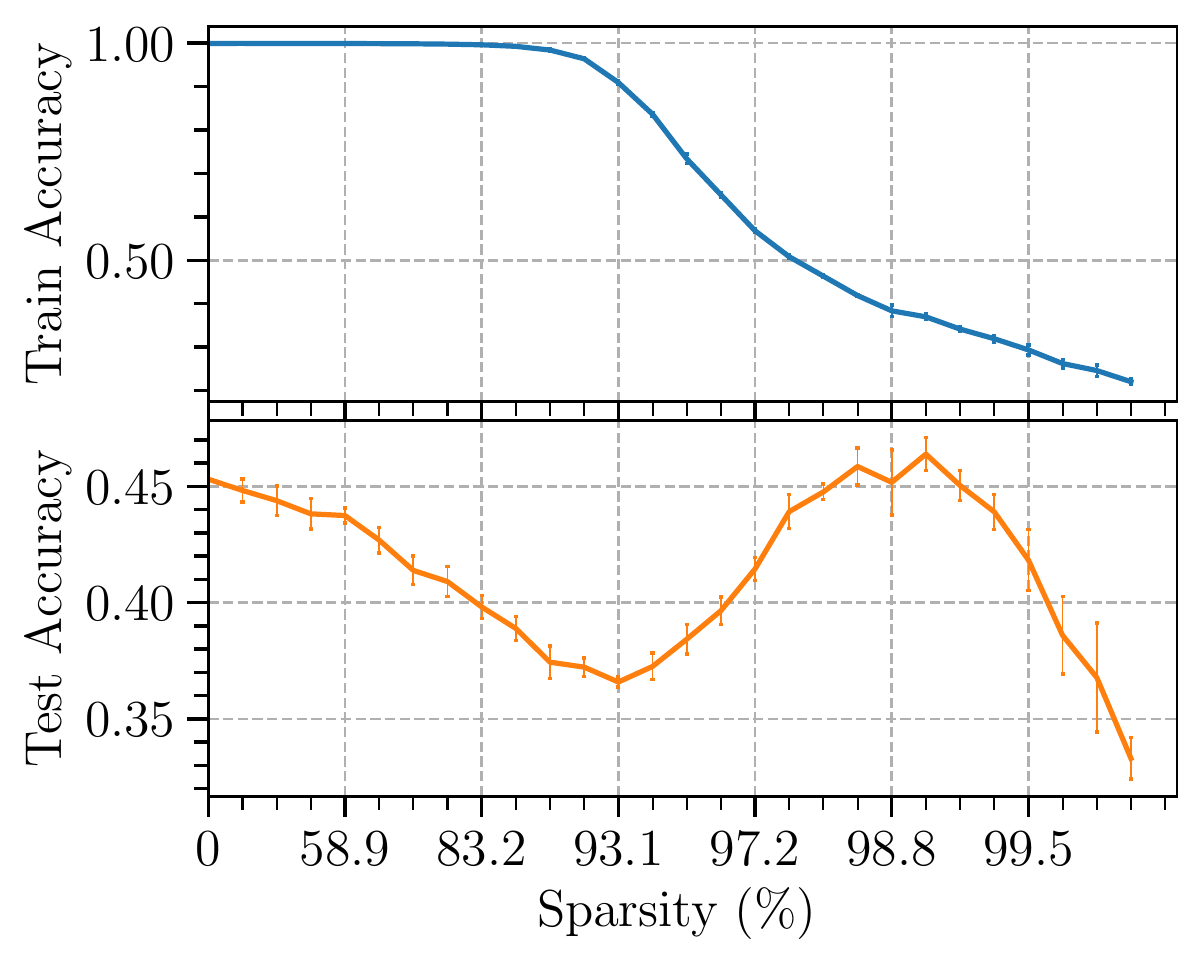}
\end{center}
\vspace{-0.3cm}
\caption{Sparse Double descend phenomenon with different retraining methods for ResNet-18 on CIFFAR-100 with symmetric label noise, $\epsilon=40\%$. \textbf{Left}: Finetuning. \textbf{Middle}: Learning Rate Rewinding. \textbf{Right}: Scratch retraining.}
\label{fig:sparsedd-cifar100-retrain-0.4}
\end{figure}

\begin{figure}[H]
\vspace{-0.3cm}
\setlength{\abovecaptionskip}{0pt} 
\setlength{\belowcaptionskip}{0pt} 
\begin{center}
    \includegraphics[width=0.3\textwidth]{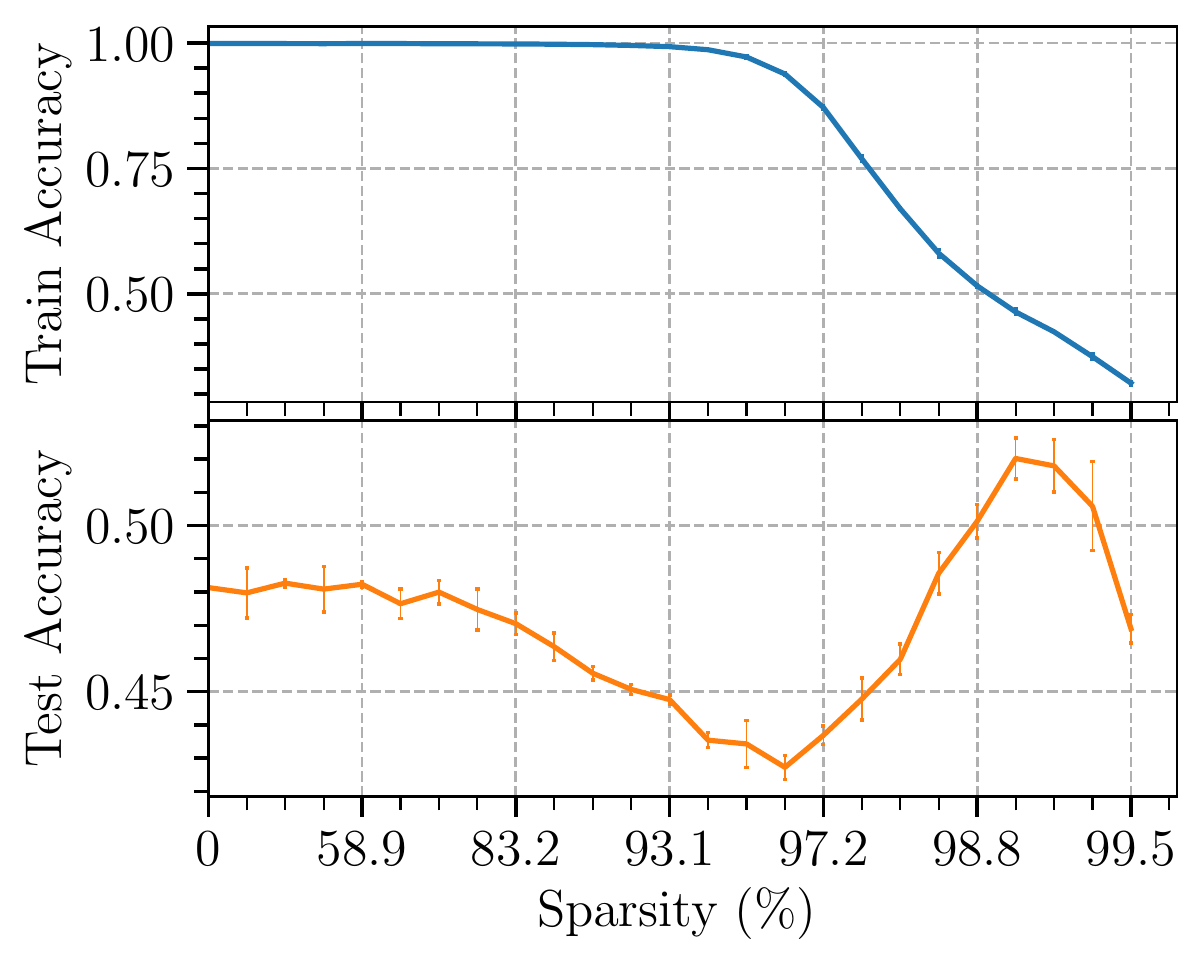}
    \includegraphics[width=0.3\textwidth]{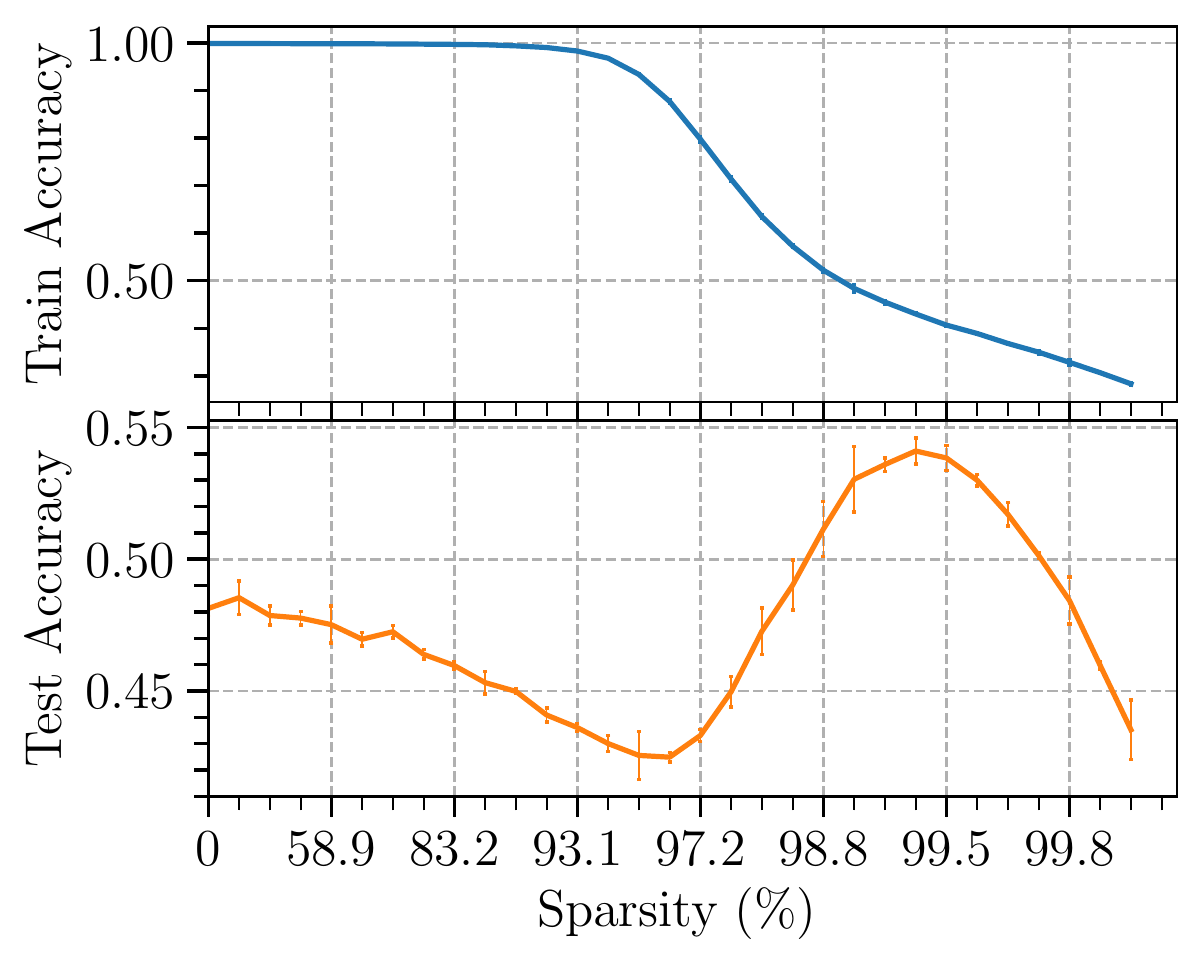}
    \includegraphics[width=0.3\textwidth]{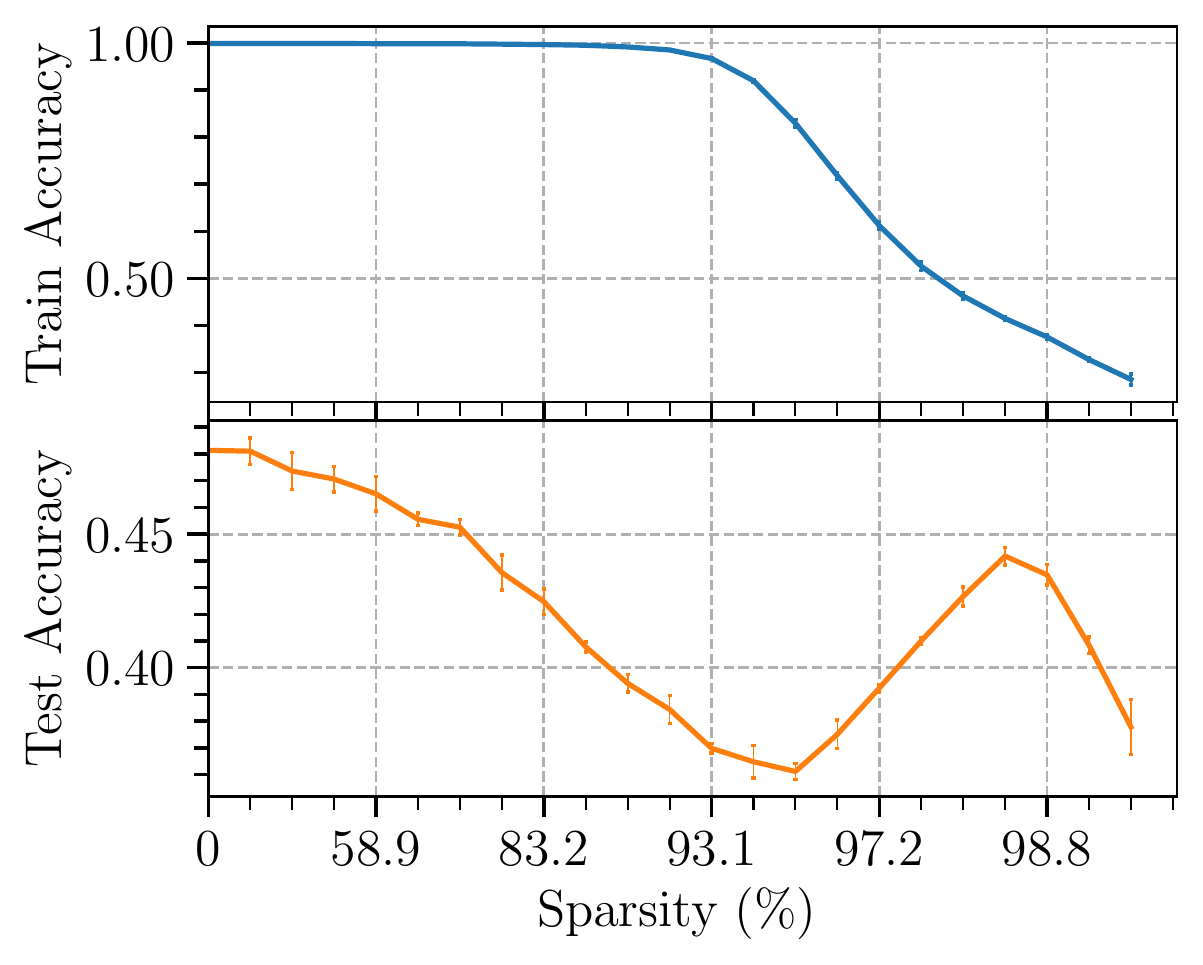}
\end{center}
\vspace{-0.3cm}
\caption{Sparse Double descend phenomenon with different pruning strategies for VGG-16 on CIFAR-100  with symmetric label noise, $\epsilon=40\%$. \textbf{Left}: Magnitude-based pruning. \textbf{Middle}: Gradient-based pruning. \textbf{Right}: Random pruning.}
\label{fig:sparsedd-cifar100-pruning-strategy-0.4-vgg}
\end{figure}

\begin{figure}[H]
\vspace{-0.3cm}
\setlength{\abovecaptionskip}{0pt} 
\setlength{\belowcaptionskip}{0pt} 
\begin{center}
    \includegraphics[width=0.3\textwidth]{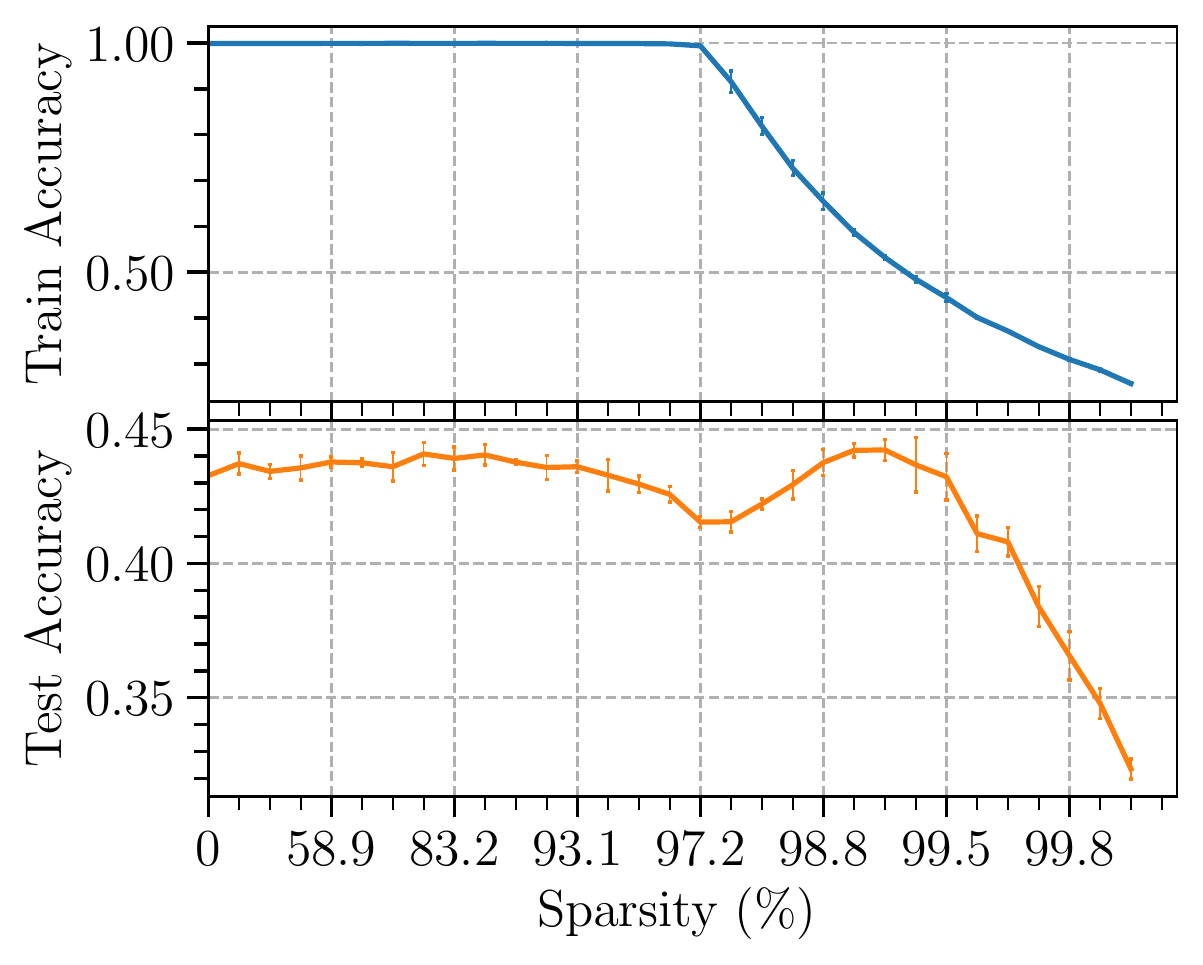}
    \includegraphics[width=0.3\textwidth]{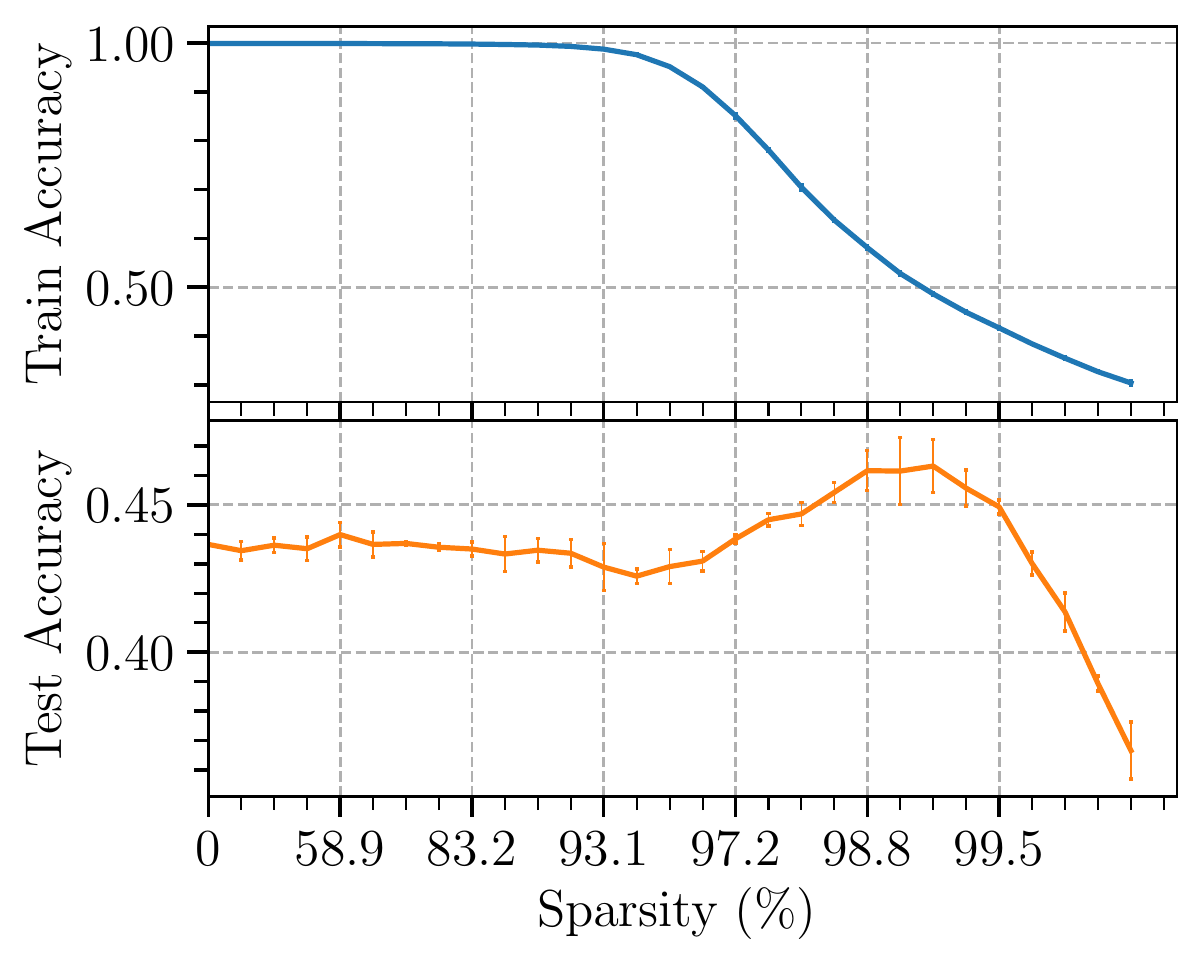}
    \includegraphics[width=0.3\textwidth]{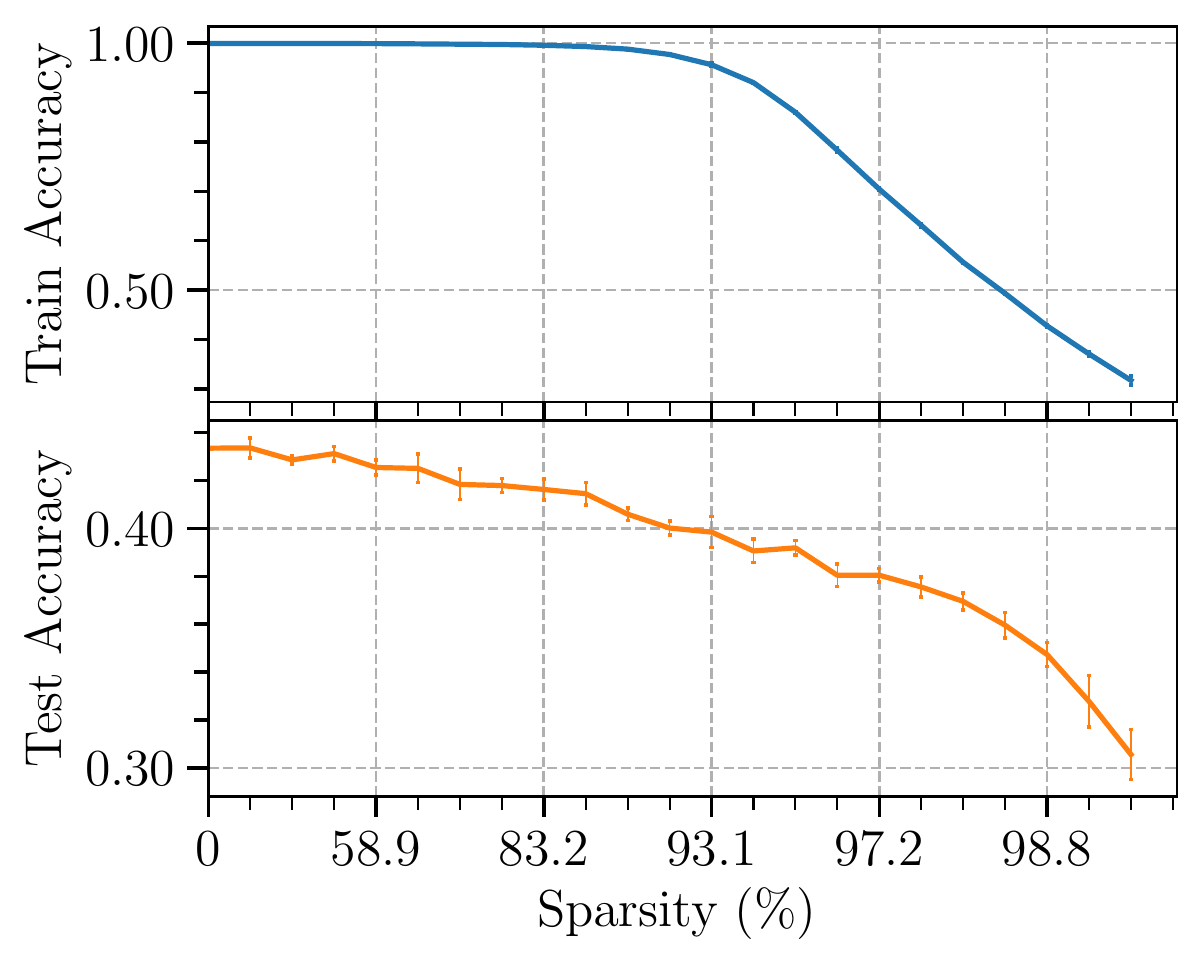}
\end{center}
\vspace{-0.3cm}
\caption{Sparse Double descend phenomenon with different pruning strategies for ResNet-18 on CIFAR-100  with asymmetric label noise, $\epsilon=40\%$. \textbf{Left}: Magnitude-based pruning. \textbf{Middle}: Gradient-based pruning. \textbf{Right}: Random pruning.}
\label{fig:sparsedd-cifar100-asy-0.4}
\end{figure}

\begin{figure}[H]
\vspace{-0.3cm}
\setlength{\abovecaptionskip}{0pt} 
\setlength{\belowcaptionskip}{0pt} 
\begin{center}
    \includegraphics[width=0.3\textwidth]{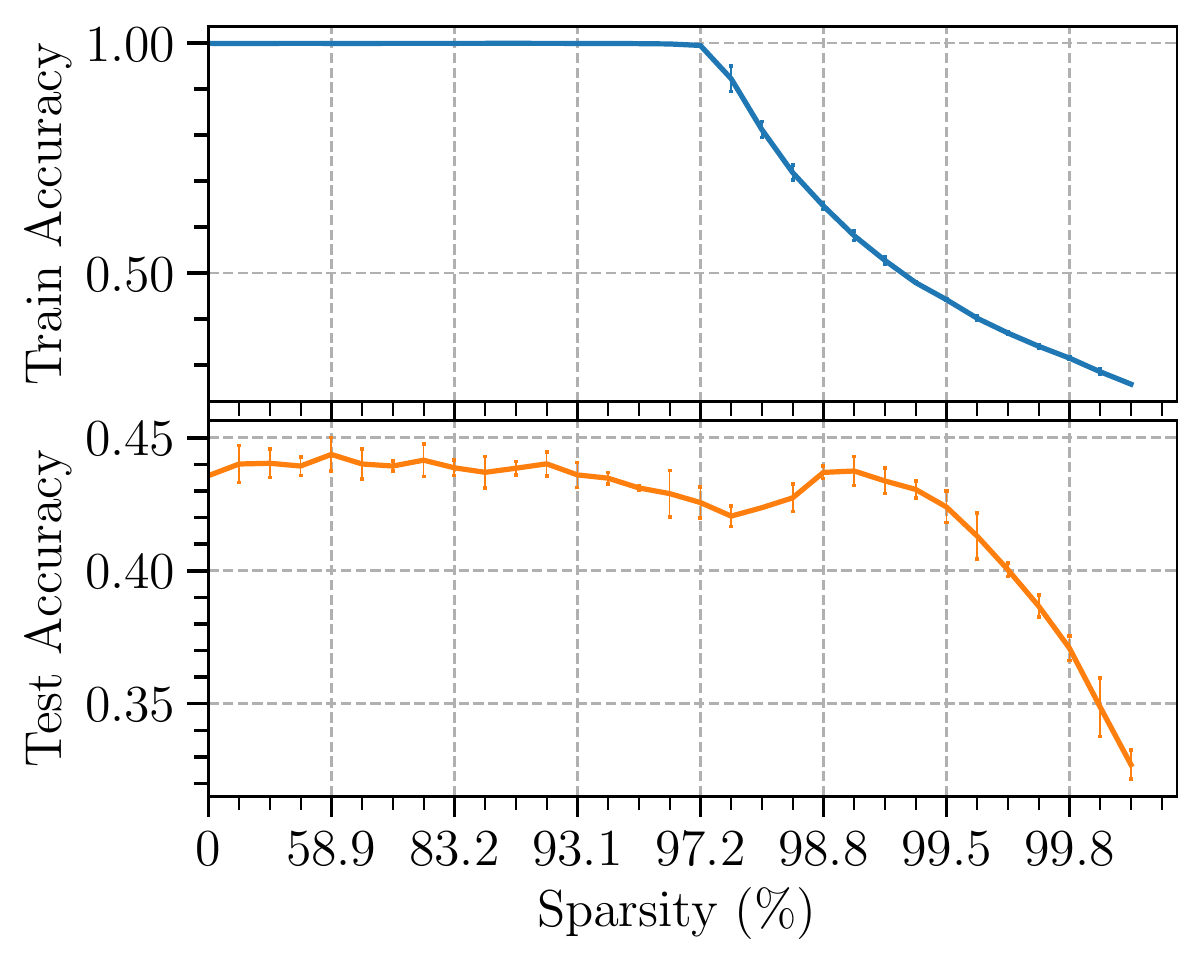}
    \includegraphics[width=0.3\textwidth]{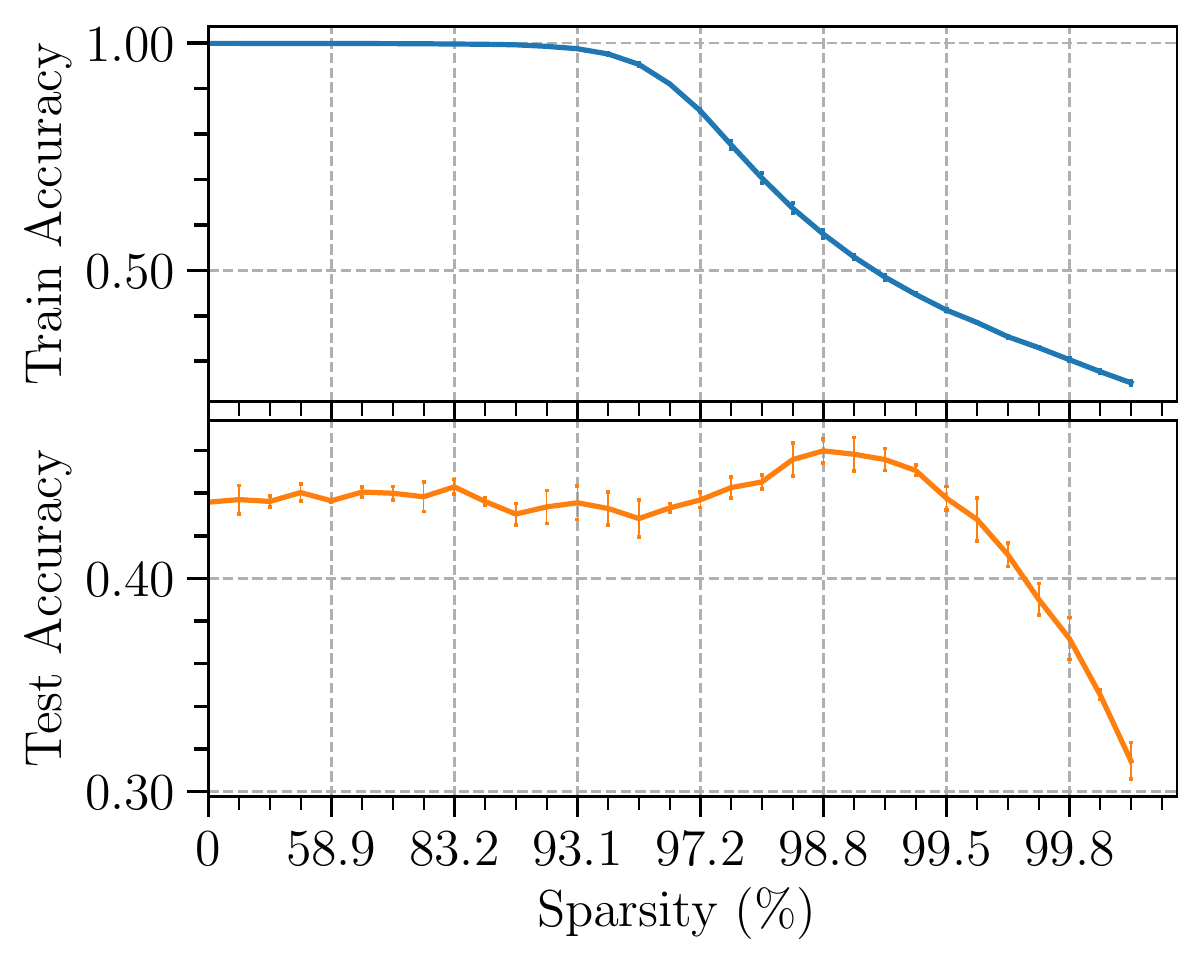}
    \includegraphics[width=0.3\textwidth]{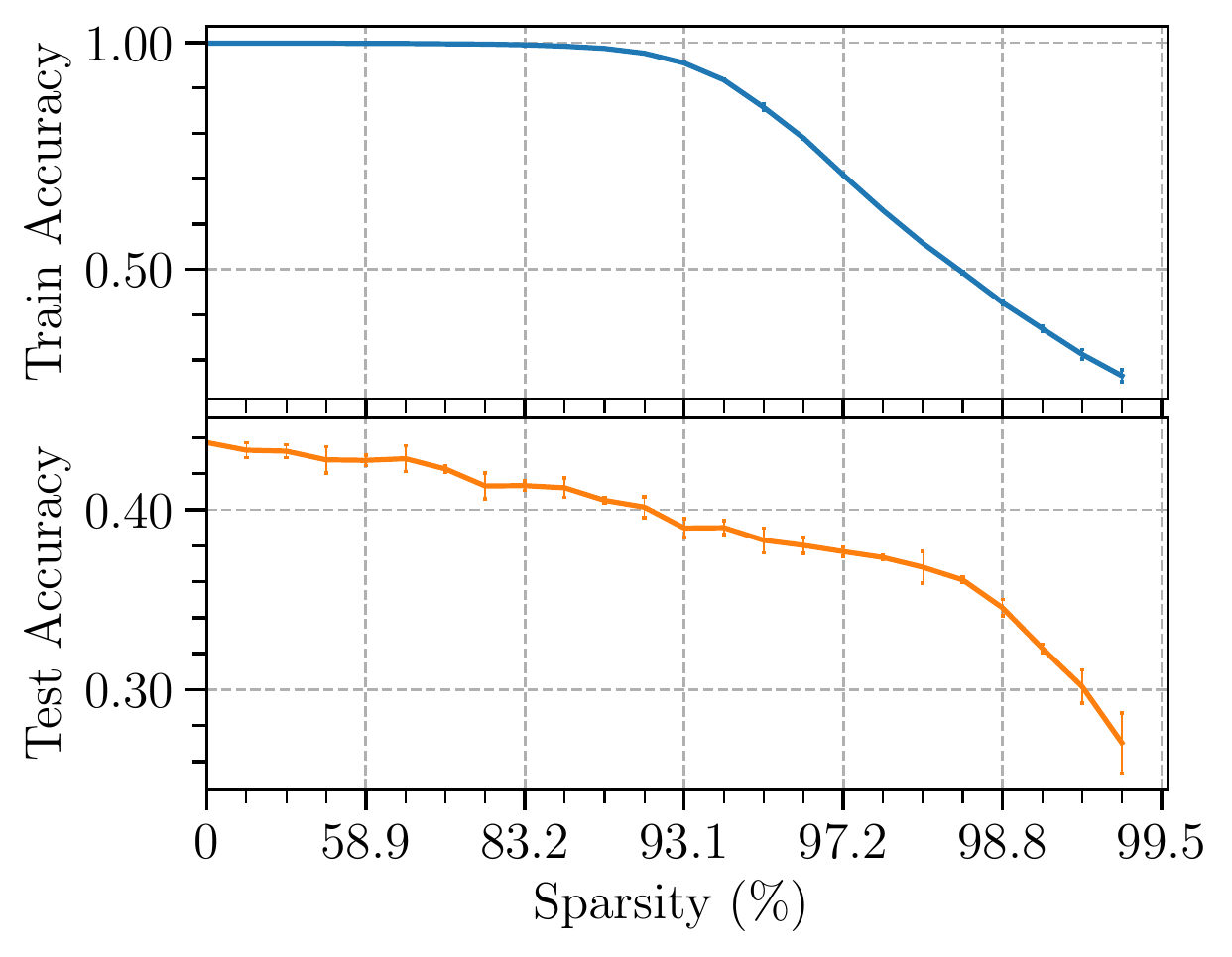}
\end{center}
\vspace{-0.3cm}
\caption{Sparse Double descend phenomenon with different pruning strategies for ResNet-18 on CIFAR-100  with pairflip label noise, $\epsilon=40\%$. \textbf{Left}: Magnitude-based pruning. \textbf{Middle}: Gradient-based pruning. \textbf{Right}: Random pruning.}
\label{fig:sparsedd-cifar100-pairflip-0.4}
\end{figure}

\begin{figure}[H]
\vspace{-0.3cm}
\setlength{\abovecaptionskip}{0pt} 
\setlength{\belowcaptionskip}{0pt} 
\begin{center}
    \includegraphics[width=0.3\textwidth]{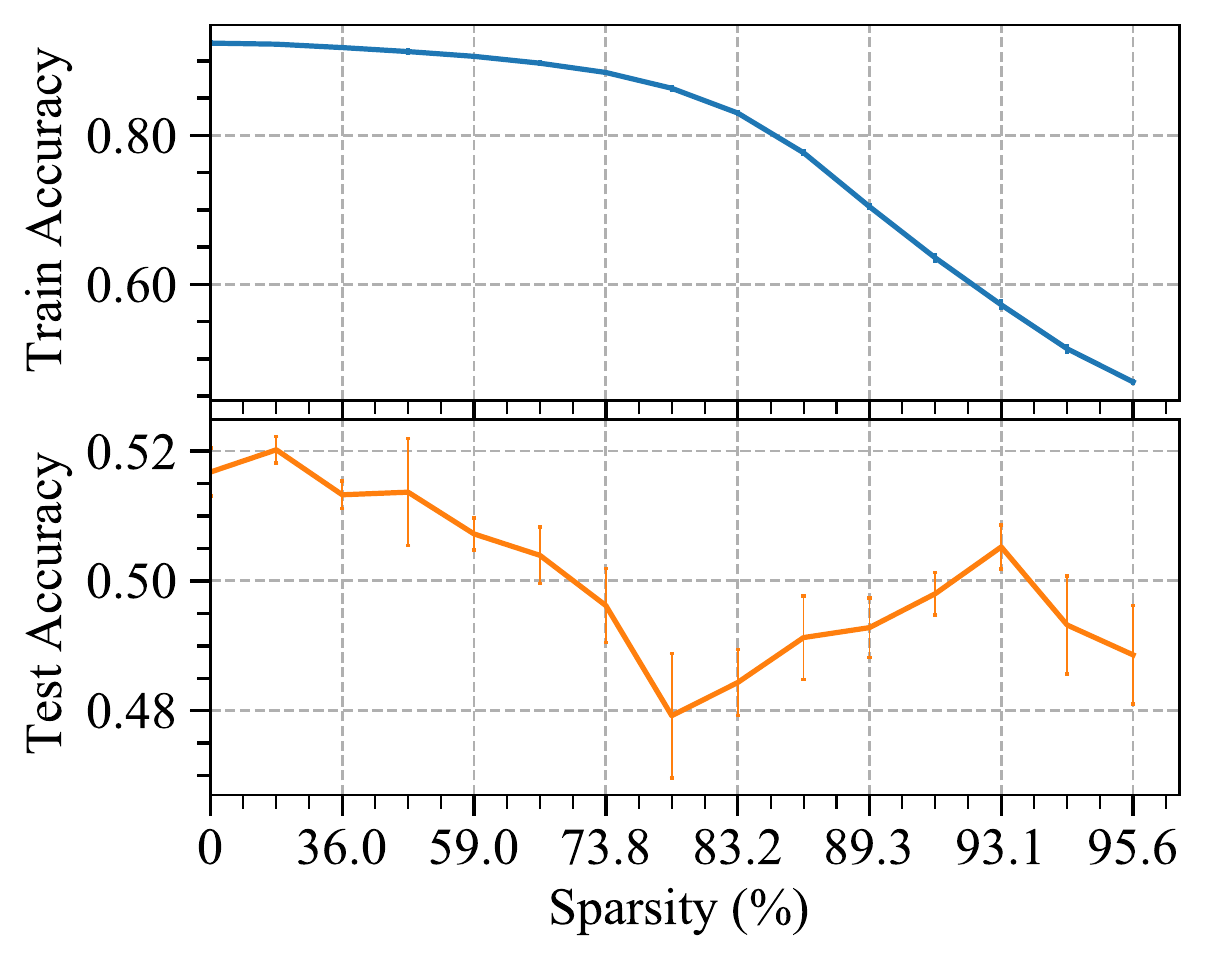}
\end{center}
\vspace{-0.3cm}
\caption{Sparse Double descend phenomenon with magnitude-based pruning for ResNet-101 on Tiny ImageNet  with symmetric label noise, $\epsilon=20\%$.}
\label{fig:sparsedd-tinyimagenet}
\end{figure}

\begin{figure}[H]
\vspace{-0.3cm}
\setlength{\abovecaptionskip}{0pt} 
\setlength{\belowcaptionskip}{0pt} 
\begin{center}
    \includegraphics[width=0.3\textwidth]{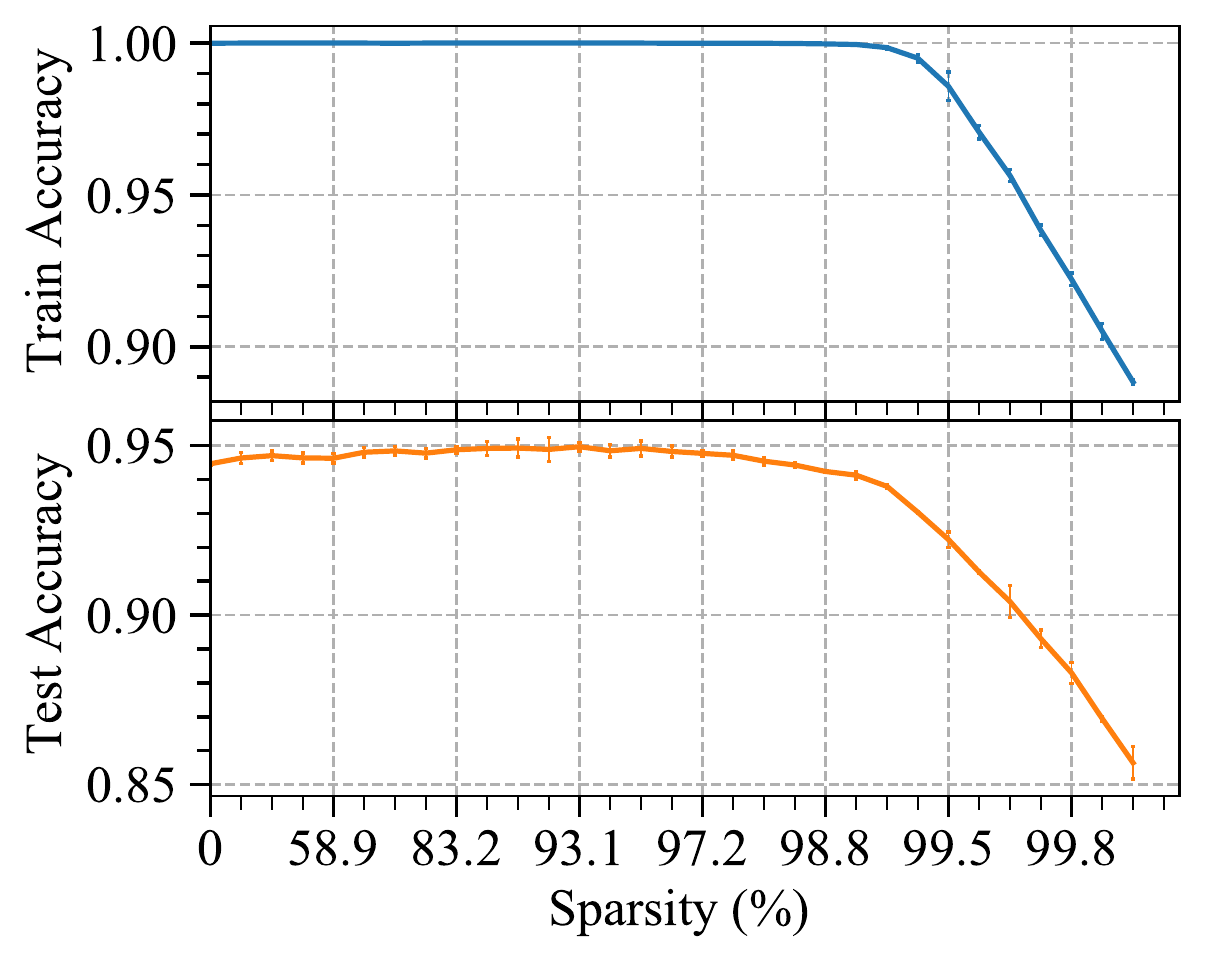}
    \includegraphics[width=0.3\textwidth]{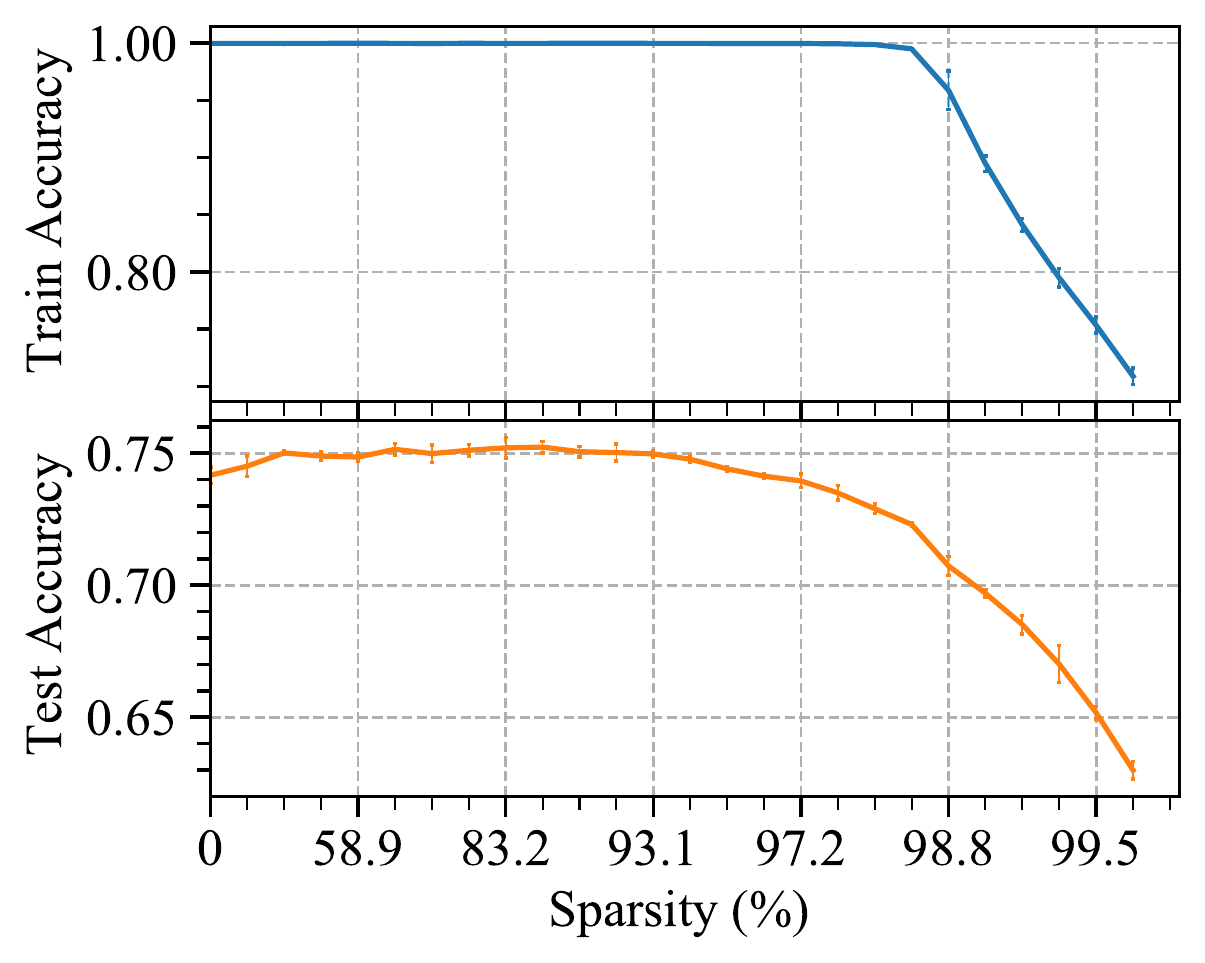}
\end{center}
\vspace{-0.3cm}
\caption{Results of magnitude-based pruning for ResNet-18 on CIFAR  with no label noise. \textbf{Left}: CIFAR-10. \textbf{Right}: CIFAR-100.}
\label{fig:sparsedd-zero-noise}
\end{figure}

\begin{figure}[H]
\center
\includegraphics[width=0.3\linewidth]{img/mnist/0.2_acc_magnitude.pdf}
\includegraphics[width=0.3\linewidth]{img/cifar10/0.2_acc_magnitude.pdf}
\includegraphics[width=0.3\linewidth]{img/cifar100/0.2_acc_magnitude.pdf}
\includegraphics[width=0.3\linewidth]{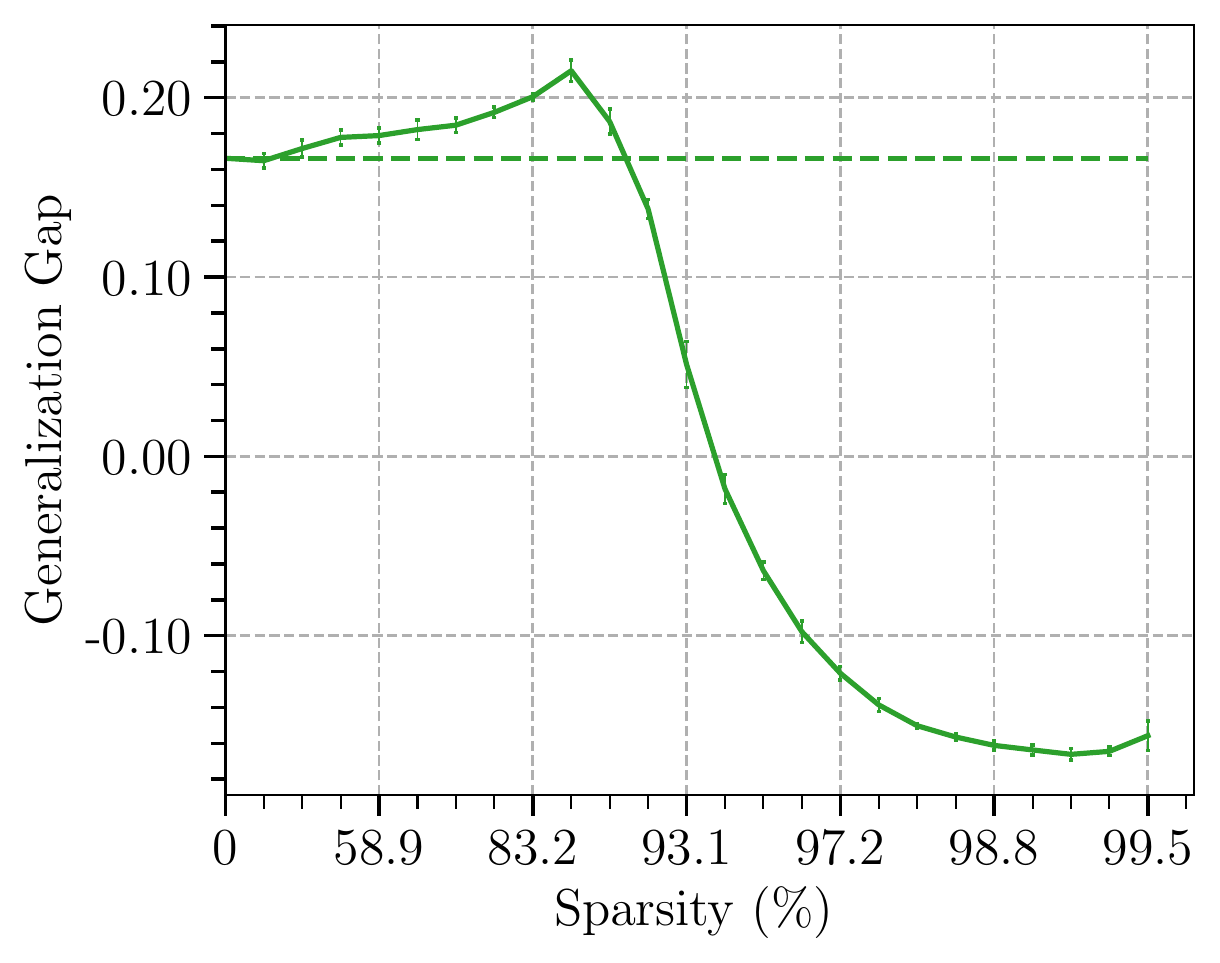}
\includegraphics[width=0.3\linewidth]{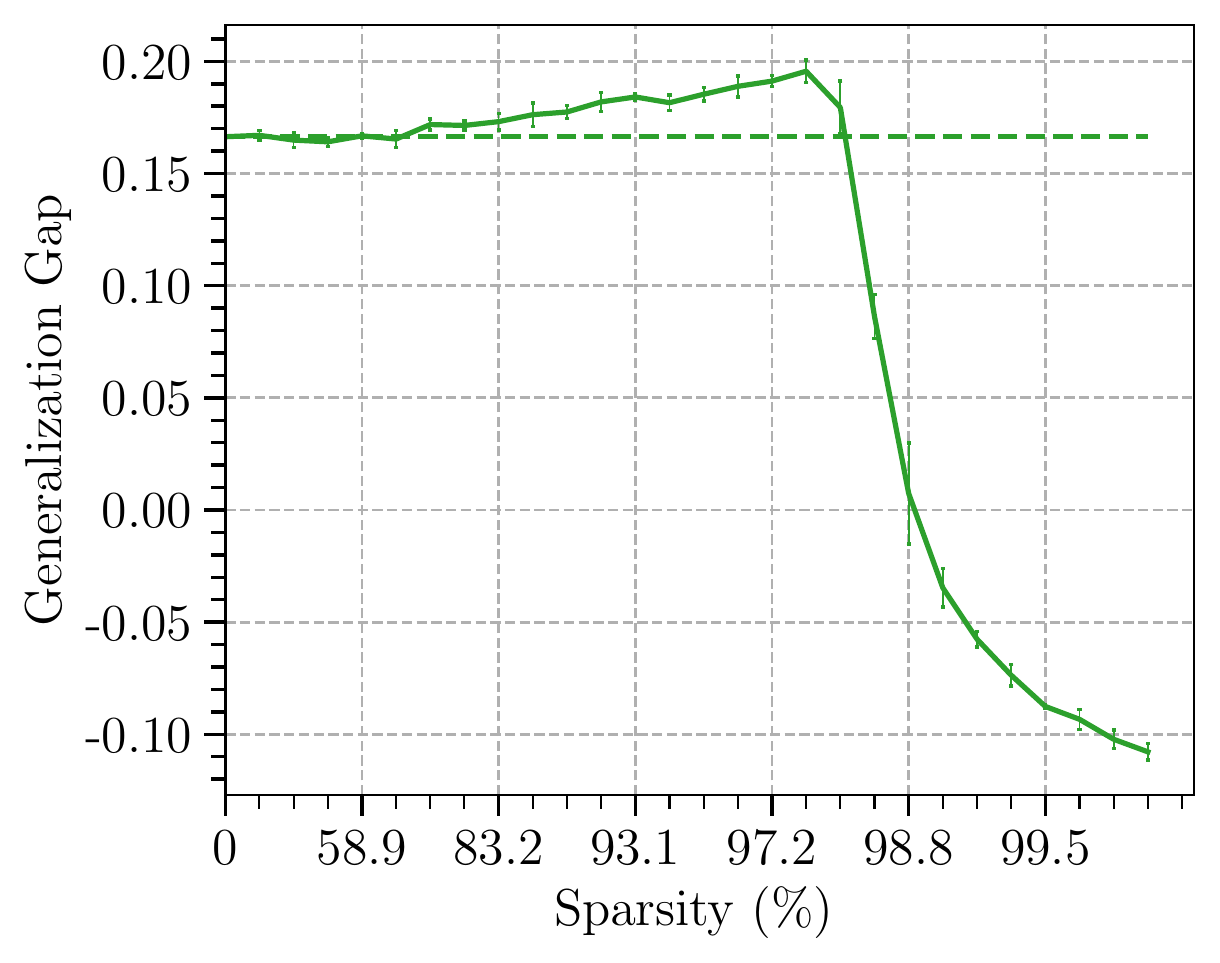}
\includegraphics[width=0.3\linewidth]{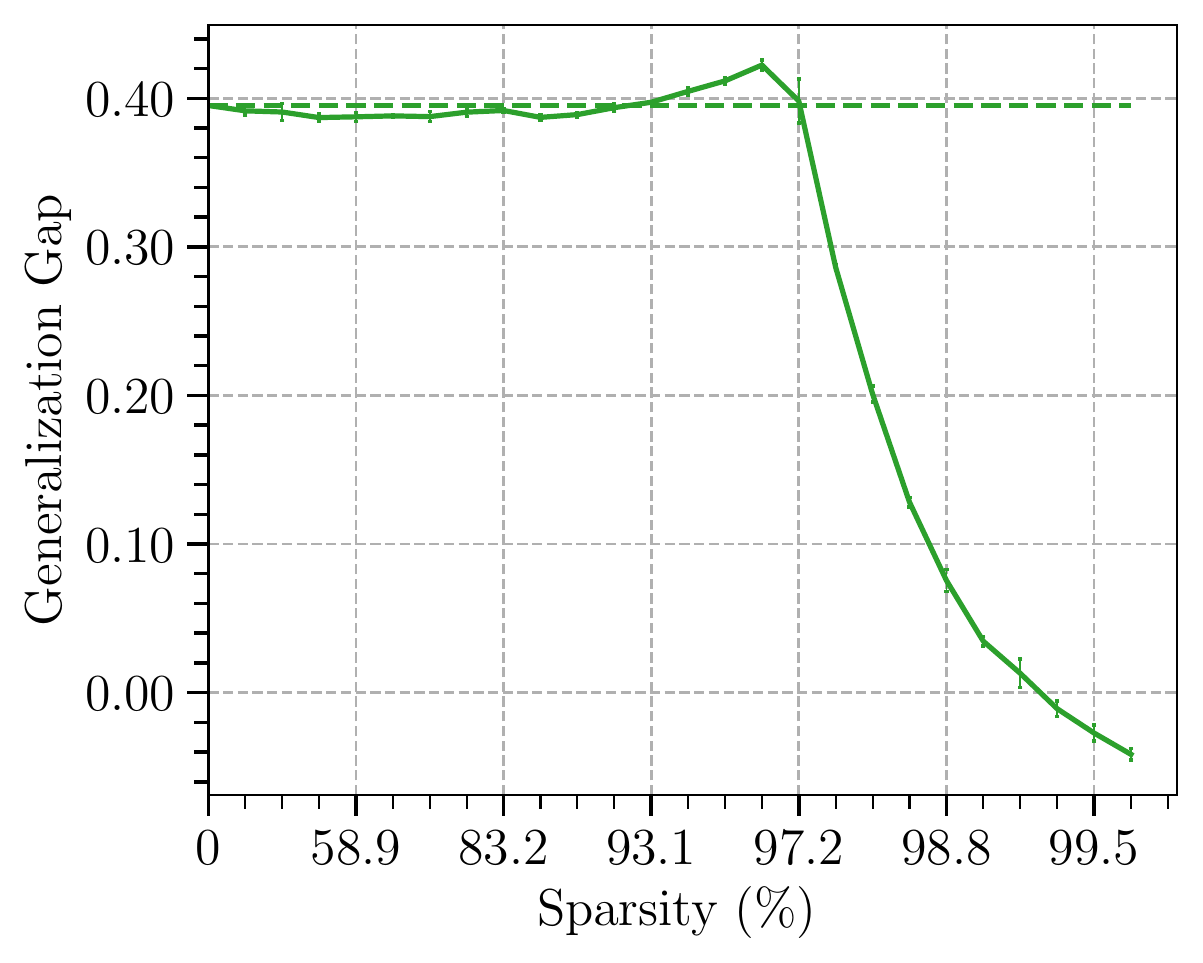}
\caption{Sparse Double Descent of LeNet-300-100 on MNIST and ResNet-18 on CIFAR with $20\%$ symmetric label noise, pruned with magnitude-based pruning, and retrained with LTR. We show how the generalization performance of networks depends on model sparsity, where generalization gap equals test error - train error. \textbf{Left}: MNIST. \textbf{Middle}: CIFAR-10. \textbf{Right}: CIFAR-100.}
\label{fig:sparsedd-mnist-cifar-gengap}
\end{figure}

\subsection{Comparison between Lottery Tickets and Random Initializations}

Here we compare the performance of lottery tickets and random initializations. Models at the same sparsity share the same mask structure. 

\begin{figure}[H]
\begin{center}
\includegraphics[width=0.3\textwidth]{img/cifar10/0.2_sym_acc_mag_lottery_reinit.pdf}
\includegraphics[width=0.3\textwidth]{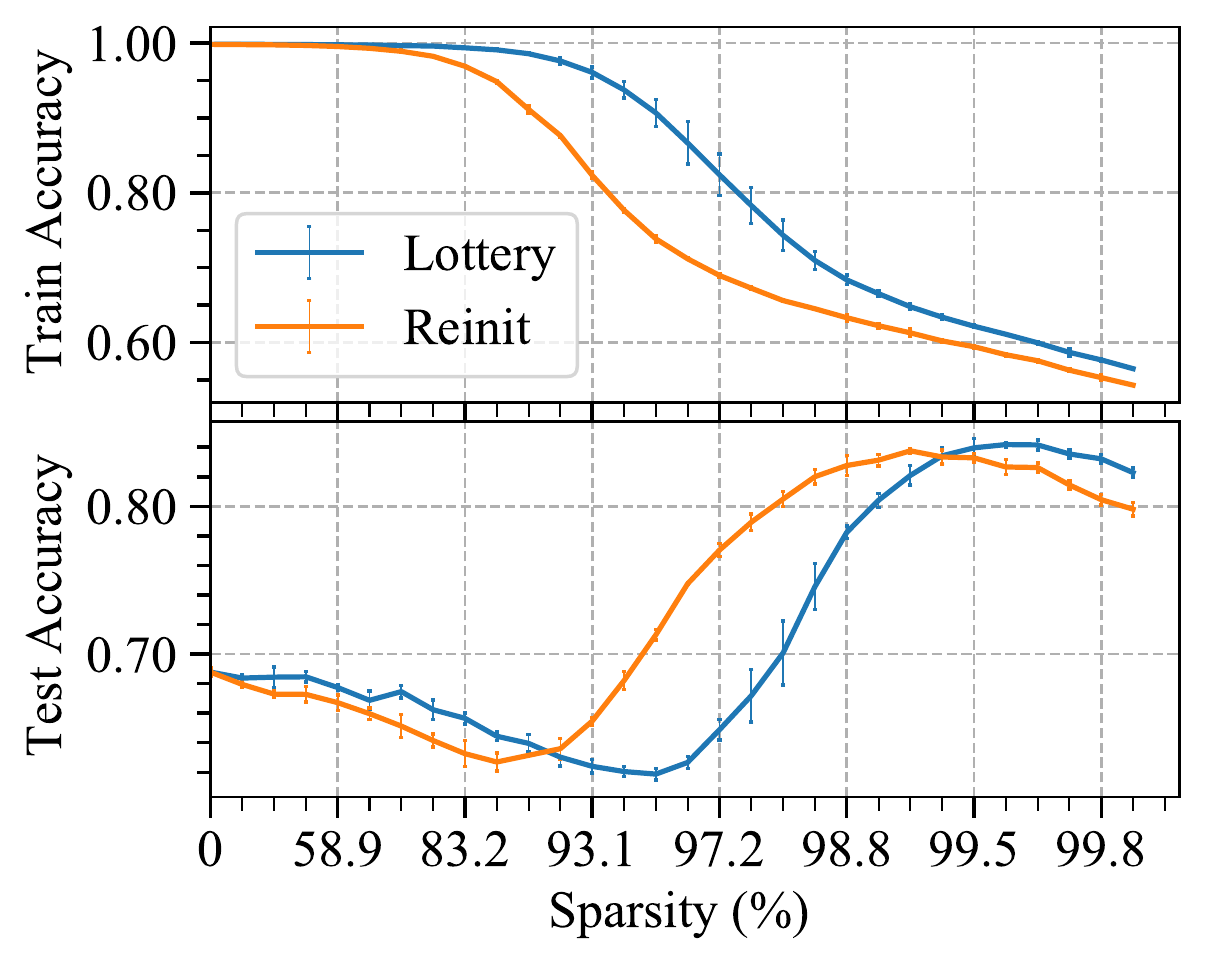}
\includegraphics[width=0.3\textwidth]{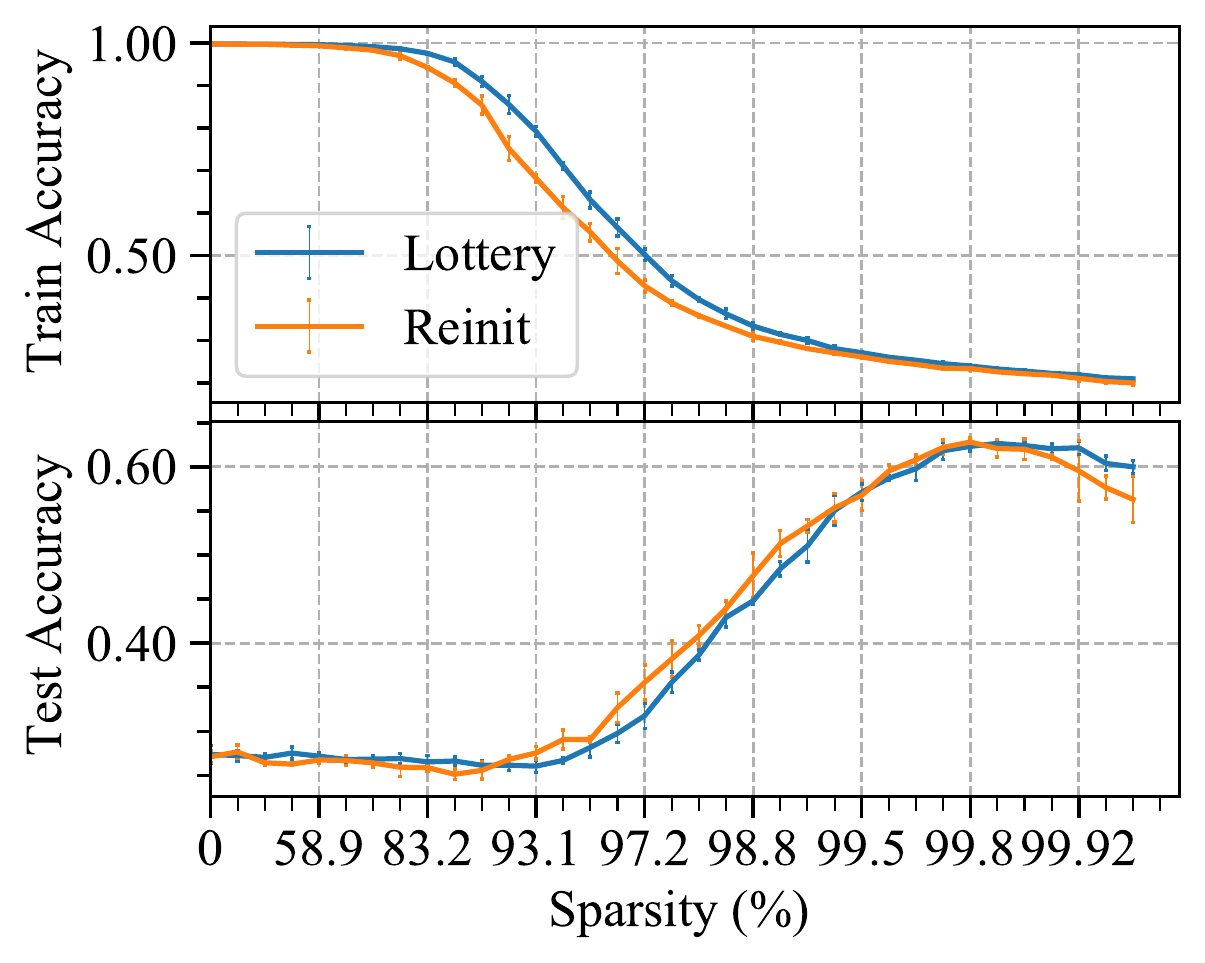}
\end{center}
\vspace{-0.5cm}
\caption{Performance of ResNet-18 on CIFAR-10 when retrained from either the original initialization (lottery tickets), or a random reinitialization. \textbf{Left}: $\epsilon=20\%$. \textbf{Middle}: $\epsilon=40\%$.  \textbf{Right}: $\epsilon=80\%$.}
\label{fig:sparsedd-cifar10-lottery-reinit}

\end{figure}
\begin{figure}[H]
\vspace{-0.3cm}
\begin{center}
\includegraphics[width=0.3\textwidth]{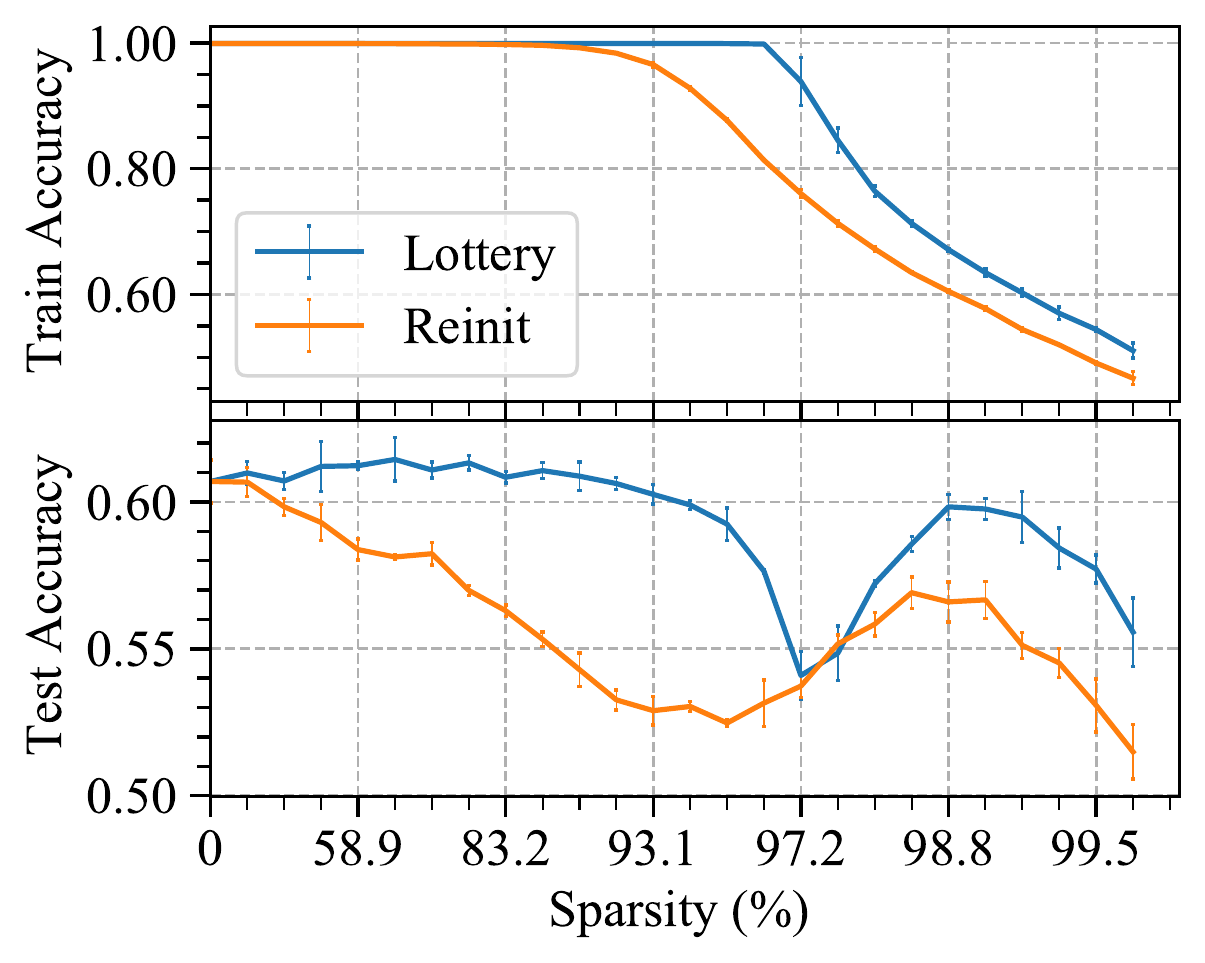}
\includegraphics[width=0.3\textwidth]{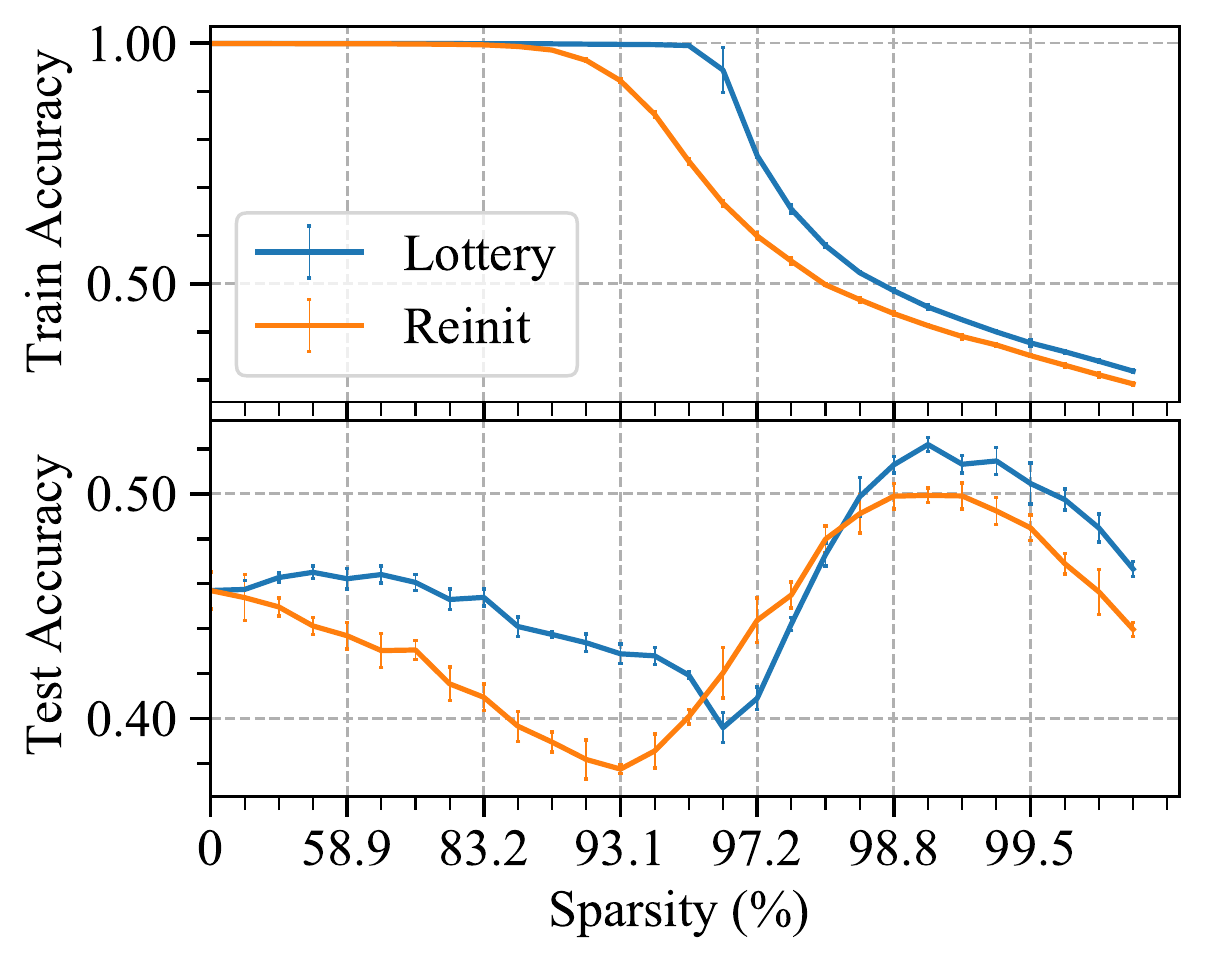}
\includegraphics[width=0.3\textwidth]{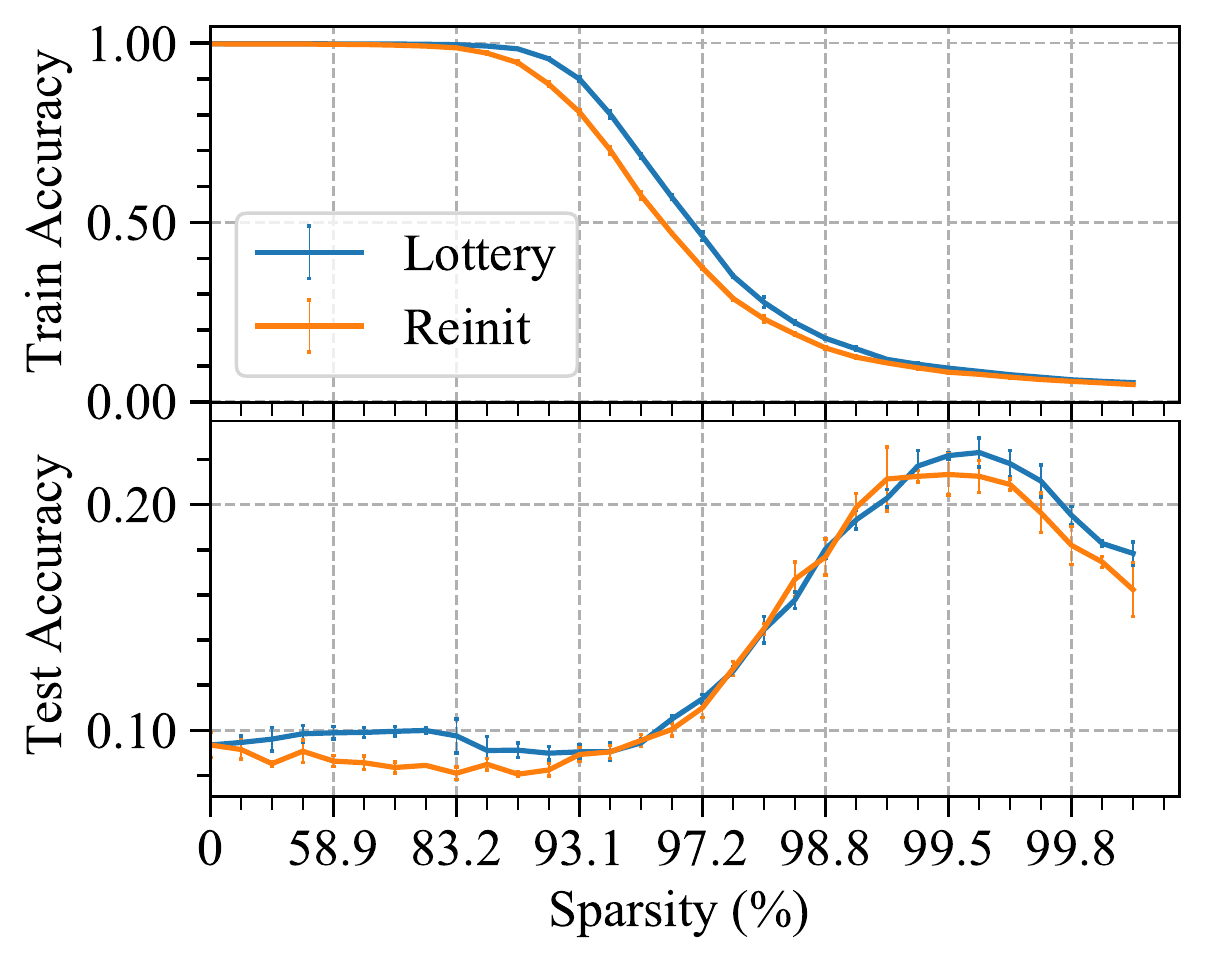}
\end{center}
\vspace{-0.5cm}
\caption{Performance of ResNet-18 on CIFAR-100 when retrained from either the original initialization (lottery tickets), or a random reinitialization. \textbf{Left}: $\epsilon=20\%$. \textbf{Middle}: $\epsilon=40\%$.  \textbf{Right}: $\epsilon=80\%$.}
\label{fig:sparsedd-cifar100-lottery-reinit}
\end{figure}

\subsection{Training Dynamics of Sparse Models}
\begin{figure}[H]
\begin{center}
\includegraphics[width=0.49\textwidth]{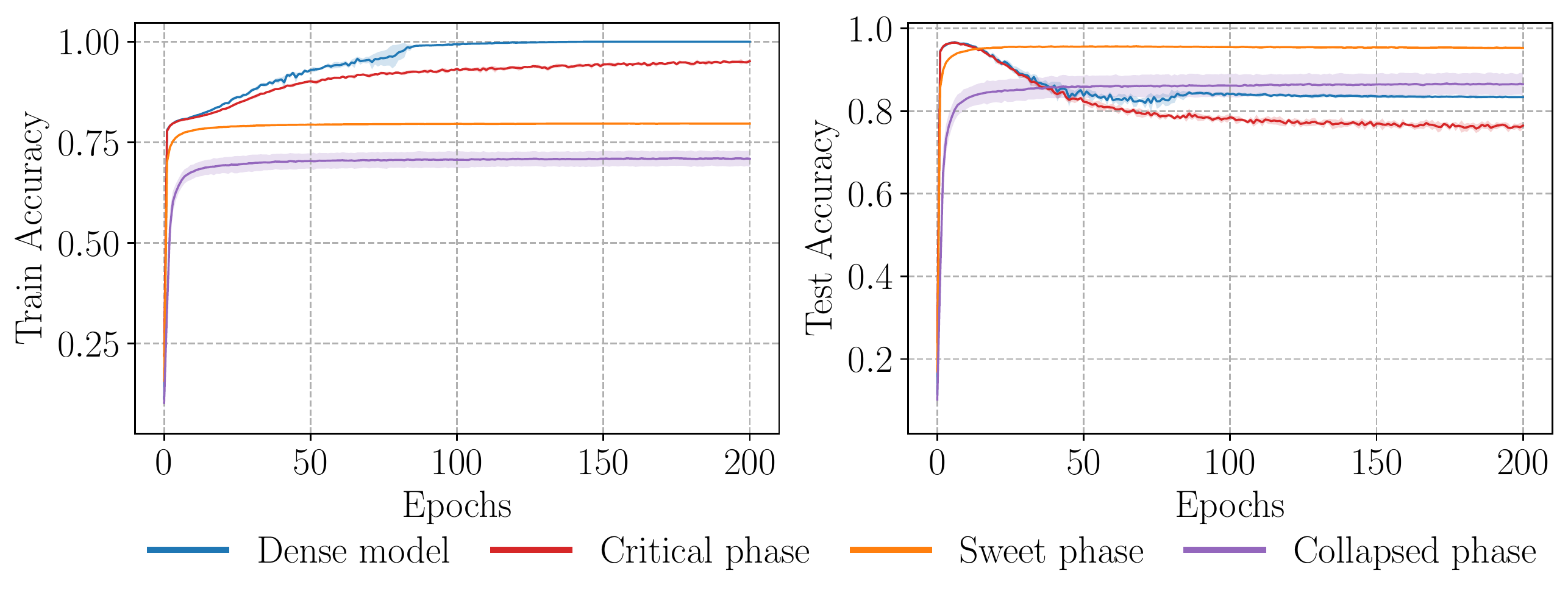}
\includegraphics[width=0.49\textwidth]{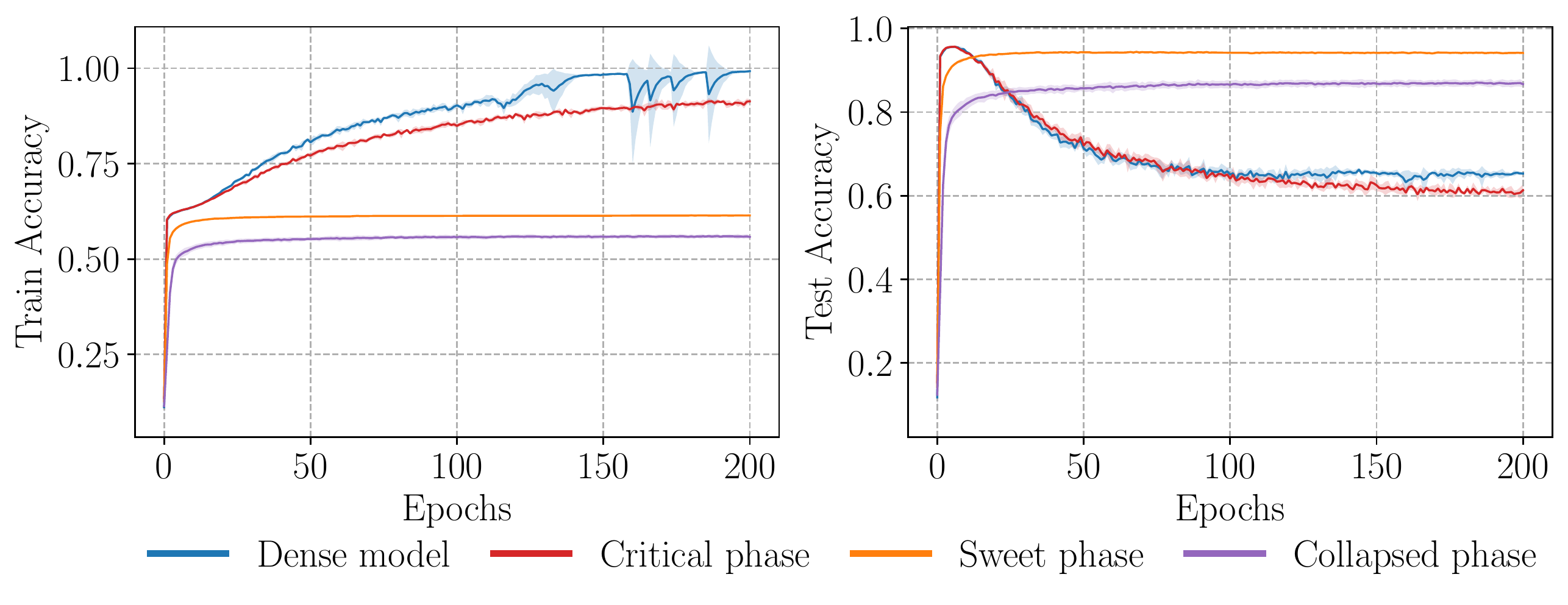}
\end{center}
\caption{Training dynamics w.r.t. epochs at four sparsities across different permuted fractions. Models are LeNet-300-100 for MNIST pruned with magnitude-based strategy. \textbf{Left}: $\epsilon=20\%$. \textbf{Right}: $\epsilon=40\%$.}
\label{fig:epoch-wies-train-test-accuracy-mnist}
\end{figure}

\begin{figure}[H]
\setlength{\abovecaptionskip}{0pt} 
\setlength{\belowcaptionskip}{0pt} 
\begin{center}
\subfigure[CIFAR-10]{
\begin{minipage}[b]{0.48\textwidth}
    \includegraphics[width=1\textwidth]{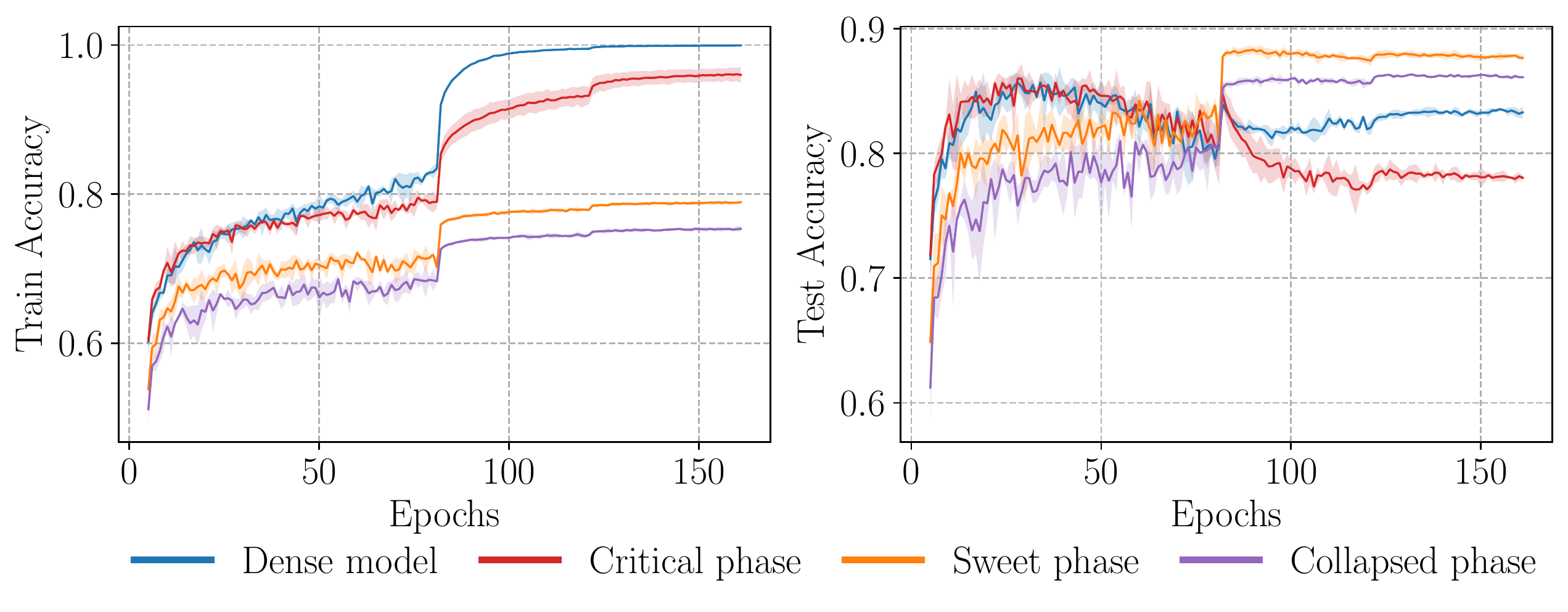}
    \includegraphics[width=1\textwidth]{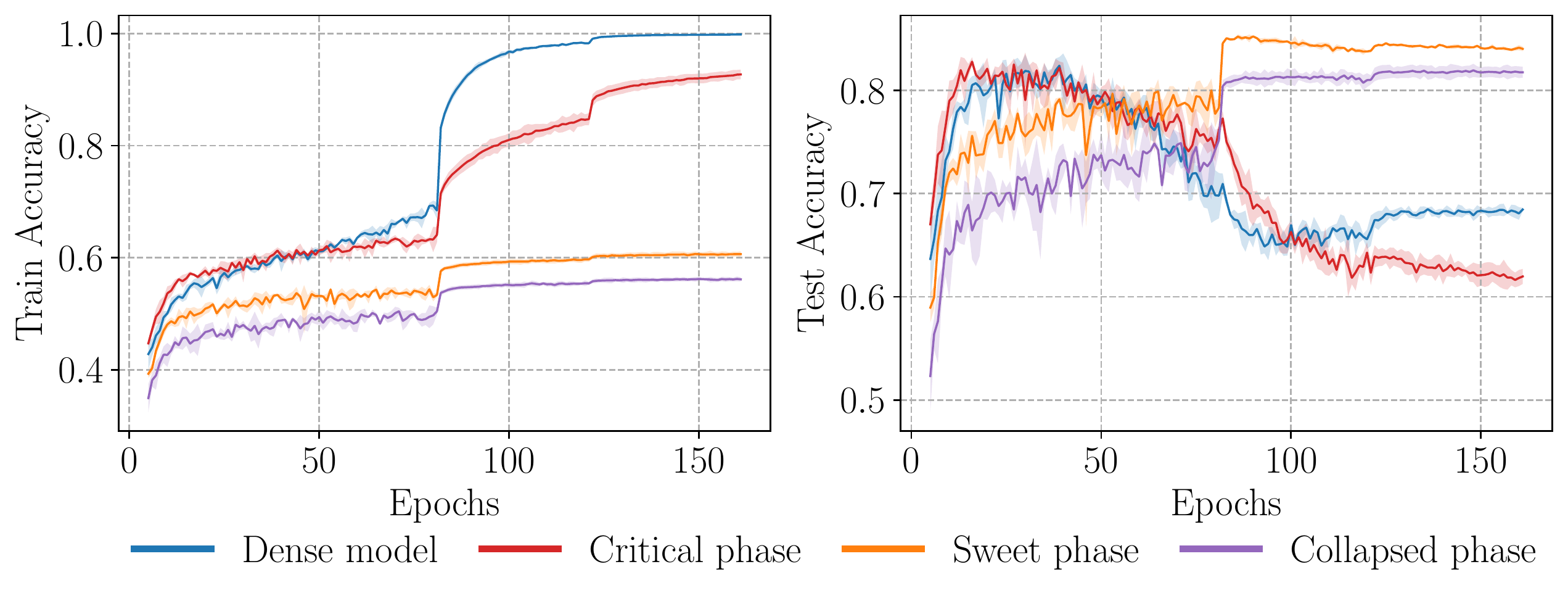}
    \includegraphics[width=1\textwidth]{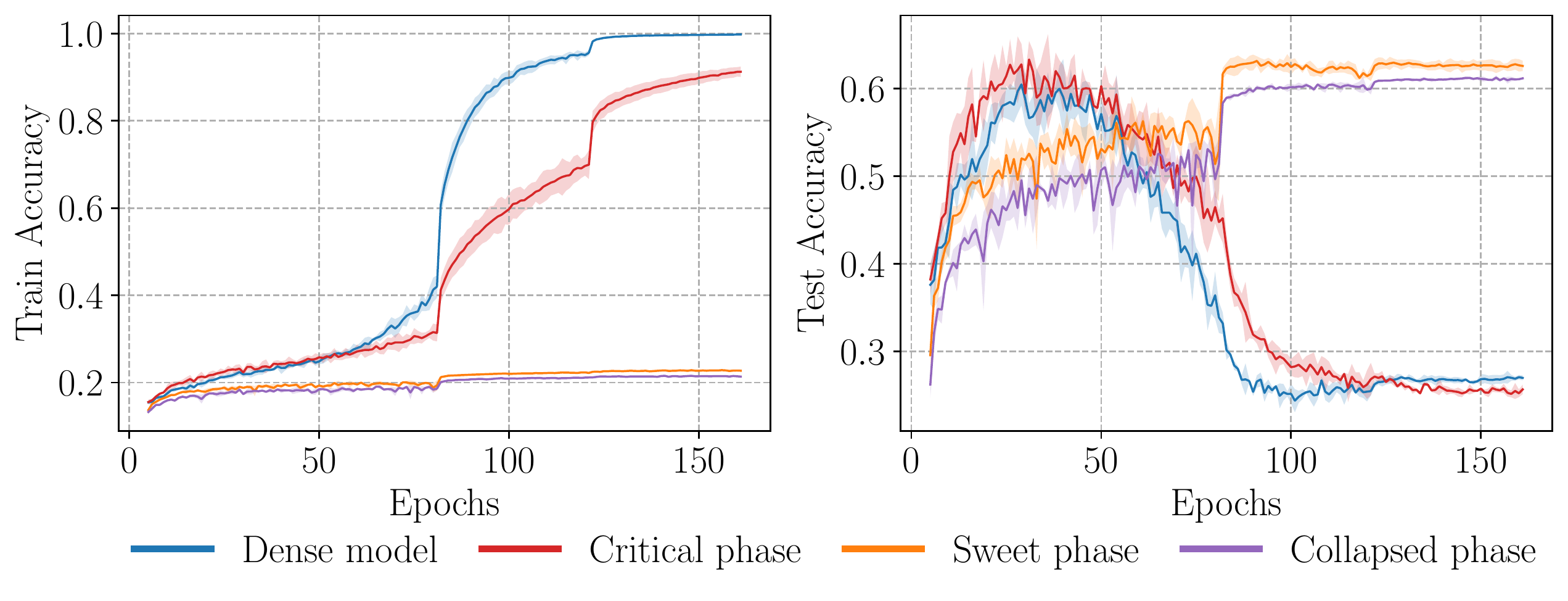}
\end{minipage}
}
\subfigure[CIFAR-100]{
\begin{minipage}[b]{0.48\textwidth}
    \includegraphics[width=1\textwidth]{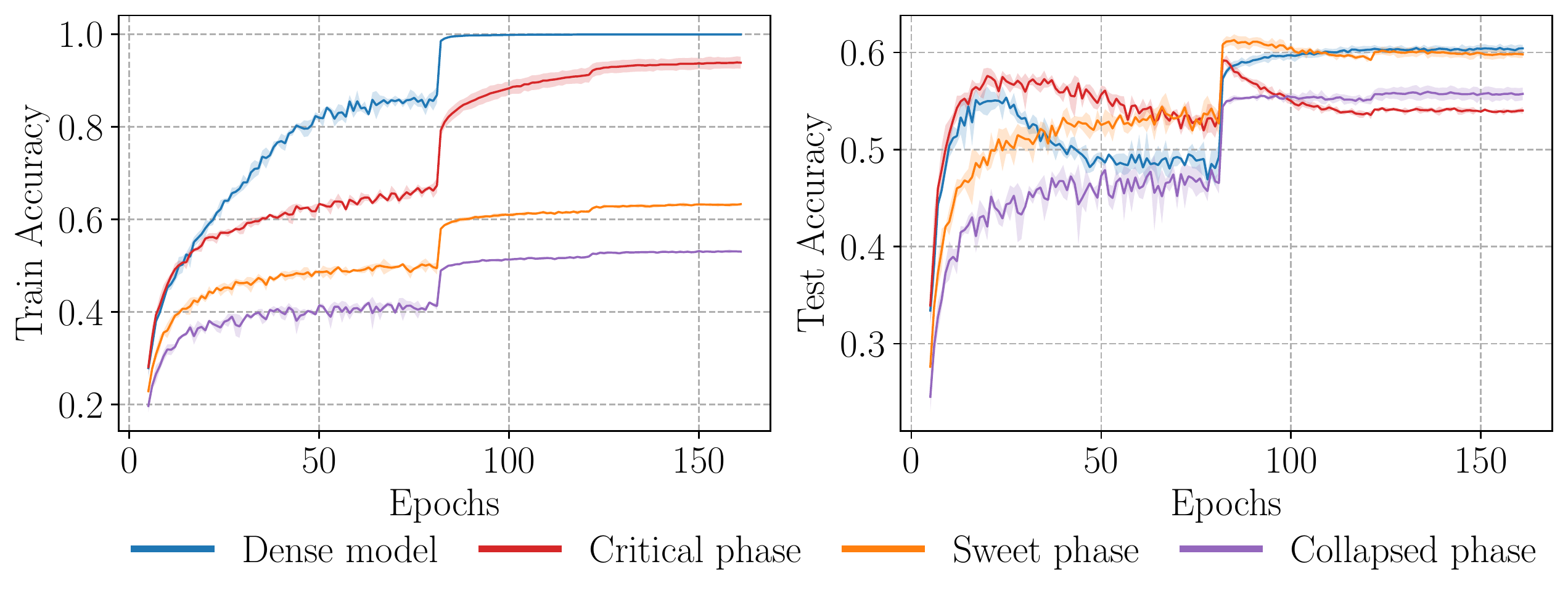}
    \includegraphics[width=1\textwidth]{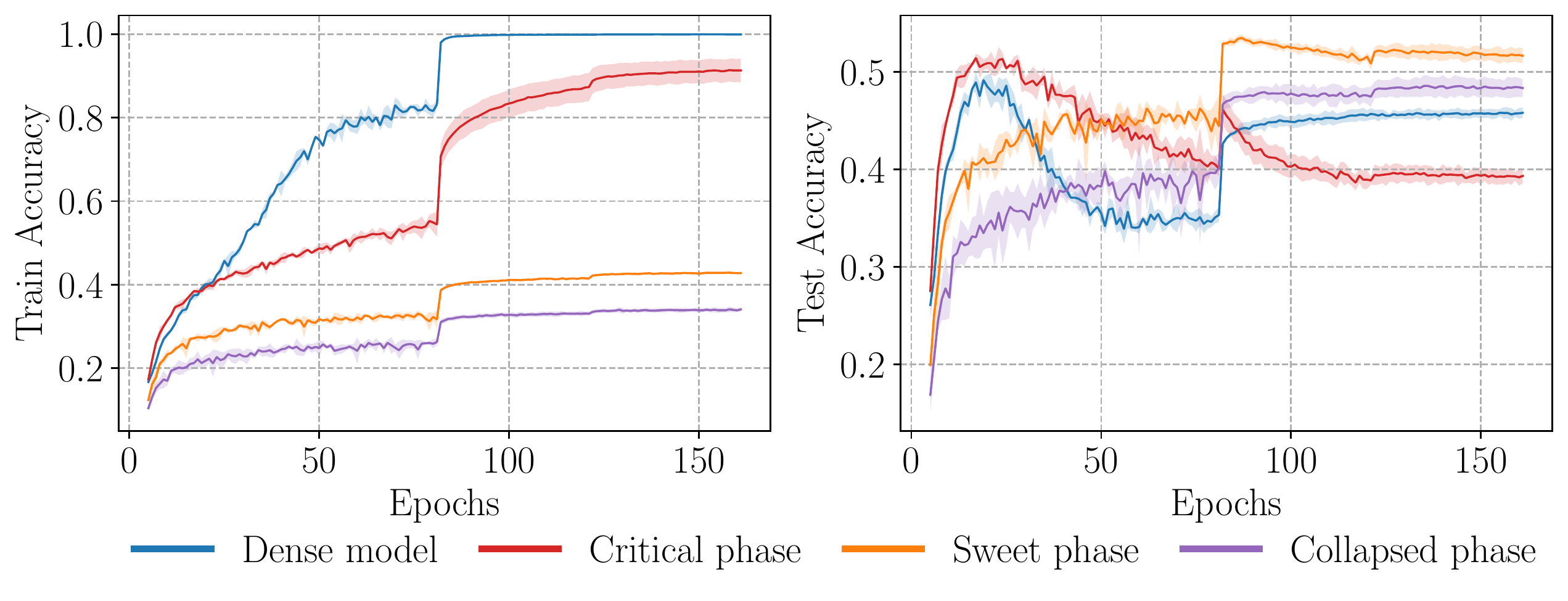}
    \includegraphics[width=1\textwidth]{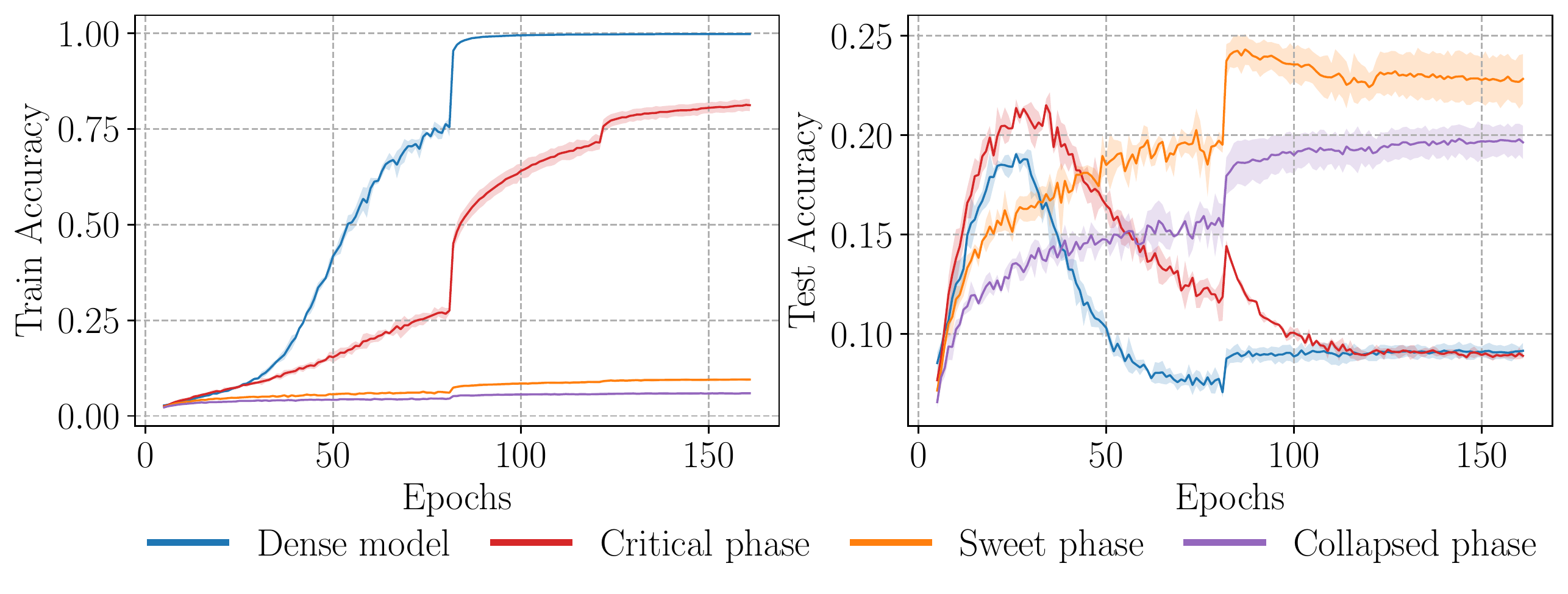}
\end{minipage}
}
\end{center}
\caption{Training dynamics w.r.t. epochs at four sparsities across different permuted fractions. Datasets are CIFAR-10/CIFAR-100 and models are ResNet-18 pruned with magnitude-based strategy. \textbf{Top}: $\epsilon=20\%$. \textbf{Middle}: $\epsilon=40\%$. \textbf{Bottom}: $\epsilon=80\%$.}
\label{fig:epoch-wies-train-test-accuracy-cifar}
\end{figure}

\subsection{Re-dense Training Results}

\begin{figure}[H]
\vspace{-0.5cm}
\begin{center}
    \includegraphics[width=1\textwidth]{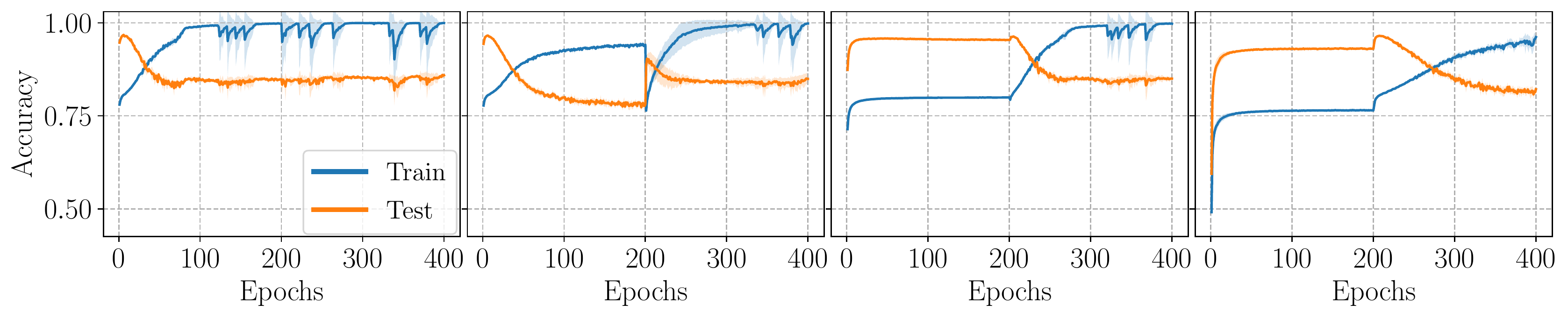}
\end{center}
\vspace{-0.5cm}
\caption{Accuracy curve of the sparse and re-dense training process. We recover pruned weights at epoch 200, and training them from value of zero for another 200 epochs using the last learning rate of sparse training, which is 0.1 for MNIST with $\epsilon=20\%$, LeNet-300-100.}
\label{fig:re-dense-training-mnist}
\end{figure}

\begin{figure}[H]
\vspace{-0.5cm}
\begin{center}
    \includegraphics[width=1\textwidth]{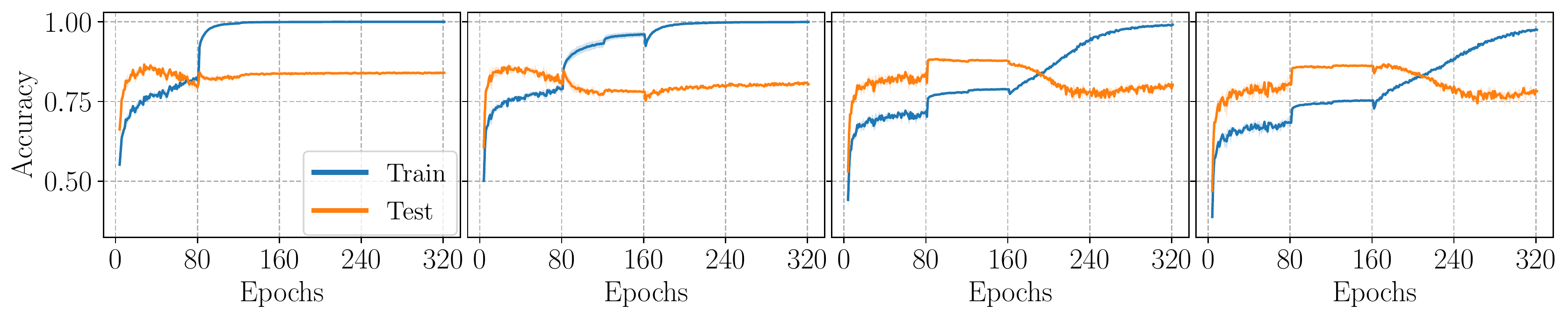}
    \includegraphics[width=1\textwidth]{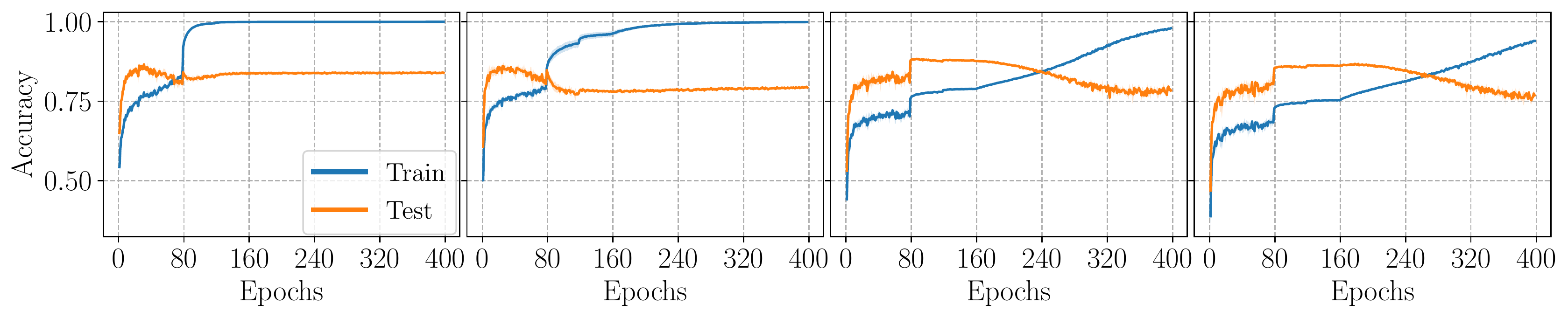}
\end{center}
\vspace{-0.5cm}
\caption{Accuracy curve of the sparse and re-dense training process. We recover pruned weights at epoch 160, and training them from value of zero for another 160 epochs using a small learning rate of sparse training, which is 0.001 as the last epoch learning rate (the top) and 0.0001 (the bottom) for CIFAR-10 with $\epsilon=20\%$, ResNet-18.}
\label{fig:re-dense-training-cifar10-2}
\end{figure}

\begin{figure}[H]
\vspace{-0.5cm}
\begin{center}
    \includegraphics[width=1\textwidth]{img/cifar100/others/0.4_regain_0.001_accuracy.pdf}
    \includegraphics[width=1\textwidth]{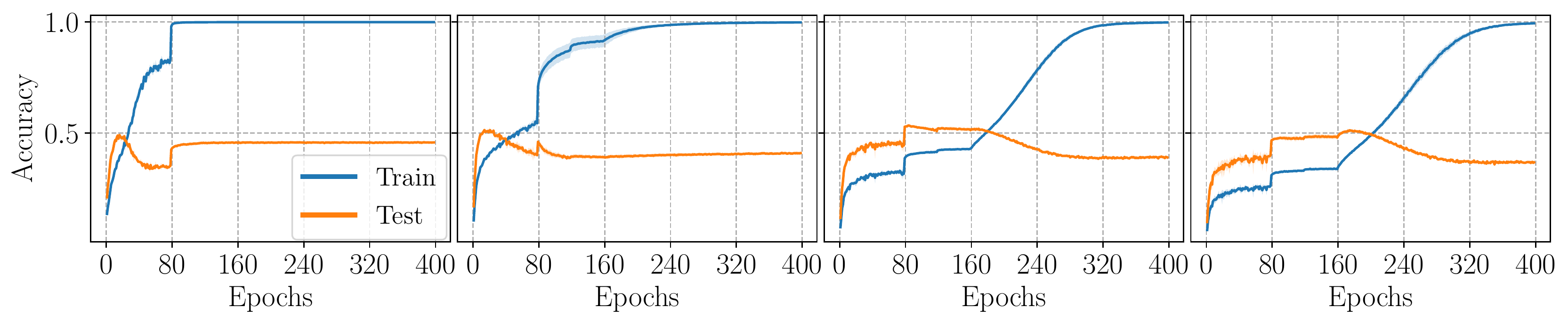}
\end{center}
\vspace{-0.5cm}
\caption{Accuracy curve of the sparse and re-dense training process. We recover pruned weights at epoch 160, and training them from value of zero for another 160 epochs using a small learning rate of sparse training, which is 0.001 as the last epoch learning rate (the top) and 0.0001 (the bottom) for CIFAR-100 with $\epsilon=40\%$, ResNet-18.}
\label{fig:re-dense-training-cifar100-2}
\end{figure}

\subsection{Linear Interpolation Results}
\label{apx:linear-interpolation}

\begin{figure}[H]
\vspace{-0.5cm}
\setlength{\abovecaptionskip}{0pt} 
\setlength{\belowcaptionskip}{0pt} 
\begin{center}
\subfigure[Critical Phase]{
\begin{minipage}[t]{0.25\textwidth}
\begin{center}
\end{center}
    \includegraphics[width=1\linewidth]{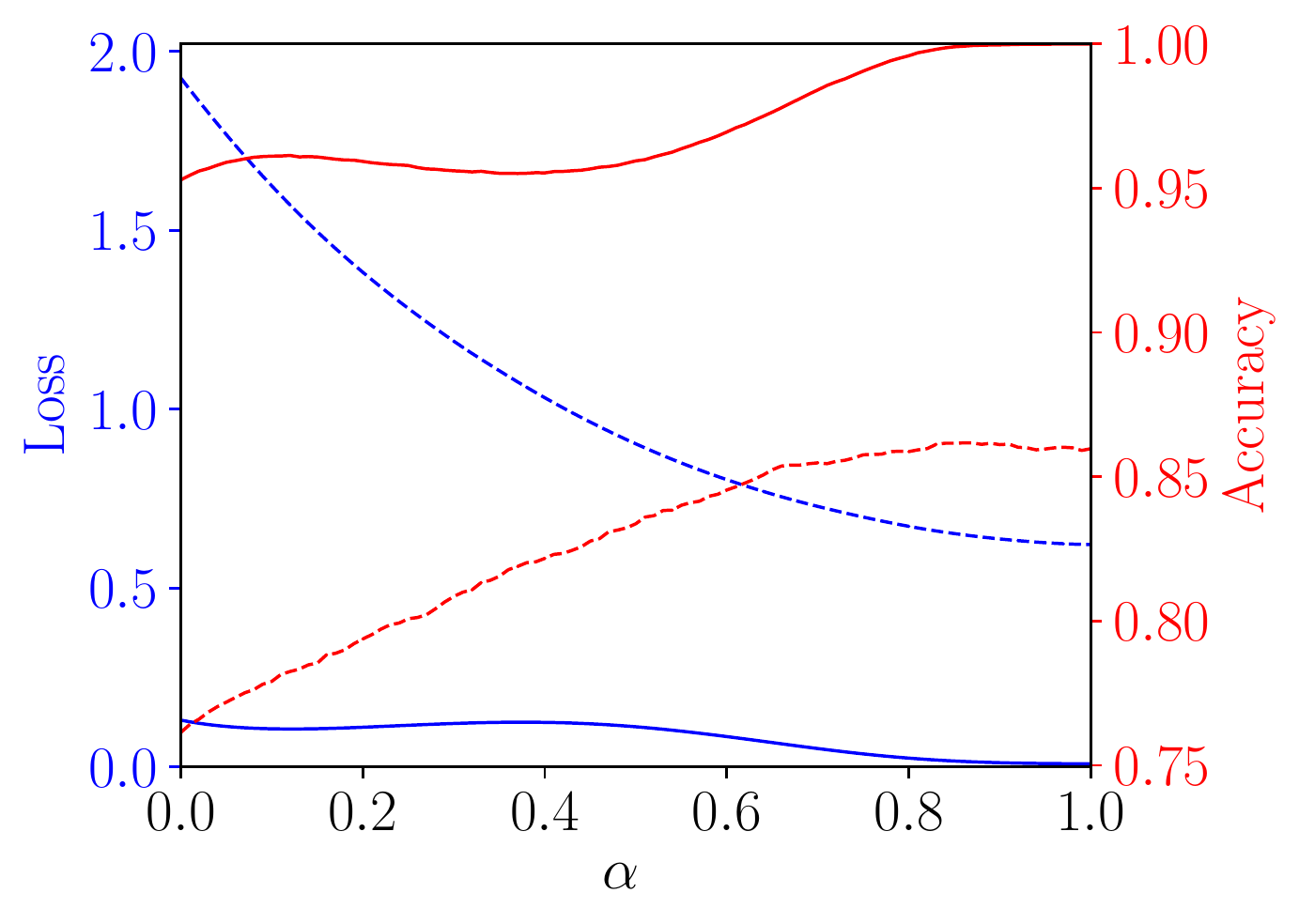}
\end{minipage}
}   
\subfigure[Sweet Phase]{
\begin{minipage}[t]{0.25\textwidth}
\begin{center}
\end{center}
    \includegraphics[width=1\linewidth]{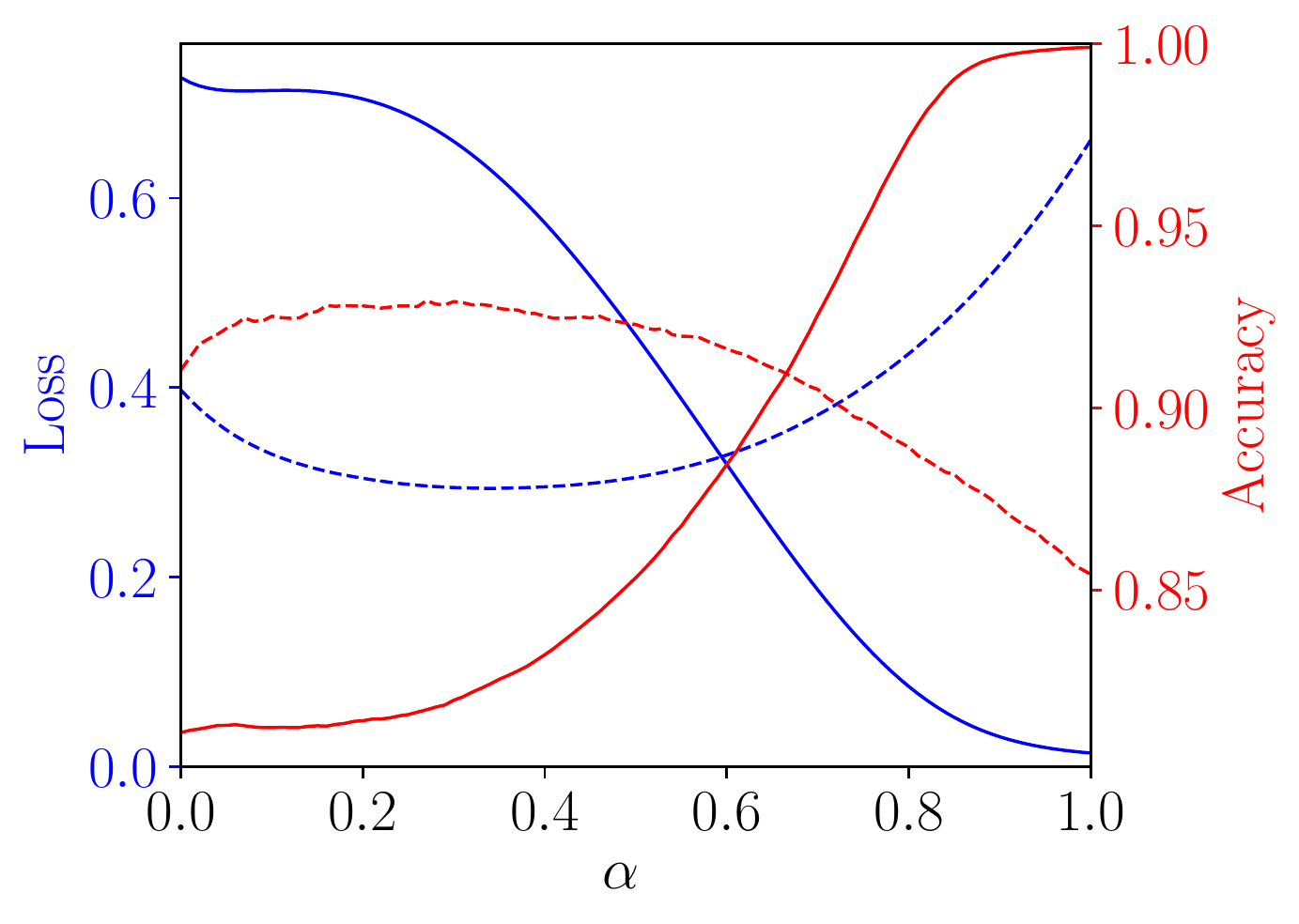}
\end{minipage}
}    
\end{center}
\vspace{-0.5cm}
\caption{Linear interpolation plots. Models are LeNet-300-100 on MNIST with $\epsilon=20\%$. $\alpha=0$ corresponds to sparse solutions and $\alpha=1$ corresponds to the re-dense solutions. The blue lines are loss curves and the red lines are accuracy curves; solid lines indicate training data set and dashed lines indicate testing data set. For re-dense models, sparsity is measured before recovering weights.}
\label{fig:linear-interpolation-mnist-apx}
\end{figure}

\begin{figure}[H]
\vspace{-0.5cm}
\setlength{\abovecaptionskip}{0pt} 
\setlength{\belowcaptionskip}{0pt} 
\begin{center}
\subfigure[Critical Phase]{
\begin{minipage}[t]{0.25\textwidth}
\begin{center}
\end{center}
    \includegraphics[width=1\linewidth]{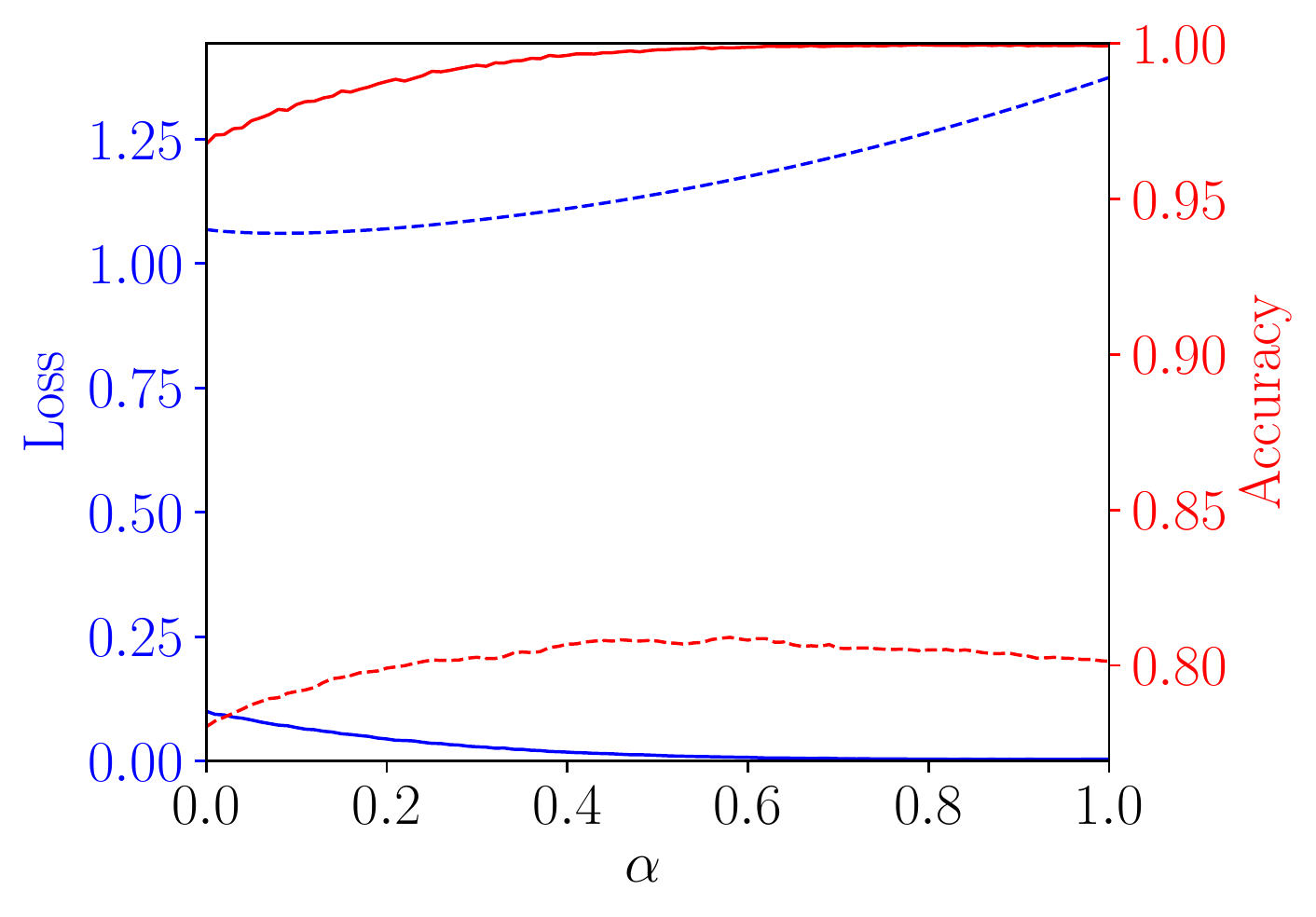}
\end{minipage}
}   
\subfigure[Sweet Phase]{
\begin{minipage}[t]{0.25\textwidth}
\begin{center}
\end{center}
    \includegraphics[width=1\linewidth]{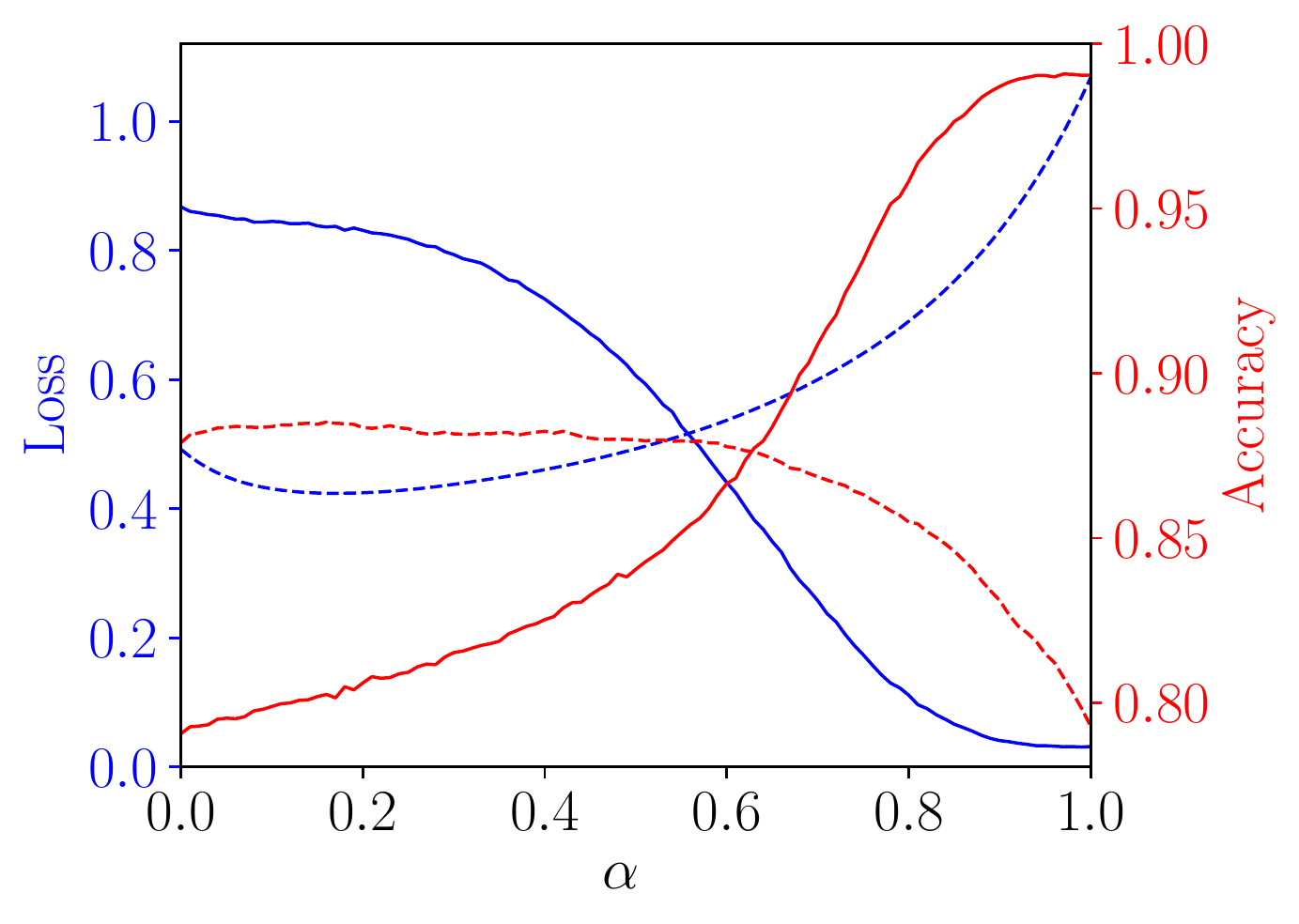}
\end{minipage}
}    
\end{center}
\vspace{-0.5cm}
\caption{Linear interpolation plots. Models are ResNet-18 on CIFAR-10 with $\epsilon=20\%$. $\alpha=0$ corresponds to sparse solutions and $\alpha=1$ corresponds to the re-dense solutions. The blue lines are loss curves and the red lines are accuracy curves; solid lines indicate training data set and dashed lines indicate testing data set. For re-dense models, sparsity is measured before recovering weights.}
\label{fig:linear-interpolation-cifar10-apx}
\end{figure}

\begin{figure}[H]
\vspace{-0.5cm}
\setlength{\abovecaptionskip}{0pt} 
\setlength{\belowcaptionskip}{0pt} 
\begin{center}
\subfigure[Critical Phase]{
\begin{minipage}[t]{0.25\textwidth}
\begin{center}
\end{center}
    \includegraphics[width=1\linewidth]{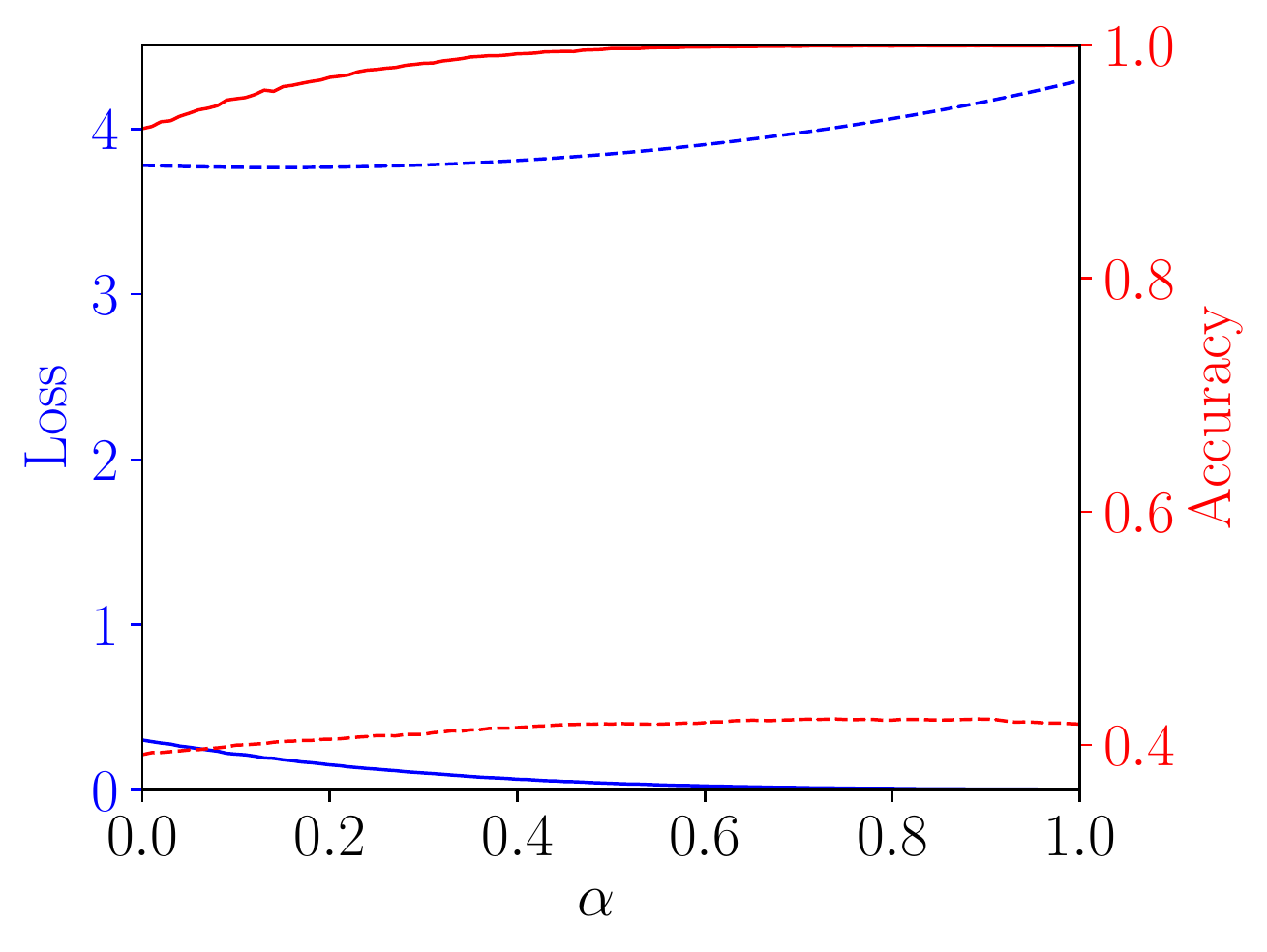}
\end{minipage}
}   
\subfigure[Sweet Phase]{
\begin{minipage}[t]{0.25\textwidth}
\begin{center}
\end{center}
    \includegraphics[width=1\linewidth]{img/cifar100/others/0.4_1_level_22_model_ep160_it0-level_22_model_ep320_it0_0.0_1.0_101.pdf}
\end{minipage}
}    
\end{center}
\vspace{-0.5cm}
\caption{Linear interpolation plots. Models are ResNet-18 on CIFAR-100 with $\epsilon=40\%$. $\alpha=0$ corresponds to sparse solutions and $\alpha=1$ corresponds to the re-dense solutions. The blue lines are loss curves and the red lines are accuracy curves; solid lines indicate training data set and dashed lines indicate testing data set. For re-dense models, sparsity is measured before recovering weights.}
\label{fig:linear-interpolation-cifar100-apx}
\end{figure}

\subsection{1-D Visualization of Re-dense Solutions}

\begin{figure}[H]
\vspace{-0.5cm}
\center
\includegraphics[width=0.3\linewidth]{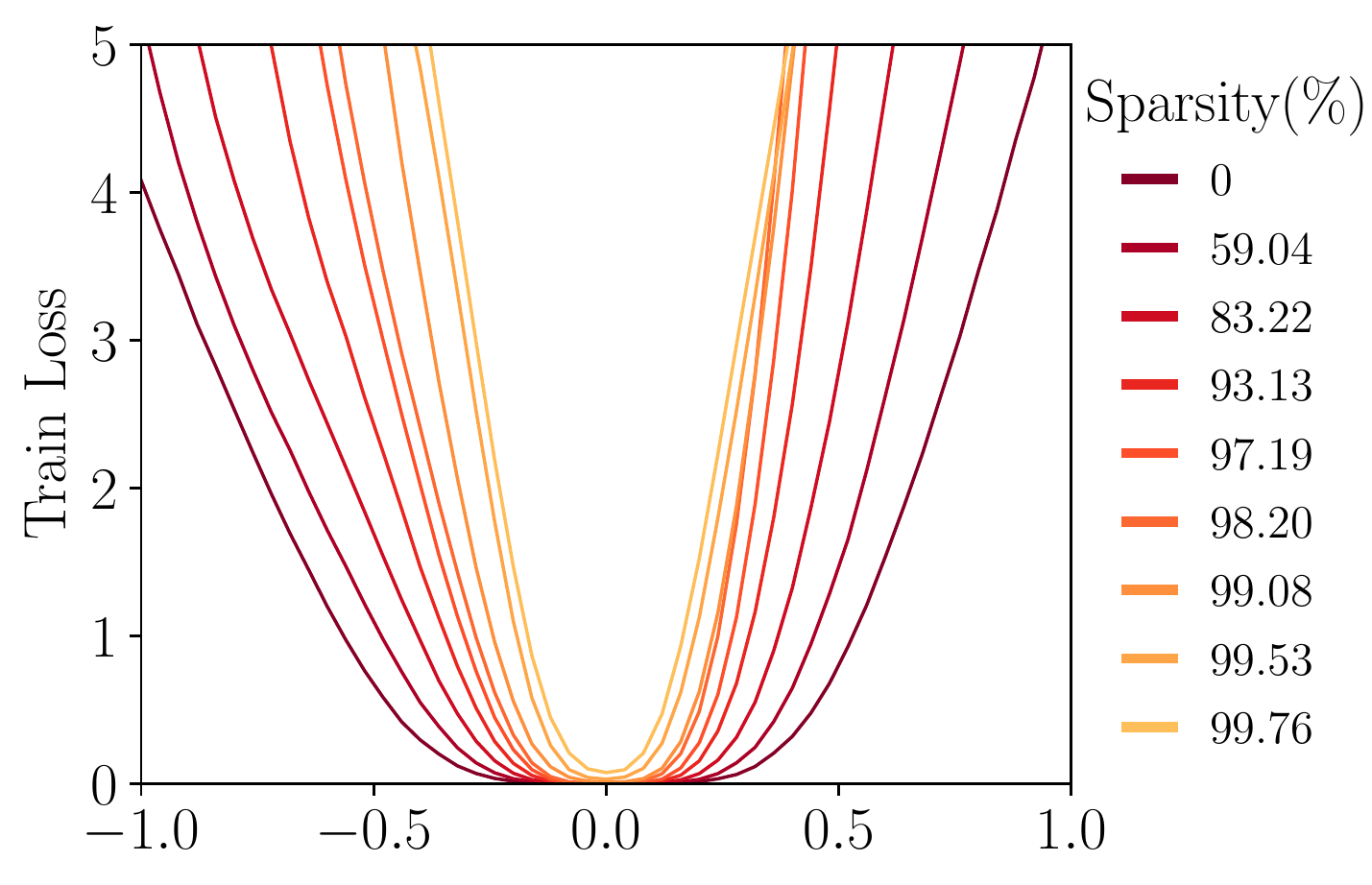}
\includegraphics[width=0.3\linewidth]{img/cifar100/others/0.4_1d_loss_repeat.pdf}
\vspace{-0.5cm}
\caption{The 1-D loss visualization of minima found by re-dense training using filter normalization \citep{li2018landscape}.The sparsity of particular model is measured before the re-dense step. The sparser the original pruned network is, the sharper minima it might converge to after re-dense training. 
\textbf{Left}: ResNet-18 on CIFAR-10 with $\epsilon=20\%$. \textbf{Right}:  ResNet-18  on CIFAR-100 with $\epsilon=40\%$.
}
\label{fig:1d-loss-cifar-apx}
\end{figure}

\subsection{Learning Distance Results}
\label{apx:distance}

\begin{figure}[H]
    \centering
    \includegraphics[width=0.3\linewidth]{img/mnist/distance/0.2_sym_mag_accuracy-distance.pdf}
    \includegraphics[width=0.3\linewidth]{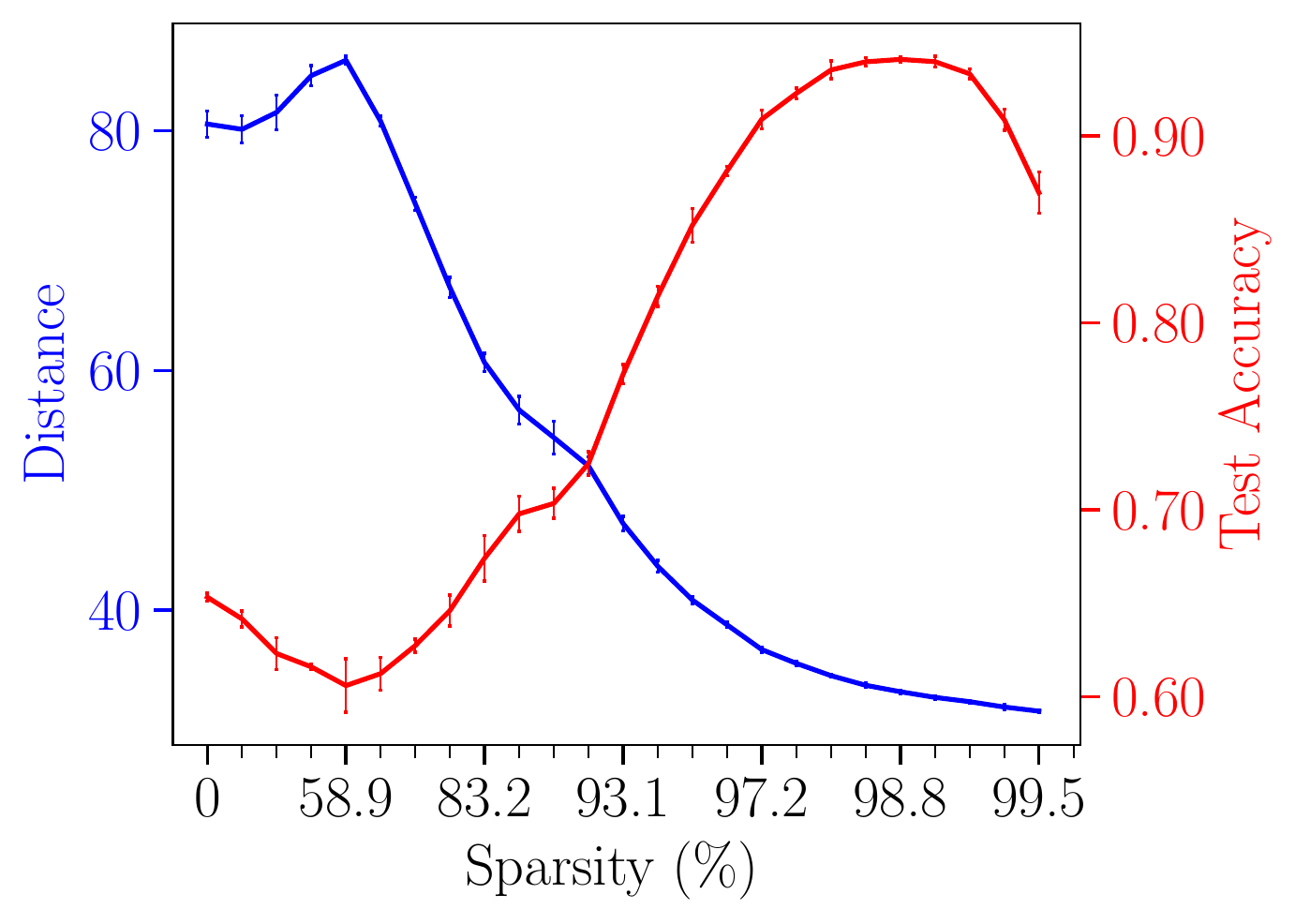}
    \includegraphics[width=0.3\linewidth]{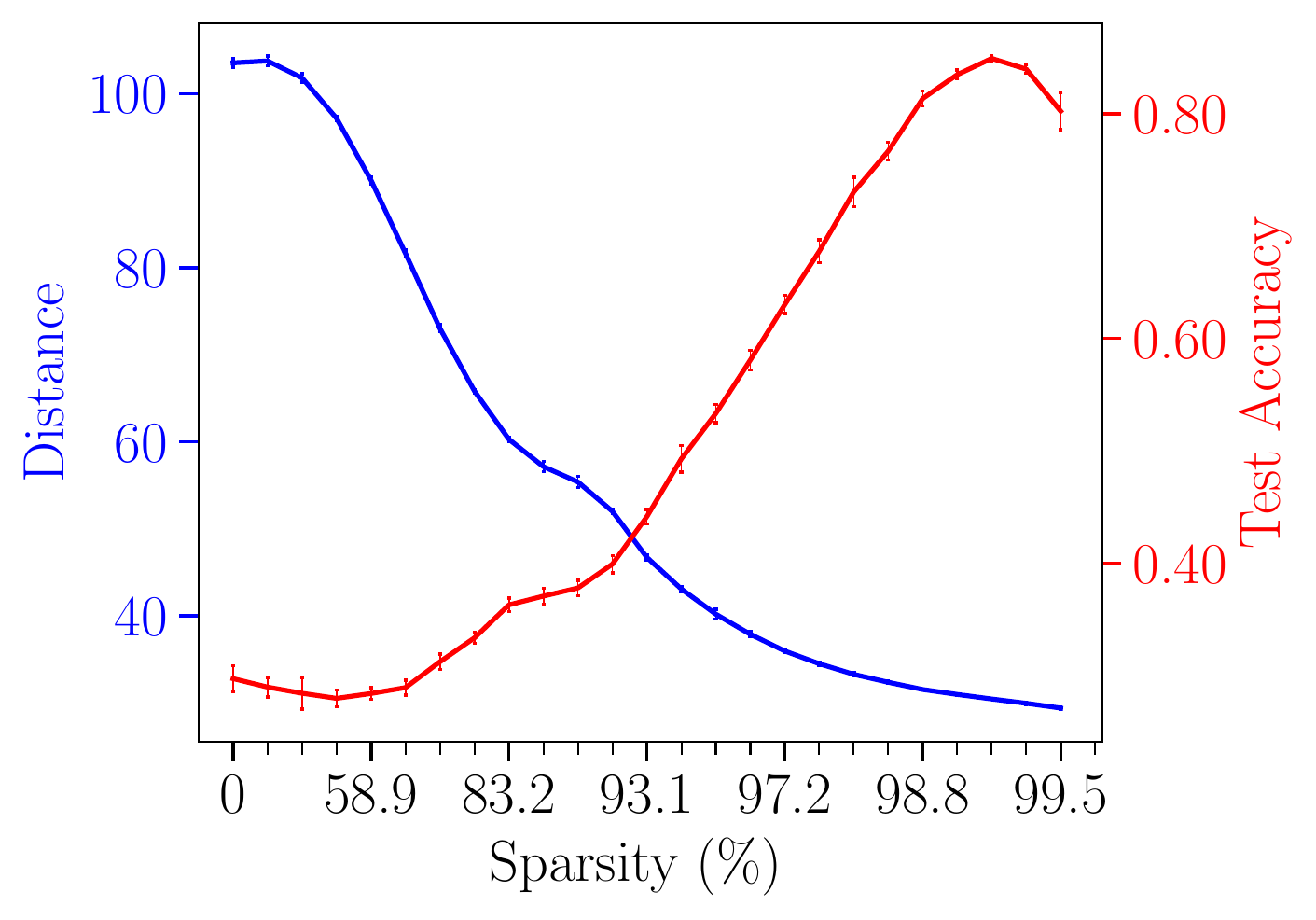}
    \caption{Learning distance for LeNet-300-100 on MNIST. The blue lines refer to $\ell_2$ learning distance and the red lines are test accuracy. \textbf{Left}: $\epsilon=20\%$. \textbf{Middle}: $\epsilon=40\%$. \textbf{Right}: $\epsilon=80\%$.}
    \label{fig:distance-sym-mnist}
\end{figure}

\begin{figure}[H]
    \centering
    \includegraphics[width=0.3\linewidth]{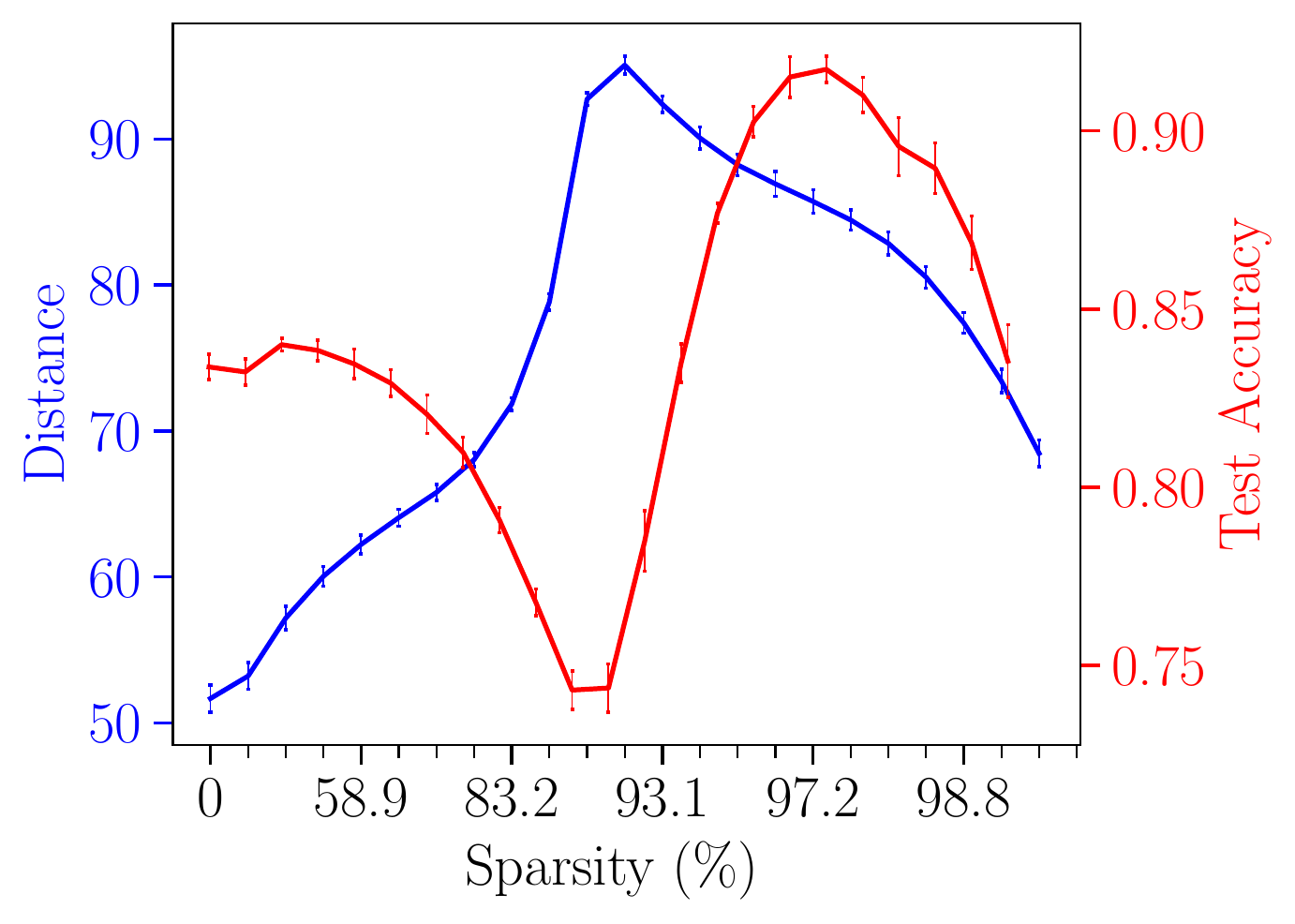}
    \includegraphics[width=0.3\linewidth]{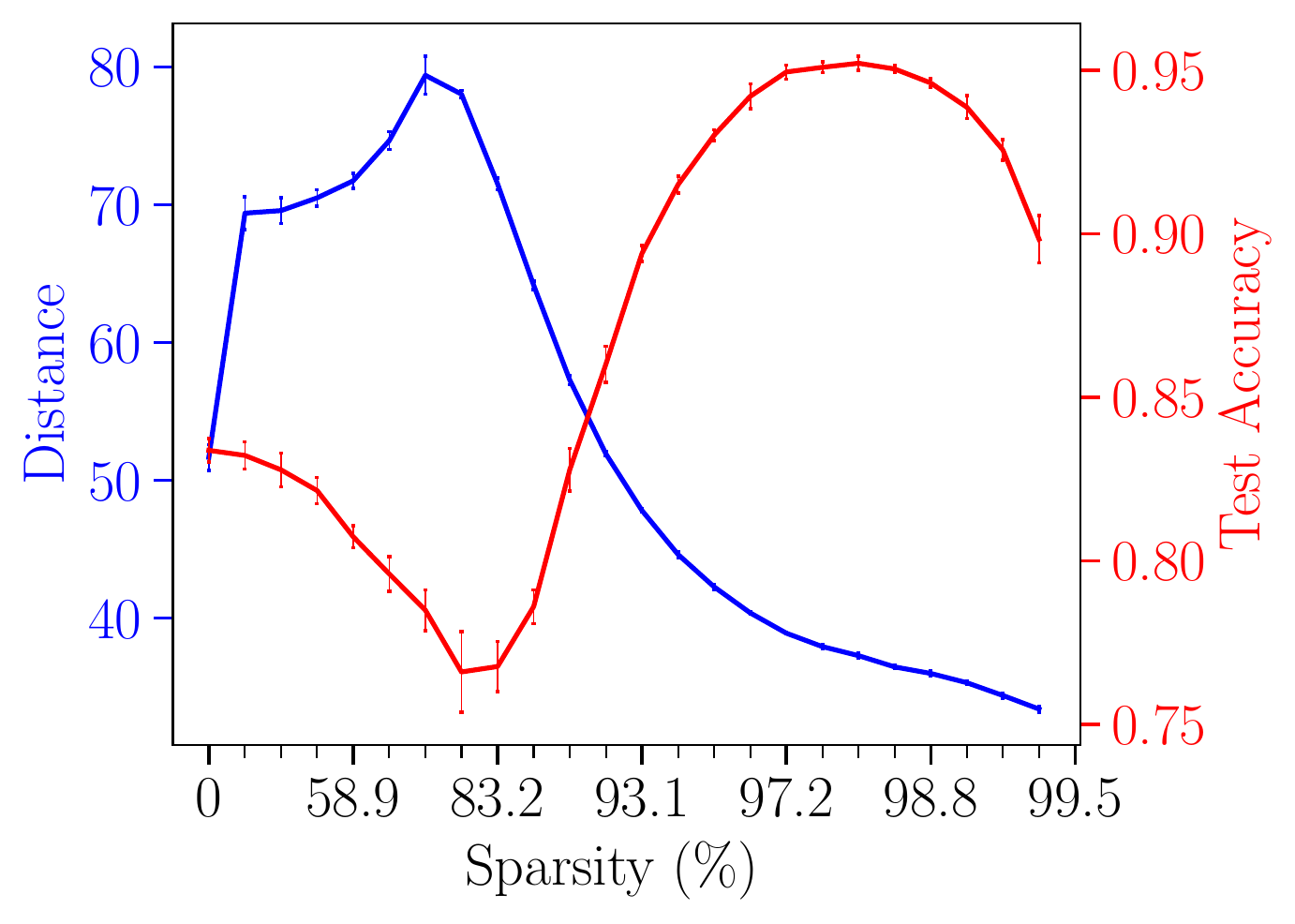}
    \caption{Learning distance for LeNet-300-100 on MNIST with $\epsilon=20\%$. The blue lines refer to $\ell_2$ learning distance and the red lines are test accuracy. \textbf{Left}: Retraining using learning rate rewinding. \textbf{Right}: Scratch retraining.}
    \label{fig:distance-retrain-mnist-0.2}
\end{figure}

\begin{figure}[H]
    \centering
    \includegraphics[width=0.3\linewidth]{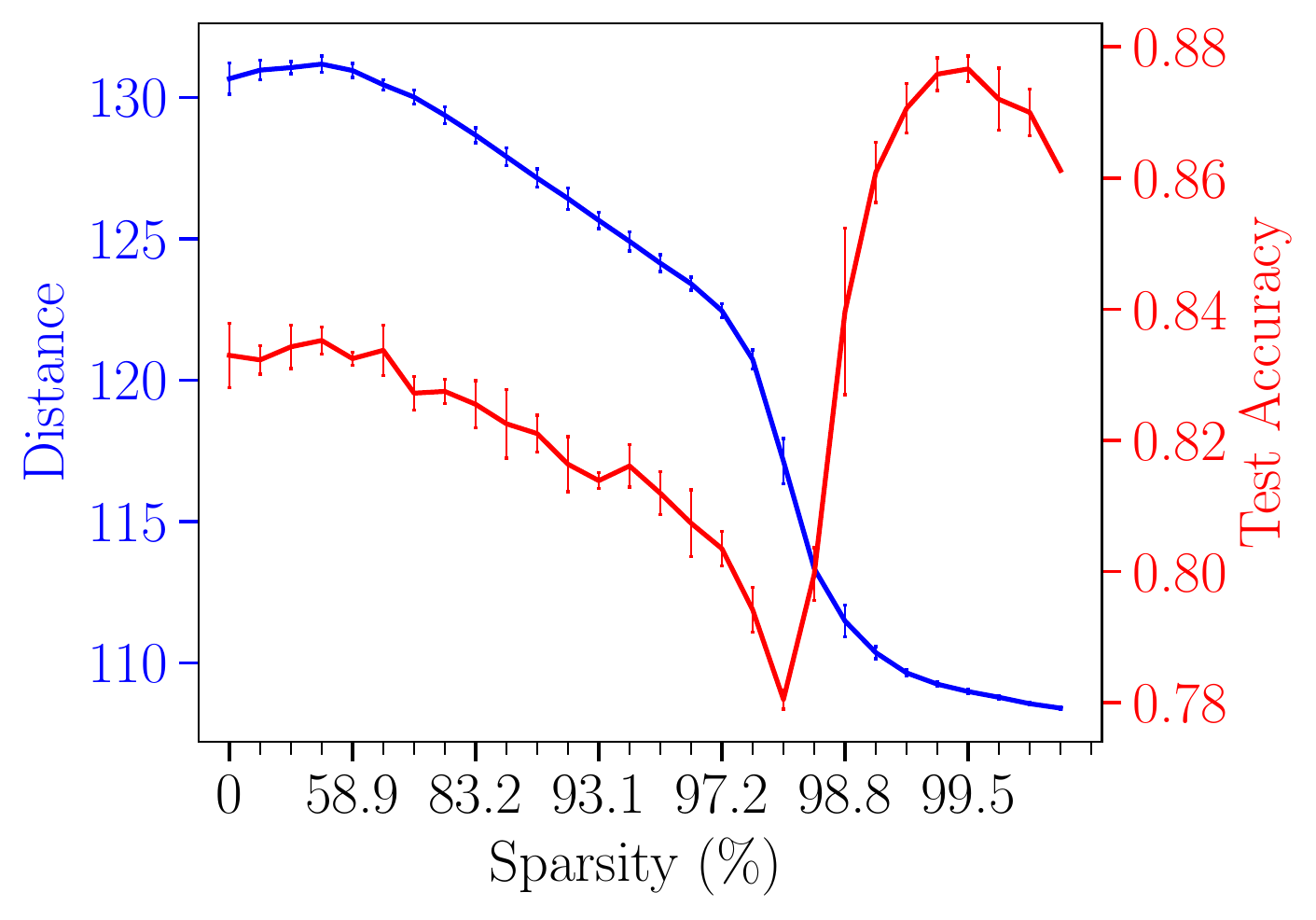}
    \includegraphics[width=0.3\linewidth]{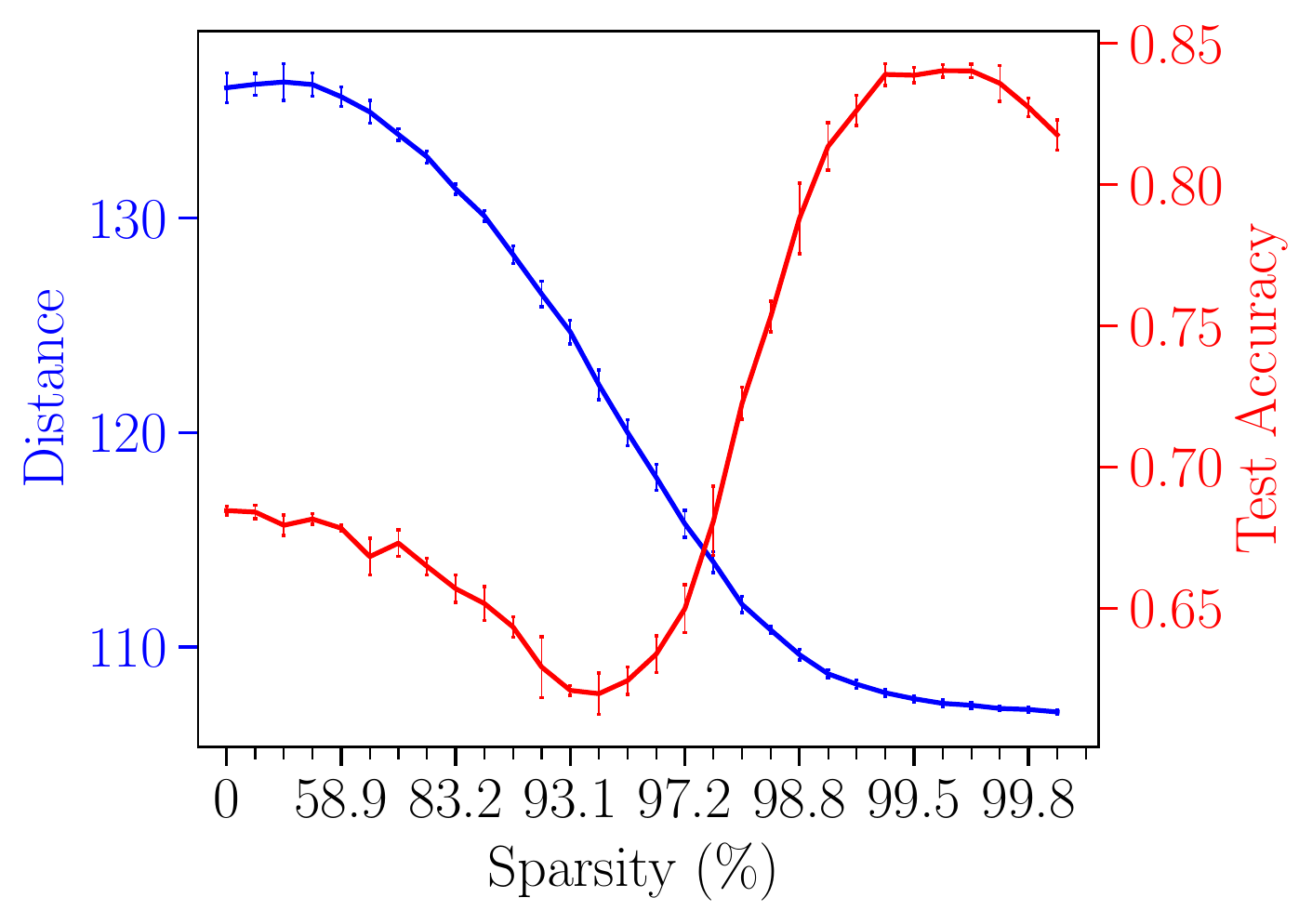}
    \includegraphics[width=0.3\linewidth]{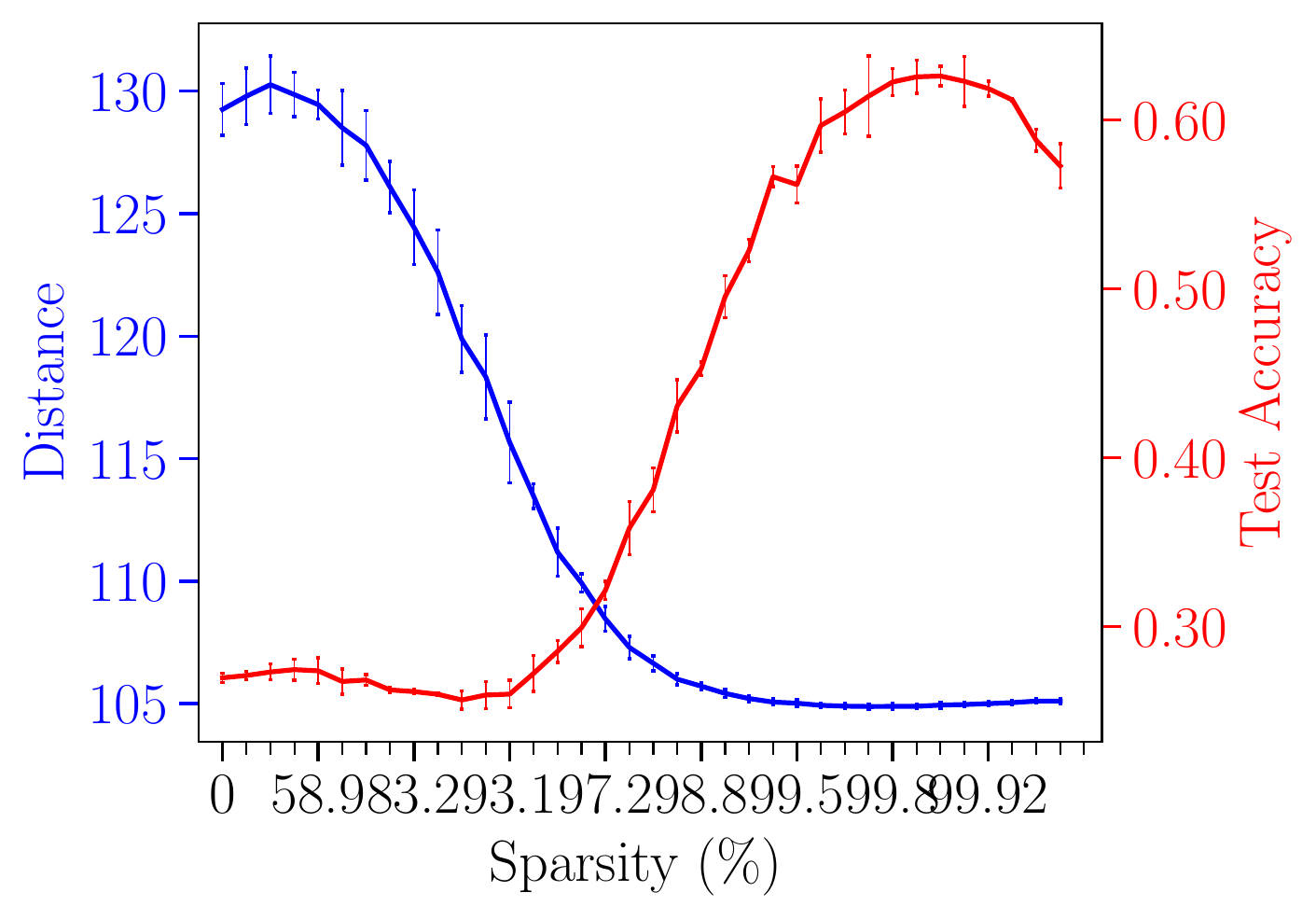}
    \caption{Learning distance for ResNet-18 on CIFAR-10. The blue lines refer to $\ell_2$ learning distance and the red lines are test accuracy. \textbf{Left}: $\epsilon=20\%$. \textbf{Middle}: $\epsilon=40\%$. \textbf{Right}: $\epsilon=80\%$}
    \label{fig:distance-sym-cifar10}
\end{figure}

\begin{figure}[H]
    \centering
    \includegraphics[width=0.3\linewidth]{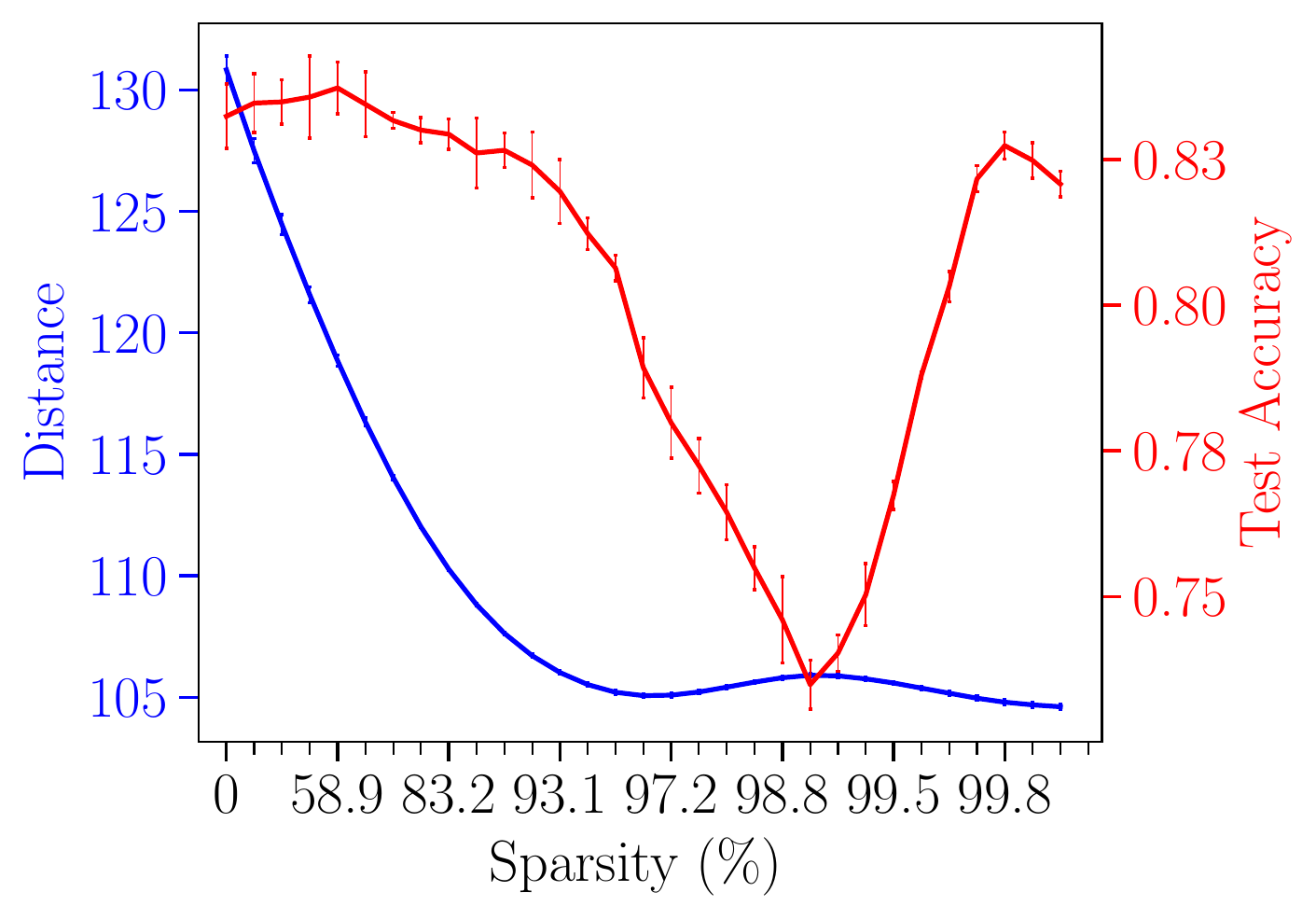}
    \includegraphics[width=0.3\linewidth]{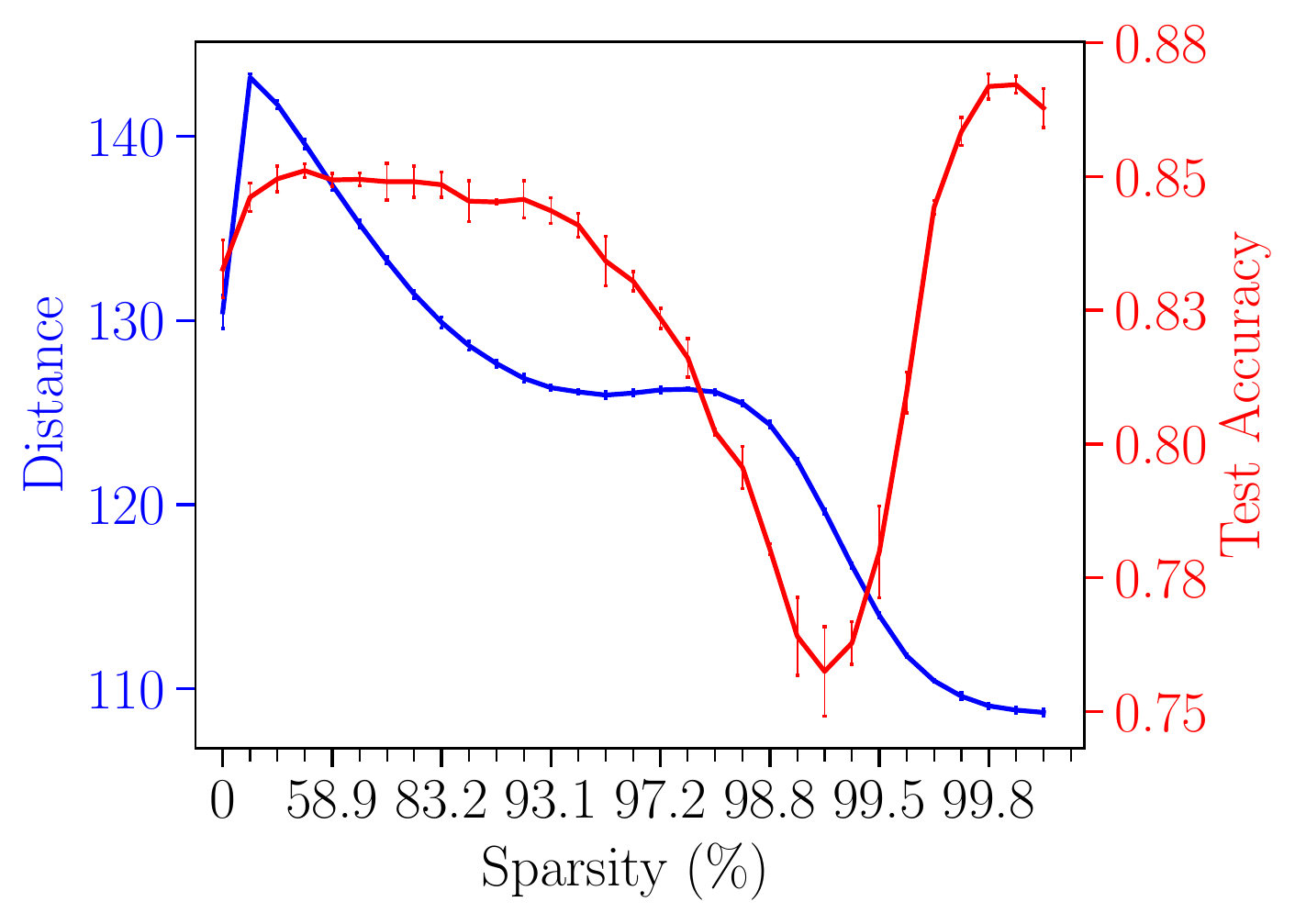}
    \includegraphics[width=0.3\linewidth]{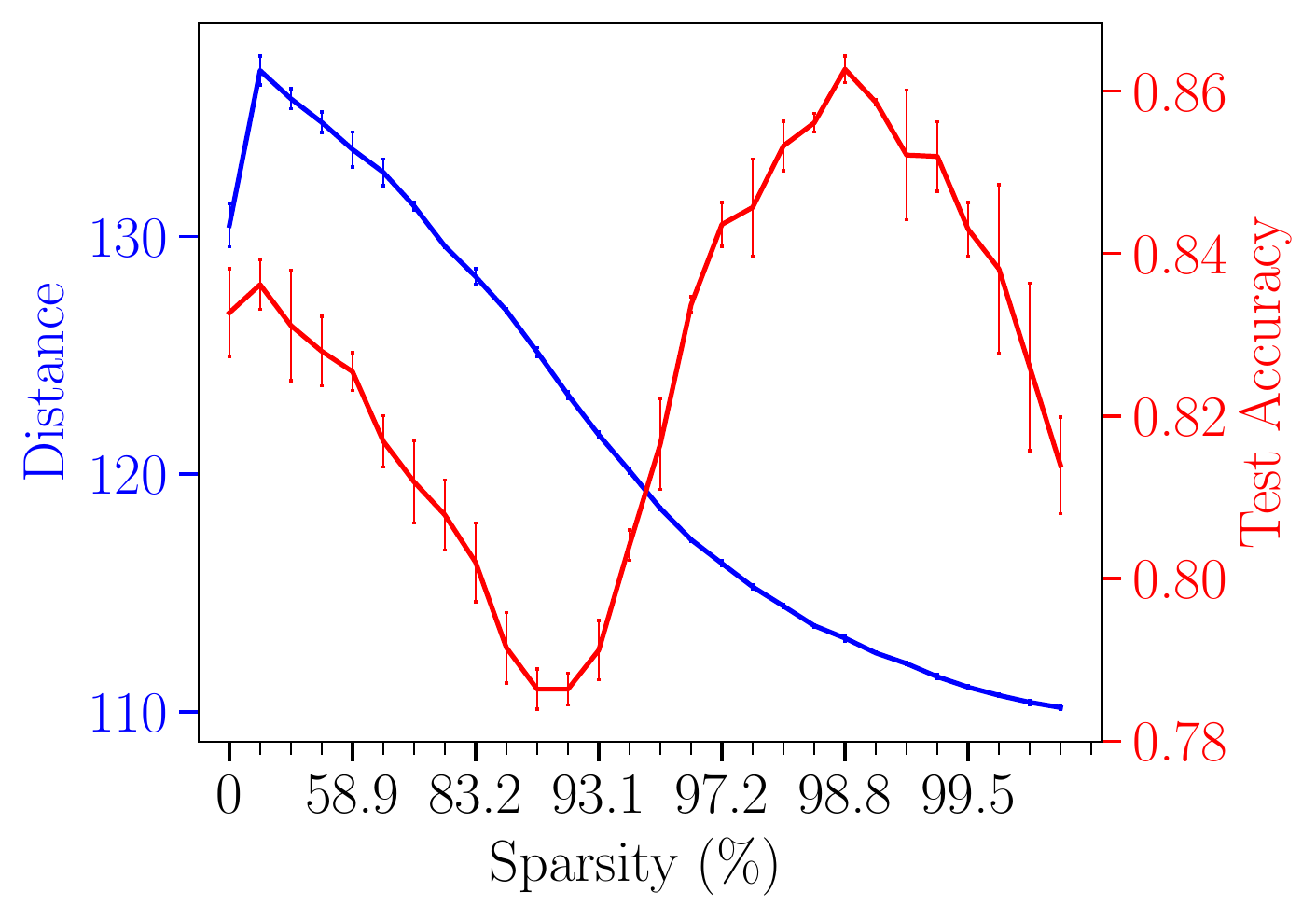}
    \caption{Learning distance for ResNet-18 on CIFAR-10 with different retraining methods. The blue lines refer to $\ell_2$ learning distance and the red lines are test accuracy. \textbf{Left}: Finetuning. \textbf{Middle}: Learning rate rewinding. \textbf{Right}: Scratch retraining.}
    \label{fig:distance-retrain-cifar10-0.2}
\end{figure}

\begin{figure}[H]
    \centering
    \includegraphics[width=0.3\linewidth]{img/cifar100/distance/0.2_sym_mag_accuracy-distance.pdf}
    \includegraphics[width=0.3\linewidth]{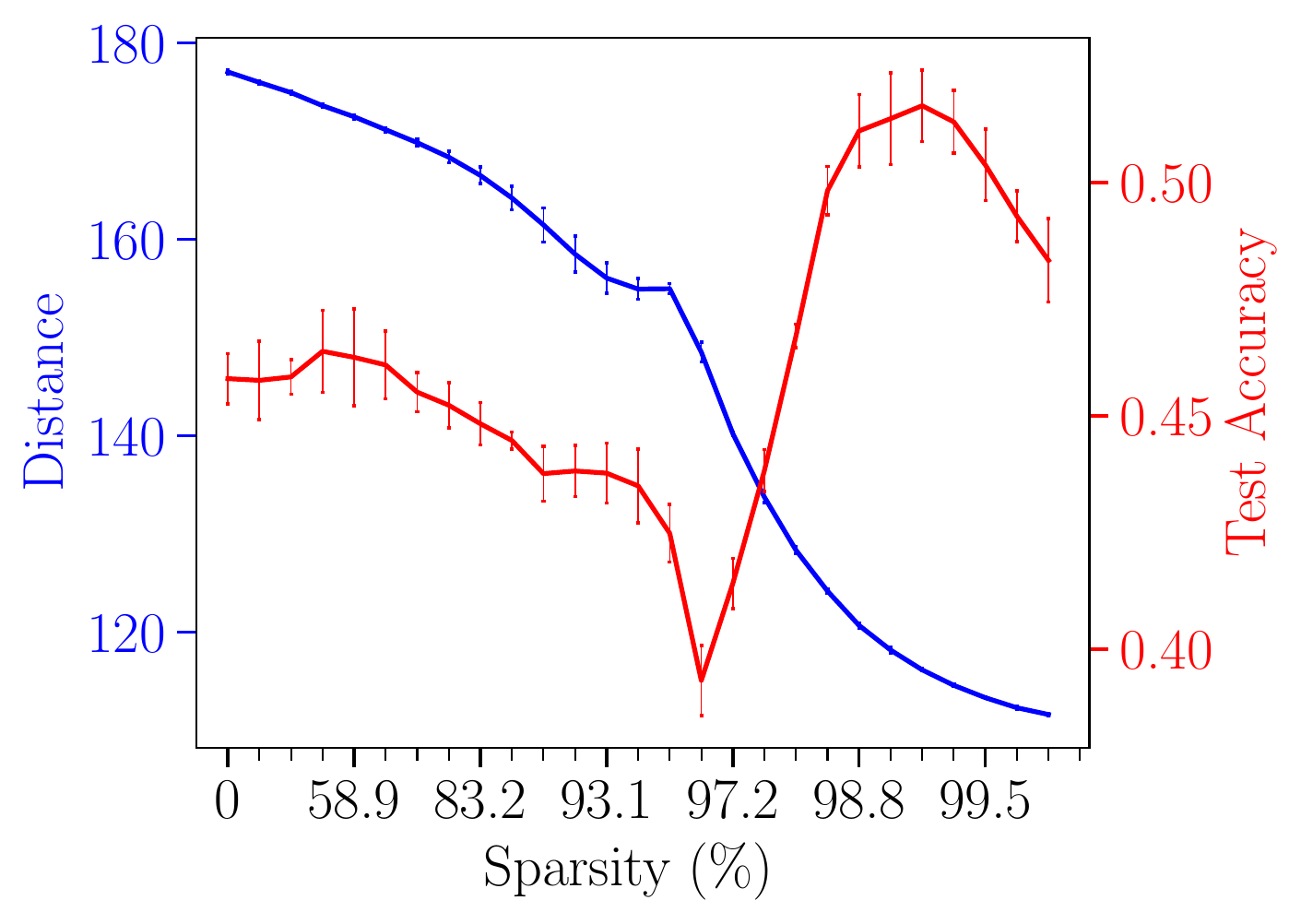}
    \includegraphics[width=0.3\linewidth]{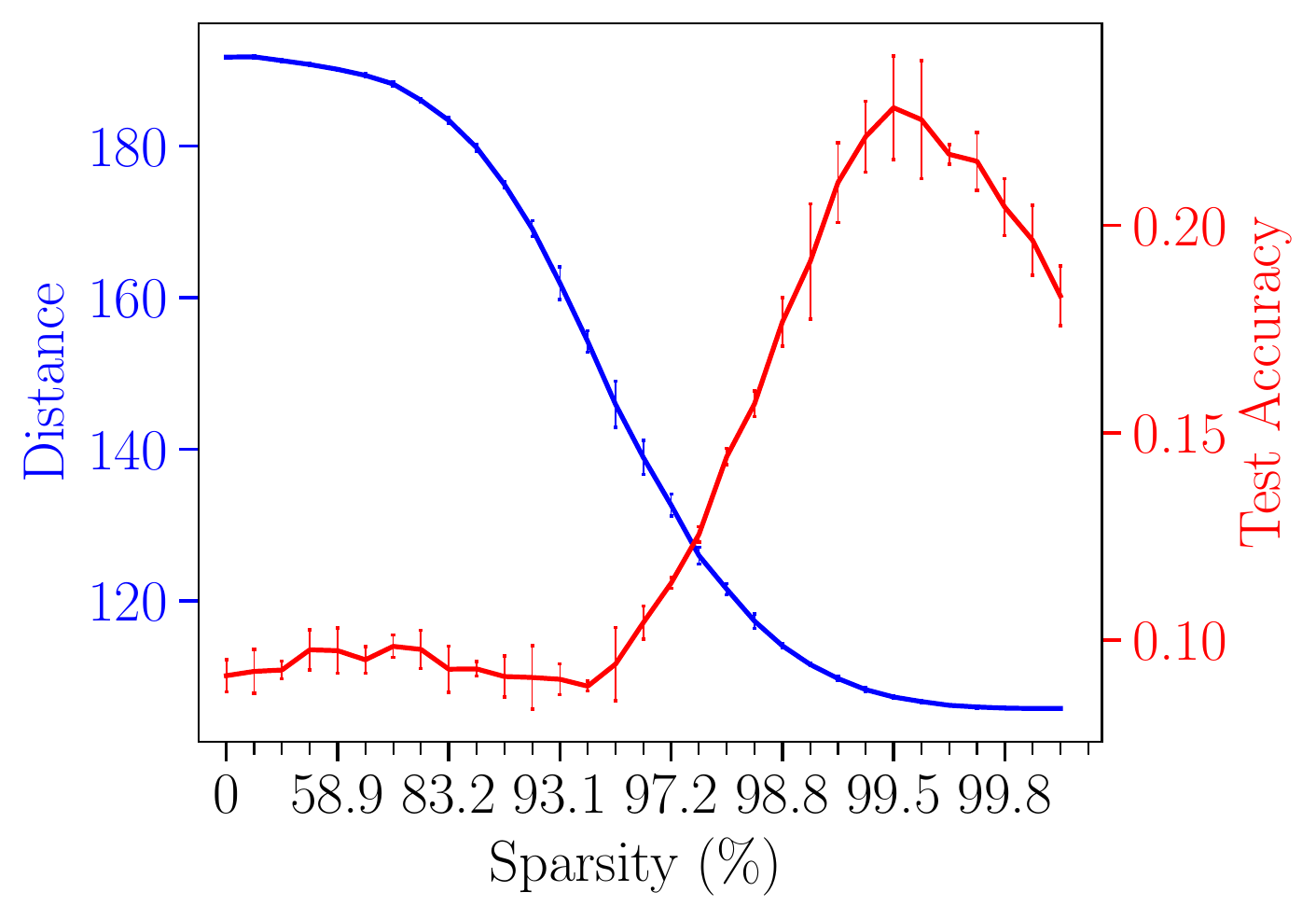}
    \caption{Learning distance for ResNet-18 on CIFAR-100. The blue lines refer to $\ell_2$ learning distance and the red lines are test accuracy. \textbf{Left}: $\epsilon=20\%$. \textbf{Middle}: $\epsilon=40\%$. \textbf{Right}: $\epsilon=80\%$}
    \label{fig:distance-sym-cifar100}
\end{figure}

\begin{figure}[H]
    \centering
    \includegraphics[width=0.3\linewidth]{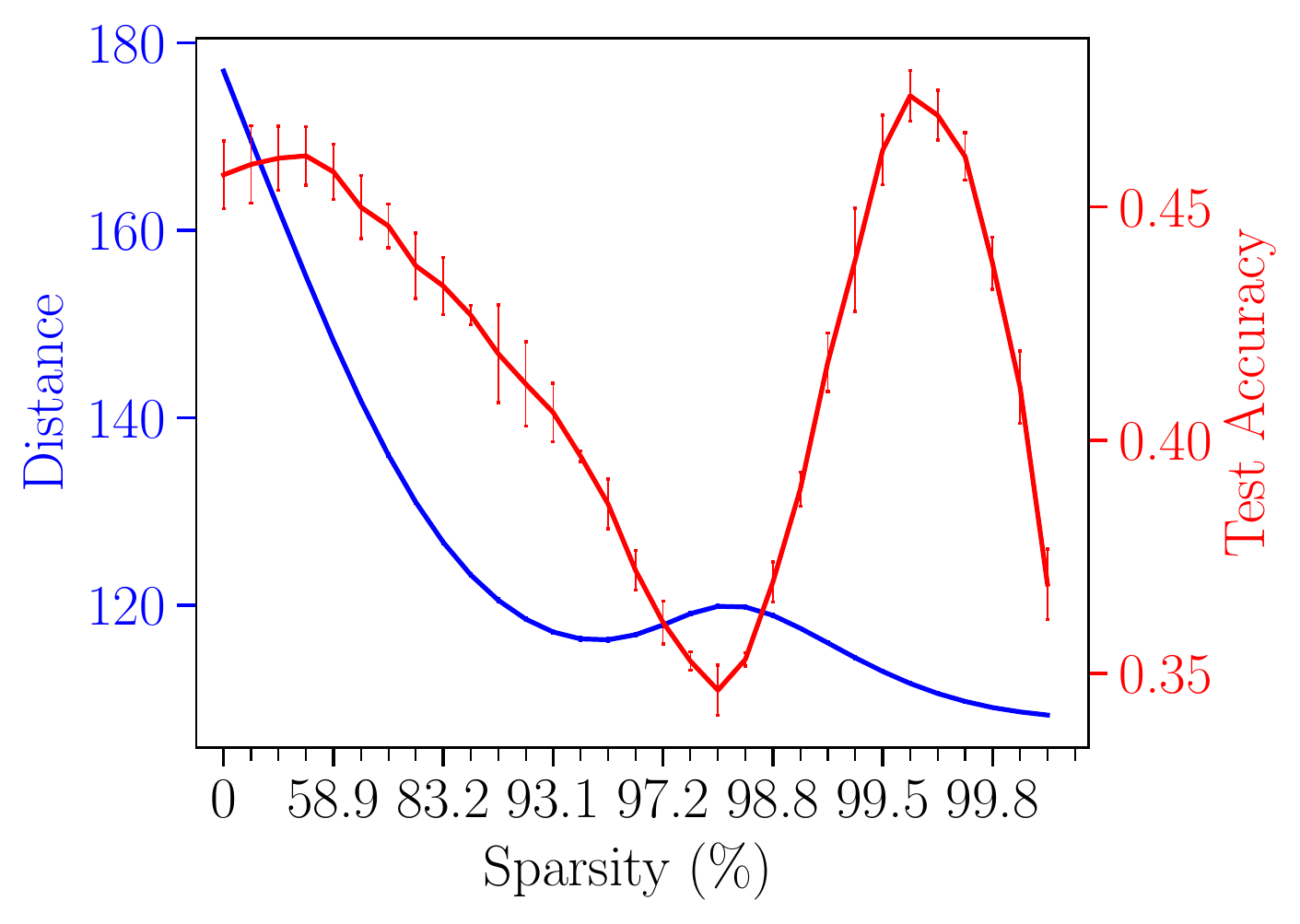}
    \includegraphics[width=0.3\linewidth]{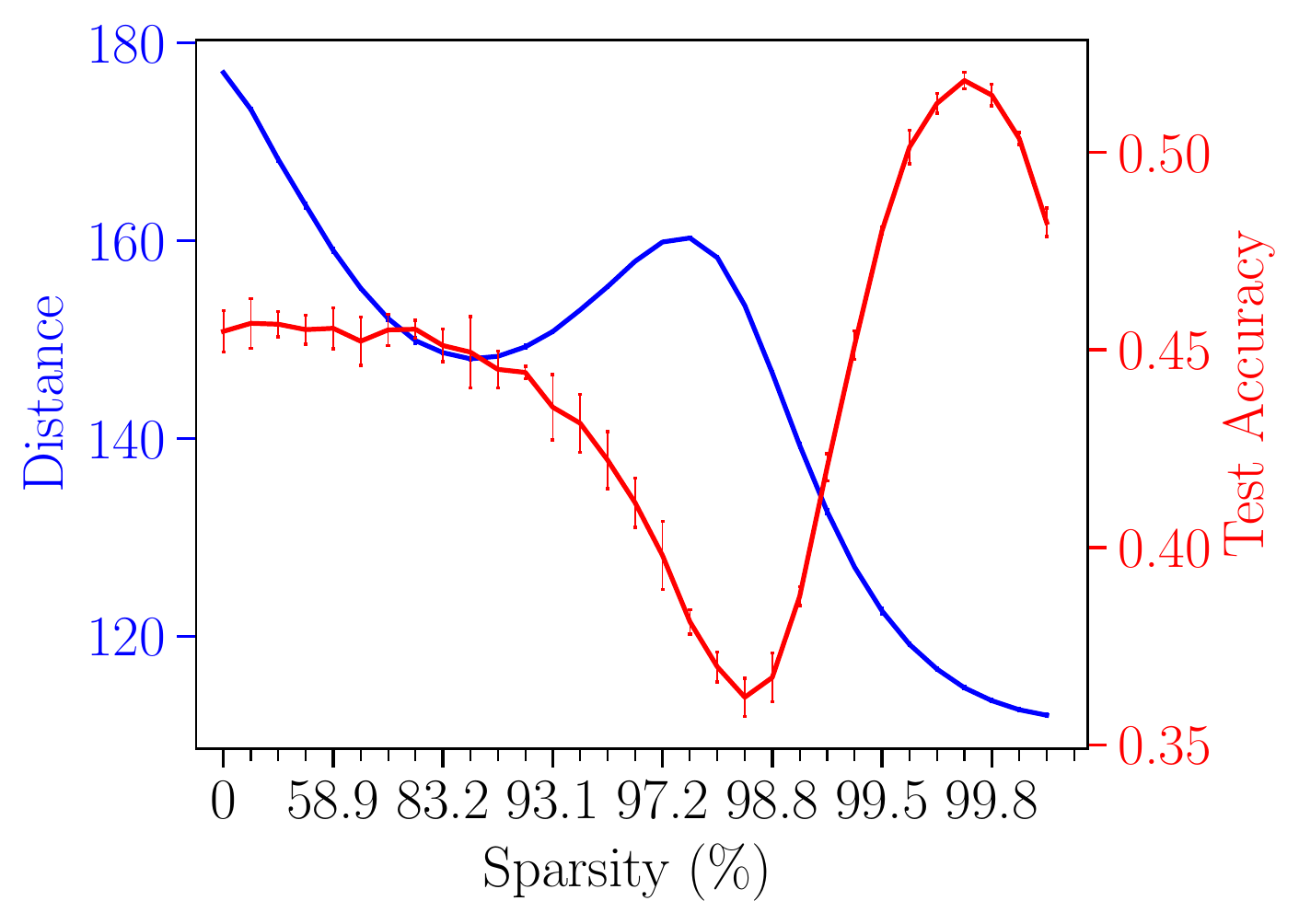}
    \includegraphics[width=0.3\linewidth]{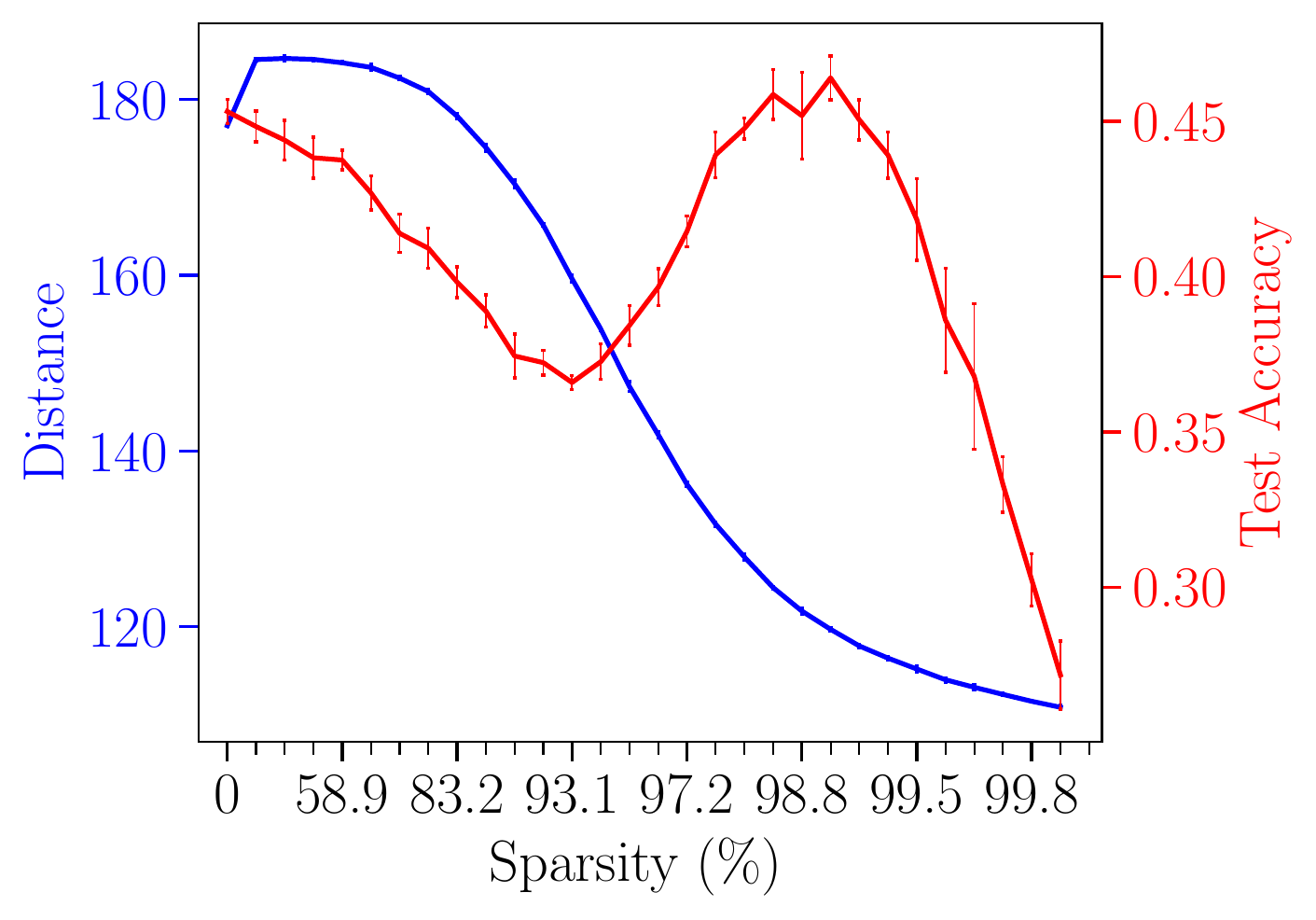}
    \caption{Learning distance for ResNet-18 on CIFAR-100 with different retraining methods. The blue lines refer to $\ell_2$ learning distance and the red lines are test accuracy. \textbf{Left}: Finetuning. \textbf{Middle}: Learning rate rewinding. \textbf{Right}: Scratch retraining.}
    \label{fig:distance-retrain-cifar100-0.4}
\end{figure}

\end{document}